\documentclass[10pt,twocolumn,letterpaper]{article}

\usepackage[pagenumbers]{wacv} %
\usepackage[accsupp]{axessibility}

\usepackage{graphicx}
\usepackage[export]{adjustbox}
\usepackage{booktabs}
\usepackage{soul}
\usepackage{enumitem}
\usepackage[table]{xcolor}
\usepackage{multirow}
\usepackage{makecell}
\usepackage{xcolor}
\usepackage{caption}
\usepackage{subcaption}
\usepackage{array}
\usepackage{algorithm}
\usepackage{algorithmicx}
\usepackage{algpseudocode}
\usepackage{listings}
\usepackage{mathtools}

\AtEndPreamble{
    \usepackage[capitalize]{cleveref}
    \crefname{section}{Sec.}{Secs.}
    \Crefname{section}{Section}{Sections}
    \crefname{table}{Tab.}{Tabs.}
    \Crefname{table}{Table}{Tables}
    \crefname{supp}{Supplementary}{Supplementary}
    \Crefname{supp}{Supplementary}{Supplementary}
}

\usepackage{calc}      %

\usepackage{amsmath}
\usepackage{amssymb}
\usepackage{amsfonts}
\usepackage{bm}

\newcommand{\mat}[1]{\mathbf{#1}}
\newcommand{\vect}[1]{\mathbf{#1}}

\newcommand{\graph}{\mathcal{G}}
\newcommand{\vertices}{\mathcal{V}}
\newcommand{\edges}{\mathcal{E}}

\newcommand{\R}{\mathbb{R}}

\newcommand{\set}[1]{\mathcal{#1}}

\usepackage{tikz}
\usepackage{pgfplots}
\pgfplotsset{compat=1.18}

\usepackage[accsupp]{axessibility}  %
\definecolor{commentgreen}{rgb}{0.1, 0.6, 0.1}
\definecolor{keywordblue}{rgb}{0.0, 0.0, 0.8}
\definecolor{stringred}{rgb}{0.8, 0.1, 0.1}

\definecolor{wacvblue}{rgb}{0.21,0.49,0.74}
\usepackage[breaklinks,colorlinks,allcolors=wacvblue]{hyperref}

\def\wacvPaperID{2725} %
\def\confName{WACV}
\def\confYear{2027}

\title{PANC: Prior-Aware Normalized Cut for Training-Free In-Context Segmentation}

\author{Juan Guti\'errez$^{1,2}$ \qquad Victor Guti\'errez-Garc\'ia$^{2}$ \qquad Jos\'e Luis Blanco-Murillo$^{1,2}$\\[4pt]
$^{1}$Information Processing and Telecommunications Center\\
$^{2}$ETSI de Telecomunicaci\'on, Universidad Polit\'ecnica de Madrid, Spain\\
{\tt\small \{juan.gutierrez, v.ggarcia, jl.blanco\}@upm.es}
}

\begin{document}
\maketitle
\begin{abstract}
We study training-free in-context segmentation with externally annotated context priors: given a target image and a small set of foreground/background tokens extracted from annotated context images, the goal is to segment the corresponding target concept without training, fine-tuning, or target-image clicks. Existing ViT-token spectral methods such as TokenCut are class-agnostic and can select the wrong object in multi-object scenes, or become unstable in low-semantic texture-dominated domains. We propose PANC, a prior-aware spectral segmentation pipeline that injects external labeled tokens into a frozen ViT affinity graph through two foreground/background anchor nodes. The anchors do not clamp target-image tokens; instead, they softly bias the Normalized Cut eigenspace while preserving the target-image affinity structure. PANC further uses prior-aware sign stabilization and prior-driven thresholding to obtain deterministic masks. Across heterogeneous saliency datasets and homogeneous low-semantic domains, PANC improves over architecture-matched no-prior token-graph baselines, with the largest gains in settings where class-agnostic saliency is ambiguous or weak. Available at \url{https://github.com/jgnav/PANC}.
\end{abstract}

\section{Introduction}
\label{sec:intro}

Annotating per-pixel segmentation masks at scale is labor-intensive and time-consuming. Building large supervised datasets demands extensive manual effort and annotation infrastructure \cite{lin2014microsoft,bearman2016s}. This motivates segmentation methods that reduce annotation burden by relying on either unsupervised discovery or sparse forms of guidance. Recent unsupervised pipelines based on self-supervised Vision Transformer (ViT \cite{dosovitskiy2021imageworth16x16words}) tokens, most notably TokenCut \cite{wang2023tokencutsegmentingobjectsimages} and related token-ranking heuristics, exploit dense frozen patch embeddings to produce class-agnostic object masks without input labels \cite{siméoni2021localizingobjectsselfsupervisedtransformers, wang2023tokencutsegmentingobjectsimages, melas2022deep}. These methods show competitive unsupervised object discovery results on standard benchmarks, yet their outputs remain underconstrained when the target is not explicitly specified. In multi-object or ambiguous scenes, \textit{post hoc} selection based on saliency, ranking, or heuristic thresholds may select different entities under minor variations. The same heuristics often break down on imagery with weak semantic cues or nearly homogeneous appearance, where saliency does not necessarily correlate with the task-relevant structure \cite{wang2023tokencutsegmentingobjectsimages, siméoni2021localizingobjectsselfsupervisedtransformers}.

\begin{figure*}[t]
\centering
    \includegraphics[width=0.7\linewidth]{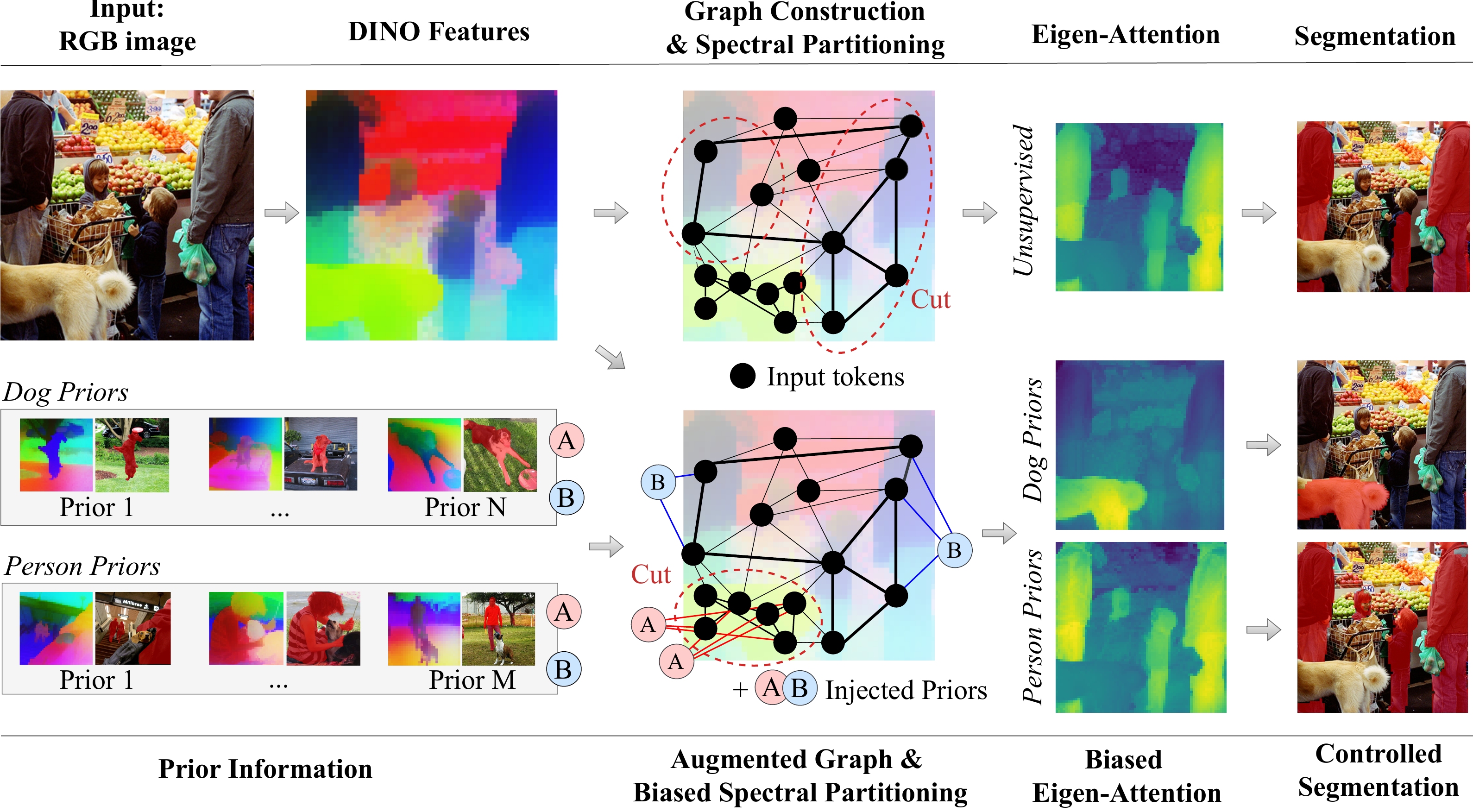}
    \caption{PANC performs external-prior guided spectral segmentation. Given a target image and sparse foreground/background tokens from annotated context images, PANC augments a frozen ViT token graph with label anchors. The resulting spectral partition is biased toward the concept specified by the external priors, enabling controlled segmentation without target-image clicks.}   
     \label{FIG:PLACEHOLDER}
\end{figure*}

Weak supervision offers a middle ground: a small amount of targeted input can resolve ambiguities and inject semantic intent into propagation or grouping algorithms \cite{bearman2016s, lin2016scribblesup, ahn2018learning}. However, different forms of guidance lead to different problem settings. Interactive segmentation methods use clicks, scribbles, or seeds placed directly on the target image \cite{rother2004grabcut, grady2006random, zhu2003semi}. PANC uses \emph{external context priors} by incorporating foreground and background tokens from annotated examples into the target token graph, guiding segmentation without interacting with the target image.

We introduce PANC, a training-free, spectral segmentation framework guided by external priors that injects a compact set of labeled context tokens into the Normalized Cut \cite{shi2000normalized} formulation built on frozen self-supervised ViT features. In Figure~\ref{FIG:PLACEHOLDER}, external prior tokens are connected to foreground/background anchors, steering the eigenspace toward prior-consistent partitions while retaining the target-image affinity structure. Leveraging DINOv3-style tokens, we avoid dataset-specific affinity learning while exploiting the stability of recent encoders \cite{siméoni2025dinov3, oquab2024dinov2learningrobustvisual}.

Our main contributions include: 
\begin{itemize}[nosep,leftmargin=*] 
    \item \textbf{External-prior guided spectral segmentation.} We study a training- and target-interaction-free setting where sparse foreground/background tokens from annotated context images guide target-image segmentation, without target-image clicks or model training. 
    \item \textbf{Anchor-augmented ViT token graphs.} We inject external prior labels into a frozen ViT affinity graph through foreground/background anchors, softly biasing the NCut solution while preserving target-image affinities. Together with prior-aware sign orientation and thresholding, this yields deterministic masks from external guidance. 
    \item \textbf{Evaluation under matched protocols.} We evaluate across heterogeneous and homogeneous domains, compare against architecture-matched token-graph baselines, and ablate prior count, anchor coupling, temperature, resolution, thresholding, and prior error. GPU memory and computational scaling are reported in Supp.~A. 
\end{itemize}
\section{Related Work}
\label{sec:related}

\paragraph{\textbf{Self-supervised Vision Transformers.}}
Self-supervised Vision Transformers provide dense token embeddings from late layers whose pairwise similarities often correlate with object parts and extents, making them a strong substrate for label-free grouping \cite{caron2021emergingpropertiesselfsupervisedvision, he2021maskedautoencodersscalablevision, zhou2022ibotimagebertpretraining, assran2023selfsupervisedlearningimagesjointembedding}. Scaling and curation (DINOv2) yield frozen features that transfer robustly to dense prediction \cite{oquab2024dinov2learningrobustvisual}, while recent refinements (DINOv3) improve cross-resolution stability and geometric consistency---properties that directly influence the quality of token affinity graphs and their spectra \cite{siméoni2025dinov3}. In practice, patch stride limits token resolution, motivating long-sequence encodings and feature upsampling to recover fine detail from frozen backbones \cite{siméoni2025dinov3, fu2024featupmodelagnosticframeworkfeatures, docherty2025upsamplingdinov2featuresunsupervised}.

\paragraph{\textbf{Unsupervised image segmentation with token graphs.}}
Unsupervised image segmentation evolved from co-localization/co-segmentation to single-image pipelines built on dense learned features \cite{joulin2010discriminative, joulin2012multi, vicente2011object, cho2015unsupervised, vo2019unsupervised, vo2020toward}. Spectral analyses on feature affinities can produce meaningful areas without labels \cite{melas2022deep}, and ViT-token methods such as LOST and TokenCut build graphs from frozen self-supervised embeddings and extract objects via simple spectral/graph criteria, achieving strong zero-shot results on DUTS-TE, DUT-OMRON, and ECSSD datasets \cite{siméoni2021localizingobjectsselfsupervisedtransformers, wang2023tokencutsegmentingobjectsimages}. However, these class-agnostic pipelines remain under-specified for \emph{targeted} segmentation (\eg, COCO/VOC datasets \cite{lin2014microsoft, everingham2010pascal}). In multi-object scenes, different choices can select different entities, and in low-contrast or near-homogeneous imagery, the affinity graph can be weakly informative. %

\paragraph{\textbf{Spectral clustering and graph-based segmentation.}}
Image segmentation can be formulated as a balanced partitioning problem on an affinity graph. This has been solved using normalized cut \cite{shi2000normalized, von2007tutorial}. 
From a spectral perspective, segmentation quality depends on the structure of the graph spectrum: when low-frequency modes are poorly separated, the resulting partitions become unstable and sensitive to noise and thresholding.
Practical systems therefore rely on constraints and scalable approximations, including sparse $k$-NN graphs, Nyström/landmark methods, and iterative eigensolvers \cite{ng2001spectral, chen2011large, pourkamali2020scalable, wang2014constrained, xu2009fast}. Nodes associated with ViT tokens enable coherent spectral groupings and remain sensitive to affinity construction and to semantic guidance. %

\paragraph{\textbf{Seeded and weakly supervised segmentation.}}
Weak cues (points, scribbles, boxes, or image-level tags) are commonly converted into dense masks by seed generation and propagation, followed by boundary refinement \cite{bearman2016s, lin2016scribblesup, dai2015boxsup, ahn2018learning, papandreou2015weakly, kolesnikov2016seed}. Classical interactive methods (GrabCut, random walks) have already shown that a few seeds can steer graph-based objectives toward high-quality masks \cite{rother2004grabcut, grady2006random, zhu2003semi}, and subsequent pipelines have improved seed quality with class activation maps (CAMs) and learned affinity networks \cite{ahn2018learning, papandreou2015weakly}. Constrained spectral formulations provide an alternative to learned affinities by encoding sparse labels directly in the graph objective while preserving global consistency \cite{wang2014constrained, xu2009fast, tang2018normalized}. These trends suggest a natural synthesis: use geometry-robust self-supervised ViT tokens as graph nodes and inject compact priors to bias the low-frequency eigenspace, enabling controllable, test-time segmentation without per-dataset affinity learning.

\paragraph{\textbf{Positioning relative to constrained and interactive graph methods.}} PANC does not claim that constrained graph partitioning or seed-guided segmentation is new: graph cuts, random walks, and constrained spectral clustering already use seeds, labels, or pairwise constraints to guide graph objectives \cite{rother2004grabcut, grady2006random, zhu2003semi, wang2014constrained, xu2009fast, tang2018normalized}. PANC differs in the source of supervision: it injects labeled tokens from external context images into a frozen ViT token graph, without target-image clicks or learned affinity models.
\section{Method}
\label{sec:method}

Let $\mat{I} \in \R^{H \times W \times 3}$ denote an input RGB image and let $\{\vect{f}_i\}_{i=1}^n$ be its frozen ViT patch tokens. PANC also receives a sparse external prior set $\set{P}=\set{P}_{+}\cup\set{P}_{-}$, where $\set{P}_{+}$ contains foreground tokens and $\set{P}_{-}$ contains background tokens extracted from annotated context images. The task is to predict a binary mask over the target-image tokens only. Priors are not clicks on the target image: they are context examples that specify the intended concept. PANC is therefore training- and target-interaction-free, but not annotation-free: priors are extracted from externally annotated context images. %

\subsection{Preliminaries}
\label{sec:preliminaries}

\paragraph{\textbf{ViT Tokenization.}}
We process $\mat{I}$ using a frozen Vision Transformer (ViT) backbone trained with self-supervision (e.g., DINO \cite{siméoni2025dinov3}). The image is partitioned into a regular grid of $n$ non-overlapping patches of fixed size, which are mapped through the transformer layers to yield a set of token embeddings in $\R^d$ as $\{\vect{f}_i\}_{i=1}^n$.

\paragraph{\textbf{Normalized Cut.}}
Unsupervised object discovery can be formulated as a graph partitioning problem over a fully connected, undirected affinity graph $\graph = (\vertices, \edges)$, where the vertices $\vertices$ represent the ViT tokens and edges $\edges$ represent their pairwise similarities. Let $\mat{W} \in \R^{n \times n}$ be the affinity matrix, where each entry $\mat{W}_{ij} \ge 0$ measures the similarity between tokens $i$ and $j$. The objective of the Normalized Cut (NCut) is to split $\graph$ into two disjoint subgraphs, $\vertices_a$ and $\vertices_b$, by minimizing the cut cost relative to the volume of the subsets: 
\begin{equation}
\operatorname{NCut}(\vertices_a, \vertices_b) = \frac{\operatorname{cut}(\vertices_a, \vertices_b)}{\operatorname{assoc}(\vertices_a, \vertices)} + \frac{\operatorname{cut}(\vertices_a, \vertices_b)}{\operatorname{assoc}(\vertices_b, \vertices)},
\label{eq:ncut}
\end{equation}
\noindent where $\operatorname{cut}(\vertices_a, \vertices_b) = \sum_{i \in \vertices_a, j \in \vertices_b} \mat{W}_{ij}$ is the total weight of edges connecting the two partitions, and $\operatorname{assoc}(\vertices_a, \vertices) = \sum_{i \in \vertices_a, k \in \vertices} \mat{W}_{ik}$ is the total connection weight from $\vertices_a$ to all nodes in the graph. Minimizing this objective is NP-hard, but it can relax into a generalized eigenvalue problem: 
\begin{equation}
(\mat{D} - \mat{W})\vect{y} = \lambda \mat{D}\vect{y},
\label{eq:ncut_eigensystem}
\end{equation}
\noindent where $\mat{D} = \operatorname{diag}(\mat{W}\vect{1})$ is the degree matrix and $\vect{1}$ is an all-ones vector \cite{shi2000normalized}. The trivial solution $\lambda_1=0$, $\vect
y = \vect{1}$ is excluded via the constraint $\vect{y}^\top \mat{D}\vect{1} = 0$, yielding the second smallest eigenpair as the solution. 

\paragraph{\textbf{Eigen-Attention.}}
The continuous solution to the NCut relaxation is given by the eigenvector $\vect{y} \in \R^{n}$ corresponding to the second smallest eigenvalue, commonly referred to as the Fiedler vector. Because the input tokens represent a spatial grid, the values of $\vect{y}$ can be reshaped back to the original patch layout to form a continuous feature map. In the context of dense self-supervised ViT features, this map---termed eigen attention---smoothly localizes the most salient semantic regions in the image.

\subsection{Prior-Aware Normalized Cut (PANC)}
\label{sec:panc_algorithm}

\paragraph{\textbf{Augmented Graph Construction.}}
To move from unsupervised discovery to controllable segmentation, one can augment the affinity graph $\graph$ (cf. \cref{sec:preliminaries}) with a small set of externally provided priors obtained from the same ViT. In addition to the image-token features, $\{\mathbf f_i\}_{i=1}^{n}$, let $\{\mathbf p_i\}_{i=1}^{m}$ be $m$ annotated prior features. We concatenate both sets to form the feature matrix $\mat{F}=[\mathbf f_1, \ldots, \mathbf f_n , \mathbf p_1, \ldots, \mathbf p_m]^\top \in\mathbb R^{N\times d}$
with $N=n+m$ and $\ell_2$-normalize the rows of $\mat{F}$ so that the cosine similarities between tokens $i$ and $j$ are $S_{ij}=(\mat{F}\mat{F}^\top)_{ij}$.

We convert similarities to positive affinities using a temperature kernel, $\mat{W}_{ij}=\exp(S_{ij}/\tau)$, with $\tau>0$, and set $\mat{W}_{ii}=0$. Setting $\mat{W}_{ii}=0$ only removes self-loops; the degree remains $\mat{D}_{ii}=\sum_{j\neq i}\mat{W}_{ij}$ and is therefore not zero. This yields a weighted graph over the $N$ tokens with degree matrix $\mat{D}=\mathrm{diag}(\mat{W}\mathbf 1)$ and Laplacian $\mat{L}=\mat{D}-\mat{W}$.

Let $\set{Q} = \{1, \dots, n\}$ be the index set of image (query) tokens, and $\set{P} = \{n+1, \dots, N\}$ be the index set of prior tokens. Let $\set{P}_{+}$ and $\set{P}_{-}$ be disjoint index sets within $\set{P}$, denoting priors labeled as the target class (foreground) and the complementary class (background). We introduce two virtual anchor vertices $u_{+}$ and $u_{-}$ and form the augmented graph
$\tilde{\graph}=(\vertices\cup\{u_+,u_-\},\tilde{\edges})$
with block affinity 
\begin{equation}
\tilde{\mat{W}}=
\begin{bmatrix}
\mat{W} & \mat{C}\\
\mat{C}^\top & \mat{0}_{2\times 2}
\end{bmatrix},
\qquad
\mat{C}\in\mathbb R^{N\times 2}.
\label{eq:anchor_block}
\end{equation}
In this structure, only priors connect directly to anchors, so that unannotated image tokens ($i \in \set{Q}$) can be influenced by anchors only through their affinities to priors in $\mat{W}$. Anchors encode foreground/background labels from external priors without clamping query tokens, so target-image affinities remain the dominant structure of the NCut problem.
We use a per-seed adaptive coupling strength $\alpha_i$ scaled to the average local affinity between each prior token $i$ and all $n$ image tokens,
\begin{equation}
\alpha_i = \kappa \cdot ({1}/{n}) \textstyle\sum_{j=1}^{n}\mat{W}_{ij}, \qquad \kappa>0,
\end{equation}
and populate $\mat{C}$ as $\mat{C}_{i,1}=\alpha_i$ if $i\in\set{P}_{+}$, or $\mat{C}_{i,2}=\alpha_i$ if $i\in\set{P}_{-}$, else $0$.

\paragraph{\textbf{Spectral partition and interpretation.}}
Let $\tilde{\mat{D}}=\mathrm{diag}(\tilde{\mat{W}}\mathbf 1)$ and
$\tilde{\mat{L}}=\tilde{\mat{D}}-\tilde{\mat{W}}$ be the degree matrix
and Laplacian of the anchor-augmented graph. PANC applies the
standard relaxed Normalized Cut formulation to this augmented graph:
\begin{equation}
\label{eq:ncutsolution}
\tilde{\mat{L}}\tilde{\mathbf y}
=
\tilde{\lambda}\,\tilde{\mat{D}}\tilde{\mathbf y},
\qquad
\tilde{\lambda}_2
=
\min_{\tilde{\mathbf y}\neq \mathbf 0,\;
\tilde{\mathbf y}^\top \tilde{\mat{D}}\mathbf 1=0}
\frac{
\tilde{\mathbf y}^\top \tilde{\mat{L}}\tilde{\mathbf y}
}{
\tilde{\mathbf y}^\top \tilde{\mat{D}}\tilde{\mathbf y}
}.
\end{equation}
As in standard NCut and spectral clustering
\cite{shi2000normalized, von2007tutorial}, the constraint
$\tilde{\mathbf y}^\top \tilde{\mat D}\mathbf 1=0$ removes the trivial
constant solution, and the second-smallest eigenvector gives a relaxed partition
over image tokens, prior tokens, and the two anchors; the final mask is
obtained only from the image-token entries.

The anchors bias this partition without hard-clamping any target-image
token. Write $\tilde{\mathbf y}=[\mathbf y;\mathbf y_u]$, where
$\mathbf y=[\mathbf y_Q;\mathbf y_P]\in\mathbb R^{N}$ contains the
query and prior scores and $\mathbf y_u=[y_{u_+};y_{u_-}]$ the anchor
scores. The graph energy decomposes as
\begin{equation}
\label{eq:priors_terms}
\begingroup
\setlength{\jot}{2pt}
\begin{aligned}
\tilde{\mathbf y}^{\top}\tilde{\mat L}\tilde{\mathbf y}
={}&
\underbrace{\tfrac12\!\sum_{i,j\in\set Q}\!
\mat W_{ij}(y_i-y_j)^2}_{\text{\scriptsize query-only}}
+
\underbrace{\tfrac12\!\sum_{i,j\in\set P}\!
\mat W_{ij}(y_i-y_j)^2}_{\text{\scriptsize prior self}} \\
&+
\underbrace{\sum_{\substack{i\in\set Q, \,j\in\set P}}\!
\mat W_{ij}(y_i-y_j)^2}_{\text{\scriptsize query--prior}}
+
\underbrace{\tilde{\mathbf y}^{\top}\mat L_C\tilde{\mathbf y}}_
{\text{\scriptsize anchor}} .
\end{aligned}
\endgroup
\end{equation}
The first term is the usual image-token NCut energy, which preserves the
spatial and semantic structure of the target image. The second and third
terms keep prior tokens mutually consistent and transfer their influence
to unlabeled image tokens only through the ViT feature affinities
$\mat W$; priors never overwrite the target partition directly. The last
term, the only one introduced by the anchors, reduces to
\begin{equation}
\label{eq:energy}
\tilde{\mathbf y}^\top \mat{L}_C\tilde{\mathbf y}
=
\sum_{i\in\set{P}_+}\alpha_i(y_i-y_{u_+})^2
+
\sum_{i\in\set{P}_-}\alpha_i(y_i-y_{u_-})^2 ,
\end{equation}
\ie, a foreground (background) prior pays energy when its score moves
away from the foreground (background) anchor. The eigensolver therefore
favors partitions in which labeled priors agree with their anchors,
while query tokens, which are not connected to anchors, remain governed
by the target-image affinities.

The coupling weights $\alpha_i$ set the strength of this bias. Small
values keep PANC close to the class-agnostic NCut solution, with priors
mainly selecting among already plausible cuts; larger values help when
the unsupervised graph is weakly separated, as in low-semantic domains.
Since the anchors also change the degree normalization in
\eqref{eq:ncutsolution}, increasing $\alpha_i$ does not simply force all
scores to follow the priors: PANC remains a balanced spectral partition,
\ie, a soft external-prior bias on the NCut eigenspace rather than a
hard seed-propagation rule.

\vspace{-0.5em}
\paragraph{\textbf{Implementation.}}
We solve \eqref{eq:ncutsolution} with a dense GPU eigensolver on
$\mat I-\tilde{\mat D}^{-1/2}\tilde{\mat W}\tilde{\mat D}^{-1/2}$ and set
$\tilde{\mathbf y}=\tilde{\mat D}^{-1/2}\mathbf v$. For numerical
stability, the anchor block in \eqref{eq:anchor_block} stores
$\epsilon=10^{-8}$ on its diagonal, $\alpha_i$ is clipped to
$[10^{-4},10^{3}]$, and the Fiedler vector is the eigenvector of the first
eigenvalue above $\epsilon$ (\ie, $\tilde\lambda_2$ for a connected graph).

\vspace{-1em}
\paragraph{\textbf{Sign Stabilization.}}
We cut the augmented graph using \cref{eq:ncut_eigensystem} substituting $\mat{W}$ with our $\tilde{\mat{W}}$ to obtain  $\tilde{\vec{y}}$. The eigenvector $\tilde{\mathbf y}$ is sign-ambiguous. We orient it deterministically using the priors:
\begin{equation}
\label{eq:mu_foreground}
\mu_{+/-}
=
\operatorname{med}\big(\{y_i : i\in\set{P}_{+/-}\}\big)
\quad
\text{[fore- / back-ground]}
\end{equation}
and flip $\tilde{\mathbf y}\leftarrow-\tilde{\mathbf y}$ if $\mu_{+}<\mu_{-}$; the median makes the orientation robust to outlying prior scores. We then discard the two anchor entries and map the remaining target/prior token scores to $[0,1]$ via min--max normalization, $ s_i=\tfrac{y_i-\min_j y_j}{\max_j y_j-\min_j y_j}$, $i=1,\ldots,N$. This deterministic orientation aligns high scores with the foreground prior class.

\begin{figure*}[t]
    \centering
    \includegraphics[width=0.7\linewidth]{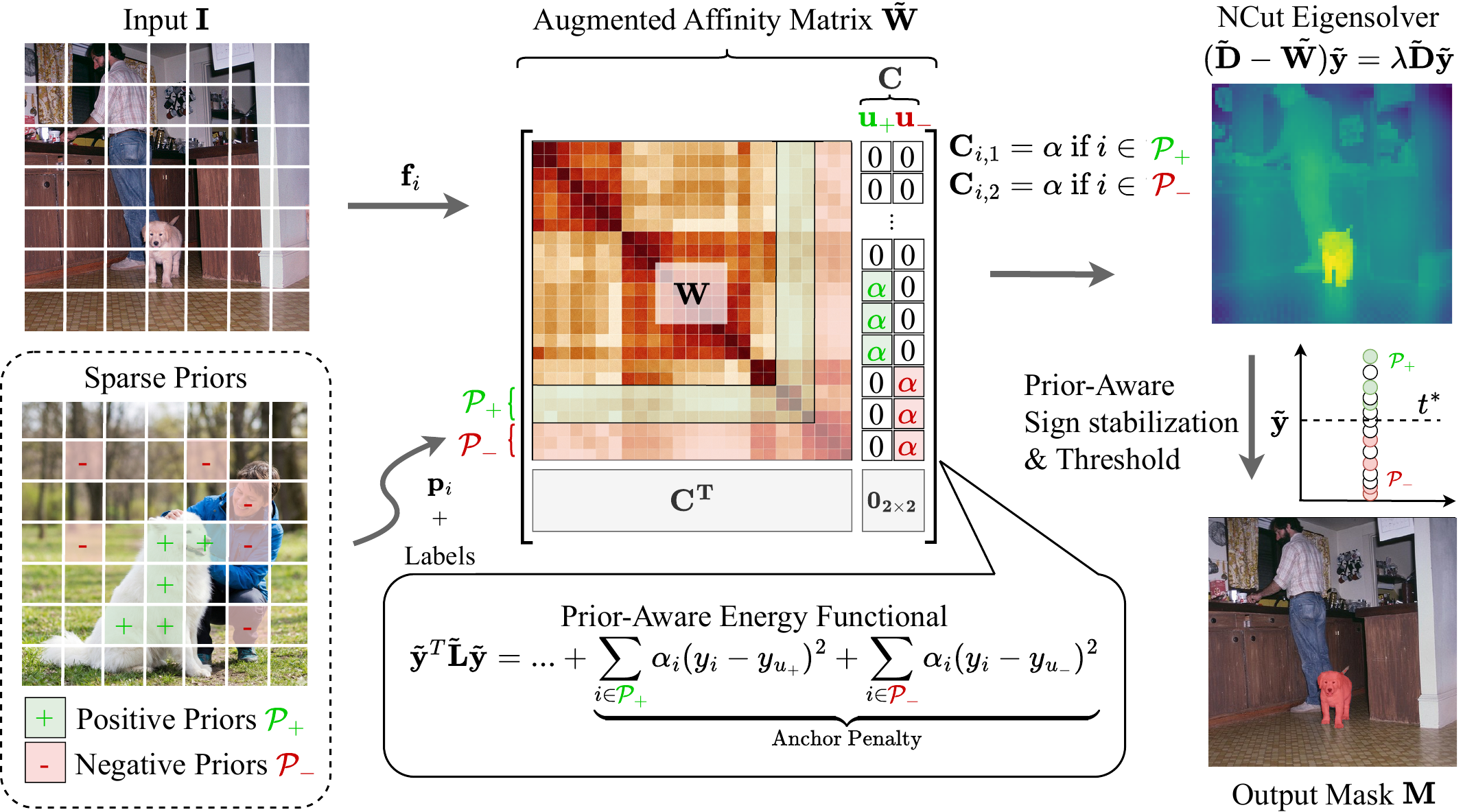}
    \caption{The input image is tokenized to extract dense features, which are concatenated with a sparse set of priors. Injected anchors bias the subsequent normalized cut toward a partition consistent with the annotations, yielding stable, controllable segmentation.}
    \label{fig:overview}
\end{figure*}

\vspace{-1em}
\paragraph{\textbf{Thresholding.}}
We binarize the continuous scores using a prior-driven threshold $t^\star$ (computed using the labeled prior nodes only). We consider four options:
\begin{itemize}[nosep,leftmargin=*]
    \item \textbf{Prior-ROC.} Choose $t^\star=\arg\max_{t\in[0,1]}(\mathrm{TPR}(t)-\mathrm{FPR}(t))$ on $\set{P}_{+}$ vs.\ $\set{P}_{-}$ (200-point grid).
    \item \textbf{Median midpoint.} $t^\star=\tfrac12\big(\mathrm{med}(\{s_i:i\in\set{P}_+\})+\mathrm{med}(\{s_i:i\in\set{P}_-\})\big)$.
    \item \textbf{GMM.} Fit a 2-component 1D GMM to $\{s_i\}_{i=1}^N$ (initialized from the prior medians) and take density intersection.
    \item \textbf{Platt scaling.} Fit logistic regression on priors and set $t^\star$ where the calibrated probability equals $0.5$.
\end{itemize}
The final segmentation mask is $M_i=\mathbf 1\{s_i>t^\star\}$, $i=1,\dots,n$. We output labels for image tokens only.  The complete sequence of these operations---from feature extraction and augmented affinity construction to biased spectral partitioning and deterministic mask generation---is visually summarized in Figure~\ref{fig:overview}.
\section{Experiments}
\label{sec:experiments}

We evaluate whether external priors improve class-agnostic spectral segmentation, where such guidance helps most, whether PANC provides gains beyond the priors, and whether context controls the concept. All masks are obtained without training, fine-tuning, or target-image clicks.

\subsection{Experimental Setup}
\label{sec:prior_bank}

\paragraph{\textbf{Implementation Details.}} We use frozen DINOv3 encoders \cite{siméoni2025dinov3} as default feature backbones for PANC: DINOv3-H for natural images, and a satellite-pretrained DINOv3-L for low-diversity, texture-dominated domains. To fully leverage the representational capacity of these models, initial features are computed at the maximum resolution supported by each backbone capability. All masks are rescaled and padded to a common comparison resolution of $1120 \times 1120$ pixels, fixed a priori and not selected on validation or test data. We set the affinity temperature to $\tau=0.7$. Experiments are run on a single NVIDIA A100 (40 GB) GPU. We use two dataset-level coupling regimes rather than per-image tuning: a saliency-selection regime for heterogeneous natural images and a structure-injection regime for homogeneous or low-semantic domains. Evaluation is carried out using per-image Intersection-over-Union (IoU) and aggregated as mean IoU (mIoU).

\paragraph{\textbf{Prior Retrieval and supervision protocol.}} PANC is agnostic to the origin of the external priors. For reproducible evaluation, we emulate external concept guidance with a fixed dataset-level protocol:
we cluster training-split CLS embeddings, select centroid representatives with annotations, and retrieve a sparse label-balanced set of foreground/background prior tokens using Maximum Marginal Relevance (MMR). TokenCut uses neither task-specific training nor external priors \cite{wang2023tokencutsegmentingobjectsimages},
whereas trained weakly supervised methods learn from dataset-level supervision or pseudo-labels \cite{lv2024weakly, xiang2024unified, li2024sg, gao2021ts}.
Context masks are converted to token labels by area-averaging each binary mask over the patch grid: a patch is a foreground prior if its foreground fraction is at least $0.6$ and a background prior if it is at most $0.05$; boundary patches in between are discarded (Supp.~B).

\vspace{-1em}
\paragraph{\textbf{Evaluation protocol.}}
Prior banks are built from annotated images disjoint from all evaluation queries (training splits where available), and evaluation queries are drawn from disjoint test sets. The annotation budget is 30 dense context masks / 1,500 retrieved prior tokens for the saliency-selection regime and 5 dense context masks / 2,500 retrieved prior tokens for the structure-injection regime. Backbone, prior count $m$, affinity temperature $\tau$, and coupling $\kappa$ are selected per regime on disjoint validation data and fixed before test evaluation; no test annotations are used for configuration selection. The COCO and controlled analyses (Secs.~\ref{sec:domain} and~\ref{sec:ablation}) give all compared methods the same DINOv3-S features at $448\times448$, 512 query-specific priors, and fixed query subsets (Supp.~B).

\subsection{External-prior guided saliency detection}
\label{sec:weak}

We evaluate the benefit of external prior guidance over class-agnostic token-graph saliency segmentation.

\vspace{-1em}
\paragraph{\textbf{Datasets.}}
We evaluate on three standard heterogeneous saliency benchmarks: ECSSD \cite{ecssd} with 1,000 complex scenes, the 5,019-image DUTS-TE test set \cite{duts}, and DUT-OMRON \cite{dut_omron} featuring 5,168 everyday scenes.

\vspace{-1em}
\paragraph{\textbf{Settings}.} These heterogeneous datasets exhibit high intra-class variance and diverse backgrounds, so retrieved priors may imperfectly represent the target. We therefore use the saliency-selection regime, where priors bias the cut among salient objects without over-constraining the graph. We fix a 30-image prior bank, 1,500 retrieved tokens, and conservative coupling ($\kappa=1.0$) across ECSSD, DUTS-TE, and DUT-OMRON. For fair comparison, we also re-evaluate the closest unsupervised token-graph baselines with the same modern DINOv3 backbone used by PANC.

\begin{table}[t]
\centering
\caption{
Comparison across heterogeneous and homogeneous segmentation datasets.
Results are mIoU (\%, higher is better); best in \textbf{bold}, second-best
\underline{underlined}. \emph{Train.}: task-specific model training.
\emph{Prior}: externally annotated context/support used at inference. For
PANC, priors are tokens from annotated context images, never from the target
image or its annotations. PANC entries report \emph{saliency-selection / structure-injection}:
first for heterogeneous datasets, second for homogeneous datasets.
$^\ast$ denotes satellite-pretrained DINOv3-L results for CFD and HAM10000. All backbones are frozen, without target-dataset fine-tuning.
}
\label{tab:comprehensive_compare}

\scriptsize
\setlength{\tabcolsep}{1.8pt}
\renewcommand{\arraystretch}{1.05}

\begin{adjustbox}{width=\columnwidth}
\begin{tabular}{lcclccc}
\toprule
\multicolumn{7}{c}{\textit{\textbf{Heterogeneous Domain}}} \\
\midrule
\textbf{Method} &
\textbf{Train.} &
\textbf{Prior} &
\textbf{Backbone} &
\textbf{ECSSD}~\cite{ecssd} &
\textbf{DUTS}~\cite{duts} &
\textbf{DUT-O}~\cite{dut_omron} \\
\midrule

\multicolumn{7}{l}{\textit{Unsupervised methods}} \\
BigBiGAN~\cite{voynov2021object}
    & \checkmark & $\times$ & BigGAN
    & 67.2 & 49.8 & 45.3 \\
E-BigBiGAN~\cite{voynov2021object}
    & \checkmark & $\times$ & BigGAN
    & 68.4 & 51.1 & 46.4 \\
FindGAN~\cite{melas2021finding}
    & \checkmark & $\times$ & GAN
    & 71.3 & 52.8 & 50.9 \\
LOST~\cite{siméoni2021localizingobjectsselfsupervisedtransformers}
    & $\times$ & $\times$ & DINOv1-S
    & 65.4 & 51.8 & 41.0 \\
DeepSpectral~\cite{melas2022deep}
    & $\times$ & $\times$ & DINOv1-S
    & 64.5 & 47.1 & 42.8 \\
TokenCut~\cite{wang2023tokencutsegmentingobjectsimages}
    & $\times$ & $\times$ & DINOv1-S
    & 71.2 & 57.6 & 53.3 \\
TokenCut~\cite{wang2023tokencutsegmentingobjectsimages}
    & $\times$ & $\times$ & DINOv3-H
    & 72.5 & 62.1 & 55.1 \\

\midrule
\multicolumn{7}{l}{\textit{Weakly supervised / external-prior guided methods}} \\
WSCUOD~\cite{lv2024weakly}
    & \checkmark & $\times$ & DINOv1-S
    & 72.7 & 59.9 & 53.6 \\
WSCUOD~\cite{lv2024weakly}
    & \checkmark & $\times$ & DINOv3-H
    & \underline{73.0}
    & \underline{64.2}
    & \underline{55.8} \\
\textbf{PANC (ours)}
    & $\times$ & \checkmark & DINOv1-S
    & $71.4/67.8$
    & $61.4/54.4$
    & $53.4/48.2$ \\
\textbf{PANC (ours)}
    & $\times$ & \checkmark & DINOv3-H
    & $\mathbf{73.3}/69.7$
    & $\mathbf{66.5}/59.5$
    & $\mathbf{58.6}/53.4$ \\

\midrule[0.8pt]
\multicolumn{7}{c}{
    \textit{\textbf{Homogeneous \& Challenging Domains}}
} \\
\midrule

\textbf{Method} &
\textbf{Train.} &
\textbf{Prior} &
\textbf{Backbone} &
\textbf{CUB}~\cite{wah2011caltech} &
\textbf{CFD}~\cite{shi2016automatic} &
\textbf{HAM}~\cite{tschandl2018ham10000} \\
\midrule

\multicolumn{7}{l}{\textit{Unsupervised methods}} \\
BigBiGAN~\cite{voynov2021object}
    & \checkmark & $\times$ & BigGAN
    & 68.3 & -- & -- \\
E-BigBiGAN~\cite{voynov2021object}
    & \checkmark & $\times$ & BigGAN
    & 71.0 & -- & -- \\
FindGAN~\cite{melas2021finding}
    & \checkmark & $\times$ & GAN
    & 66.4 & -- & -- \\
LOST~\cite{siméoni2021localizingobjectsselfsupervisedtransformers}
    & $\times$ & $\times$ & DINOv1-S
    & 68.8 & -- & -- \\
DeepSpectral~\cite{melas2022deep}
    & $\times$ & $\times$ & DINOv1-S
    & 66.7 & 82.3 & \underline{78.4} \\
UP-CrackNet~\cite{ma2024up}
    & \checkmark & $\times$ & U-Net
    & -- & 30.5 & -- \\
TokenCut~\cite{wang2023tokencutsegmentingobjectsimages}
    & $\times$ & $\times$ & DINOv1-S
    & 74.8 & 30.1 & 67.5 \\
TokenCut~\cite{wang2023tokencutsegmentingobjectsimages}
    & $\times$ & $\times$ & DINOv3-H/Sat$^\ast$
    & 75.5 & 46.2$^*$ & 55.4$^*$ \\

\midrule
\multicolumn{7}{l}{
    \textit{Weakly/external-prior guided methods}
} \\
PFENet~\cite{tian2020prior}
    & \checkmark & \checkmark & ResNet/VGG
    & 72.4 & -- & -- \\
WSCUOD~\cite{lv2024weakly}
    & \checkmark & $\times$ & DINOv1-S
    & \underline{77.8} & -- & -- \\
UWSCS~\cite{xiang2024unified}
    & \checkmark & $\times$ & ViT+ResNet
    & -- & 74.5 & -- \\
SG-MIAN~\cite{li2024sg}
    & \checkmark & $\times$ & MIAN
    & -- & -- & 74.3 \\
TS-CAM~\cite{gao2021ts}
    & \checkmark & $\times$ & DeiT
    & -- & -- & 67.5 \\
\textbf{PANC (ours)}
    & $\times$ & \checkmark & DINOv1-S
    & $73.5/75.7$
    & $77.4/84.1$
    & $72.0/76.2$ \\
\textbf{PANC (ours)}
    & $\times$ & \checkmark & DINOv3-H/Sat$^\ast$
    & $75.8/\mathbf{78.0}$
    & $\underline{84.3}/\mathbf{91.0}^{*}$
    & $74.6/\mathbf{78.8}^{*}$ \\

\bottomrule
\end{tabular}
\end{adjustbox}

\vspace{-0.4em}
\end{table}

\vspace{-1em}
\paragraph{\textbf{Results}.} Table~\ref{tab:comprehensive_compare} shows that PANC improves over architecture-matched unsupervised token-graph baselines across all three saliency datasets. Compared with TokenCut using the same DINOv3-H backbone, PANC gains +0.8\% on ECSSD, +4.4\% on DUTS, and +3.5\% on DUT-OMRON. %
These gains quantify the benefit of introducing
external annotated context relative to no-prior token-graph baselines. Supervision-matched comparisons isolating the contribution of PANC itself are reported in Sec.~\ref{sec:ablation}. Additional examples are provided in Supp.~C.

\subsection{Segmentation of homogeneous and challenging-domains}
\label{sec:domain}

Homogeneous, texture-dominated, and low-semantic images provide limited visual cues, yielding poorly differentiated self-supervised patch embeddings and weakly informative affinity graphs. We evaluate this regime on CUB-200-2011 \cite{wah2011caltech} (5,794 bird images; homogeneous baseline), HAM10000 \cite{tschandl2018ham10000} (10,015 weakly semantic dermoscopic images with low inter-class contrast), and CrackForest (CFD) \cite{shi2016automatic} (152 textureless images with thin cracks blending into the background). PANC uses external priors to amplify subtle differences and guide
the spectral partition.

\vspace{-1em}
\paragraph{\textbf{Settings.}}
For HAM10000 and CFD, we use a satellite-pretrained DINOv3-L as a generic texture-oriented frozen encoder, without training or fine-tuning on the target datasets. Because their unsupervised graphs are weakly separated, we
use the structure-injection regime with a 5-image prior bank, 2,500 retrieved tokens per test sample, and strong anchor coupling ($\kappa = 1000$). This regime-level configuration is fixed for the homogeneous-domain experiments and analyzed in the ablations.

\begin{figure}[!t]
\centering

\newlength{\rowlabw}
\newlength{\imgsep}
\newlength{\rowlabelsep}
\newlength{\imgw}

\setlength{\rowlabw}{8pt}
\setlength{\imgsep}{2pt}
\setlength{\rowlabelsep}{4pt}

\setlength{\imgw}{\dimexpr(\linewidth-\rowlabw-\rowlabelsep-5\imgsep)/6\relax}

\scriptsize
\setlength{\tabcolsep}{0pt}
\renewcommand{\arraystretch}{0}

\begin{tabular}{
@{}>{\centering\arraybackslash}m{\rowlabw}@{\hspace{\rowlabelsep}}
>{\centering\arraybackslash}m{\imgw}@{\hspace{\imgsep}}
>{\centering\arraybackslash}m{\imgw}@{\hspace{\imgsep}}
>{\centering\arraybackslash}m{\imgw}@{\hspace{\imgsep}}
>{\centering\arraybackslash}m{\imgw}@{\hspace{\imgsep}}
>{\centering\arraybackslash}m{\imgw}@{\hspace{\imgsep}}
>{\centering\arraybackslash}m{\imgw}@{}
}

\rotatebox{90}{\textit{HAM}} &
\includegraphics[width=\imgw]{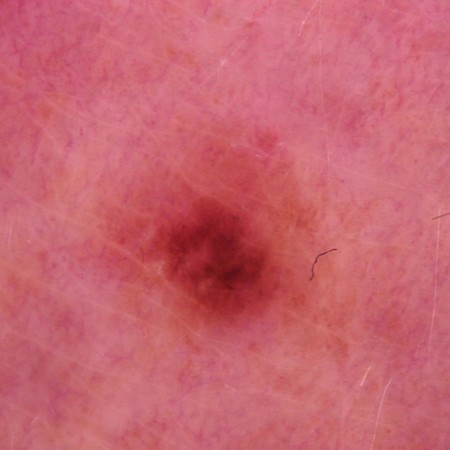} &
\includegraphics[width=\imgw]{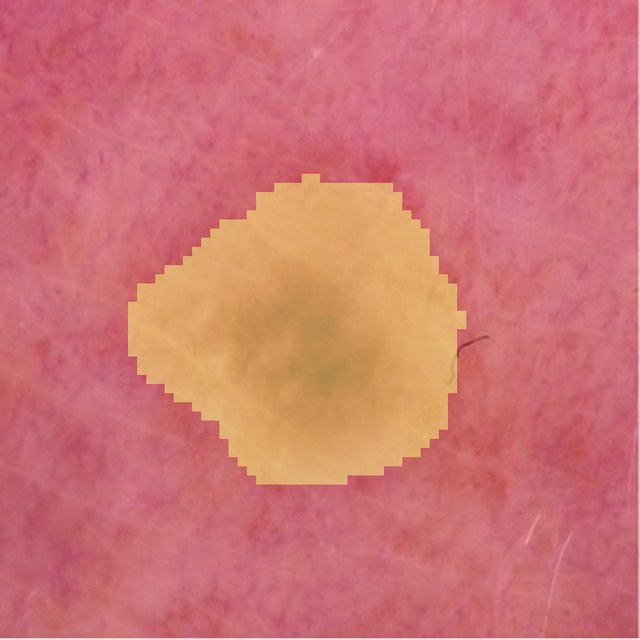} &
\includegraphics[width=\imgw]{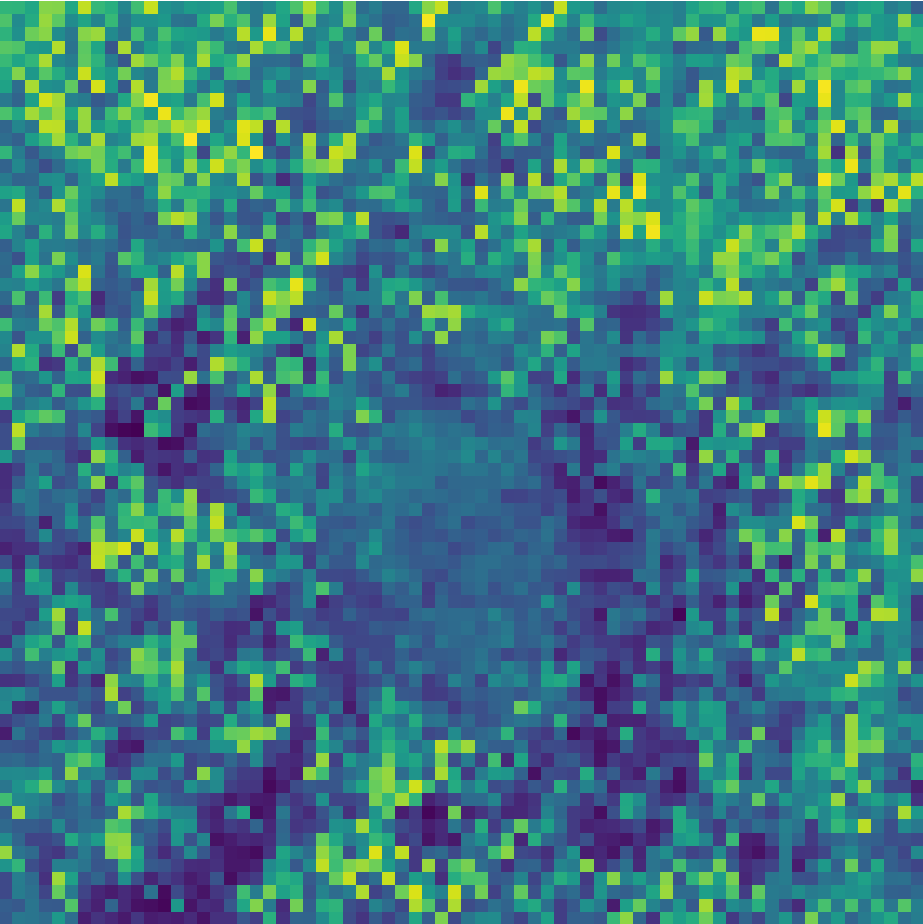} &
\includegraphics[width=\imgw]{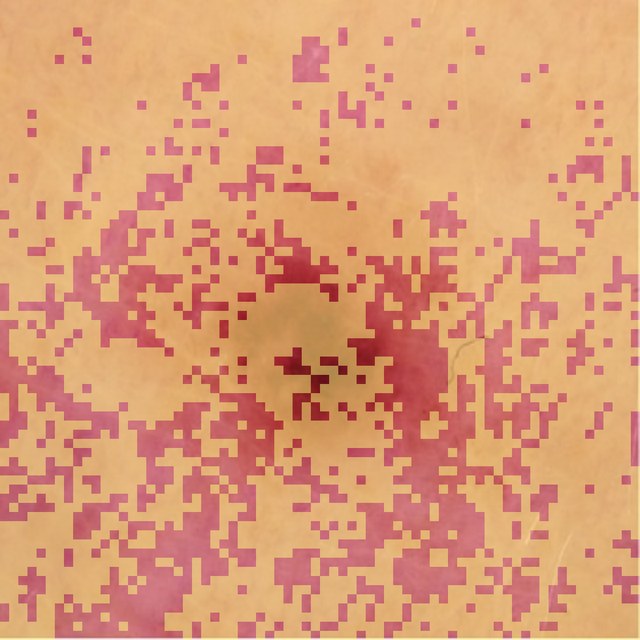} &
\includegraphics[width=\imgw]{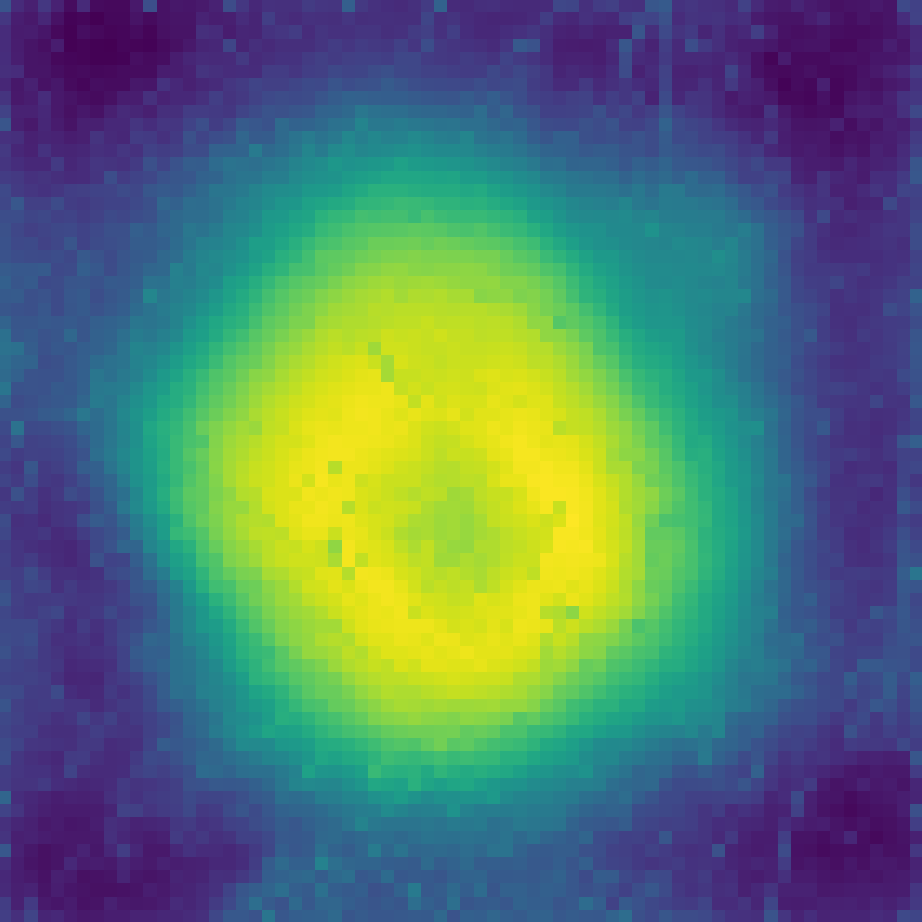} &
\includegraphics[width=\imgw]{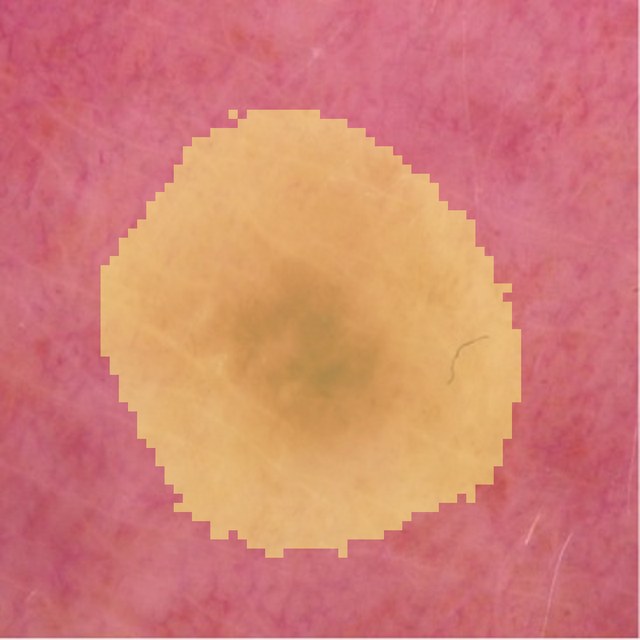} \\
\noalign{\vspace{2pt}}

\rotatebox{90}{\textit{CFD}} &
\includegraphics[width=\imgw]{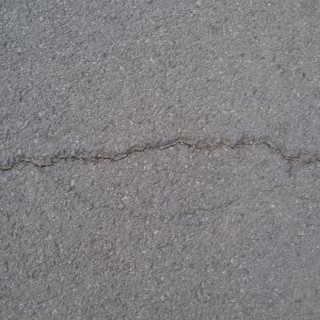} &
\includegraphics[width=\imgw]{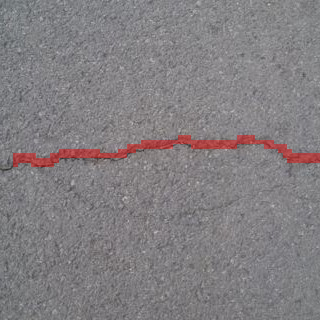} &
\includegraphics[width=\imgw]{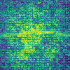} &
\includegraphics[width=\imgw]{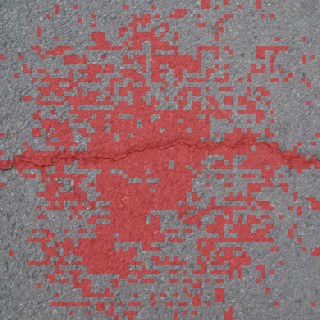} &
\includegraphics[width=\imgw]{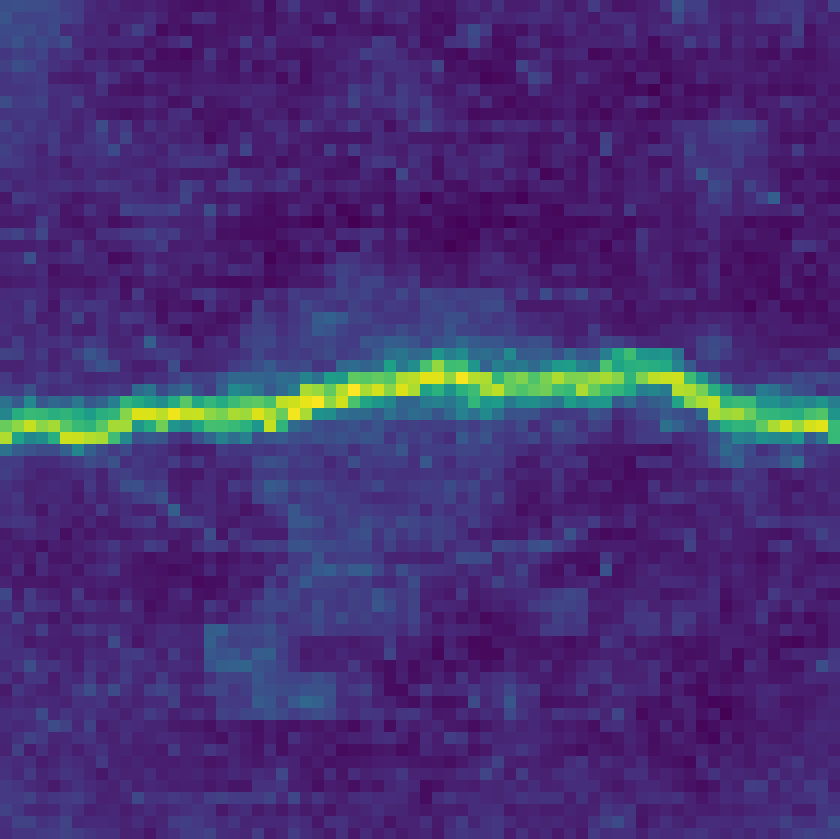} &
\includegraphics[width=\imgw]{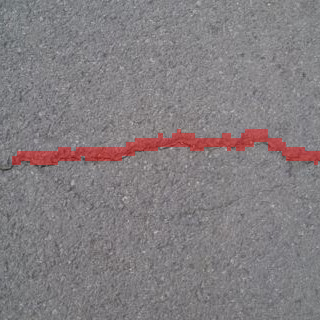} \\
\noalign{\vspace{2pt}}

\rotatebox{90}{\textit{CUB}} &
\includegraphics[width=\imgw]{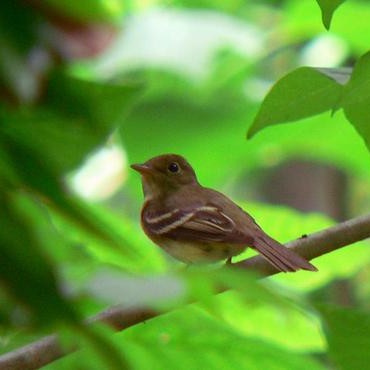} &
\includegraphics[width=\imgw]{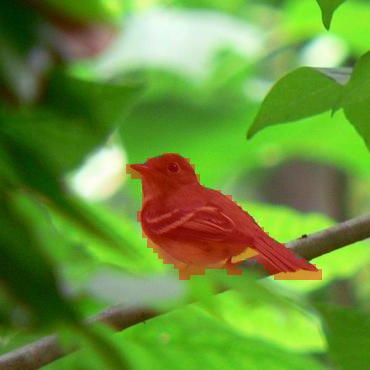} &
\includegraphics[width=\imgw]{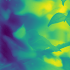} &
\includegraphics[width=\imgw]{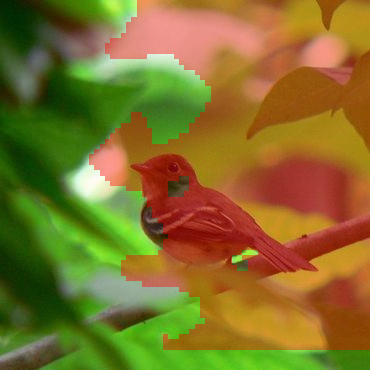} &
\includegraphics[width=\imgw]{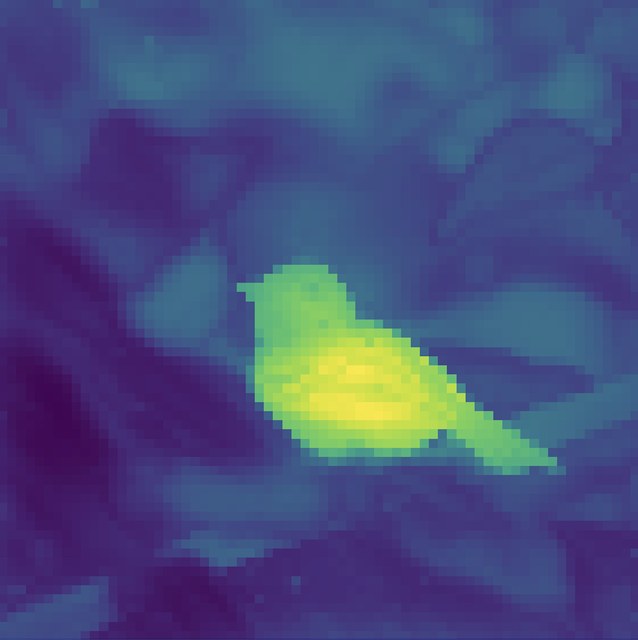} &
\includegraphics[width=\imgw]{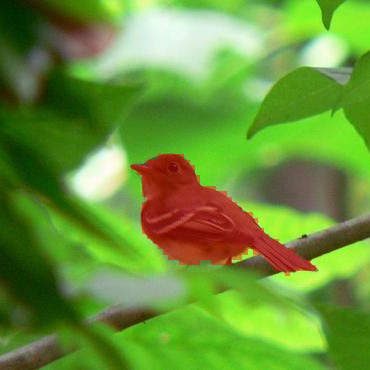} \\
\noalign{\vspace{3pt}}

&
\makebox[\imgw][c]{Input} &
\makebox[\imgw][c]{GT} &
\makebox[\imgw][c]{\shortstack{Unsup.\\Eigen Attn.}} &
\makebox[\imgw][c]{\shortstack{Unsup.\\Mask}} &
\makebox[\imgw][c]{\shortstack{PANC\\Eigen Attn.}} &
\makebox[\imgw][c]{\shortstack{PANC\\Mask}} \\

\end{tabular}

\caption{Qualitative comparison on challenging specialized datasets, HAM (light mask), CFD (red mask), CUB (red mask). PANC excels where unsupervised baselines fail by utilizing strong priors to resolve weak feature differentiation.}
\label{fig:homog_and_textureless}
\end{figure}

\vspace{-1em}
\paragraph{\textbf{Results.}}
Table~\ref{tab:comprehensive_compare} shows that PANC leads the reported baselines on all three datasets, reaching 78.0\% mIoU on CUB-200-2011, 78.8\% on HAM10000, and 91.0\% on CFD, with the largest improvement (+8.7\%) on CFD. Figure~\ref{fig:homog_and_textureless} illustrates how priors suppress surface artifacts in HAM10000, recover thin crack structures missed by the unsupervised baseline in CFD, and steer CUB partitions toward the intended object class.

\begin{figure}[t]
\centering

\newlength{\ctrlleftw}
\newlength{\ctrlrightw}
\newlength{\ctrloutersep}
\newlength{\ctrlinputw}
\newlength{\ctrlgridw}
\newlength{\ctrlinputgridsep}
\newlength{\ctrlrowlabw}
\newlength{\ctrlrowlabsep}
\newlength{\ctrlimgsep}
\newlength{\ctrlsegimgw}
\newlength{\ctrlpriorimgw}

\setlength{\ctrlleftw}{0.64\linewidth}
\setlength{\ctrlrightw}{0.30\linewidth}
\setlength{\ctrloutersep}{3pt}

\setlength{\ctrlinputw}{0.26\ctrlleftw}
\setlength{\ctrlgridw}{0.72\ctrlleftw}
\setlength{\ctrlinputgridsep}{\dimexpr\ctrlleftw-\ctrlinputw-\ctrlgridw\relax}

\setlength{\ctrlrowlabw}{7pt}
\setlength{\ctrlrowlabsep}{3pt}
\setlength{\ctrlimgsep}{2pt}

\setlength{\ctrlsegimgw}{\dimexpr(\ctrlgridw-\ctrlrowlabw-\ctrlrowlabsep-2\ctrlimgsep)/3\relax}

\setlength{\ctrlpriorimgw}{\dimexpr(\ctrlrightw-\ctrlimgsep)/2\relax}

\setlength{\tabcolsep}{0pt}
\renewcommand{\arraystretch}{0}

\raisebox{-12pt}{%
\begin{subfigure}[t]{\ctrlleftw}
\vspace{0pt}
    \centering

    \begin{minipage}[c]{\ctrlinputw}
        \centering
        \includegraphics[width=\linewidth]{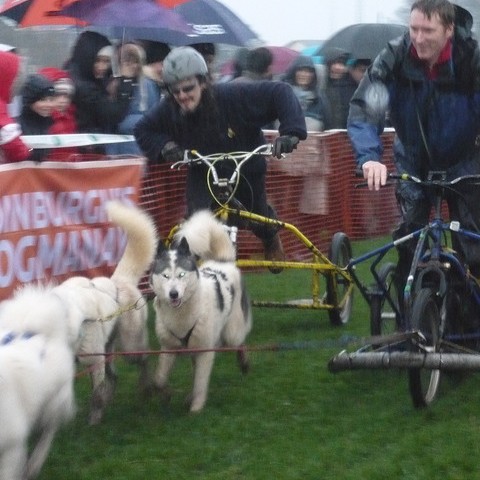}\\[-1pt]
        {\tiny Input}
    \end{minipage}%
    \hspace{\ctrlinputgridsep}%
    \begin{minipage}[c]{\ctrlgridw}
        \centering
        \begin{tabular}{
            @{}
            >{\centering\arraybackslash}m{\ctrlrowlabw}
            @{\hspace{\ctrlrowlabsep}}
            >{\centering\arraybackslash}m{\ctrlsegimgw}
            @{\hspace{\ctrlimgsep}}
            >{\centering\arraybackslash}m{\ctrlsegimgw}
            @{\hspace{\ctrlimgsep}}
            >{\centering\arraybackslash}m{\ctrlsegimgw}
            @{}
        }
            \rotatebox{90}{\tiny\textit{Person}} &
            \includegraphics[width=\ctrlsegimgw]{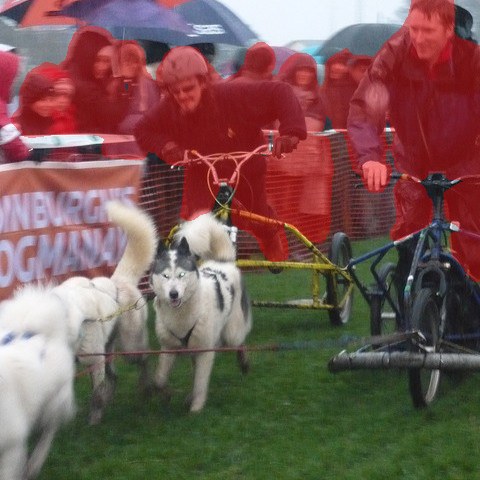} &
            \includegraphics[width=\ctrlsegimgw]{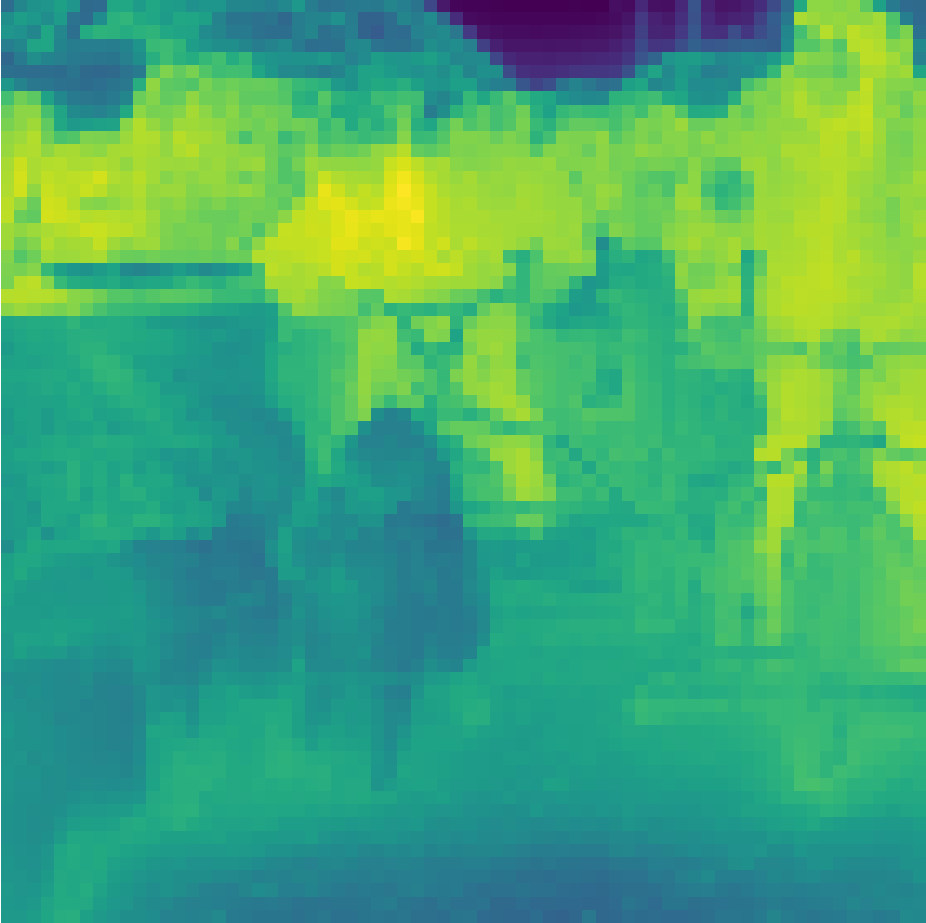} &
            \includegraphics[width=\ctrlsegimgw]{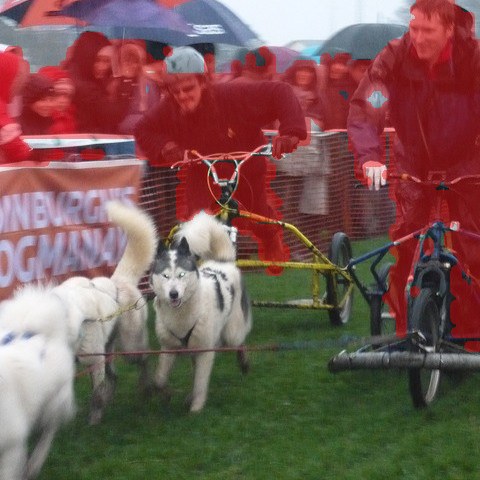} \\
            \noalign{\vspace{2pt}}

            \rotatebox{90}{\tiny\textit{Dog}} &
            \includegraphics[width=\ctrlsegimgw]{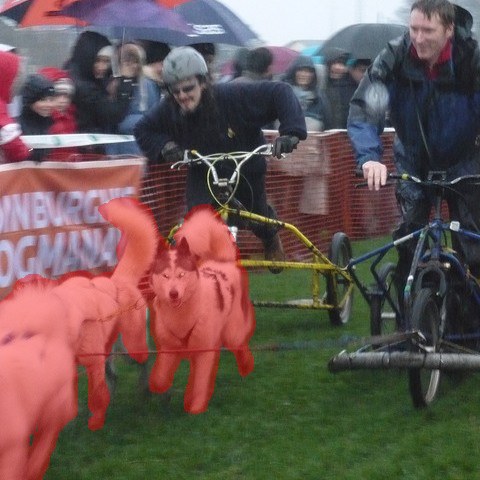} &
            \includegraphics[width=\ctrlsegimgw]{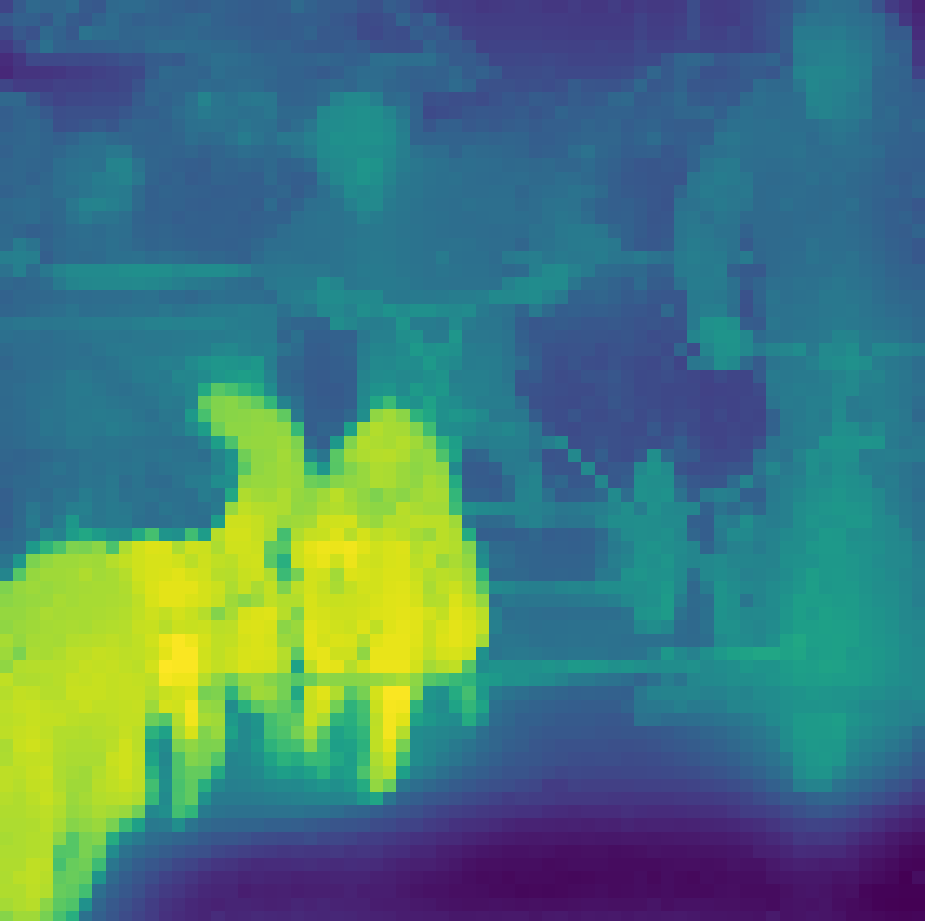} &
            \includegraphics[width=\ctrlsegimgw]{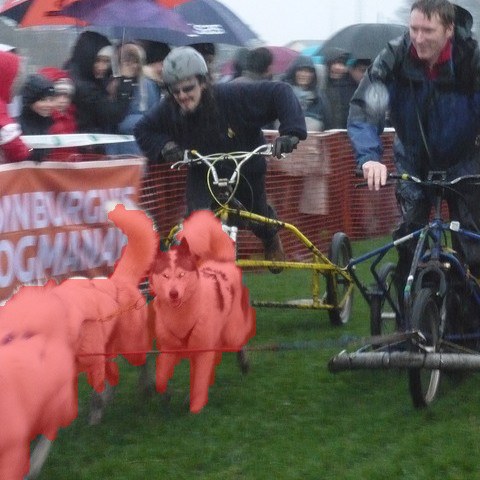} \\
        \end{tabular}
    \end{minipage}

    \par\nointerlineskip\vspace{2pt}\noindent

    \begin{minipage}[c]{\ctrlinputw}
        \centering
        \includegraphics[width=\linewidth]{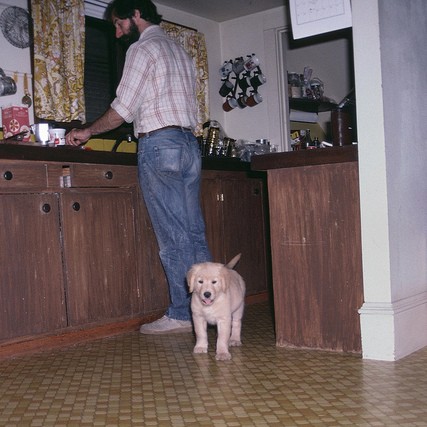}\\[-1pt]
        {\tiny Input}
    \end{minipage}%
    \hspace{\ctrlinputgridsep}%
    \begin{minipage}[c]{\ctrlgridw}
        \centering
        \begin{tabular}{
            @{}
            >{\centering\arraybackslash}m{\ctrlrowlabw}
            @{\hspace{\ctrlrowlabsep}}
            >{\centering\arraybackslash}m{\ctrlsegimgw}
            @{\hspace{\ctrlimgsep}}
            >{\centering\arraybackslash}m{\ctrlsegimgw}
            @{\hspace{\ctrlimgsep}}
            >{\centering\arraybackslash}m{\ctrlsegimgw}
            @{}
        }
            \rotatebox{90}{\tiny\textit{Person}} &
            \includegraphics[width=\ctrlsegimgw]{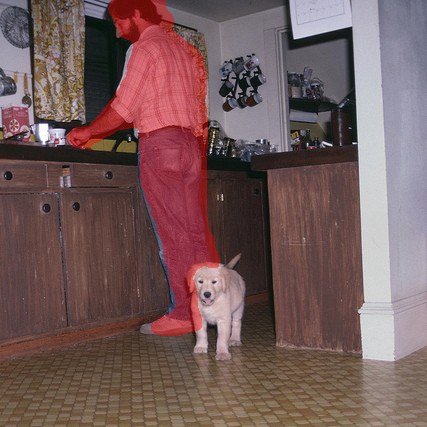} &
            \includegraphics[width=\ctrlsegimgw]{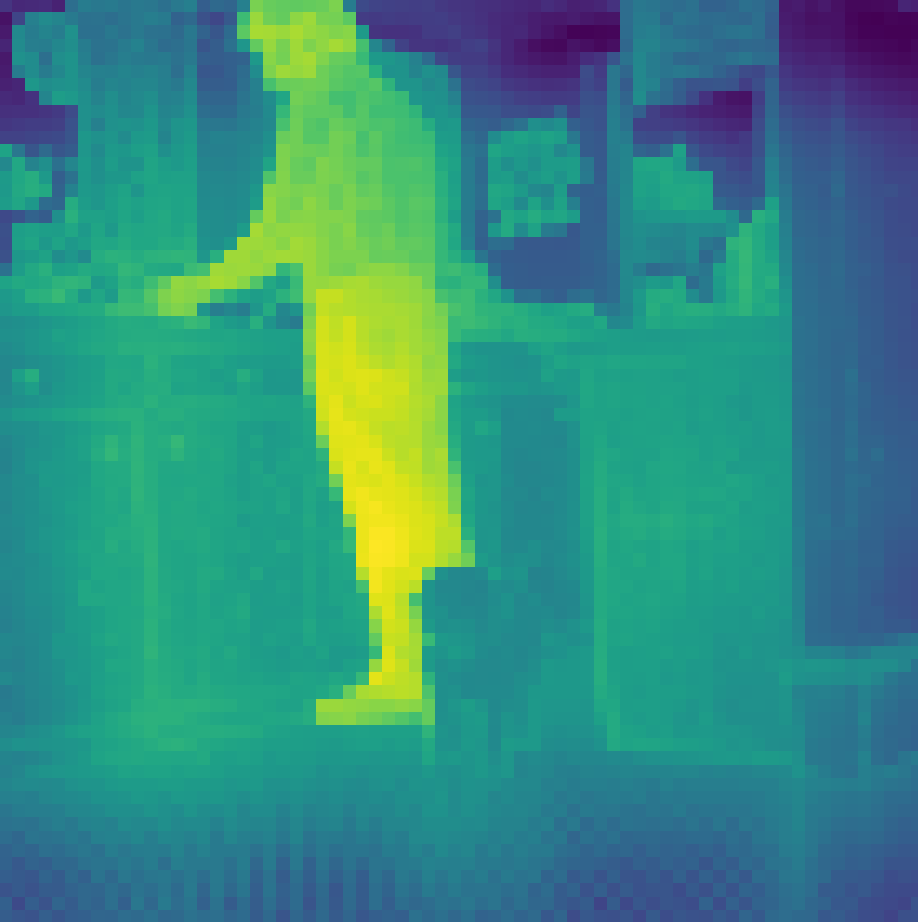} &
            \includegraphics[width=\ctrlsegimgw]{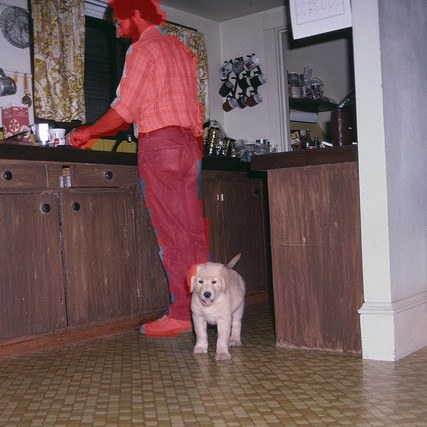} \\
            \noalign{\vspace{2pt}}

            \rotatebox{90}{\tiny\textit{Dog}} &
            \includegraphics[width=\ctrlsegimgw]{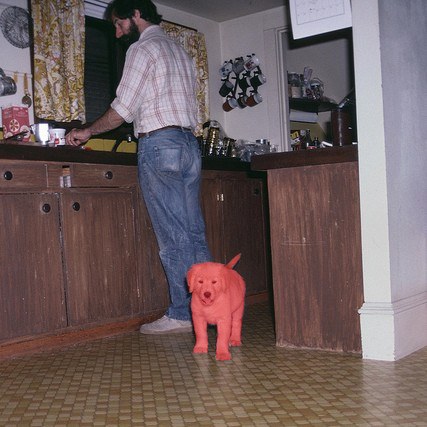} &
            \includegraphics[width=\ctrlsegimgw]{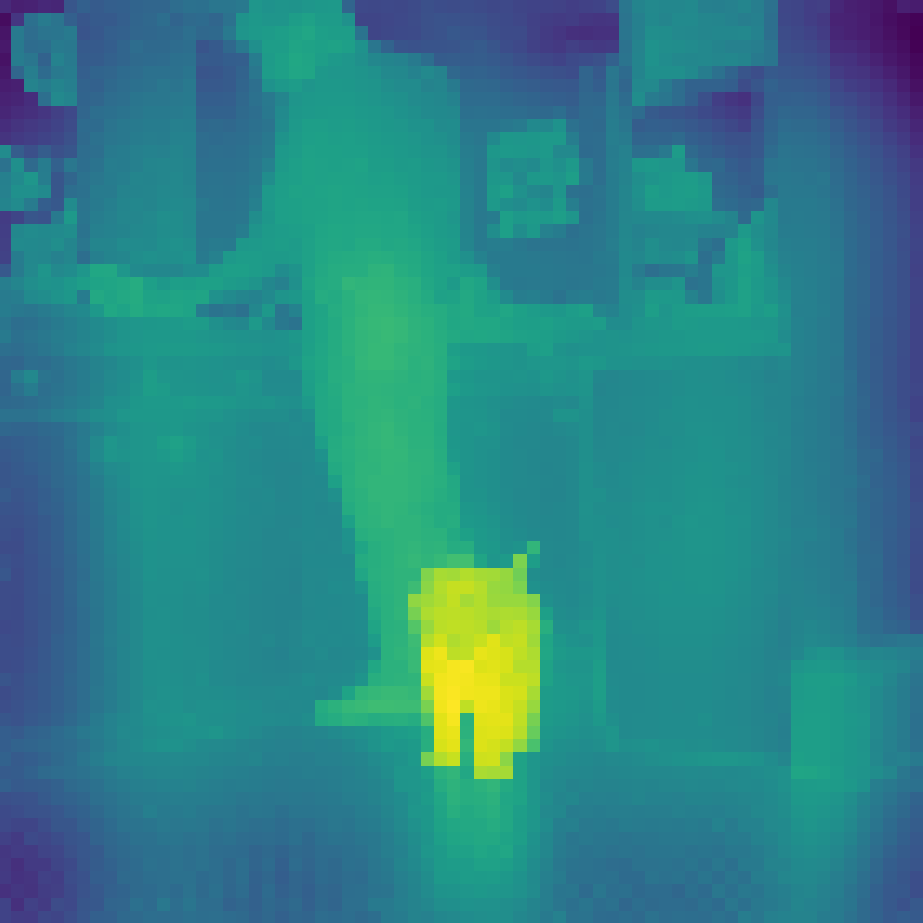} &
            \includegraphics[width=\ctrlsegimgw]{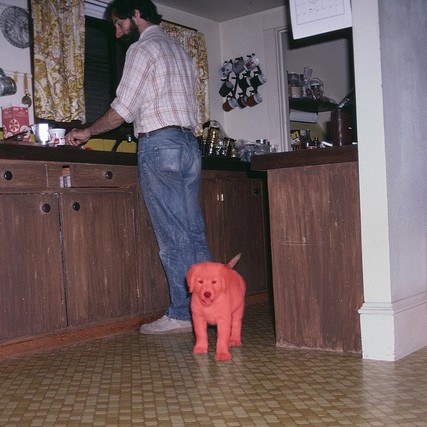} \\
            \noalign{\vspace{2pt}}

            &
            {\tiny GT} &
            {\tiny \shortstack{Eigen\\Attn.}} &
            {\tiny Output} \\
        \end{tabular}
    \end{minipage}

    \caption{Controlled Segmentation Results}
    \label{fig:results_a}
\end{subfigure}%
}%
\hspace{\ctrloutersep}%
\begin{minipage}[t]{\ctrlrightw}
\vspace{0pt}
    \centering

    \begin{subfigure}[t]{\linewidth}
        \centering
        \begin{tabular}{
            @{}
            c@{\hspace{\ctrlimgsep}}c
            @{}
        }
            \includegraphics[width=\ctrlpriorimgw]{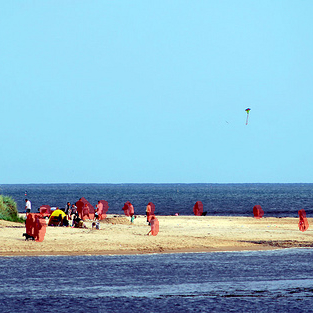} &
            \includegraphics[width=\ctrlpriorimgw]{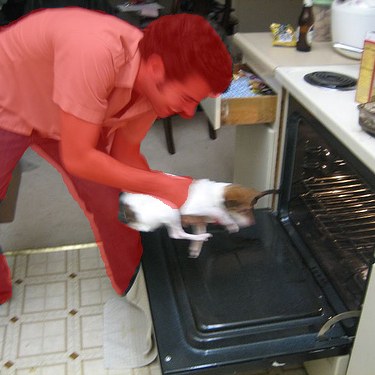} \\
            \noalign{\vspace{2pt}}
            \includegraphics[width=\ctrlpriorimgw]{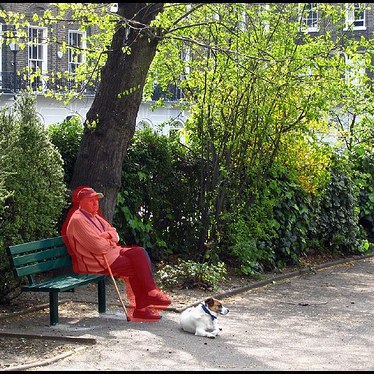} &
            \includegraphics[width=\ctrlpriorimgw]{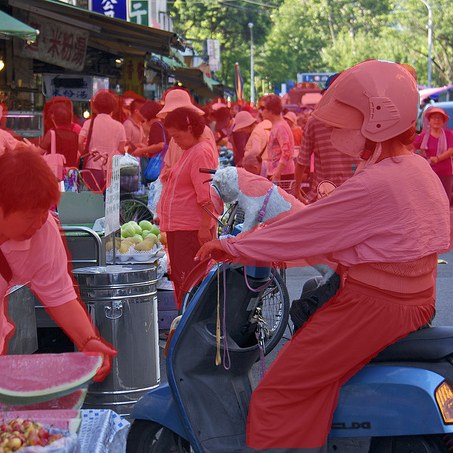}
        \end{tabular}
        \caption{Person Priors}
        \label{fig:priors_person}
    \end{subfigure}

    \vspace{3pt}

    \begin{subfigure}[t]{\linewidth}
        \centering
        \begin{tabular}{
            @{}
            c@{\hspace{\ctrlimgsep}}c
            @{}
        }
            \includegraphics[width=\ctrlpriorimgw]{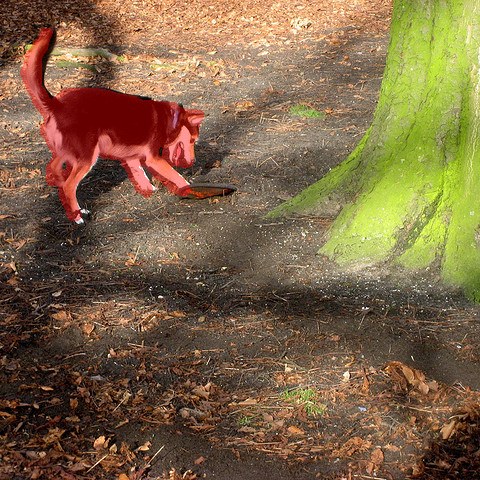} &
            \includegraphics[width=\ctrlpriorimgw]{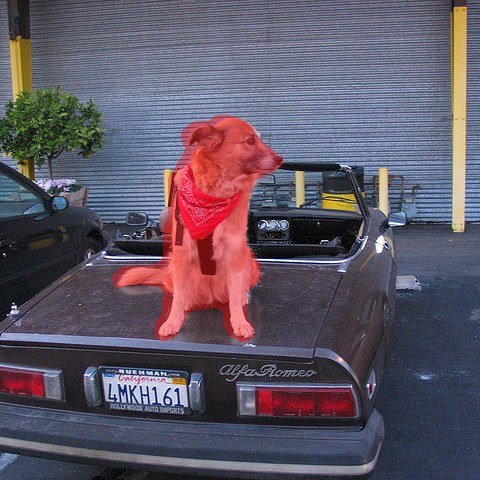} \\
            \noalign{\vspace{2pt}}
            \includegraphics[width=\ctrlpriorimgw]{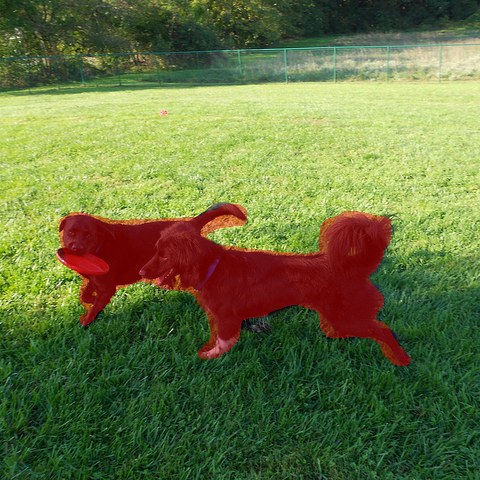} &
            \includegraphics[width=\ctrlpriorimgw]{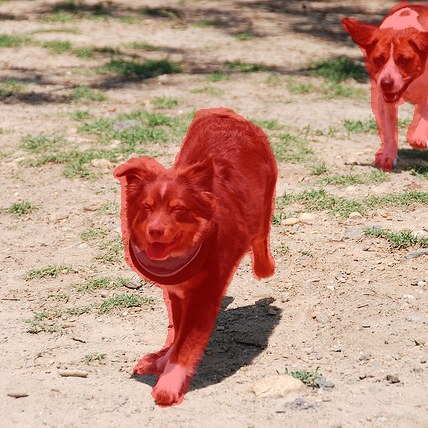}
        \end{tabular}
        \caption{Dog Priors}
        \label{fig:priors_dog}
    \end{subfigure}

    \vspace{6pt}
\end{minipage}

\vspace{2pt}
\caption{Explicit class controllability in multi-object scenes: \textbf{(a)} for a given input image, the semantic focus of the Fiedler vector (Eigen-Attn.) and the resulting output mask shift deterministically depending on the prior bank injected. \textbf{(b)} and \textbf{(c)} display the respective prior banks used to guide the target classes.}
\label{fig:controllability_aligned_grid}
\end{figure}

To quantify PANC's ability to resolve semantic ambiguity, we evaluate multi-object MS COCO val2017 scenes \cite{lin2014microsoft} containing both people and dogs, keeping the target image and backbone representation fixed while changing only the external context. With the protocol of Sec.~\ref{sec:prior_bank} and $\kappa=1.0$, we use 8 queries in which each class covers at least 1\% of the image; the target is the union of all instances of the class. Prior conditions are averaged over 3 independently sampled context banks of 5 annotated images per class, disjoint from the queries (Supp.~B). \emph{Unrelated priors} swap the banks (dog context for the person target and vice versa) with the same retrieval and token count, without matching their similarity to the query. Compared with no priors, matched priors improve target-class mIoU from 35.6\% to 60.6\% for Person and from 48.8\% to 56.8\% for Dog (see Table~\ref{tab:incontext}). Relative to unrelated priors, the gains are $+31.6$ and $+24.7$ points, respectively. Figure~\ref{fig:controllability_aligned_grid} shows the corresponding shift of the Fiedler vector and output mask toward the concept specified by the priors.

\begin{table}[t]
\centering
\caption{In-context segmentation on COCO Person \& Dog (target-class mIoU, \%; 8 queries, averaged over 3 context banks).}
\label{tab:incontext}
\footnotesize
\begin{tabular}{c c c}
\midrule
{\textit{Context}} & {\textit{Person}} & {\textit{Dog}}\\
\midrule
No priors       & 35.6 & 48.8 \\
Unrelated priors  & 29.0 & 32.1 \\
Matched priors     & 60.6 & 56.8 \\
\midrule
\end{tabular}
\end{table}

\subsection{Spectral Diagnostics}
\label{sec:diagnostics}
We use spectral quantities as diagnostics. For each image, we compute the two smallest non-trivial generalized eigenvalues of the NCut relaxation, the eigengap $\Delta=\lambda_3-\lambda_2$, and the quotient $\Lambda=\lambda_3/\lambda_2$, and analogously $\tilde\lambda_2,\tilde\lambda_3,\tilde\Delta,\tilde\Lambda$ with correct and imperfect priors. On DUTS and CFD (Fig.~\ref{fig:spectral_diag}), correct priors tend to increase the eigengap and improve mIoU, while imperfect priors weaken these trends and reduce segmentation quality.

\begin{figure}[t]
\centering
\begingroup

\captionsetup[subfigure]{
    font=scriptsize,
    labelfont=bf,
    skip=1pt,
    justification=centering,
    singlelinecheck=true
}

\def\legdot#1{%
  \tikz[baseline=-0.55ex]\node[circle,fill=#1,draw=#1,inner sep=0pt,minimum size=3.2pt] {};%
}
\def\legcircle{%
  \tikz[baseline=-0.55ex]\node[circle,draw=black,fill=none,inner sep=0pt,minimum size=4.0pt] {};%
}
\def\legsquare{%
  \tikz[baseline=-0.55ex]\node[rectangle,draw=black,fill=none,inner sep=0pt,minimum size=4.0pt] {};%
}

{\scriptsize
\begin{tabular}{@{}c@{\hspace{0.8em}}c@{\hspace{0.8em}}c@{}}
\legdot{black}\,Unsupervised &
\legdot{blue}\,Correct priors &
\legdot{red}\,Imperfect priors \\[-0.2ex]
\multicolumn{3}{c}{
\legcircle\,DUTS
\hspace{1.2em}
\legsquare\,CrackForest
}
\end{tabular}
}

\begin{subfigure}[t]{\linewidth}
\centering
\begin{tikzpicture}[trim axis left, trim axis right]
\begin{axis}[
    scale only axis,
    width=0.78\linewidth,
    height=0.34\linewidth,
    xlabel={$\lambda_2$},
    ylabel={$\Delta=\lambda_3-\lambda_2$},
    grid=both,
    ymin=-0.03, ymax=0.92,
    xmin=0.0, xmax=0.93,
    ticklabel style={font=\scriptsize},
    label style={font=\scriptsize},
    xlabel style={yshift=-0.1em},
    ylabel style={yshift=-0.2em},
    clip=true,
    clip marker paths=true,
]

\addplot[only marks, mark=*, mark size=1.4pt, color=black, opacity=0.18] coordinates {
(0.2420,0.1657)
(0.4735,0.0580)
(0.1291,0.3456)
(0.4616,0.1471)
(0.0774,0.3165)
(0.5016,0.0511)
(0.4242,0.0934)
(0.3389,0.1637)
(0.6157,0.0384)
(0.4242,0.0678)
};
\addplot[only marks, mark=*, mark size=2.8pt, color=black] coordinates {(0.3102,0.2503)};

\addplot[only marks, mark=*, mark size=1.4pt, color=blue, opacity=0.18] coordinates {
(0.3724,0.0444)
(0.1333,0.5080)
(0.2280,0.5376)
(0.0353,0.8453)
(0.5298,0.1960)
(0.5364,0.0246)
(0.2550,0.0240)
(0.0397,0.5802)
(0.0346,0.6200)
(0.8110,0.0200)
};
\addplot[only marks, mark=*, mark size=2.8pt, color=blue] coordinates {(0.2656,0.4019)};

\addplot[only marks, mark=*, mark size=1.4pt, color=red, opacity=0.18] coordinates {
(0.6276,0.1924)
(0.1367,0.5046)
(0.2966,0.4691)
(0.0380,0.8427)
(0.1513,0.3785)
(0.4636,0.4289)
(0.7450,0.1898)
(0.0471,0.5728)
(0.6546,0.0276)
(0.1890,0.5410)
};
\addplot[only marks, mark=*, mark size=2.8pt, color=red] coordinates {(0.3508,0.2126)};

\addplot[only marks, mark=square, mark size=1.4pt, color=black, opacity=0.18] coordinates {
(0.0808,0.3731)
(0.2851,0.1774)
(0.0372,0.7519)
(0.1999,0.3027)
(0.0541,0.6255)
(0.2796,0.1685)
(0.0318,0.4176)
(0.0603,0.7593)
(0.4812,0.0712)
(0.5418,0.0381)
};
\addplot[only marks, mark=square, mark size=2.8pt, color=black] coordinates {(0.2789,0.2033)};

\addplot[only marks, mark=square, mark size=1.4pt, color=blue, opacity=0.18] coordinates {
(0.3963,0.0365)
(0.1599,0.5786)
(0.0292,0.0930)
(0.0260,0.8882)
(0.2648,0.0755)
(0.1575,0.3742)
(0.0286,0.0720)
(0.1099,0.0291)
(0.2026,0.0407)
(0.0948,0.0014)
};
\addplot[only marks, mark=square, mark size=2.8pt, color=blue] coordinates {(0.1268,0.4037)};

\addplot[only marks, mark=square, mark size=1.4pt, color=red, opacity=0.18] coordinates {
(0.3538,0.0411)
(0.1778,0.0453)
(0.8778,0.0158)
(0.1142,0.8000)
(0.0295,0.7057)
(0.8425,0.0277)
(0.0842,0.0165)
(0.8969,0.0396)
(0.1029,0.0357)
(0.9052,0.0816)
};
\addplot[only marks, mark=square, mark size=2.8pt, color=red] coordinates {(0.5250,0.1241)};

\end{axis}
\end{tikzpicture}
\caption{Eigengap $\Delta=\lambda_3-\lambda_2$ as a function of the second non-trivial eigenvalue $\lambda_2$.}
\label{fig:spectral_diag_a}
\end{subfigure}

\begin{subfigure}[t]{\linewidth}
\centering
\begin{tikzpicture}[trim axis left, trim axis right]
\begin{axis}[
    scale only axis,
    width=0.78\linewidth,
    height=0.34\linewidth,
    xlabel={mIoU},
    ylabel={$\Lambda=\lambda_3/\lambda_2$},
    grid=both,
    ymin=0.0, ymax=25.5,
    xmin=0, xmax=100,
    ticklabel style={font=\scriptsize},
    label style={font=\scriptsize},
    xlabel style={yshift=-0.1em},
    ylabel style={yshift=-0.2em},
    clip=true,
    clip marker paths=true,
]

\addplot[only marks, mark=*, mark size=1.4pt, color=black, opacity=0.18] coordinates {
(41.9,1.68)
(45.2,1.12)
(51.1,3.68)
(56.3,1.32)
(46.6,5.09)
(28.2,1.10)
(55.3,1.22)
(62.8,1.48)
(35.5,1.06)
(61.1,1.16)
};
\addplot[only marks, mark=*, mark size=2.8pt, color=black] coordinates {(45.3,3.87)};

\addplot[only marks, mark=*, mark size=1.4pt, color=blue, opacity=0.18] coordinates {
(74.2,1.12)
(79.0,4.81)
(9.0,3.36)
(49.3,24.94)
(95.7,1.37)
(63.5,1.05)
(100.0,1.09)
(75.4,15.61)
(80.5,18.90)
(33.0,1.02)
};
\addplot[only marks, mark=*, mark size=2.8pt, color=blue] coordinates {(66.5,5.48)};

\addplot[only marks, mark=*, mark size=1.4pt, color=red, opacity=0.18] coordinates {
(25.8,1.31)
(14.8,4.69)
(12.0,2.58)
(73.3,23.19)
(19.6,3.50)
(31.2,1.93)
(0.2,1.25)
(21.6,13.17)
(0.2,1.04)
(64.3,3.86)
};
\addplot[only marks, mark=*, mark size=2.8pt, color=red] coordinates {(25.7,4.68)};

\addplot[only marks, mark=square, mark size=1.4pt, color=black, opacity=0.18] coordinates {
(39.5,5.62)
(38.5,1.62)
(49.6,21.23)
(66.9,2.51)
(45.6,12.57)
(67.6,1.60)
(60.0,14.12)
(59.0,13.59)
(51.8,1.15)
(44.9,1.07)
};
\addplot[only marks, mark=square, mark size=2.8pt, color=black] coordinates {(50.4,4.36)};

\addplot[only marks, mark=square, mark size=1.4pt, color=blue, opacity=0.18] coordinates {
(69.6,1.09)
(100.0,4.62)
(48.6,4.19)
(18.8,35.13)
(100.0,1.28)
(100.0,3.38)
(62.6,3.51)
(100.0,1.26)
(60.7,1.20)
(100.0,1.01)
};
\addplot[only marks, mark=square, mark size=2.8pt, color=blue] coordinates {(91.0,7.42)};

\addplot[only marks, mark=square, mark size=1.4pt, color=red, opacity=0.18] coordinates {
(0.1,1.12)
(0.1,1.25)
(1.1,1.02)
(29.0,8.00)
(42.2,24.92)
(1.4,1.03)
(91.3,1.20)
(0.4,1.04)
(0.1,1.35)
(1.0,1.09)
};
\addplot[only marks, mark=square, mark size=2.8pt, color=red] coordinates {(17.4,3.92)};

\end{axis}
\end{tikzpicture}
\caption{Spectral ratio $\Lambda=\lambda_3/\lambda_2$ as a function of segmentation quality.}
\label{fig:spectral_diag_b}
\end{subfigure}

\caption{Spectral diagnostics on DUTS and CFD. Colors show condition, shapes show dataset, larger markers show aggregates. Priors reshape the low-frequency spectrum and relate to mIoU.}
\label{fig:spectral_diag}

\endgroup
\end{figure}

These diagnostics support a two-regime interpretation. On heterogeneous data (\eg, DUTS), the token graph already has a meaningful low-frequency partition, and priors mainly disambiguate the target while leaving the global spectrum largely similar (saliency selection). On homogeneous or low-contrast data (\eg, CFD), the graph is weakly separated, and priors and anchors have a stronger effect on the low-frequency modes, stabilizing the partition (structure injection). The component ablations in Sec.~\ref{sec:ablation} and Supp.~C further support this behavior.

\subsection{Controlled Analysis \& Ablation studies}
\label{sec:ablation}

We ablate PANC on DUTS \cite{duts} (heterogeneous saliency) and CrackForest (CFD) \cite{shi2016automatic} (homogeneous segmentation), varying one parameter at a time from each dataset's default configuration (Table~\ref{tab:ablation}). These post-hoc sensitivity analyses on the test splits were not used to select the configurations in Table~\ref{tab:comprehensive_compare}. The same-prior, component, and prior-error blocks use the protocol of Sec.~\ref{sec:prior_bank} on 24 DUTS-TE and 20 CFD queries (the first 12 and 6 of them for components and prior error; variants in Supp.~C).

\vspace{-1em}
\paragraph{Controlled same-prior baselines.}
To isolate the contribution of the proposed spectral formulation from that of the annotated priors themselves, we compare PANC against prototype classification, $k$-NN, and label propagation that receive exactly the same features, retrieved priors, and resolution as PANC. The baselines use standard settings that are not tuned per dataset and may thus be suboptimal ($k=5$ for $k$-NN; a symmetric 30-NN graph with $\alpha=0.95$ and 40 iterations for label propagation; scores thresholded at 0.5; details in Supp.~B). On DUTS/CFD, PANC achieves 66.5/91.0\% mIoU, outperforming the strongest same-prior baseline by $+2.0$ on DUTS (Label Propagation, 64.5\%) and $+11.9$ on CFD ($k$-NN, 79.1\%). Removing the virtual anchors (with sign stabilization and prior-ROC thresholding unchanged) reduces performance to 58.0/72.5\% mIoU, showing that the gain is not explained by access to the annotated priors alone.

\vspace{-1em}
\paragraph{\textbf{Impact of injected prior tokens ($m$).}}
Homogeneous datasets (\eg, CFD) benefit from more exemplars that reinforce the target cluster, whereas heterogeneous ones (\eg, DUTS) degrade as less relevant priors introduce noise.

\vspace{-1em}
\paragraph{\textbf{Impact of anchor coupling ($\kappa$).}}
The coupling multiplier $\kappa$ is dataset-dependent: on weakly differentiated CFD, performance improves monotonically with $\kappa$, peaking at $\kappa=1000$ as stronger token--anchor links enforce the correct partition; on heterogeneous DUTS the optimum is moderate ($\kappa=1.0$), and rigid constraints override semantic variation.

\vspace{-1em}
\paragraph{\textbf{Impact of affinity temperature ($\tau$).}}
$\tau=0.70$ yields top results (66.5\% DUTS, 91.0\% CFD), while performance remains stable across moderate variations.

\vspace{-1em}
\paragraph{Resolutions.}
This block isolates the effect of input resolution. The $1120\times1120$ resolution of Table~\ref{tab:comprehensive_compare} was fixed in advance; since this sweep is computed on the test split, its best point ($480\times480$ on DUTS) is not used to update the main comparison. The DUTS optimum at $480 \times 480$ is likely because DINOv3's pretraining on 256- and 112-pixel crops favors intermediate scales over extreme upscaling; on CFD, accuracy grows with resolution as thin structures are better resolved, at a higher memory cost.

\vspace{-1em}
\paragraph{Prior error.}
We simulate annotation noise by randomly flipping a fraction of prior labels. Priors are retrieved once and kept fixed; for each rate $r$ we flip exactly $\mathrm{round}(r\,m)$ randomly chosen labels and report mean$\pm$std over five random draws with fixed seeds (6 queries per dataset). PANC remains stable under low noise, degrades as noise increases, and collapses when conflicting anchors override affinities.

\vspace{-1em}
\paragraph{\textbf{Thresholding strategies.}}
Prior-ROC thresholding is best on CFD (91.0\%) and ties for best on DUTS (66.5\%), so we adopt it as default; all four strategies stay within 0.9 points.

\begin{table}[t]
\centering
\caption{Ablations on DUTS and CFD (mIoU, \%; best in \textbf{bold}). Same-prior, component, and prior-error blocks use the DINOv3-S setting of Sec.~\ref{sec:prior_bank} (prior error: mean$\pm$std over 5 draws); other blocks vary one parameter of the Table~\ref{tab:comprehensive_compare} configuration.}
\label{tab:ablation}
\footnotesize
\setlength{\tabcolsep}{3pt}
\renewcommand{\arraystretch}{0.90}
\begin{adjustbox}{width=\columnwidth}
\begin{tabular}{lcc|lcc}
\toprule
\textbf{Configuration} & \textbf{DUTS}~\cite{duts} & \textbf{CFD}~\cite{shi2016automatic} &
\textbf{Configuration} & \textbf{DUTS}~\cite{duts} & \textbf{CFD}~\cite{shi2016automatic} \\
\midrule

\multicolumn{3}{l|}{\textit{Controlled same-prior}} &
\multicolumn{3}{l}{\textit{Components ablation}} \\
Prototype                & 50.1 & 72.3 &
No priors (NCut)         & 62.1 & 46.2 \\
$k$-NN                   & 61.9 & 79.1 &
Priors, no anchors  & 58.0 & 72.5 \\
Label Propagation        & 64.5 & 76.6 &
No sign stabilization & 53.2 & 77.7 \\
                         &      &      &
No thres. calibration & 62.5 & 87.0 \\
                         &      &      &
Fixed anchor coupling    & 66.4 & 90.9 \\
                         &      &      &
Full PANC                & \textbf{66.5} & \textbf{91.0} \\
\midrule

\multicolumn{3}{l|}{\textit{Injected prior tokens} ($m$)} &
\multicolumn{3}{l}{\textit{Anchor contribution / coupling} ($\kappa$)} \\
100              & 64.5 & 65.9 &
$\kappa=1$       & \textbf{66.5} & 84.3 \\
1500             & \textbf{66.5} & 83.4 &
$\kappa=10$      & 64.9 & 86.2 \\
2500             & 64.3 & \textbf{91.0} &
$\kappa=100$     & 63.2 & 90.2 \\
5000             & 62.1 & 85.6 &
$\kappa=1000$    & 59.5 & \textbf{91.0} \\
\midrule

\multicolumn{3}{l|}{\textit{Affinity temperature} ($\tau$)} &
\multicolumn{3}{l}{\textit{Resolution} ($H \times W$)} \\
$\tau=0.10$      & 64.1 & 89.3 &
$160\times160$   & 42.2 & 89.1 \\
$\tau=0.40$      & 66.3 & 90.7 &
$480\times480$   & \textbf{74.8} & 89.8 \\
$\tau=0.70$      & \textbf{66.5} & \textbf{91.0} &
$880\times880$   & 61.3 & 90.7 \\
$\tau=1.00$      & 66.4 & 90.0 &
$1120\times1120$ & 66.5 & \textbf{91.0} \\
\midrule

\multicolumn{3}{l|}{\textit{Prior error} (\%)} &
\multicolumn{3}{l}{\textit{Thresholding strategy}} \\
5\%              & \textbf{66.5$\pm$2.8} & \textbf{91.0$\pm$1.9} &
Median            & \textbf{66.5} & 90.6 \\
10\%              & 66.5$\pm$4.1 & 91.0$\pm$3.0 &
Prior-ROC         & \textbf{66.5} & \textbf{91.0} \\
20\%              & 63.8$\pm$7.2& 88.3$\pm$5.8 &
GMM               & 66.3 & 90.1 \\
50\%              & 29.2$\pm$13.5 & 53.7$\pm$14.2 &
Platt             & 66.2 & \textbf{91.0} \\
\bottomrule
\end{tabular}
\end{adjustbox}
\end{table}

\vspace{-1em}
\paragraph{Computational cost.}
At the default saliency-selection configuration, end-to-end online processing takes 0.521~s per query (1.9 queries/s on one A100): 388.7~ms for prior retrieval and 132.0~ms for core inference. Within core inference, feature extraction, graph construction, eigendecomposition, and post-processing take 41.0, 35.1, 55.1, and 0.84~ms on average, respectively. Since the dense affinity grows quadratically and the eigendecomposition cubically with the total token count $N=n+m$, cost grows sharply with resolution and prior count (Supp.~A).

\section{Conclusions}
\label{sec:conclusions}

We presented PANC, a training-free method that injects labeled context tokens into a frozen ViT token graph through foreground--background anchors, biasing a Normalized Cut partition toward a specified concept without training or target-image clicks. PANC improves over architecture-matched token-graph baselines on heterogeneous saliency benchmarks and is particularly effective on homogeneous or low-contrast domains. Priors act as a \emph{selection} mechanism on general-purpose datasets and as \emph{structure injection} on homogeneous ones, which is reflected in the coupling strength: CrackForest benefits from strong anchoring ($\kappa=1000$, 91.0\% mIoU), whereas DUTS peaks at $\kappa\approx 1$ and degrades when over-constrained.

\noindent\textbf{Limitations.} PANC requires externally annotated context, and its accuracy depends on the relevance and correctness of the retrieved priors (\cref{tab:ablation}). Same-prior and component analyses use query subsets of one dataset per regime, controllability covers two COCO classes, and we report only mIoU; broader evaluations, scalable prior search, and graph-scale accelerations are left for future work.

\clearpage
{
    \small
    \bibliographystyle{ieeenat_fullname}
    \bibliography{main}
}

\clearpage
\setcounter{section}{0}

\renewcommand{\thesection}{\Alph{section}}
\renewcommand{\theHsection}{\Alph{section}}

\section{GPU-Accelerated Spectral Partitioning}
\label[supp]{supp:a}

This section details the implementation and computational requirements of the PANC framework, focusing on a single-GPU design that mitigates the delays and computational overhead of standard CPU-based solvers.
By keeping all tensors, features, affinities, and eigenvectors in VRAM—and using vectorized matrix operations—our implementation avoids host--device transfers on the default path.
Here, we provide a detailed breakdown of floating-point operations (FLOPs) and memory footprint to demonstrate PANC's predictable memory and runtime trade-offs, alongside an evaluation of how backbone selection, prior strategies, and image resolution impact overall computational demand. The complete source code is available at \url{https://github.com/jgnav/PANC}.

\subsection{Algorithm Overview and Implementation}

Algorithm~\ref{alg:panc_pipeline} outlines the core logic of the PANC framework. The pipeline avoids explicit token loops, using batched tensor operations to maximize GPU occupancy. Furthermore, in our tests, all DINO features for query image patches, $f$, and priors, $p$, were also computed directly on the GPU to optimize efficiency.
\begin{algorithm}[ht]
\caption{PANC Pipeline}\label{alg:panc_pipeline}
\begin{algorithmic}[1]
\State \textbf{input:} $f \in \mathbb{R}^{n \times d}$, $p \in \mathbb{R}^{m \times d}$, $\mathcal{P}_+, \mathcal{P}_-$, $\tau, \kappa$
\State $F \leftarrow [f; p] \in \mathbb{R}^{N \times d}$
\State $F \leftarrow \text{L}_2\text{-normalize rows of } F$
\State $S \leftarrow F F^\top$ 
\State $W \leftarrow \exp(S / \tau)$
\State $\tilde{W} \leftarrow \text{augment } W \text{ with anchors using coupling } \kappa$
\State $\tilde{D} \leftarrow \operatorname{diag}(\tilde{W}\mathbf{1})$
\State $\tilde{L} \leftarrow \tilde{D} - \tilde{W}$
\State Solve $\tilde{L}\tilde{y} = \lambda\tilde{D}\tilde{y}$ for the 2nd smallest eigenpair $(\lambda_2, \tilde{y})$
\State \textbf{if} $\operatorname{median}(\tilde{y}_{\mathcal{P}_+}) < \operatorname{median}(\tilde{y}_{\mathcal{P}_-})$ \textbf{then} $\tilde{y} \leftarrow -\tilde{y}$
\State $s \leftarrow \text{min--max normalize } \tilde{y}$ over image and prior tokens (anchors excluded)
\State $t^* \leftarrow \operatorname{threshold}(s_{\mathcal{P}}, \mathcal{P}_+, \mathcal{P}_-)$
\State \textbf{output:} Mask $M = \mathbf{1}\{s_{1:n} > t^*\}$
\end{algorithmic}
\end{algorithm}

To implement this, we map all operations into batched PyTorch routines. The main steps are detailed in the following code segments.

\paragraph{\textbf{Affinity Construction.}} 
Constructing the base affinity matrix $W$ requires computing pairwise similarities between all tokens. By operating entirely in VRAM with optimized General Matrix Multiplication (GEMM) kernels (see \cref{lst:affinity}), we avoid the iterator overhead of CPU construction.

\begin{lstlisting}[language=Python, frame=single, basicstyle=\ttfamily\footnotesize, breaklines=true, caption={Dense base affinity matrix construction.}, label={lst:affinity}]
normed = F.normalize(features, p=2, dim=1)
sim = normed @ normed.T
aff = torch.exp(sim / tau)
aff.fill_diagonal_(0.0)
\end{lstlisting}

\paragraph{\textbf{Graph Augmentation.}}
To form the augmented graph $\tilde{\mathcal{G}}$, we introduce two virtual anchors (positive and negative) and link them exclusively to the labeled prior tokens. In \cref{lst:augment}, we scale the connection weights by a coupling factor $\kappa$ multiplied by the average local affinity. We then assemble the augmented block affinity matrix $\tilde{W}$.

\begin{lstlisting}[language=Python, frame=single, basicstyle=\ttfamily\footnotesize, breaklines=true, caption={Anchor augmentation and construction of block affinity matrix $\tilde{W}$.}, label={lst:augment}]
# Calculate adaptive coupling weights \alpha_i
local_mean = aff[prior_idx, :num_query].mean(dim=1).clamp_min(eps)
anchor_w = (kappa * local_mean).clamp(min=1e-4, max=1e3)

# Build prior-anchor connection matrix C (N x 2)
connection = torch.zeros(N, 2, device=aff.device)
connection[prior_idx[pos_mask], 0] = anchor_w[pos_mask]
connection[prior_idx[neg_mask], 1] = anchor_w[neg_mask]

# Assemble augmented block matrix
anchor_block = torch.diag(torch.tensor([eps, eps], device=aff.device))
top = torch.cat([aff, connection], dim=1)
bot = torch.cat([connection.T, anchor_block], dim=1)
aug_aff = torch.cat([top, bot], dim=0)

# Ensure strict symmetry
aug_aff = 0.5 * (aug_aff + aug_aff.T)
\end{lstlisting}

\paragraph{\textbf{Spectral Eigensolver.}} 
PANC solves the generalized eigenvalue problem for the augmented graph, $\tilde{L}\tilde{y} = \lambda\tilde{D}\tilde{y}$, the problem associated with the Normalized Cut relaxation. In \cref{lst:ncut}, we convert this problem %
into a standard eigenproblem on the symmetric normalized Laplacian. The anchor block holds $\epsilon=10^{-8}$ on its diagonal (\cref{lst:augment}) for numerical safety; this self-loop is negligible relative to the anchor degrees $\sum_i\alpha_i$, with $\alpha_i\geq10^{-4}$. Selecting the first eigenvalue larger than $\epsilon$ skips the trivial eigenvalue $\lambda_1=0$; for a connected graph this returns the Fiedler vector of $\tilde\lambda_2$. 

\begin{lstlisting}[language=Python, frame=single, basicstyle=\ttfamily\footnotesize, breaklines=true, caption={Solving the NCut generalized eigenproblem.}, label={lst:ncut}]
deg = aug_aff.sum(dim=1).clamp_min(eps)
inv_sqrt_d = torch.rsqrt(deg)
norm_aff = inv_sqrt_d[:, None] * aug_aff * inv_sqrt_d[None, :]
L = torch.eye(aug_aff.size(0), device=aug_aff.device) - norm_aff

evals, evecs = torch.linalg.eigh(L)
nz = torch.nonzero(evals > eps, as_tuple=False).view(-1)
idx2 = int(nz[0].item()) if nz.numel() > 0 else min(1, evals.numel() - 1)

fiedler = inv_sqrt_d * evecs[:, idx2]
\end{lstlisting}

\paragraph{\textbf{Deterministic Orientation.}} 
Because the continuous Fiedler vector is sign-ambiguous, we utilize the sparse priors to deterministically orient it. In \cref{lst:sign}, we compare the median scores of the positive and negative subsets to correct the sign if necessary.

\begin{lstlisting}[language=Python, frame=single, basicstyle=\ttfamily\footnotesize, breaklines=true, caption={Prior-aware sign stabilization.}, label={lst:sign}]
fg_med = fiedler_vec[num_query : num_query + num_prior][pos_mask].median()
bg_med = fiedler_vec[num_query : num_query + num_prior][neg_mask].median()
if fg_med < bg_med:
    fiedler_vec = -fiedler_vec
\end{lstlisting}

\paragraph{\textbf{Thresholding Strategies.}} 
To binarize the continuous eigenvector scores into discrete masks, PANC supports several data-driven thresholding strategies. While the default is an optimized ROC analysis mapping to Youden's J-statistic, the framework also implements Median Midpoint, 1D Gaussian Mixture Model (GMM) intersection, and Platt Scaling (logistic regression). \cref{lst:thresh} outlines these implementations. The default prior-ROC path runs entirely on the GPU; the optional GMM path transfers the scores to the CPU and uses scikit-learn, and the Platt-scaling parameters are created on the same device as the scores.

\begin{lstlisting}[language=Python, frame=single, basicstyle=\ttfamily\footnotesize, breaklines=true, caption={Vectorized thresholding strategies.}, label={lst:thresh}]
# 1. ROC (Default): Maximize TPR - FPR
cands = torch.linspace(0, 1, 200, device=device)
tpr = (prior_scores[pos_mask, None] > cands[None, :]).float().mean(0)
fpr = (prior_scores[neg_mask, None] > cands[None, :]).float().mean(0)
t_roc = cands[torch.argmax(tpr - fpr)]

# 2. Median Midpoint
t_med = 0.5 * (prior_scores[pos_mask].median() + prior_scores[neg_mask].median())

# 3. Platt Scaling (Logistic Regression)
x = prior_scores.view(-1, 1)
y = torch.zeros_like(prior_scores)
y[pos_mask] = 1.0

w = torch.zeros(1, device=device, requires_grad=True)
b = torch.zeros(1, device=device, requires_grad=True)
opt = torch.optim.Adam([w, b], lr=1e-2)
for _ in range(max_iter):
    opt.zero_grad()
    torch.nn.BCEWithLogitsLoss()(x * w + b, y).backward()
    opt.step()
t_platt = (-b / (w + eps)).clamp(0.0, 1.0) # Boundary where probability = 0.5

# 4. GMM (using sklearn backend for 1D density intersection)
s = all_scores.cpu().numpy().reshape(-1, 1)
gmm = GaussianMixture(n_components=2, covariance_type="full")
gmm.means_init = np.array([[prior_scores[neg_mask].median().item()], 
                           [prior_scores[pos_mask].median().item()]])
gmm.fit(s)
grid = np.linspace(0, 1, 1000).reshape(-1, 1)
post = gmm.predict_proba(grid)[:, np.argmax(gmm.means_.ravel())]
t_gmm = torch.tensor(grid[np.argmin(np.abs(post - 0.5))])
\end{lstlisting}

\subsection{Performance Assessment}

\paragraph{\textbf{Hardware and Profiling.}}
All computational benchmarks ran on a single NVIDIA A100 GPU with 40GB of VRAM. Profiling focused on the inference stage of the pipeline, encompassing feature extraction, affinity graph construction, eigensolving, and mask binarization.

\paragraph{\textbf{Evaluation Metrics.}}
To assess efficiency, we utilize the following metrics:
\begin{itemize}
    \item FLOPs (Floating Point Operations): Measures the total computational cost of the inference pass. This is dominated by the dense matrix multiplication for affinity construction and the eigendecomposition.
    \item Peak Memory (MB): The maximum VRAM allocated during the forward pass. The peak matches the storage of the $N \times N$ augmented affinity matrix, scaling quadratically with the number of tokens.
\end{itemize}

\paragraph{\textbf{Component Cost \& Scalability Analysis.}}
Table~\ref{tab:component_analysis} presents a comprehensive ablation study quantifying the computational overhead of the PANC framework. We analyze three critical scaling dimensions:
\begin{enumerate}
    \item Number of Injected Priors ($m$): We evaluate the impact of augmenting the graph with an increasing number of annotated vertices, scaling from $m=0$ (unsupervised baseline ) up to $m=5,000$. For typical usage (e.g., $m \le 1,500$), the overhead is well-managed. However, injecting a massive number of priors ($M=5,000$) drastically expands the graph connectivity, resulting in a severe spike in GFLOPs. 
    \item Resolution Scaling: We evaluate input resolutions scaling from $224 \times 224$ up to $1344 \times 1344$.
    \item Backbone Efficiency: We also benchmark the DINOv3 family against the DINOv2-L standard. Our results indicate that DINOv3-L matches the computational footprint of legacy DINOv2-L (306 GFLOPs) while offering improvements in cross-resolution stability and geometric consistency. 
\end{enumerate}

Prior-bank construction is performed once offline and requires 3.180~s for the 30-image saliency-selection bank. Query-specific prior retrieval, including relevance scoring, prefiltering, and MMR selection, requires $388.7 \pm 4.8$~ms per query and is reported separately from the core PANC inference time.

Core inference requires 132.0~ms per query on average: $41.0 \pm 2.1$~ms for query feature extraction, $35.1 \pm 0.7$~ms for graph construction, $55.1 \pm 6.2$~ms for eigendecomposition, and $0.84 \pm 0.16$~ms for sign stabilization and thresholding. The PANC segmentation stage after feature extraction requires $91.0 \pm 6.4$~ms.

Eigendecomposition accounts for 41.8\% of the reported core inference time and 60.6\% of the PANC segmentation stage after feature extraction. This distinction is relevant because query-specific prior retrieval is external to the spectral partitioning procedure and is profiled separately.

\begin{table*}[ht]
\centering
\caption{
Computational scaling of PANC with prior count, resolution, and backbone. We report GFLOPs, peak GPU memory (MB), and estimated core inference time (s). Runtime estimates are calibrated on a single NVIDIA A100 40GB GPU. Eig. denotes the estimated fraction of core inference time spent in eigendecomposition. The resolution and backbone blocks use $m=1{,}000$ to isolate scaling trends; they are estimates, not measurements of the main configurations ($m=1{,}500$ or $2{,}500$ at $1120\times1120$).
}
\label{tab:component_analysis}

\begin{tabular}{l l c r r c c c r}
\toprule
\textbf{Method} &
\textbf{Backbone} &
\textbf{Resolution} &
\textbf{Tokens ($N$)} &
\textbf{Priors ($m$)} &
\textbf{GFLOPs} &
\textbf{Mem} &
\textbf{Core Time} &
\textbf{Eig. (\%)} \\
\midrule

\multicolumn{9}{c}{\textit{Injected Prior Tokens} ($m$)} \\
\midrule
TokenCut & DINOv3-H & $480 \times 480$ & 1,156 & 0
& 567 & 6,705 & 0.056 & 11.1 \\
PANC & DINOv3-H & $480 \times 480$ & 1,156 & 10
& 567 & 6,705 & 0.056 & 11.3 \\
PANC & DINOv3-H & $480 \times 480$ & 1,156 & 100
& 572 & 6,705 & 0.059 & 13.4 \\
PANC & DINOv3-H & $480 \times 480$ & 1,156 & 1,000
& 1,086 & 6,705 & 0.110 & 36.3 \\
PANC & DINOv3-H & $480 \times 480$ & 1,156 & 5,000
& 12,916 & 16,218 & 1.204 & 77.3 \\

\midrule
\multicolumn{9}{c}{\textit{Resolution ($H \times W$)}} \\
\midrule
PANC & DINOv3-H & $224 \times 224$ & 256 & 1,000
& 677 & 6,217 & 0.027 & 29.4 \\
PANC & DINOv3-H & $480 \times 480$ & 1,156 & 1,000
& 1,085 & 6,705 & 0.110 & 36.3 \\
PANC & DINOv3-H & $896 \times 896$ & 4,096 & 1,000
& 2,522 & 8,823 & 0.831 & 63.6 \\
PANC & DINOv3-H & $1120 \times 1120$ & 6,400 & 1,000
& 3,674 & 10,493 & 2.175 & 74.3 \\
PANC & DINOv3-H & $1344 \times 1344$ & 9,216 & 1,000
& 5,103 & 12,534 & 5.213 & 81.6 \\

\midrule
\multicolumn{9}{c}{\textit{Backbone}} \\
\midrule
PANC & DINOv2-L & $480 \times 480$ & 1,156 & 1,000
& 306 & 1,849 & 0.085 & 46.8 \\
PANC & DINOv3-S & $480 \times 480$ & 1,156 & 1,000
& 122 & 603 & 0.058 & 68.9 \\
PANC & DINOv3-B & $480 \times 480$ & 1,156 & 1,000
& 223 & 948 & 0.074 & 54.1 \\
PANC & DINOv3-L & $480 \times 480$ & 1,156 & 1,000
& 306 & 1,849 & 0.085 & 46.8 \\

\bottomrule
\end{tabular}
\end{table*}

In Table~\ref{tab:component_analysis}, the injection of priors introduces manageable computational overhead compared to the purely unsupervised baseline when using optimal configuration thresholds. The transition from an intermediate $480 \times 480$ resolution to the $1120 \times 1120$ comparison resolution results in a substantial increase in GFLOPs. While memory growth is sub-quadratic due to fixed backbone overheads, computational demand scales sharply with token count.

In summary, high-resolution processing increases memory demand, yet it remains completely feasible on modern dense hardware architectures, avoiding the immediate need for graph-scale accelerations such as sparsification.

\section{Prior Retrieval and Evaluation Protocol}
\label[supp]{supp:b}

This section details the prior retrieval protocol used to systematically generate the sparse supervision signals required by the PANC framework. To emulate selective user intent without coupling PANC to dense retrieval or large-scale indexing, we automatically select a compact and diverse set of representative tokens from a dataset to form the injected prior bank.

\subsection{Representative Image Selection}

We construct a compact prior bank from a small set of representative images. We choose these by clustering image-level descriptors produced by the frozen, self-supervised Vision Transformer. Each training image $I$ is encoded to obtain the $\ell_2$-normalized CLS token $c(I) \in \mathbb{R}^{d}$, which serves as a deterministic global semantic embedding. 

Let $C \in \mathbb{R}^{N_{train} \times d}$ stack all CLS embeddings from the training set. On the $K_{\text{clusters}}$-means we obtain a set of centroids $\{\mu_k\}_{k=1}^{K_{\text{clusters}}}$. For each cluster $k$, the most representative image is defined as the one whose embedding is closest to the cluster centroid in the Euclidean space:
$$I_k^\star = \arg\min_{I \in \mathcal{S}_k} \| c(I) - \mu_k \|_2$$
where $\mathcal{S}_k$ is the set of images assigned to cluster $k$. This step yields $K_{\text{clusters}}$ exemplar images that cover the dominant appearance modes of the dataset with minimal redundancy. From each selected representative image $I_k^\star$, we extract its dense token grid, denoted as the set of candidate prior embeddings $\{p_i\}_{i=1}^{n_k} \in \mathbb{R}^d$. Let the total pool of candidate prior tokens across all classes be denoted as $\mathcal{T}_B$.

\subsection{ Diversity-Aware Token Selection}

At inference time, for a given query image tokenized into features $\mathcal{Q} = \{f_1, \dots, f_n\}$ where $f_i \in \mathbb{R}^d$, we must retrieve a sparse, label-balanced set of $m$ prior tokens from $\mathcal{T}_B$. To ensure these injected priors are both highly relevant to the target image and mutually diverse, we apply a two-stage Maximum Marginal Relevance (MMR) selection.

\paragraph{\textbf{Stage 1: Relevance Scoring and Prefiltering.}}
We first compute a localized relevance score $r(p)$ for every candidate token $p \in \mathcal{T}_B$. Instead of global image similarity, we measure token-to-token semantic affinity. We compute the cosine similarity between $p$ and all query tokens $f \in \mathcal{Q}$, defining $\mathcal{N}_{K_{\text{sim}}}(p, \mathcal{Q})$ as the subset of the $K_{\text{sim}}$ query tokens most similar to $p$. The relevance score is the average of these top similarities:
$$r(p) = \frac{1}{K_{\text{sim}}} \sum_{f \in \mathcal{N}_{K_{\text{sim}}}(p, \mathcal{Q})} \frac{p^\top f}{\|p\|_2 \|f\|_2}$$
Using this score, we prefilter the massive token bank down to a tractable candidate pool $\mathcal{C}$ by retaining only the top $M'$ candidates per semantic label.

\paragraph{\textbf{Stage 2: Maximum Marginal Relevance (MMR).}}
From the prefiltered pool $\mathcal{C}$, we greedily construct the final set of selected priors, $\mathcal{P}_{\text{sel}}$, such that $|\mathcal{P}_{\text{sel}}| = m$. We initialize $\mathcal{P}_{\text{sel}} = \emptyset$. At each iterative step, we select the candidate token $p^* \in \mathcal{C} \setminus \mathcal{P}_{\text{sel}}$ that maximizes the marginal objective:
$$p^* = \arg\max_{p \in \mathcal{C} \setminus \mathcal{P}_{\text{sel}}} \left[ r(p) - \lambda \max_{s \in \mathcal{P}_{\text{sel}}} \left( \frac{p^\top s}{\|p\|_2 \|s\|_2} \right) \right]$$
Once identified, we update the selected set: $\mathcal{P}_{\text{sel}} \leftarrow \mathcal{P}_{\text{sel}} \cup \{p^*\}$. 

Here, the hyperparameter $\lambda \in [0,1]$ explicitly controls the trade-off between cross-image relevance and intra-prior diversity. A higher $\lambda$ heavily penalizes the insertion of prior tokens that are semantically redundant with those already selected, in $\mathcal{P}_{\text{sel}}$. This systematic procedure returns a compact, label-balanced set of prior vertices ($\mathcal{P}_+$ and $\mathcal{P}_-$) that effectively spans the feature variance of the target object without interfering the solution for the augmented graph with redundant constraints.

\subsection{Patch Labels and Retrieval Hyperparameters}
\label[supp]{supp:b_hparams}
\paragraph{\textbf{Pixel masks to token labels.}} Each binary context mask is area-averaged onto the patch grid of the backbone, giving the foreground fraction of every patch. In the main experiments, a patch becomes a foreground prior if this fraction is at least $0.6$ and a background prior if it is at most $0.05$; patches in between, which straddle object boundaries, are discarded and never used as priors. The controlled analyses (Sec.~\ref{supp:b_controlled}) use $0.5$ and $0.05$, and $0.02$ and $0.001$ on CFD, whose cracks are thinner than a $16\times16$ patch at $448\times448$. Token labels are only extracted from context images; target-query masks are used exclusively for evaluation.

\paragraph{\textbf{Retrieval.}} We use $K_{\text{sim}}=8$ query neighbours for the relevance score and set $\lambda=0.35$. Before MMR, the main experiments keep the $M'=2{,}048$ most relevant candidates per label ($M'=512$ in the controlled analyses). MMR is run separately for each label and selects the same number of foreground and background tokens ($m/2$ each, unless fewer candidates are available), so every prior set is exactly label-balanced. The first token of each label is its most relevant candidate.

\subsection{Evaluation Protocol of the Controlled Analyses}
\label[supp]{supp:b_controlled}
The same-prior baselines, the component ablation, the prior-corruption study, and the COCO controllability experiment were run with the \texttt{quick\_*.py} scripts of the code release (commit \texttt{98c115a}). Within each experiment, all compared methods share the same inputs: frozen DINOv3-S/16 features computed at $448\times448$ ($28\times28$ patch grid), the same query-specific priors ($m=512$, 256 per label), and $\tau=0.7$. PANC uses prior-ROC thresholding with $\kappa=1$ on DUTS and COCO and $\kappa=1000$ on CFD. Patch scores are bilinearly upsampled to the image resolution before thresholding, and each query is scored by the IoU of its foreground mask at full resolution. On DUTS, the 12 context images are drawn from DUTS-TR and the 24 queries from DUTS-TE; on CFD, the 5 context and 20 query images are drawn disjointly from the dataset. Both draws use fixed seeds. The component ablation uses the first 12 queries of each subset and the prior-corruption study the first 6.

\subsection{COCO Controllability Protocol}
\label[supp]{supp:b_coco}
Queries are 8 COCO val2017 images, drawn with a fixed seed among the images that contain both \emph{person} and \emph{dog} with each class covering at least 1\% of the image. Context images are sampled from val2017 images in which the class covers at least 2\% of the image, preferring images that do not contain the other class, and are disjoint from all queries. We draw 3 independent context banks with 5 images per class; matched and unrelated results are averaged over all $8\times3$ query--bank pairs. The target mask is the union of all non-crowd instances of the selected class, so the evaluation targets the class rather than a specific instance. \emph{Matched priors} use the bank of the target class, and \emph{unrelated priors} use the bank of the other class (dog context for the person target and vice versa), with the same retrieval procedure and token count ($m=512$, $\kappa=1$); the similarity of the retrieved tokens to the query is not matched across conditions. \emph{No priors} is query-only NCut thresholded at its mean score, with the sign fixed so that the entry of largest magnitude is positive. The query images, backbone, and $\tau$ are identical in all three conditions.

\subsection{Same-Prior Baselines}
\label[supp]{supp:b_baselines}
In the setting of Sec.~\ref{supp:b_controlled}, all baselines receive exactly the same query tokens and query-specific retrieved priors as PANC, and their scores are bilinearly upsampled and thresholded at $0.5$. \emph{Prototype}: cosine similarities between each query token and the $\ell_2$-normalized means of the foreground and background priors are passed through a softmax with temperature $0.07$. \emph{$k$-NN}: the score is the fraction of foreground labels among the $k=5$ most cosine-similar priors. \emph{Label propagation}: we build a symmetric $30$-NN cosine graph over query and prior tokens with weights $\exp(S_{ij}/\tau)$, $\tau=0.7$, row-normalize it into $\mat P$, and iterate $\mathbf F\leftarrow\alpha\mat P\mathbf F+(1-\alpha)\mathbf Y$ for 40 steps with $\alpha=0.95$, clamping the prior rows to their labels after every step. These are standard settings that were not tuned per dataset; the baselines may therefore not be at their individual optimum.

\subsection{Prior-Corruption Protocol}
\label[supp]{supp:b_corruption}
We use the first 6 queries per dataset of Sec.~\ref{supp:b_controlled}. For each query, priors are retrieved once from the clean bank and kept fixed. For a corruption rate $r\in\{5,10,20,50\}\%$, we flip the labels of exactly $\mathrm{round}(r\,m)$ randomly chosen priors (redrawing in the rare case that one label disappears) and rerun PANC. We use five draws with fixed seeds and report the mean and standard deviation of the per-draw mIoU.

\section{Extended Evaluation and Examples}
\label[supp]{supp:c}

\begin{table}[t]
    \centering
    \caption{
        Component ablation of PANC on DUTS and CFD
        (setting of Sec.~\ref{supp:b_controlled}, first 12 queries per dataset).
        Results are mIoU (\%, higher is better).
    }
    \label{tab:component_ablation}
    \setlength{\tabcolsep}{5pt}
    \begin{tabular}{lcc}
        \toprule
        Configuration
            & DUTS
            & CFD \\
        \midrule
        No priors (NCut)
            & $62.1$
            & $46.2$ \\
        Prior nodes, no anchors
            & $58.0$
            & $72.5$ \\
        No prior-aware sign stabilization
            & $53.2$
            & $77.7$ \\
        No threshold calibration
            & $62.5$
            & $87.0$ \\
        Fixed anchor coupling
            & $66.4$
            & $90.9$ \\
        \midrule
        \textbf{Full PANC}
            & $\mathbf{66.5}$
            & $\mathbf{91.0}$ \\
        \bottomrule
    \end{tabular}
\end{table}

\paragraph{\textbf{Component variants.}} Each row of \cref{tab:component_ablation} changes one component of full PANC in the setting of Sec.~\ref{supp:b_controlled}. \emph{No priors (NCut)}: query-only NCut with the mean score as threshold; since no labels are available, the sign is fixed so that the entry of largest magnitude is positive. \emph{Prior nodes, no anchors}: the prior tokens are added to the graph without anchors, and prior-aware sign stabilization and prior-ROC thresholding are kept unchanged. \emph{No prior-aware sign stabilization}: the full graph is used, but the sign follows the query-only rule above. \emph{No threshold calibration}: full PANC scores thresholded at their mean. \emph{Fixed anchor coupling}: every prior uses the same weight $\kappa$ times the average of the per-prior local affinities, with the same clipping as $\alpha_i$.

\subsection{Multi-Object Controllability}

We extend our class controllability tests on the MS COCO dataset, demonstrating that efficient prior bank generation enables high-quality segmentation of a class object in never-seen images.

We divided the examples of the validation set into two main categories based on the geometric deformability of the target class: rigid classes that mainly change their form with the perspective (see Figs.~\ref{fig:rigid_airplane} and~\ref{fig:rigid_boat}) and non-rigid classes with a deformable structure (see Figs.~\ref{fig:non_rigid_banana}, \ref{fig:non_rigid_tie}, and~\ref{fig:non_rigid_suitcase}). 

Despite the variety of airplanes and boats, the results are consistent from all points of view. A similar situation was identified on bananas, ties, and suitcases. In contrast, among the bananas, PANC was able to identify a printed one that was not included in the ground truth---see Fig.~\ref{fig:non_rigid_banana}, second row from the bottom.

\captionsetup[figure]{skip=3pt}

\newcommand{\qimgfour}[1]{%
  \raisebox{-0.5\height}{\includegraphics[width=0.20\linewidth]{#1}}%
}
\newcommand{\qimgsix}[1]{%
  \raisebox{-0.5\height}{\includegraphics[width=0.16\linewidth]{#1}}%
}

\begin{figure*}[p]
\centering
\begingroup
\setlength{\tabcolsep}{0pt}
\begin{adjustbox}{max width=\textwidth,max totalheight=0.95\textheight,center}
\begin{tabular}{@{}c @{\hspace{2pt}} c @{\hspace{2pt}} c @{\hspace{2pt}} c@{}}
    \qimgfour{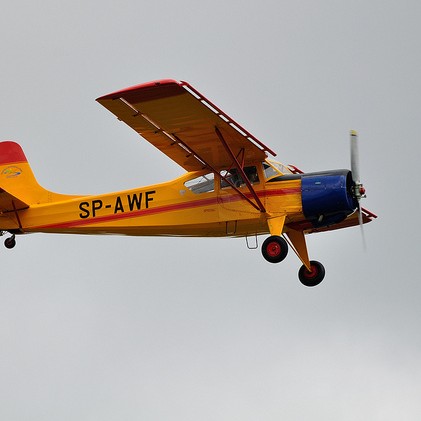} &
    \qimgfour{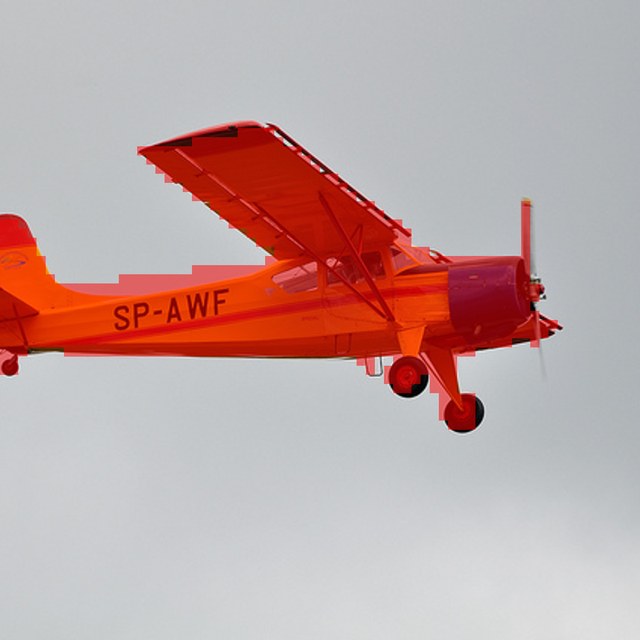} &
    \qimgfour{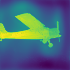} &
    \qimgfour{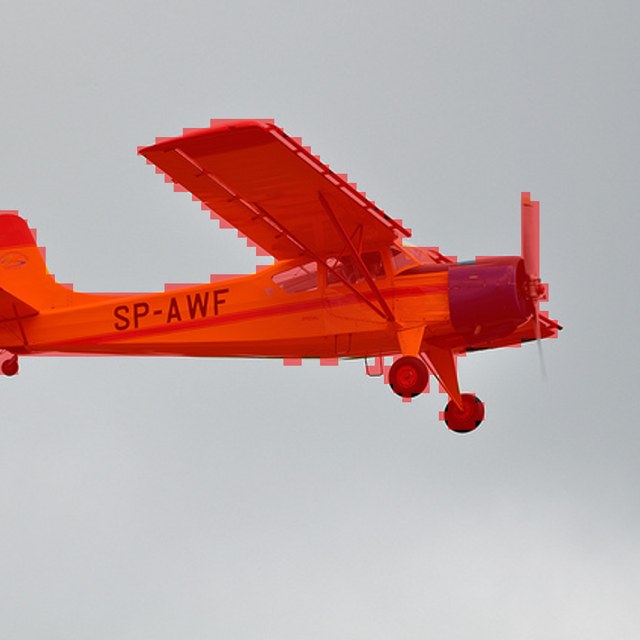} \\
    \noalign{\vspace{1pt}}

    \qimgfour{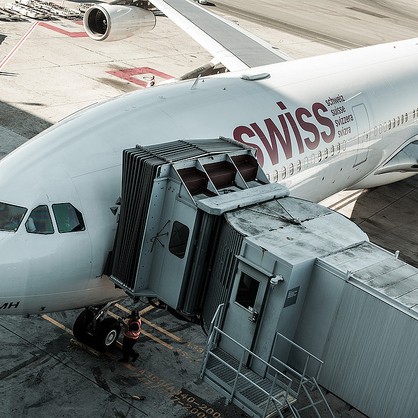} &
    \qimgfour{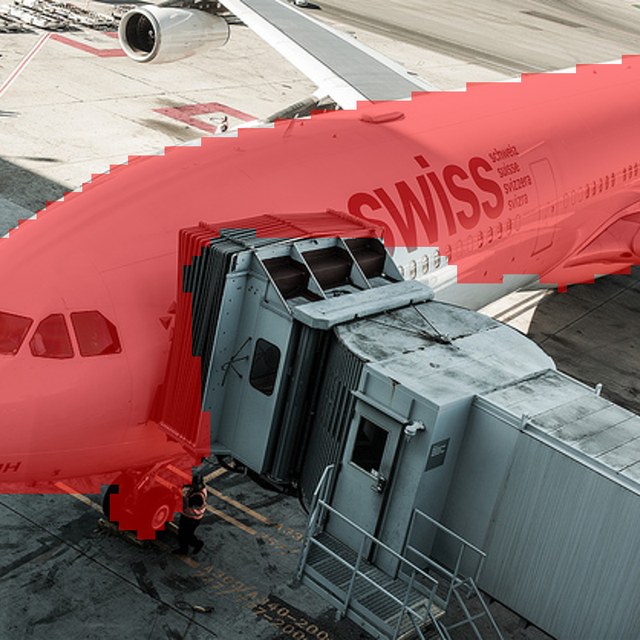} &
    \qimgfour{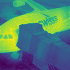} &
    \qimgfour{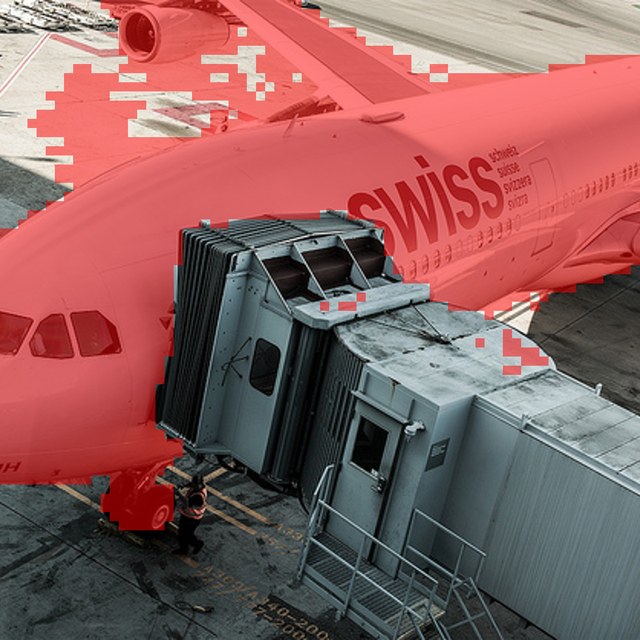} \\
    \noalign{\vspace{1pt}}

    \qimgfour{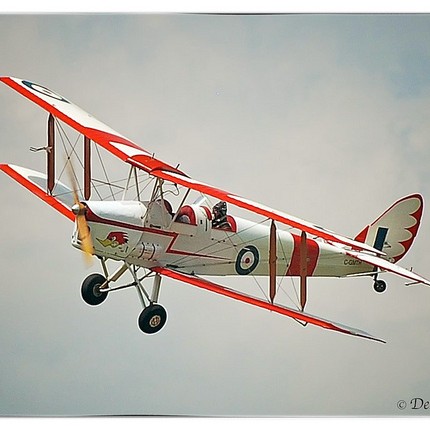} &
    \qimgfour{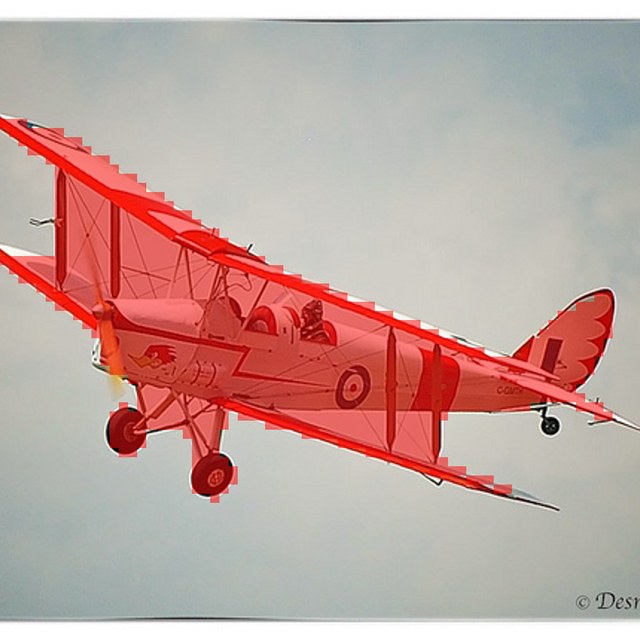} &
    \qimgfour{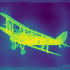} &
    \qimgfour{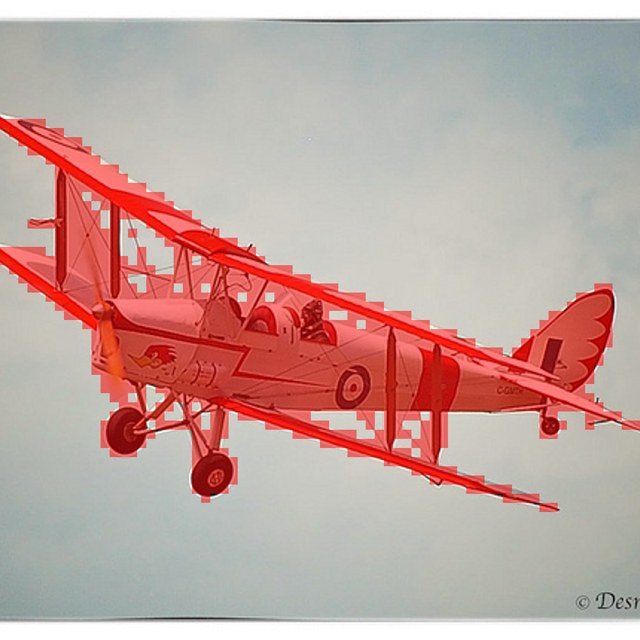} \\
    \noalign{\vspace{1pt}}

    \qimgfour{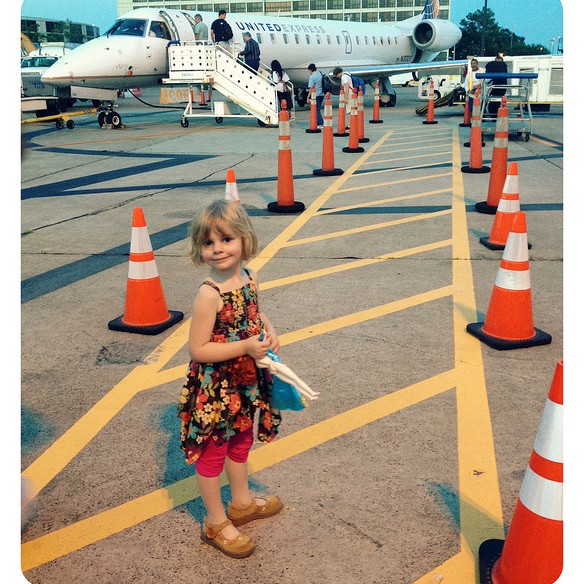} &
    \qimgfour{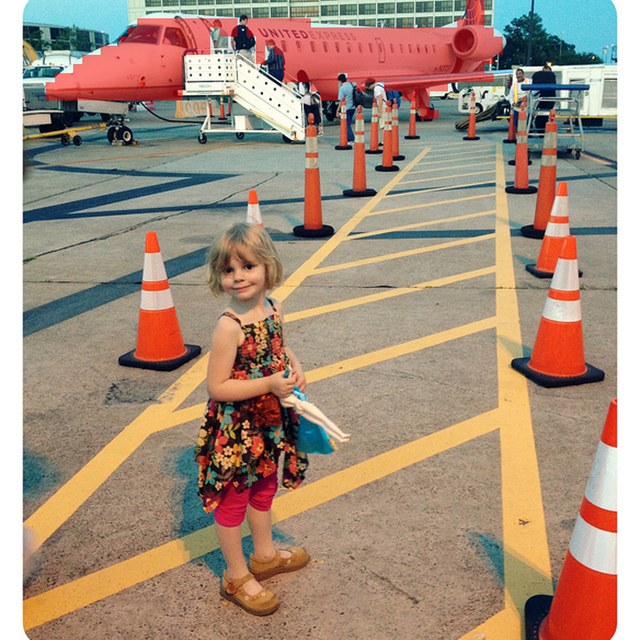} &
    \qimgfour{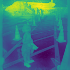} &
    \qimgfour{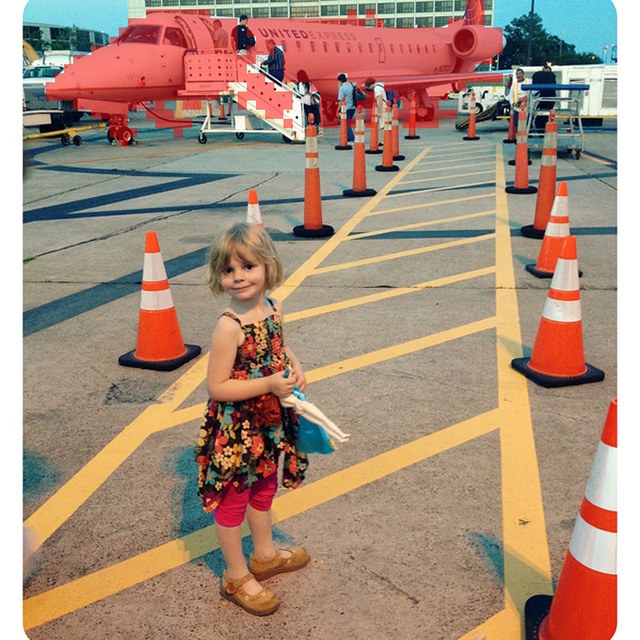} \\
    \noalign{\vspace{1pt}}

    \qimgfour{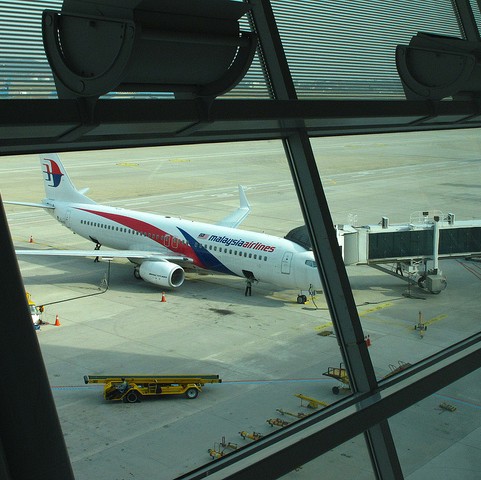} &
    \qimgfour{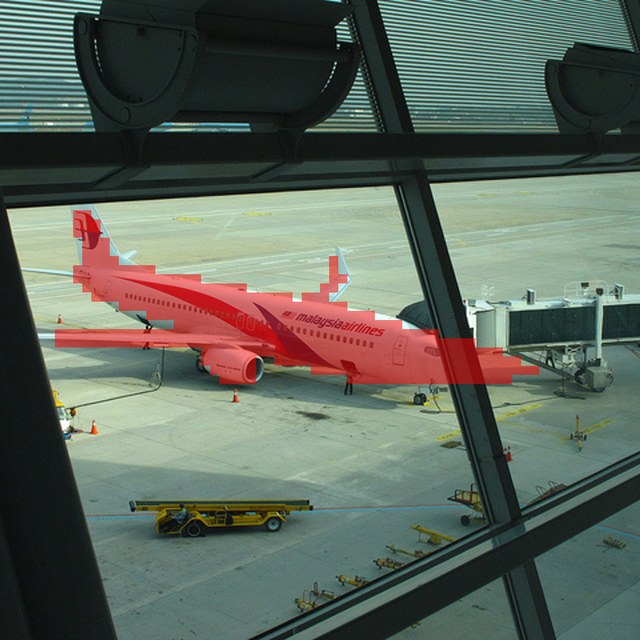} &
    \qimgfour{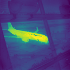} &
    \qimgfour{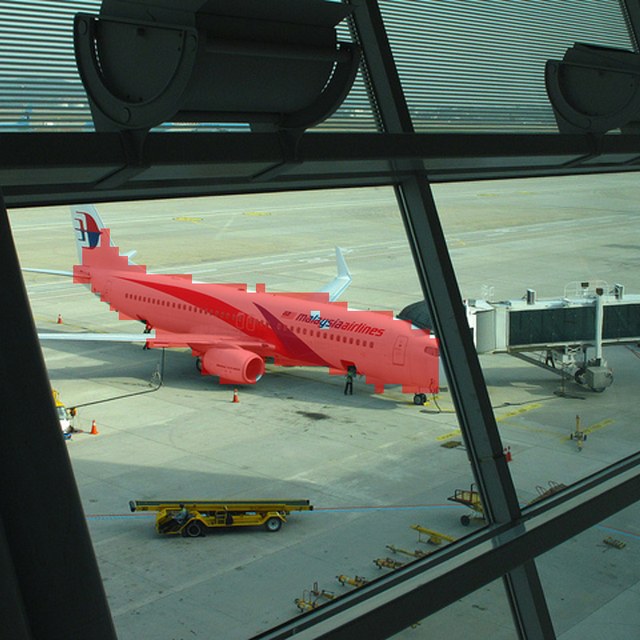} \\
    \noalign{\vspace{1pt}}

    \qimgfour{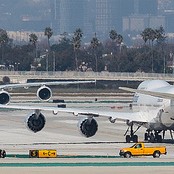} &
    \qimgfour{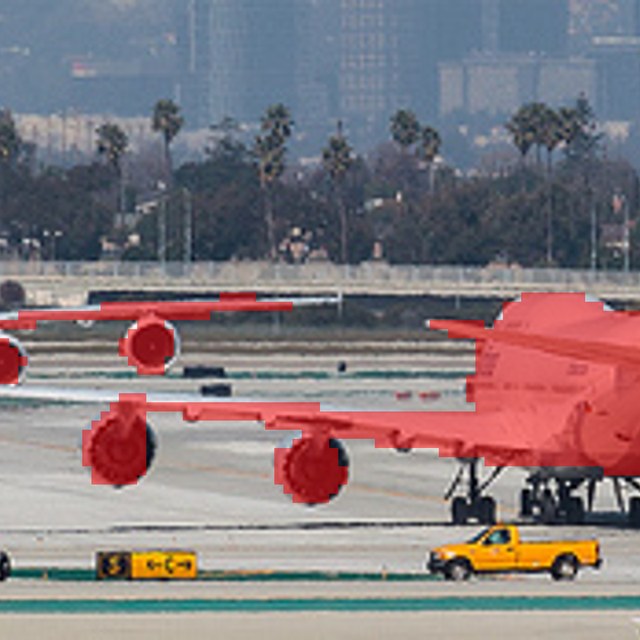} &
    \qimgfour{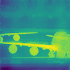} &
    \qimgfour{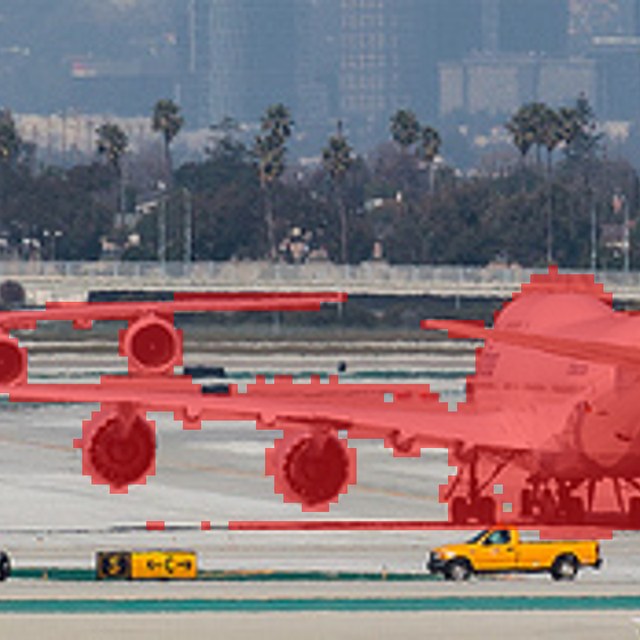} \\
    \noalign{\vspace{1pt}}

    \qimgfour{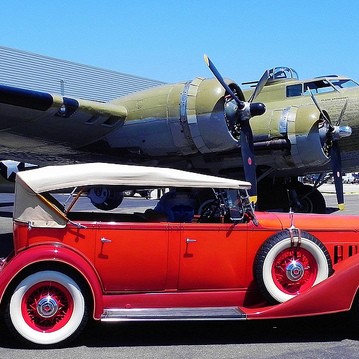} &
    \qimgfour{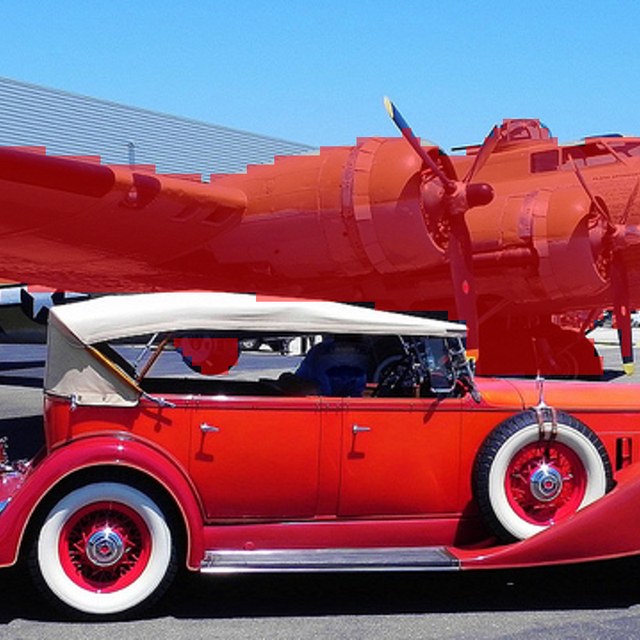} &
    \qimgfour{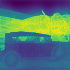} &
    \qimgfour{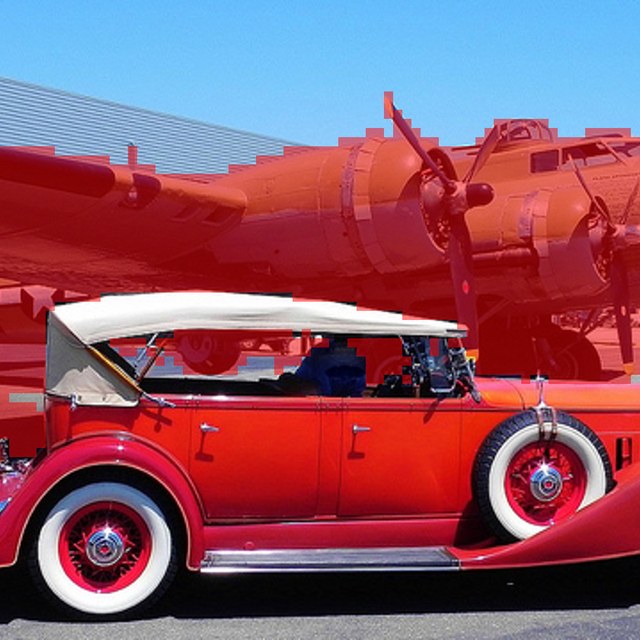} \\
    \noalign{\vspace{2pt}}

    \raisebox{-0.5\height}{\small Input} &
    \raisebox{-0.5\height}{\small GT} &
    \raisebox{-0.5\height}{\small\shortstack{PANC\\Eigen Attn.}} &
    \raisebox{-0.5\height}{\small\shortstack{PANC\\Mask}} \\
\end{tabular}
\end{adjustbox}
\endgroup
\caption{Additional qualitative comparison on the rigid MS COCO \textbf{Airplane} class.}
\label{fig:rigid_airplane}
\end{figure*}

\begin{figure*}[p]
\centering
\begingroup
\setlength{\tabcolsep}{0pt}
\begin{adjustbox}{max width=\textwidth,max totalheight=0.95\textheight,center}
\begin{tabular}{@{}c @{\hspace{2pt}} c @{\hspace{2pt}} c @{\hspace{2pt}} c@{}}
    \qimgfour{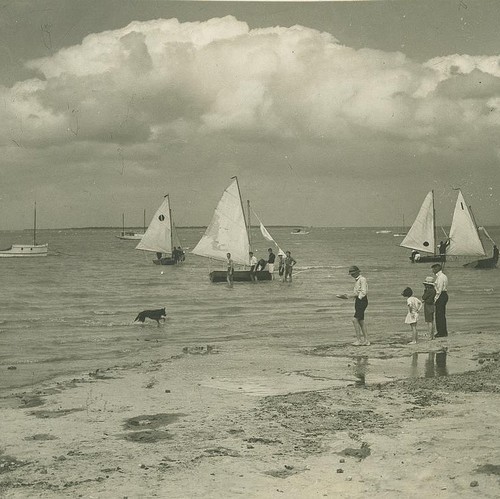} &
    \qimgfour{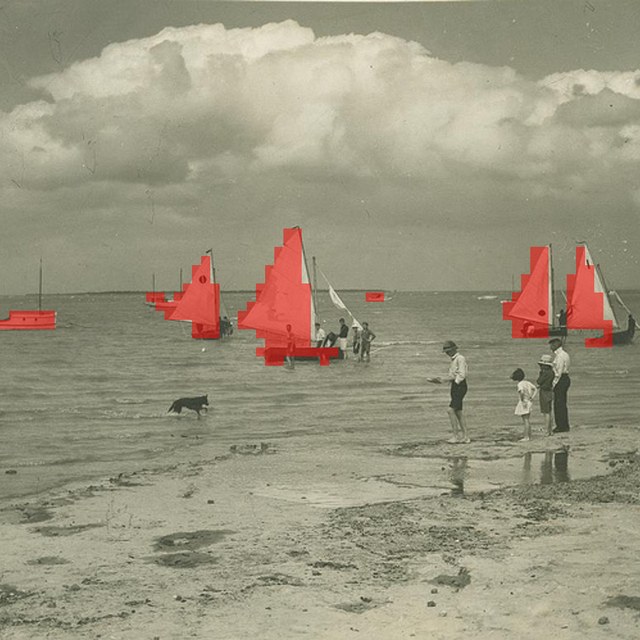} &
    \qimgfour{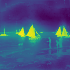} &
    \qimgfour{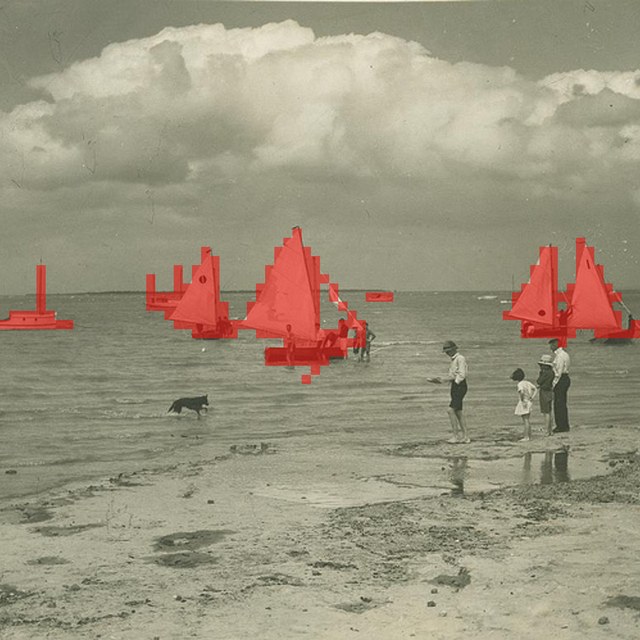} \\
    \noalign{\vspace{1pt}}

    \qimgfour{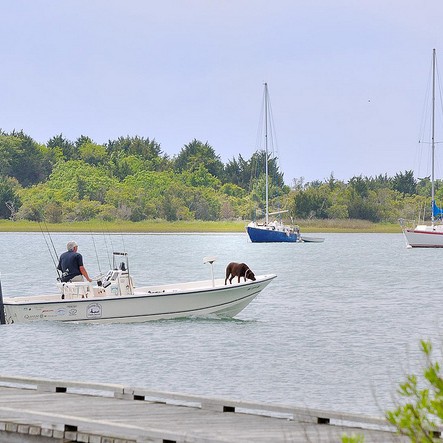} &
    \qimgfour{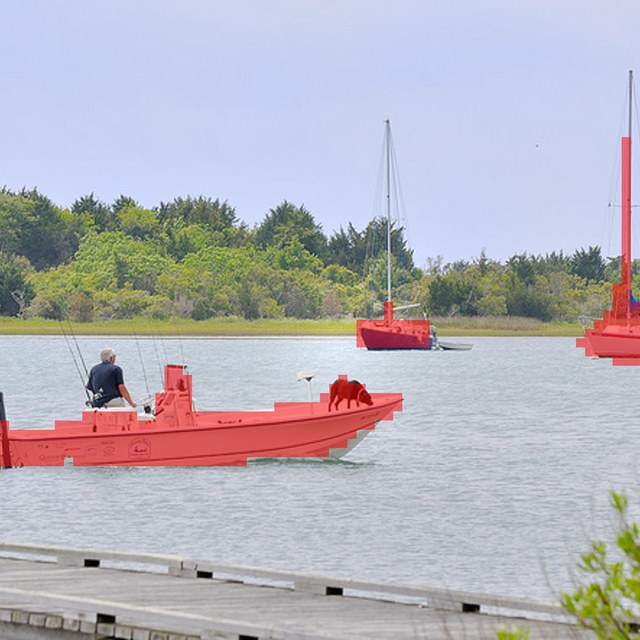} &
    \qimgfour{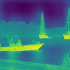} &
    \qimgfour{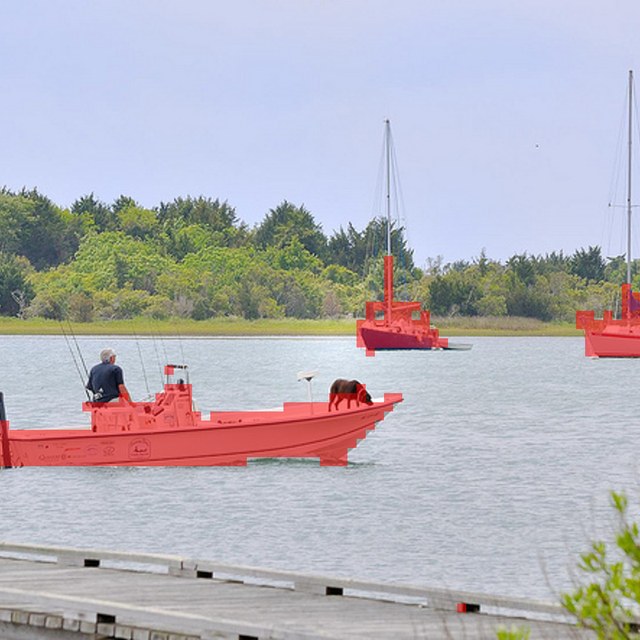} \\
    \noalign{\vspace{1pt}}

    \qimgfour{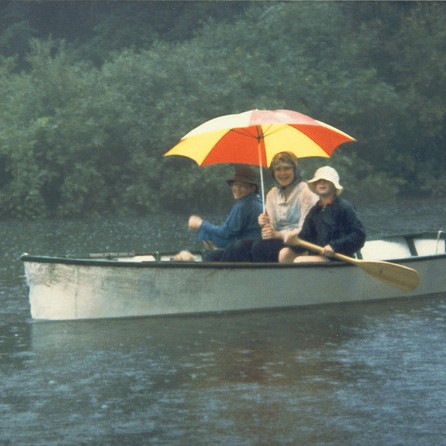} &
    \qimgfour{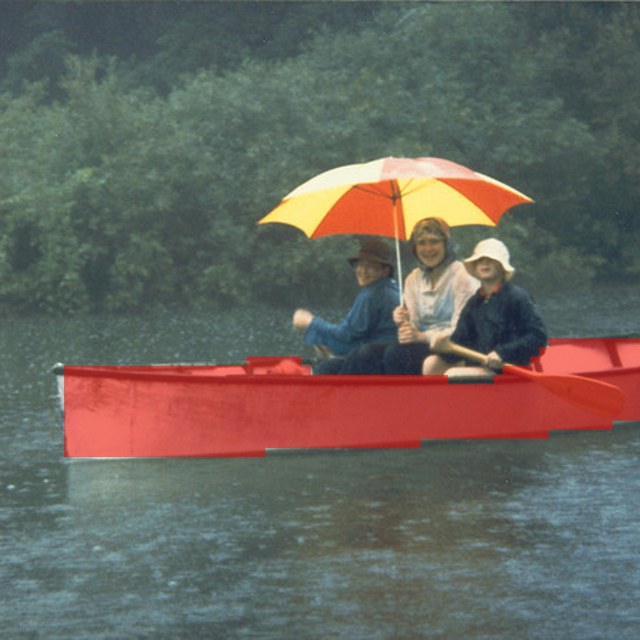} &
    \qimgfour{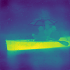} &
    \qimgfour{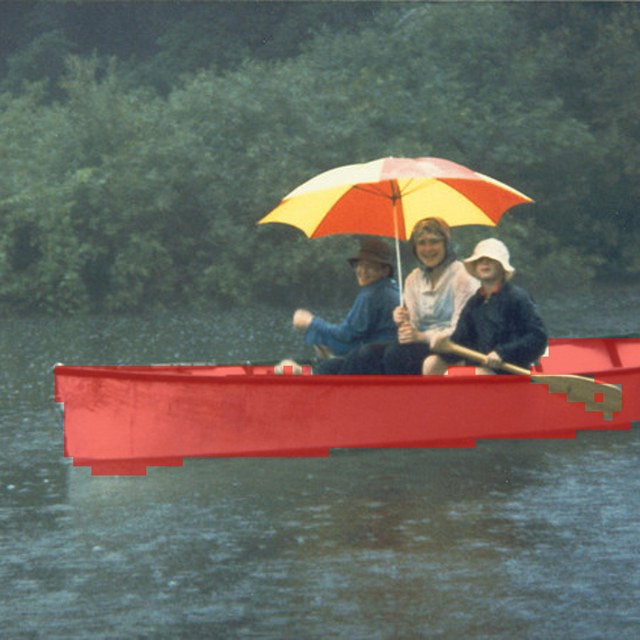} \\
    \noalign{\vspace{1pt}}

    \qimgfour{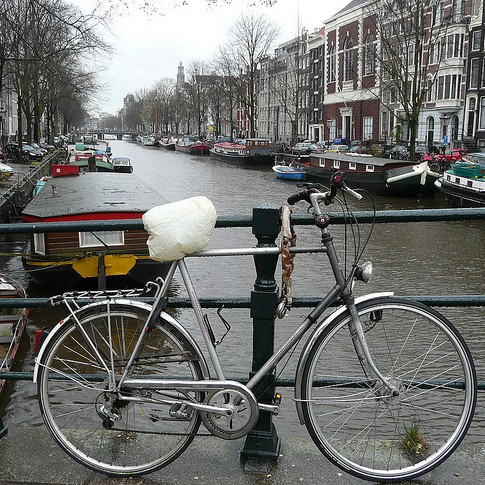} &
    \qimgfour{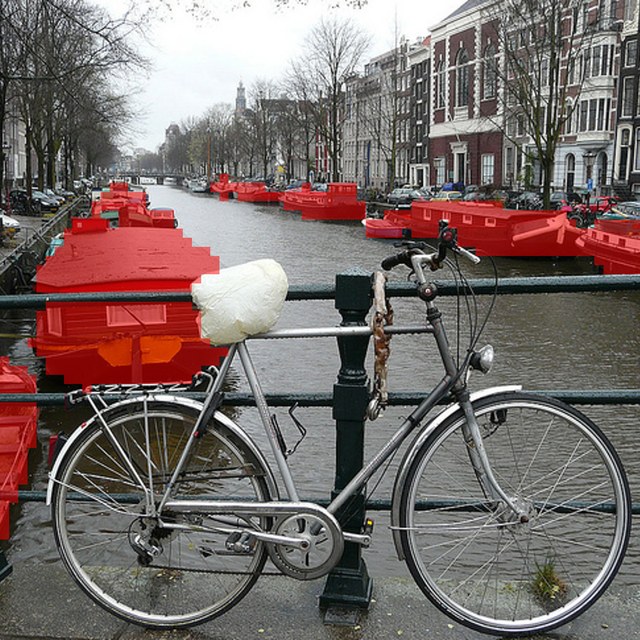} &
    \qimgfour{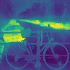} &
    \qimgfour{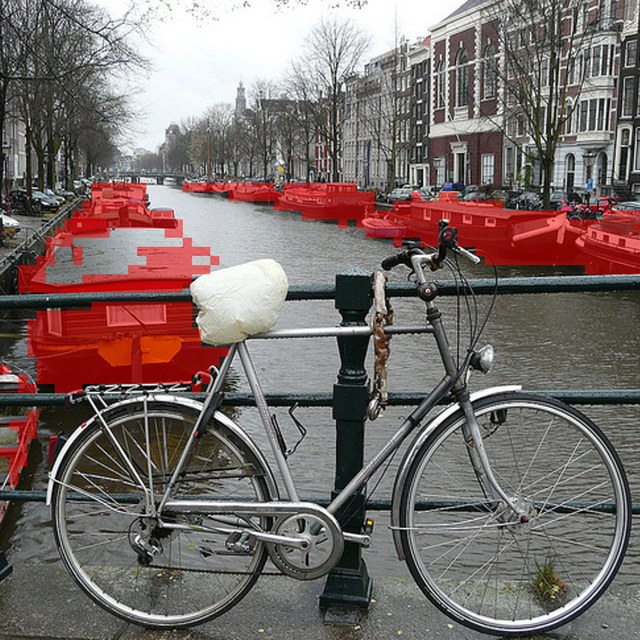} \\
    \noalign{\vspace{1pt}}

    \qimgfour{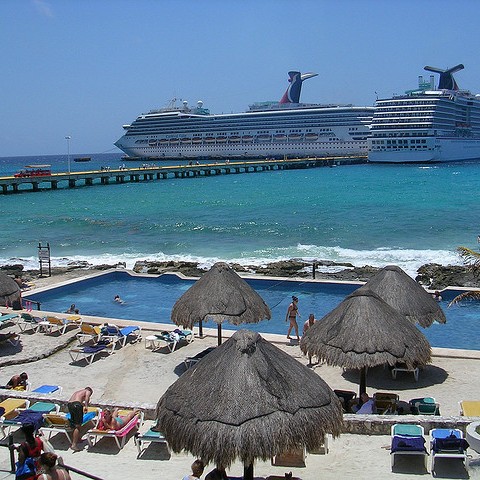} &
    \qimgfour{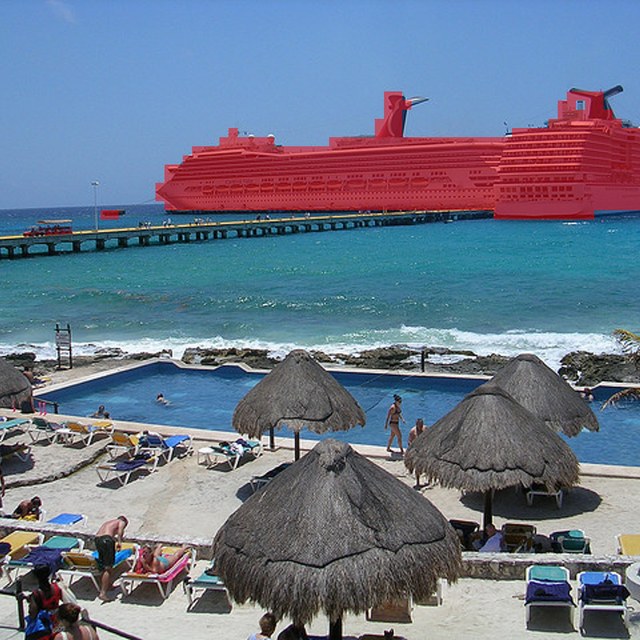} &
    \qimgfour{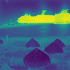} &
    \qimgfour{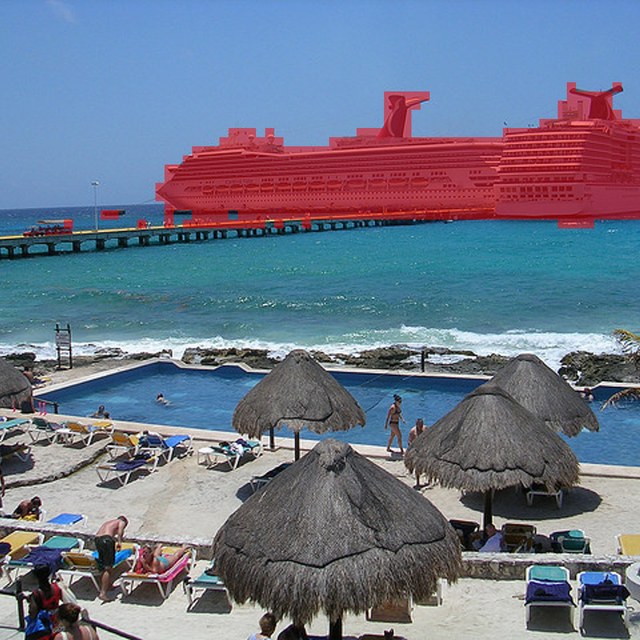} \\
    \noalign{\vspace{1pt}}

    \qimgfour{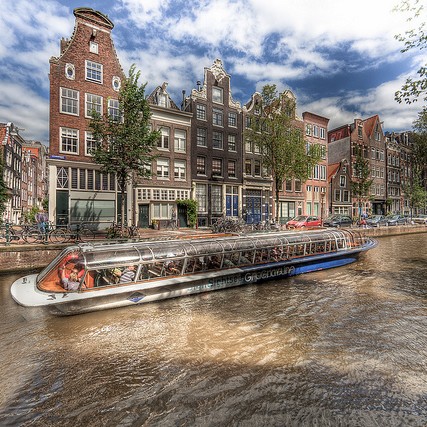} &
    \qimgfour{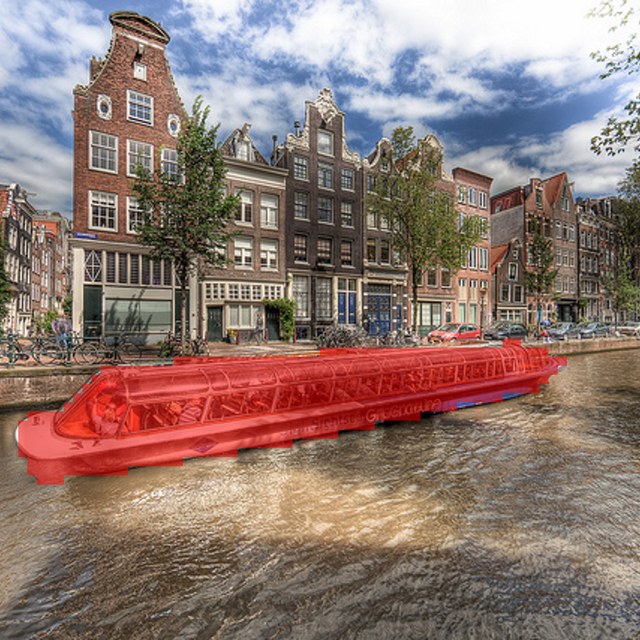} &
    \qimgfour{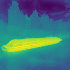} &
    \qimgfour{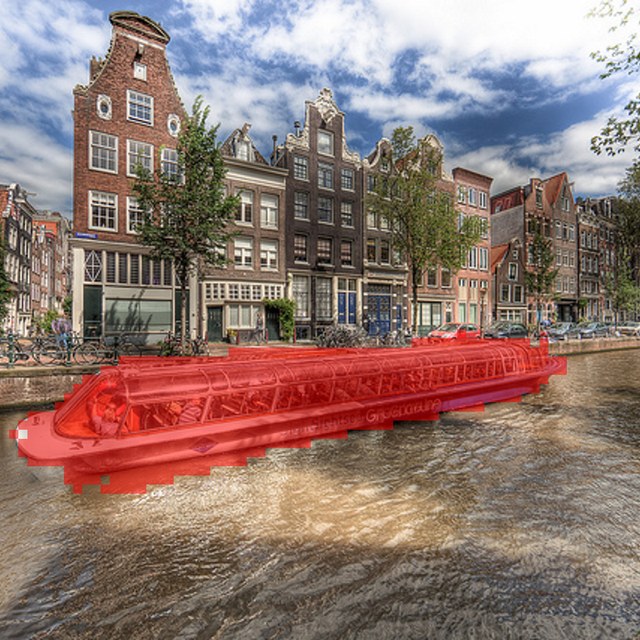} \\
    \noalign{\vspace{1pt}}

    \qimgfour{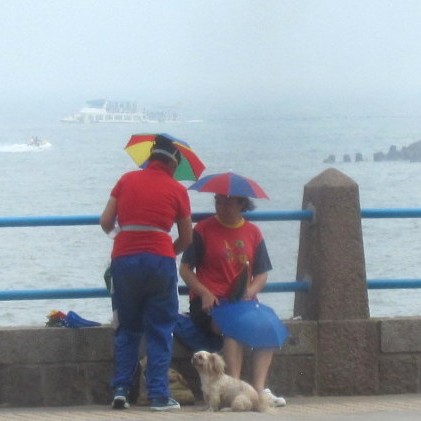} &
    \qimgfour{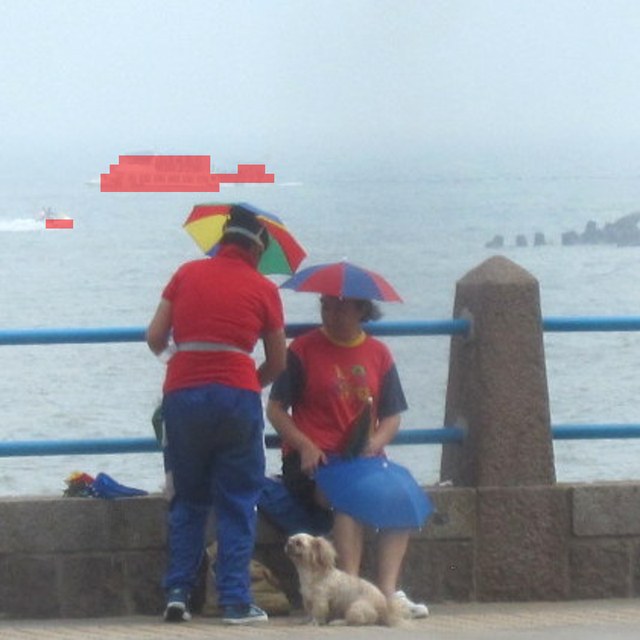} &
    \qimgfour{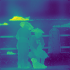} &
    \qimgfour{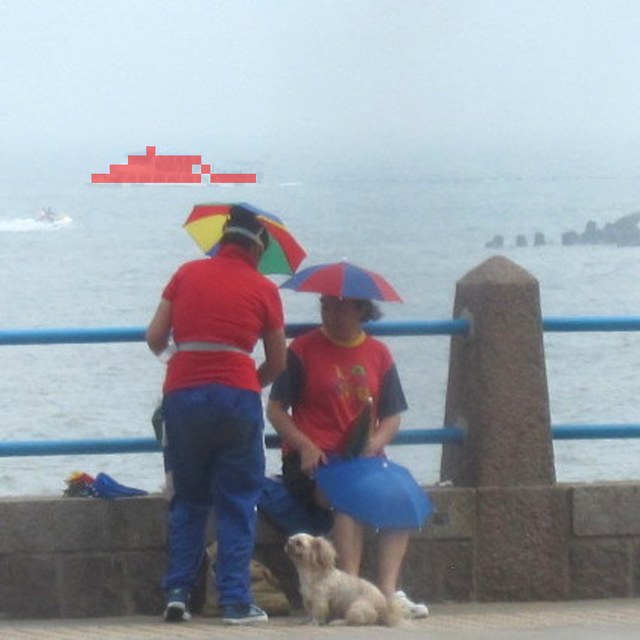} \\
    \noalign{\vspace{2pt}}

    \raisebox{-0.5\height}{\small Input} &
    \raisebox{-0.5\height}{\small GT} &
    \raisebox{-0.5\height}{\small\shortstack{PANC\\Eigen Attn.}} &
    \raisebox{-0.5\height}{\small\shortstack{PANC\\Mask}} \\
\end{tabular}
\end{adjustbox}
\endgroup
\caption{Additional qualitative comparison on the rigid MS COCO \textbf{Boat} class.}
\label{fig:rigid_boat}
\end{figure*}

\begin{figure*}[p]
\centering
\begingroup
\setlength{\tabcolsep}{0pt}
\begin{adjustbox}{max width=\textwidth,max totalheight=0.95\textheight,center}
\begin{tabular}{@{}c @{\hspace{2pt}} c @{\hspace{2pt}} c @{\hspace{2pt}} c@{}}
    \qimgfour{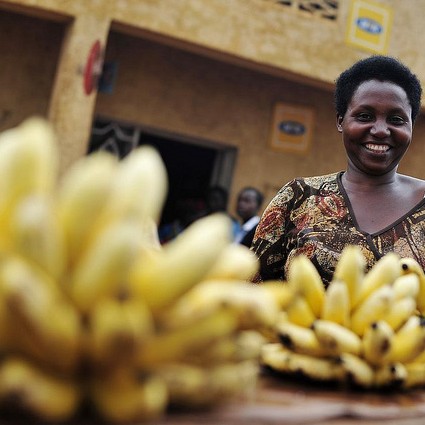} &
    \qimgfour{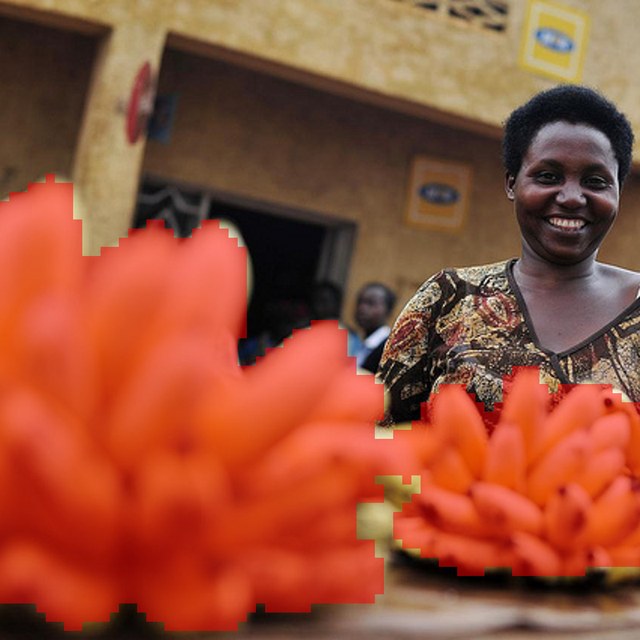} &
    \qimgfour{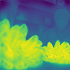} &
    \qimgfour{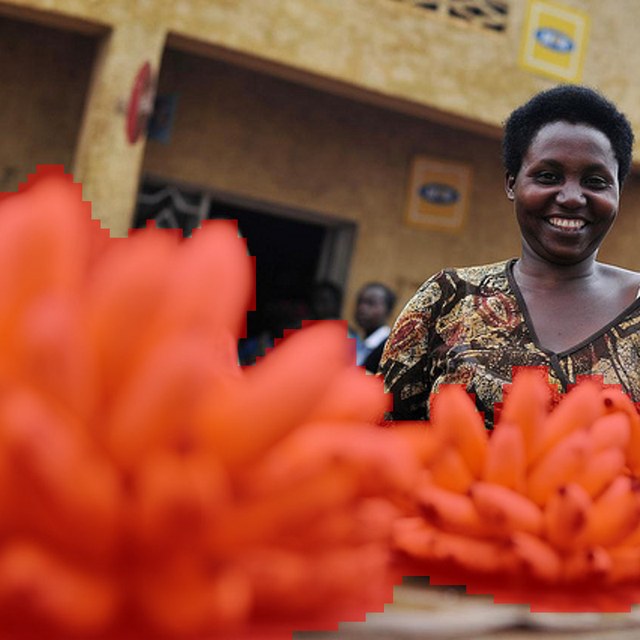} \\
    \noalign{\vspace{1pt}}

    \qimgfour{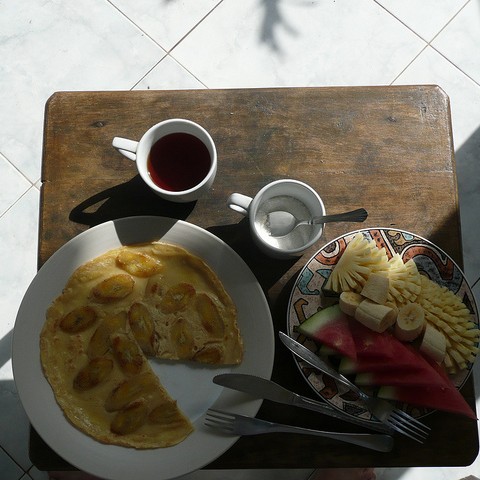} &
    \qimgfour{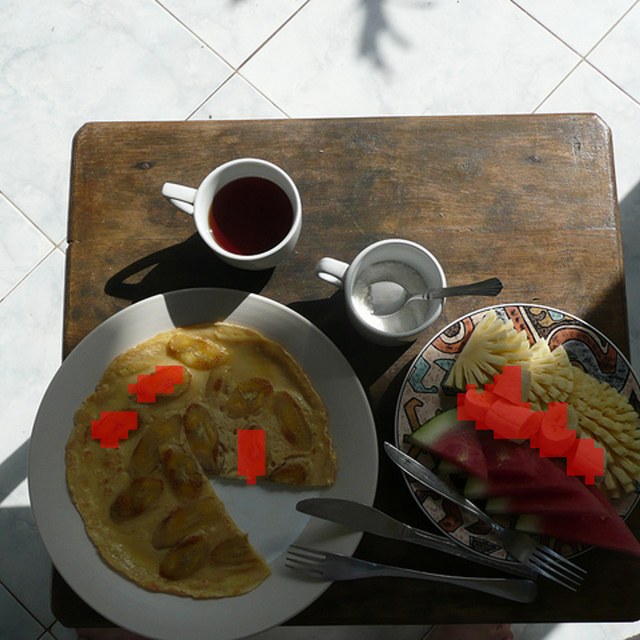} &
    \qimgfour{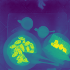} &
    \qimgfour{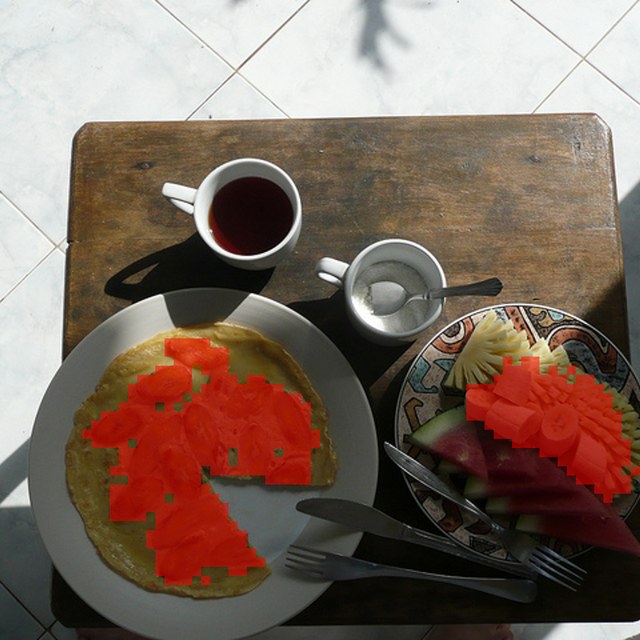} \\
    \noalign{\vspace{1pt}}

    \qimgfour{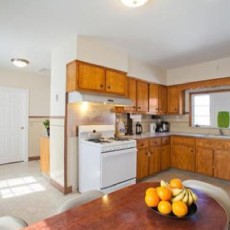} &
    \qimgfour{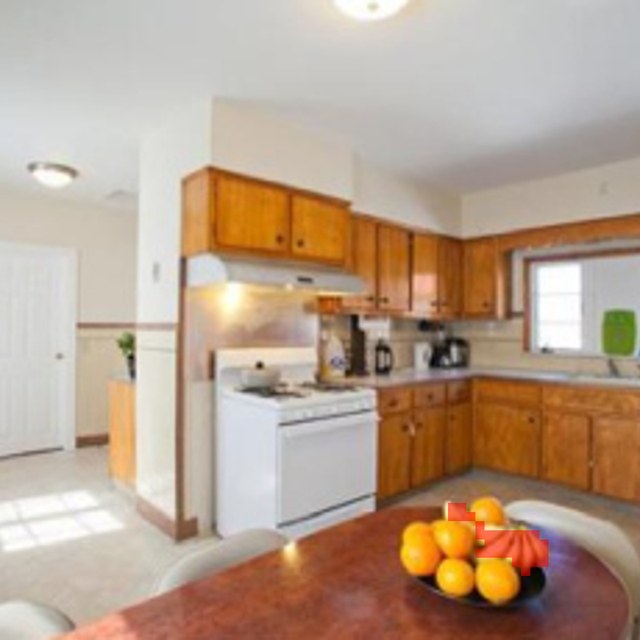} &
    \qimgfour{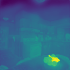} &
    \qimgfour{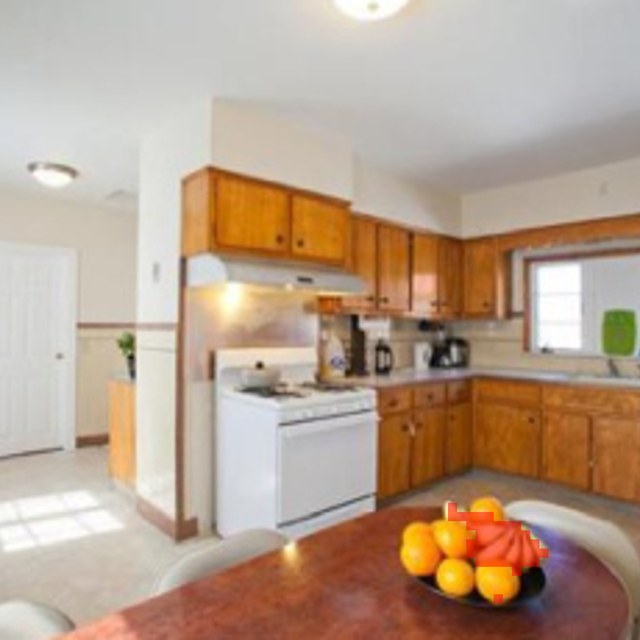} \\
    \noalign{\vspace{1pt}}

    \qimgfour{} &
    \qimgfour{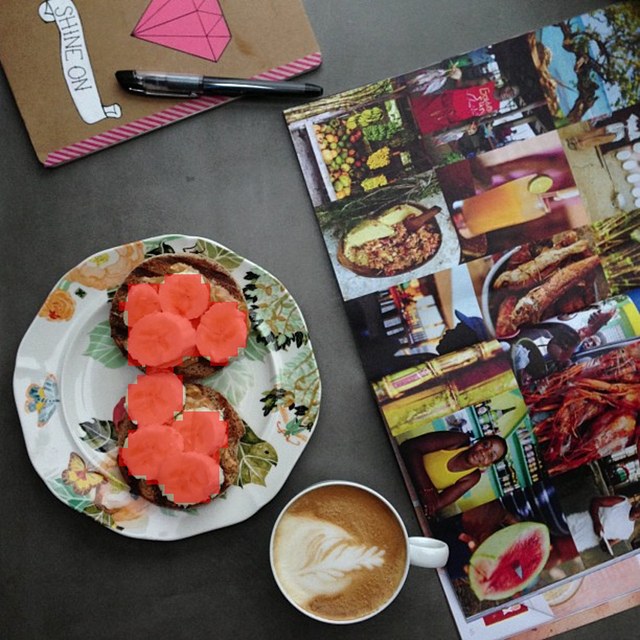} &
    \qimgfour{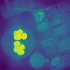} &
    \qimgfour{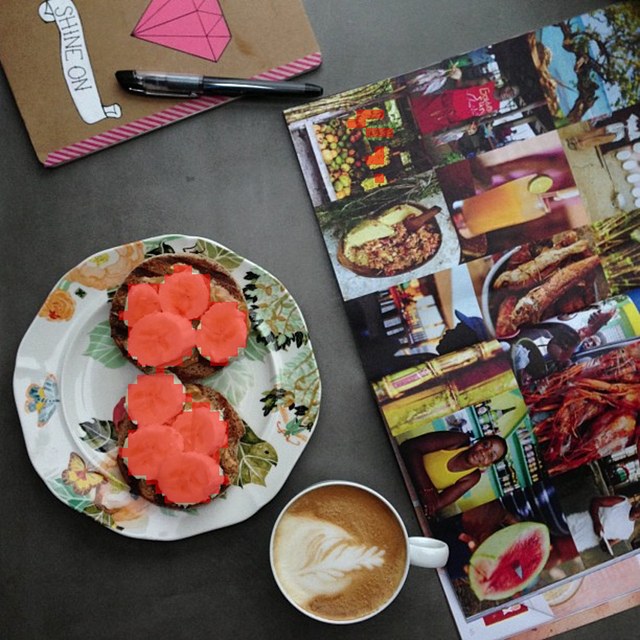} \\
    \noalign{\vspace{1pt}}

    \qimgfour{} &
    \qimgfour{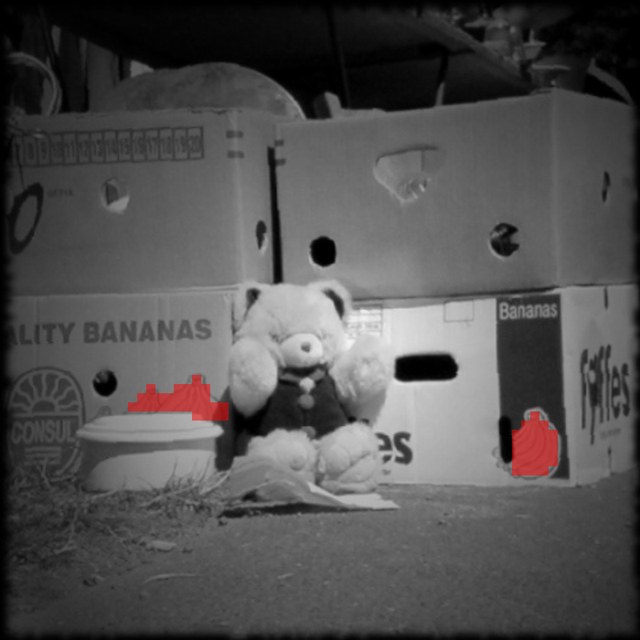} &
    \qimgfour{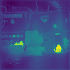} &
    \qimgfour{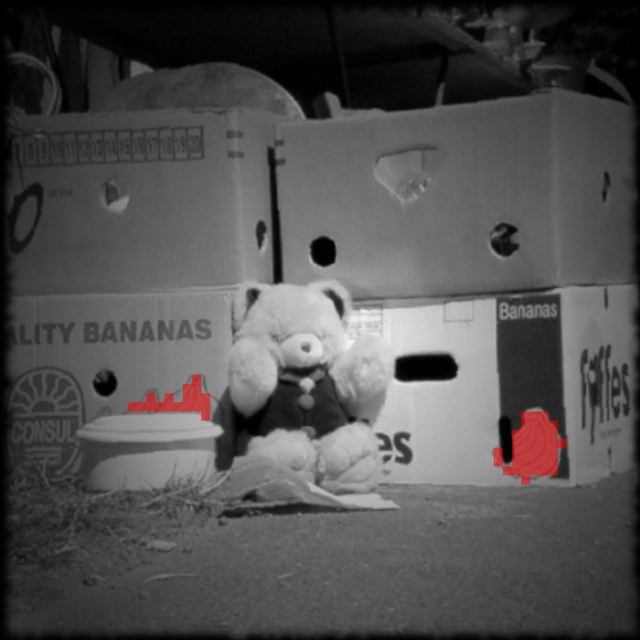} \\
    \noalign{\vspace{1pt}}

    \qimgfour{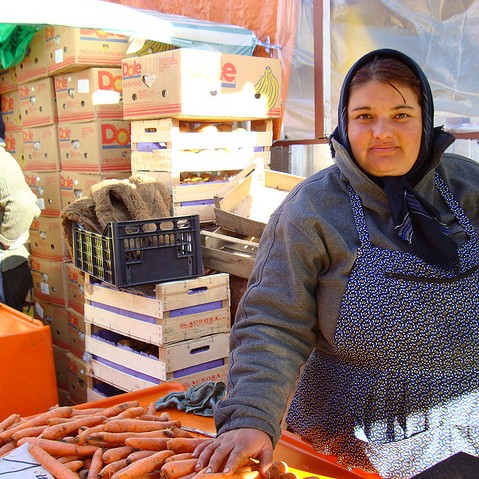} &
    \qimgfour{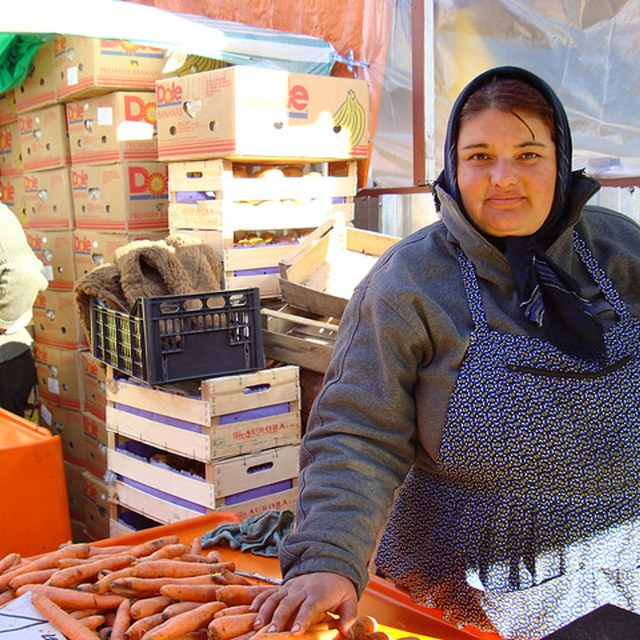} &
    \qimgfour{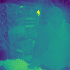} &
    \qimgfour{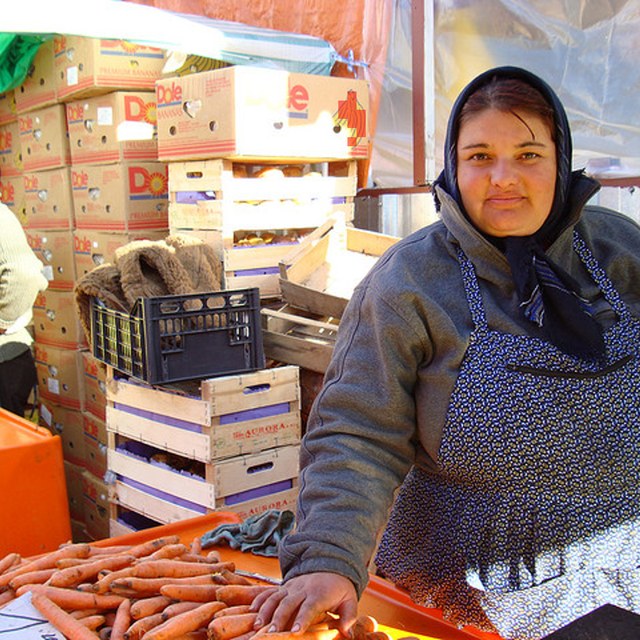} \\
    \noalign{\vspace{1pt}}

    \qimgfour{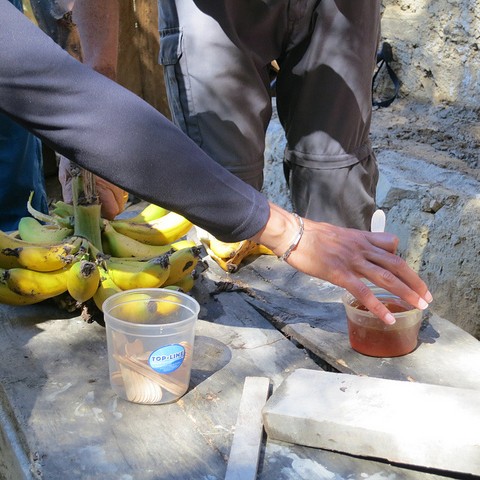} &
    \qimgfour{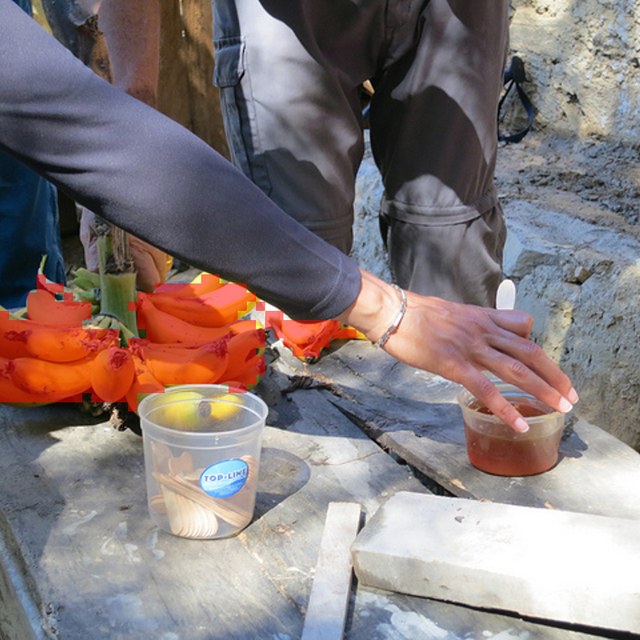} &
    \qimgfour{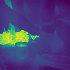} &
    \qimgfour{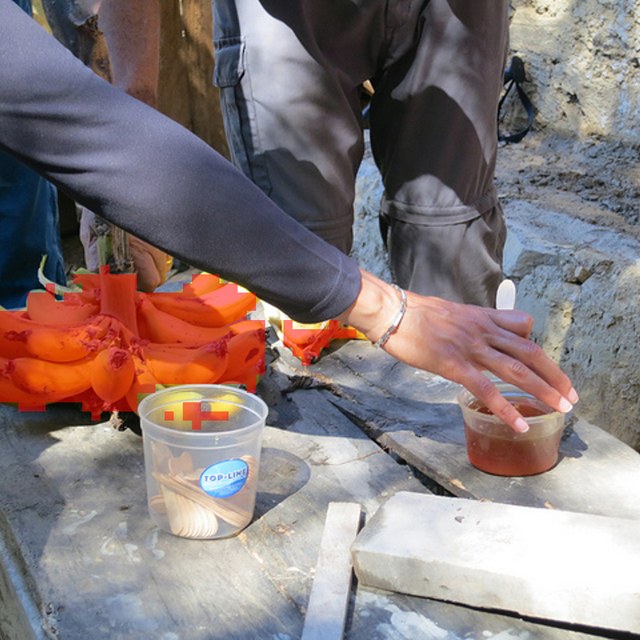} \\
    \noalign{\vspace{2pt}}

    \raisebox{-0.5\height}{\small Input} &
    \raisebox{-0.5\height}{\small GT} &
    \raisebox{-0.5\height}{\small\shortstack{PANC\\Eigen Attn.}} &
    \raisebox{-0.5\height}{\small\shortstack{PANC\\Mask}} \\
\end{tabular}
\end{adjustbox}
\endgroup
\caption{Additional qualitative comparison on the non-rigid MS COCO \textbf{Banana} class.}
\label{fig:non_rigid_banana}
\end{figure*}

\begin{figure*}[p]
\centering
\begingroup
\setlength{\tabcolsep}{0pt}
\begin{adjustbox}{max width=\textwidth,max totalheight=0.95\textheight,center}
\begin{tabular}{@{}c @{\hspace{2pt}} c @{\hspace{2pt}} c @{\hspace{2pt}} c@{}}
    \qimgfour{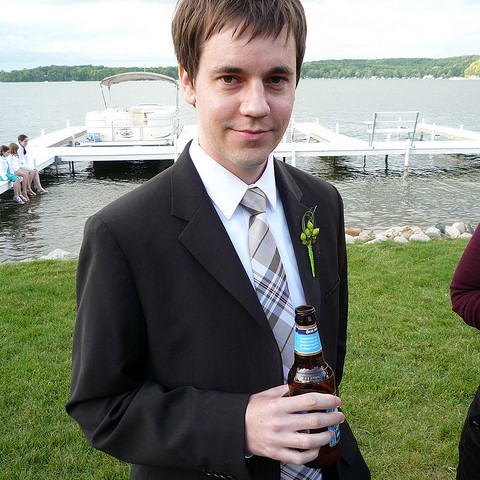} &
    \qimgfour{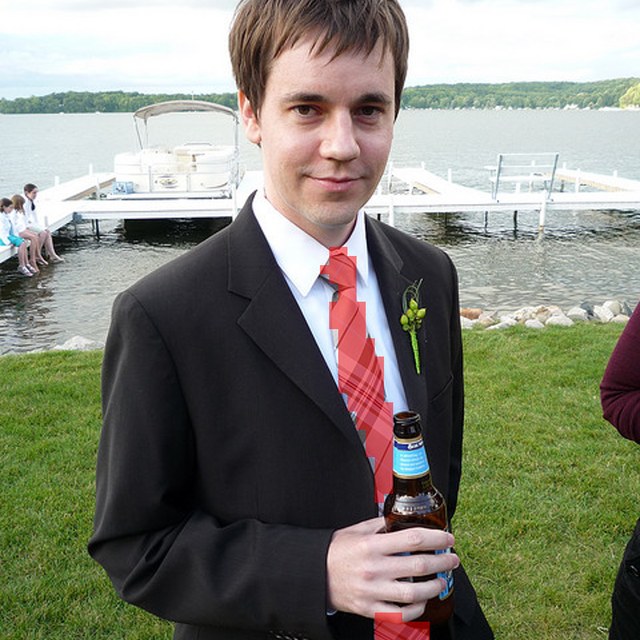} &
    \qimgfour{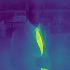} &
    \qimgfour{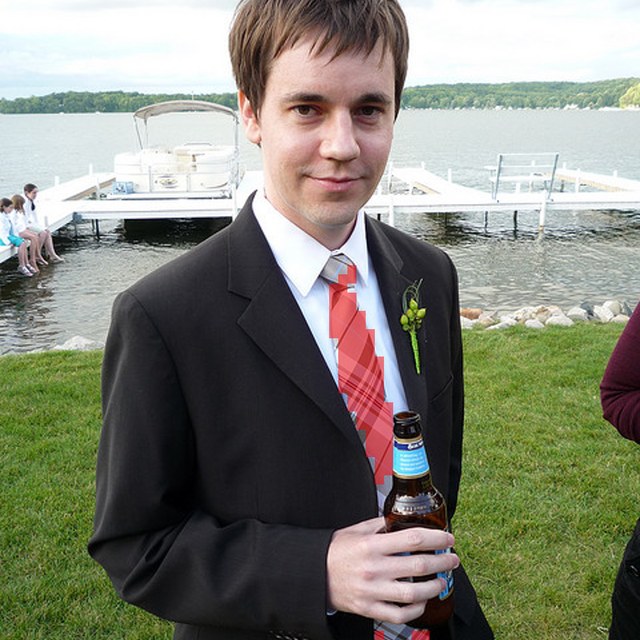} \\
    \noalign{\vspace{1pt}}

    \qimgfour{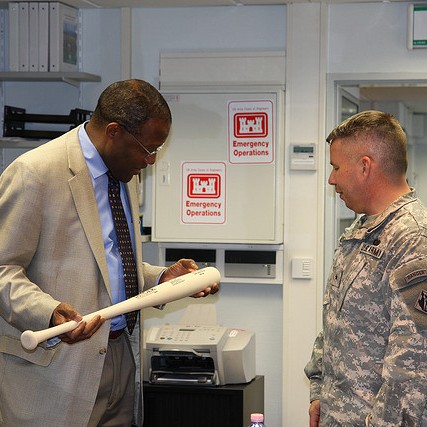} &
    \qimgfour{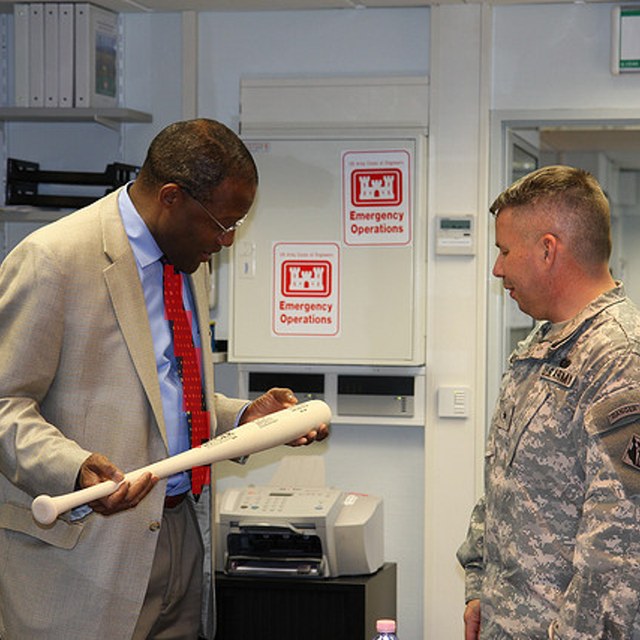} &
    \qimgfour{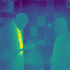} &
    \qimgfour{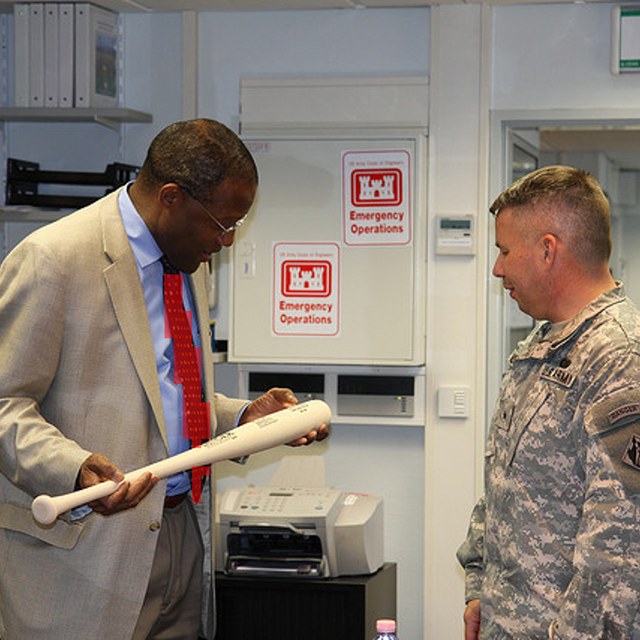} \\
    \noalign{\vspace{1pt}}

    \qimgfour{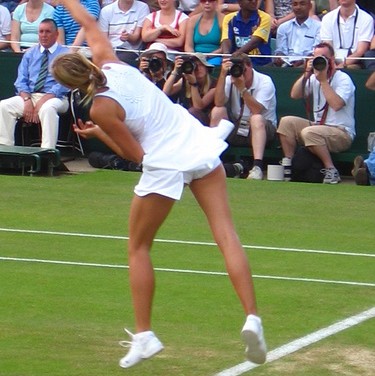} &
    \qimgfour{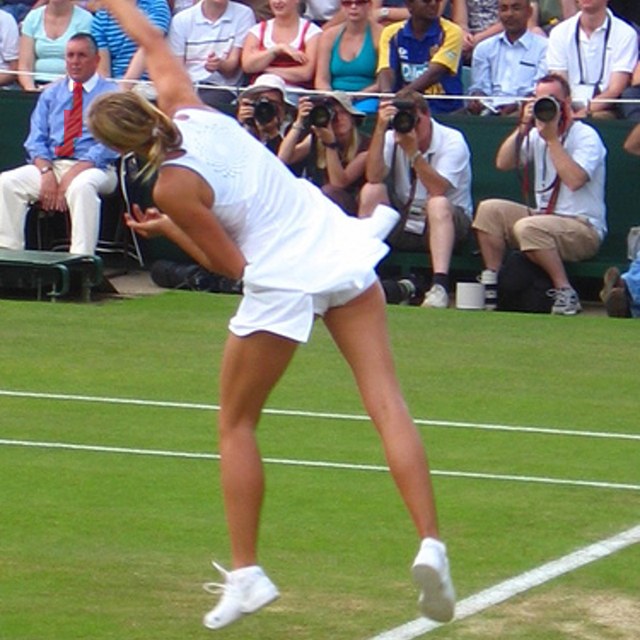} &
    \qimgfour{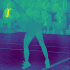} &
    \qimgfour{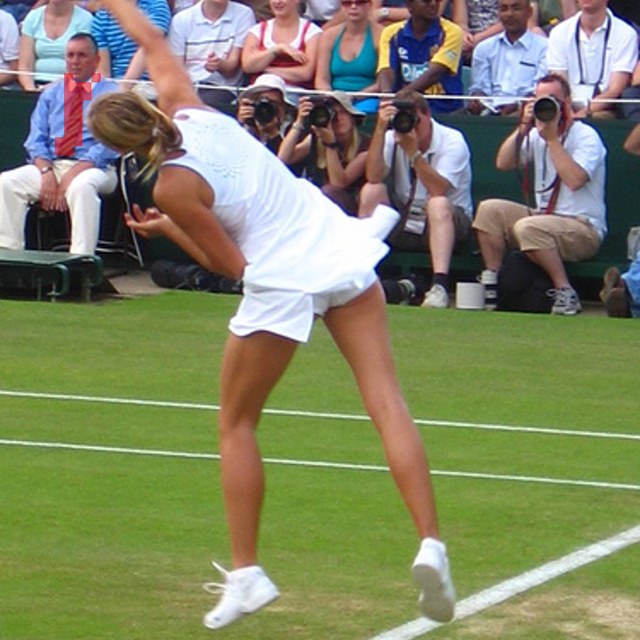} \\
    \noalign{\vspace{1pt}}

    \qimgfour{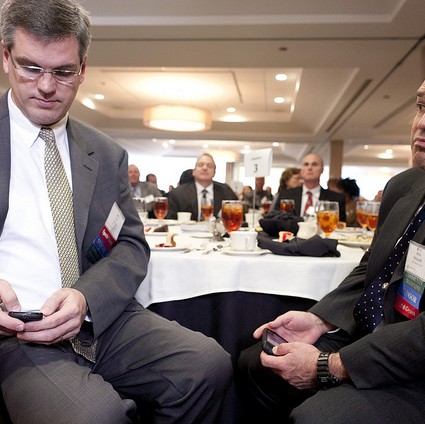} &
    \qimgfour{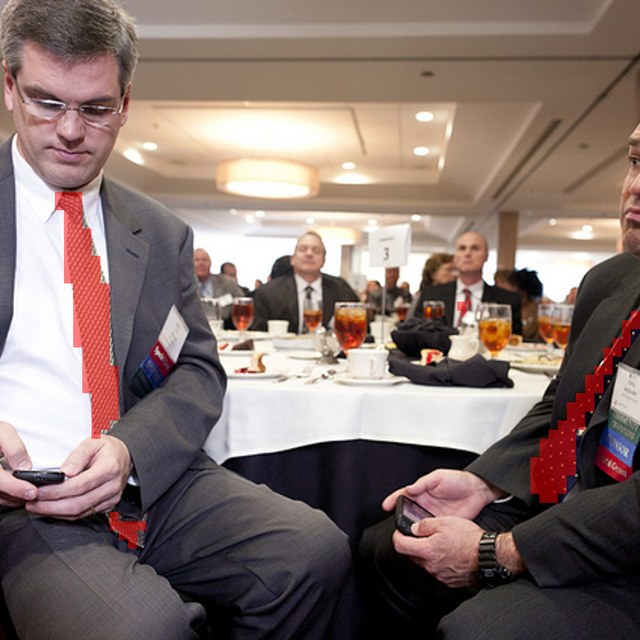} &
    \qimgfour{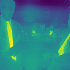} &
    \qimgfour{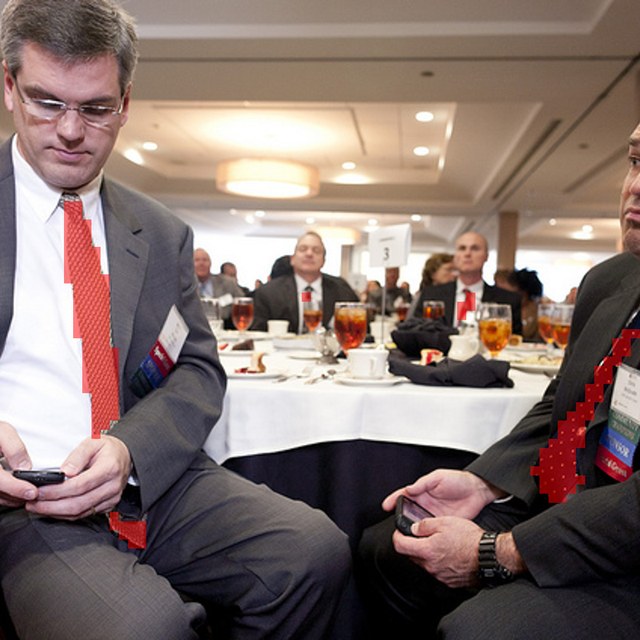} \\
    \noalign{\vspace{1pt}}

    \qimgfour{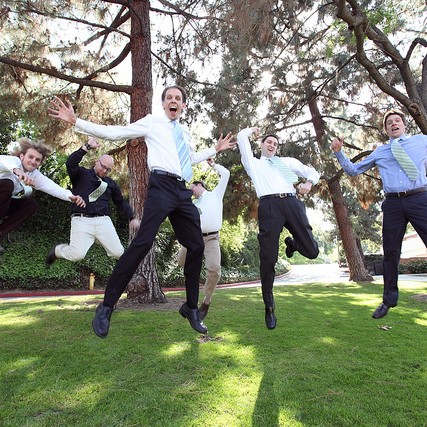} &
    \qimgfour{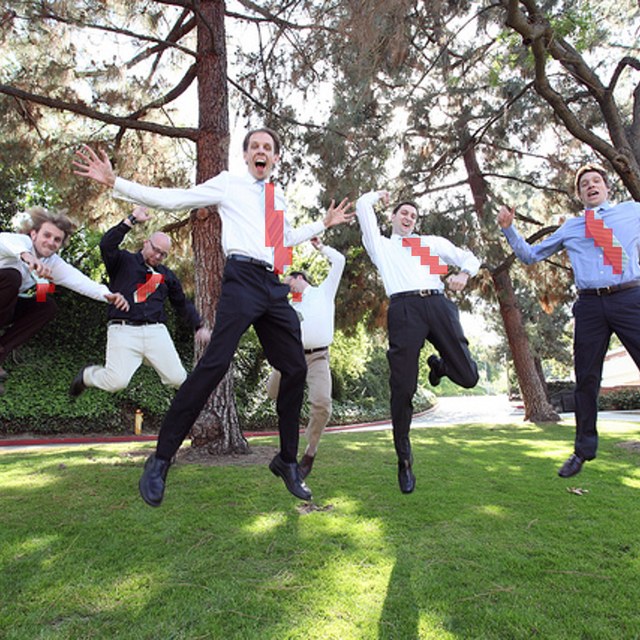} &
    \qimgfour{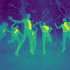} &
    \qimgfour{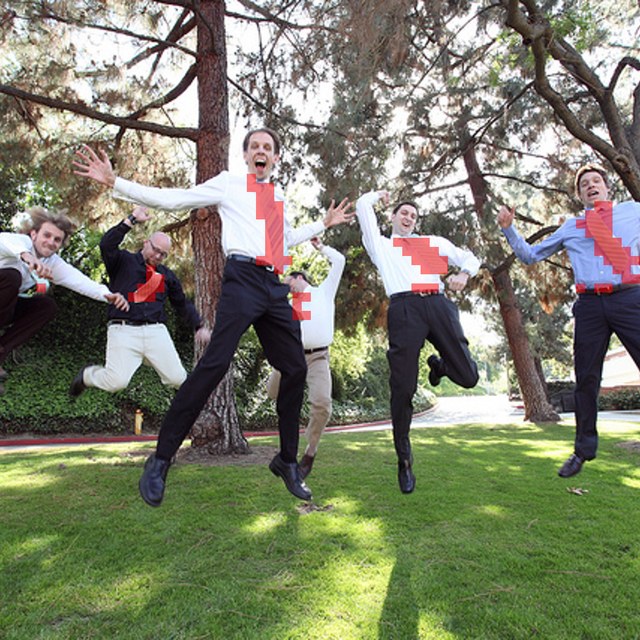} \\
    \noalign{\vspace{1pt}}

    \qimgfour{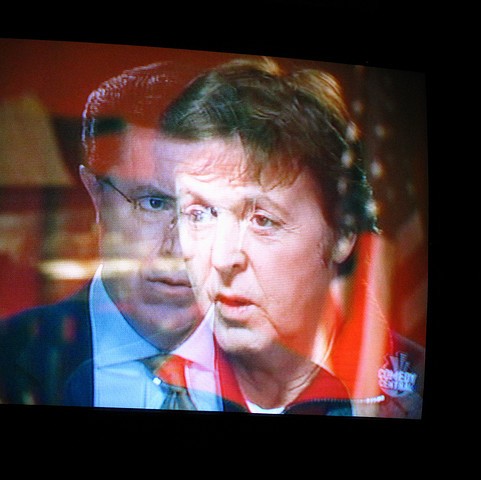} &
    \qimgfour{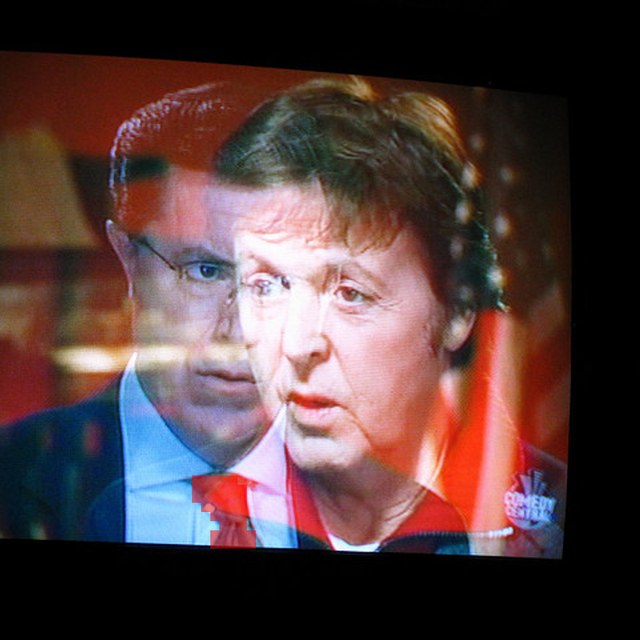} &
    \qimgfour{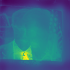} &
    \qimgfour{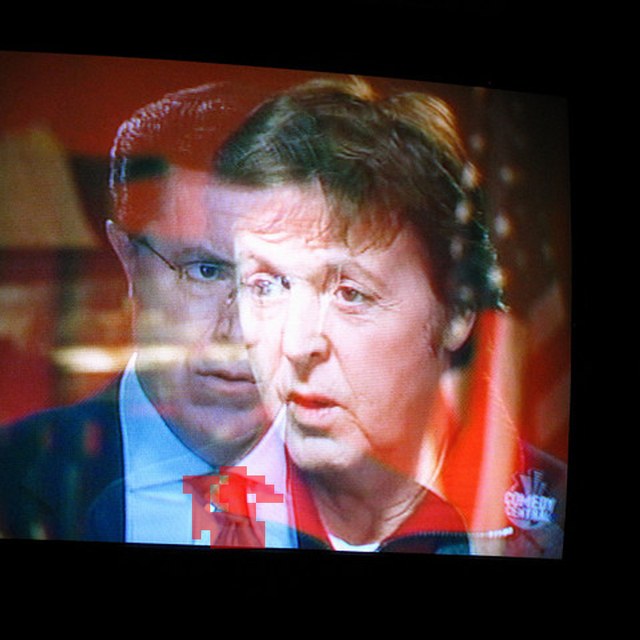} \\
    \noalign{\vspace{1pt}}

    \qimgfour{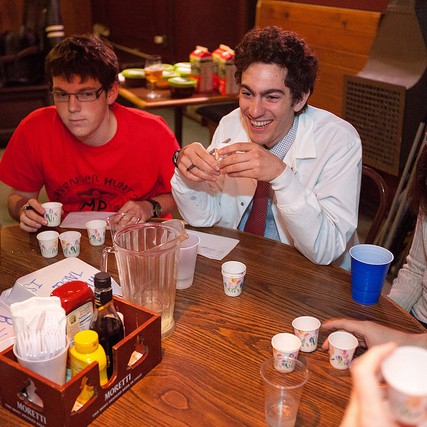} &
    \qimgfour{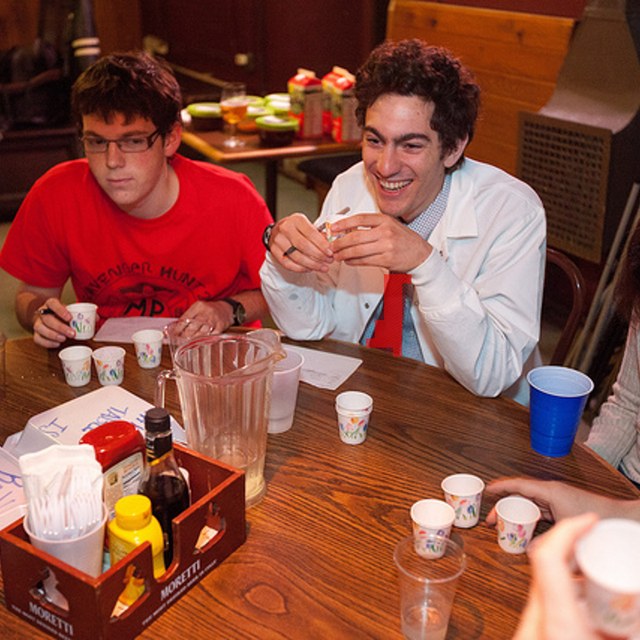} &
    \qimgfour{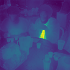} &
    \qimgfour{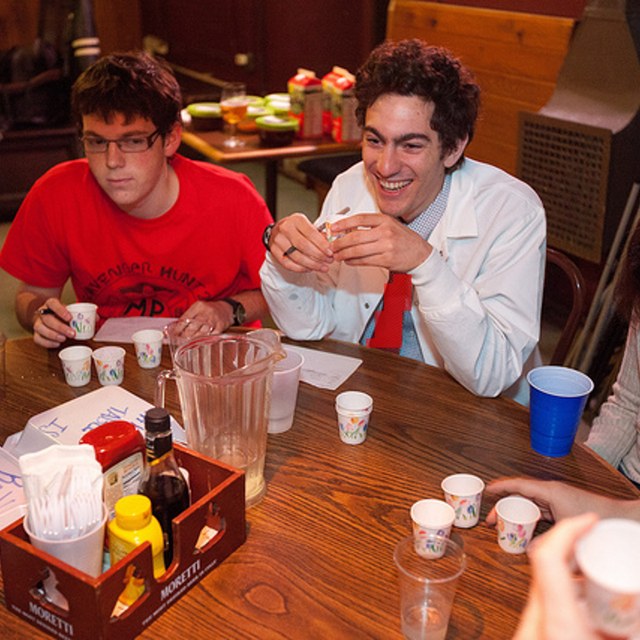} \\
    \noalign{\vspace{2pt}}

    \raisebox{-0.5\height}{\small Input} &
    \raisebox{-0.5\height}{\small GT} &
    \raisebox{-0.5\height}{\small\shortstack{PANC\\Eigen Attn.}} &
    \raisebox{-0.5\height}{\small\shortstack{PANC\\Mask}} \\
\end{tabular}
\end{adjustbox}
\endgroup
\caption{Additional qualitative comparison on the non-rigid MS COCO \textbf{Tie} class.}
\label{fig:non_rigid_tie}
\end{figure*}

\begin{figure*}[p]
\centering
\begingroup
\setlength{\tabcolsep}{0pt}
\begin{adjustbox}{max width=\textwidth,max totalheight=0.95\textheight,center}
\begin{tabular}{@{}c @{\hspace{2pt}} c @{\hspace{2pt}} c @{\hspace{2pt}} c@{}}
    \qimgfour{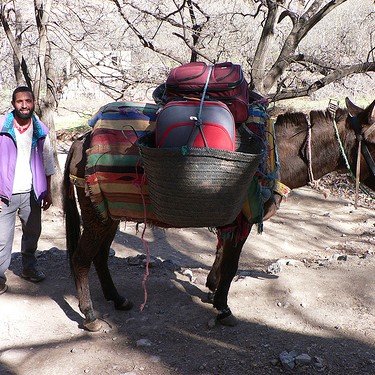} &
    \qimgfour{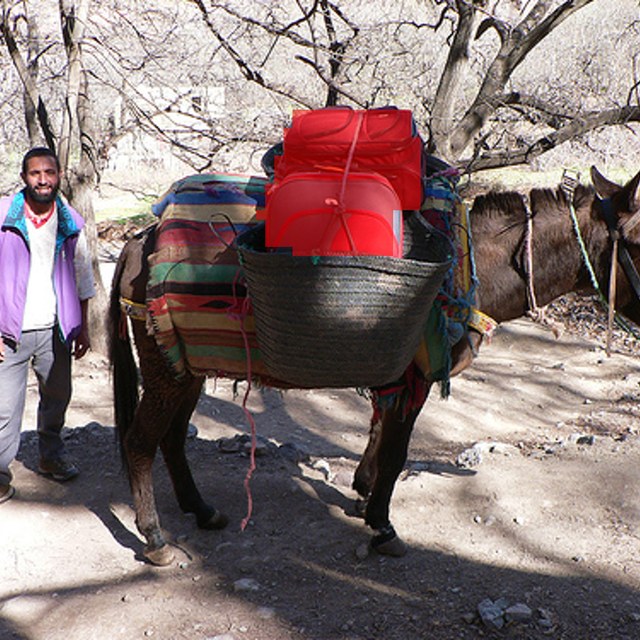} &
    \qimgfour{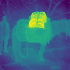} &
    \qimgfour{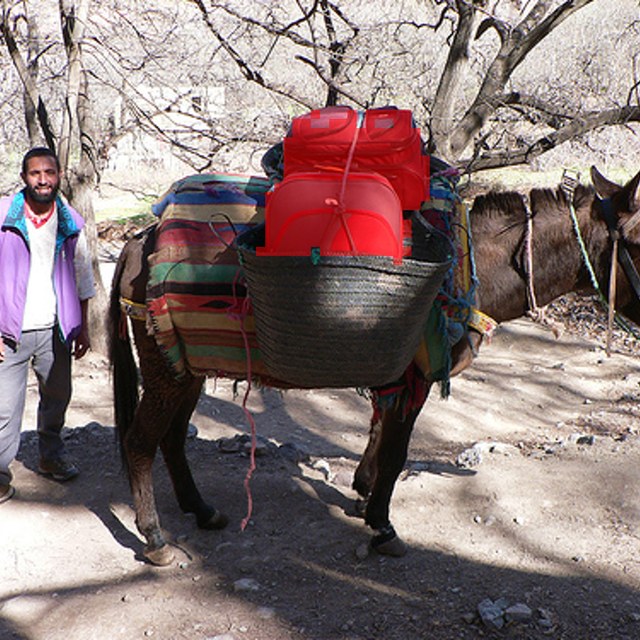} \\
    \noalign{\vspace{1pt}}

    \qimgfour{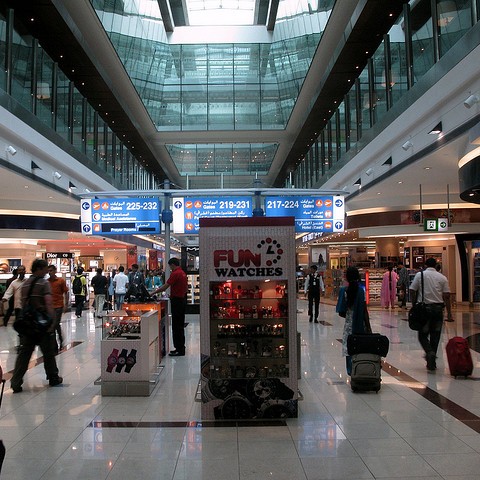} &
    \qimgfour{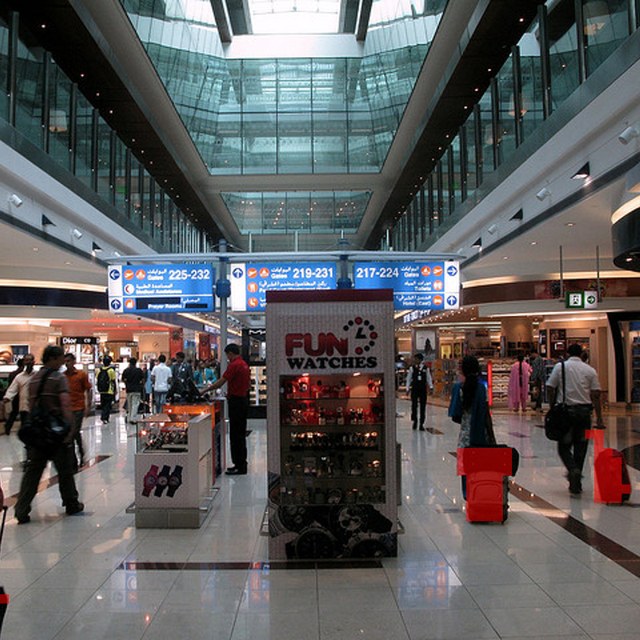} &
    \qimgfour{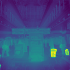} &
    \qimgfour{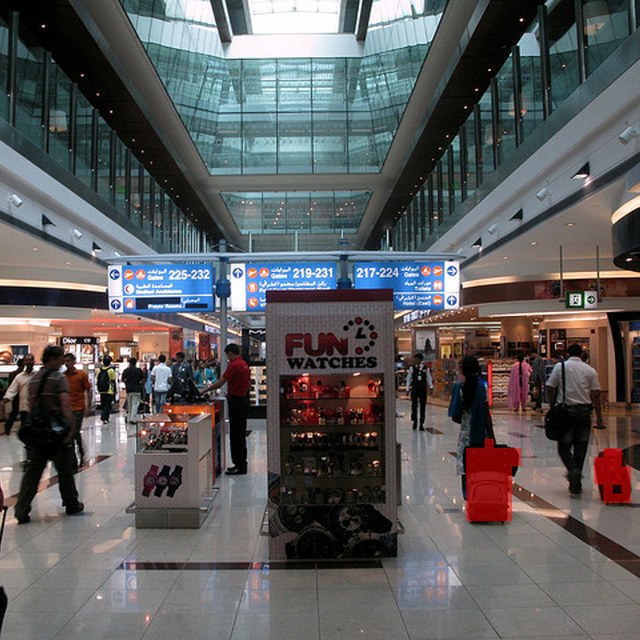} \\
    \noalign{\vspace{1pt}}

    \qimgfour{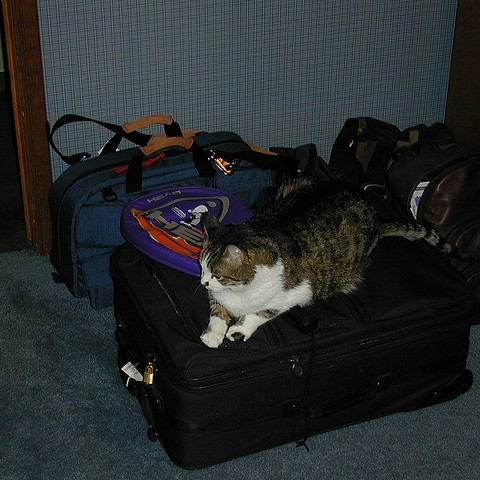} &
    \qimgfour{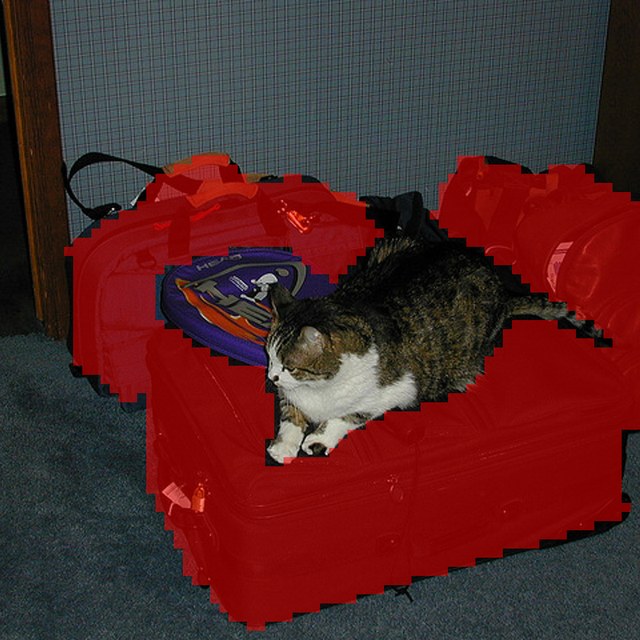} &
    \qimgfour{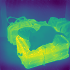} &
    \qimgfour{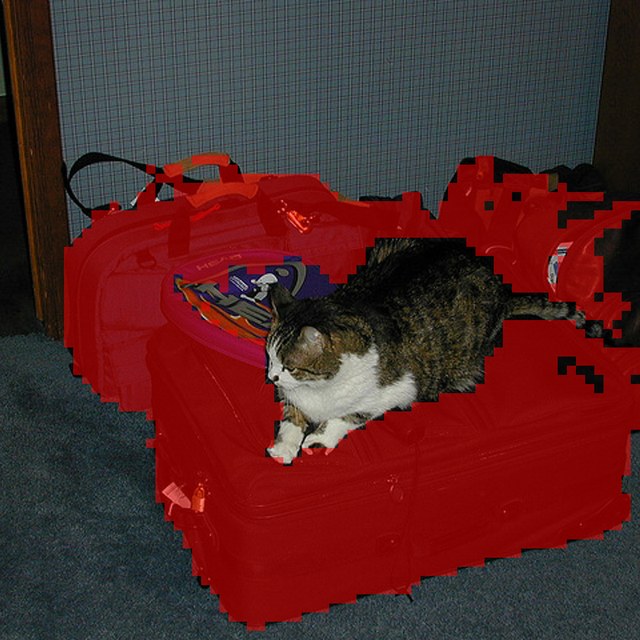} \\
    \noalign{\vspace{1pt}}

    \qimgfour{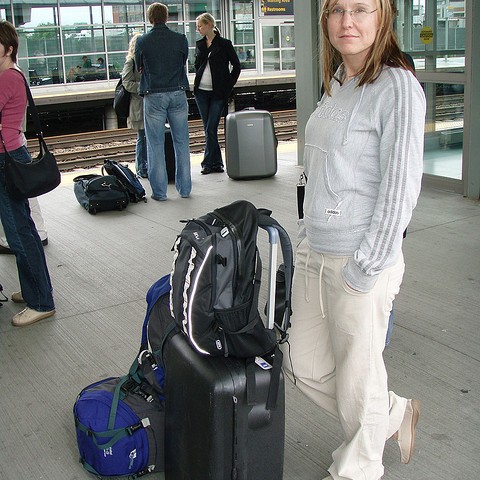} &
    \qimgfour{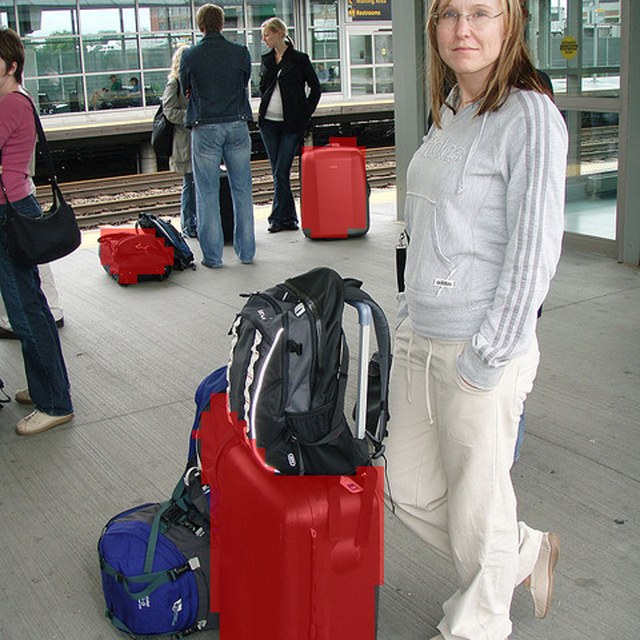} &
    \qimgfour{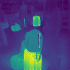} &
    \qimgfour{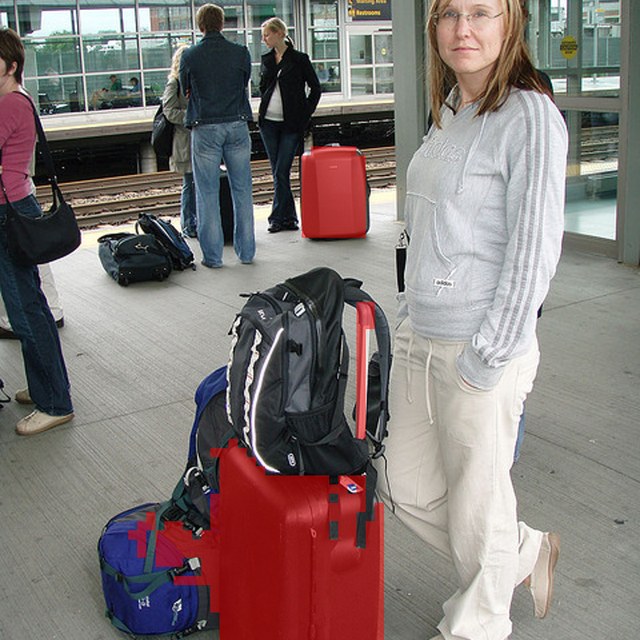} \\
    \noalign{\vspace{1pt}}

    \qimgfour{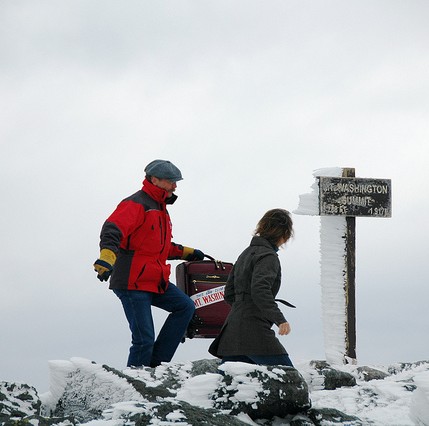} &
    \qimgfour{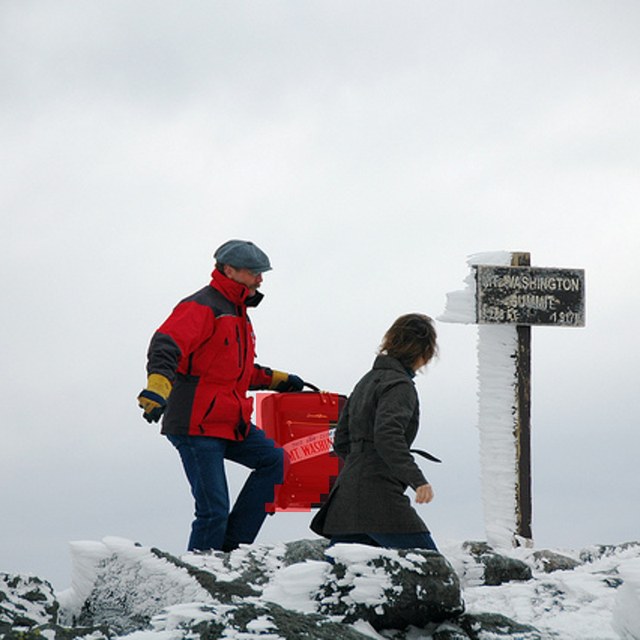} &
    \qimgfour{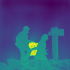} &
    \qimgfour{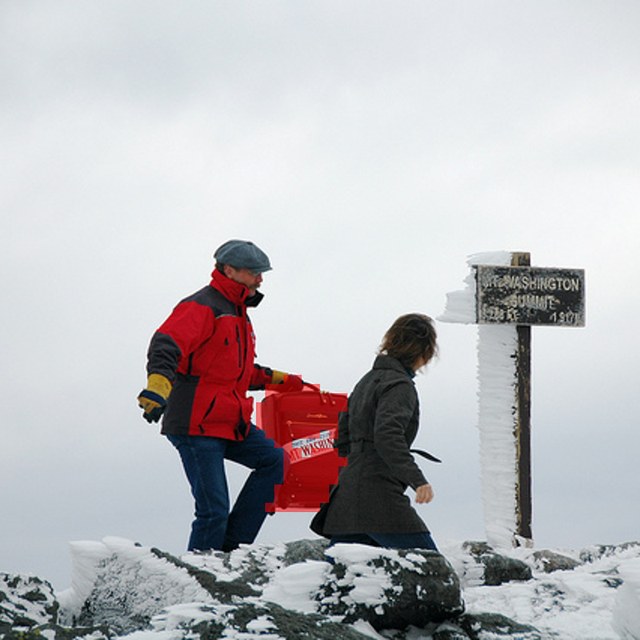} \\
    \noalign{\vspace{1pt}}

    \qimgfour{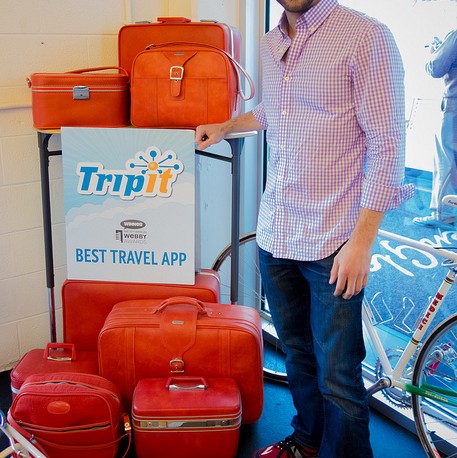} &
    \qimgfour{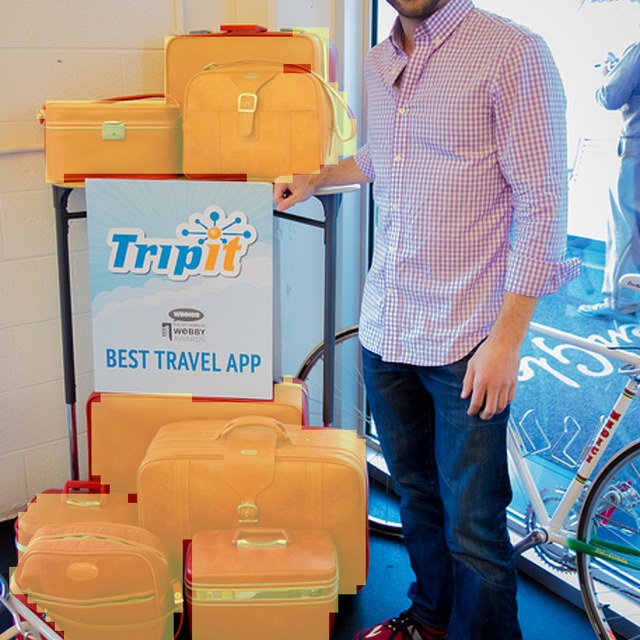} &
    \qimgfour{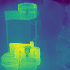} &
    \qimgfour{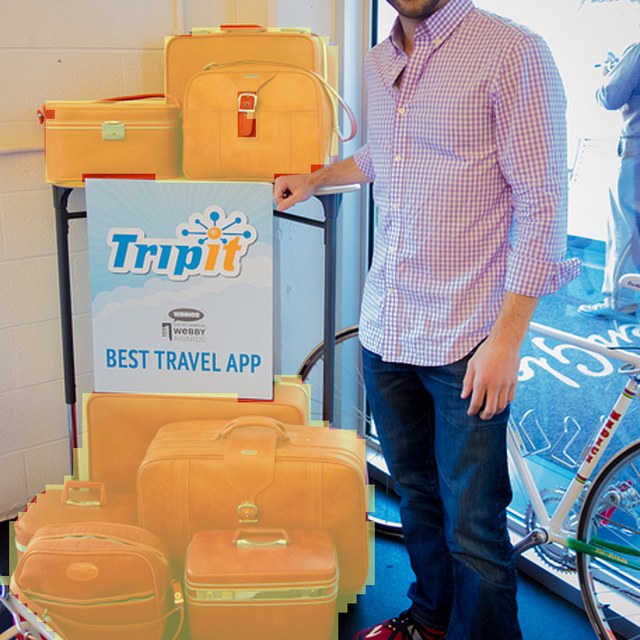} \\
    \noalign{\vspace{1pt}}

    \qimgfour{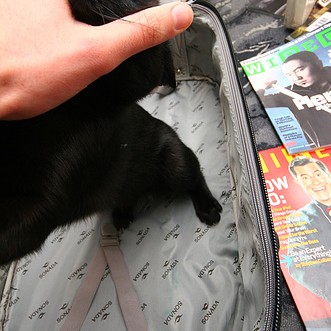} &
    \qimgfour{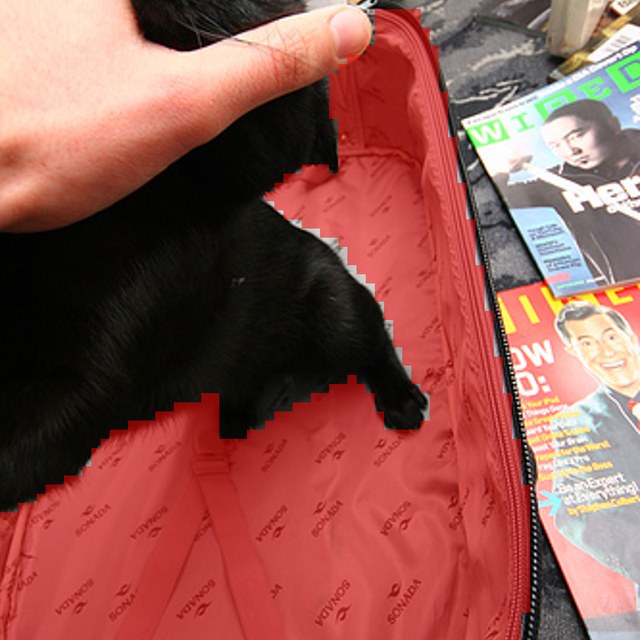} &
    \qimgfour{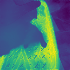} &
    \qimgfour{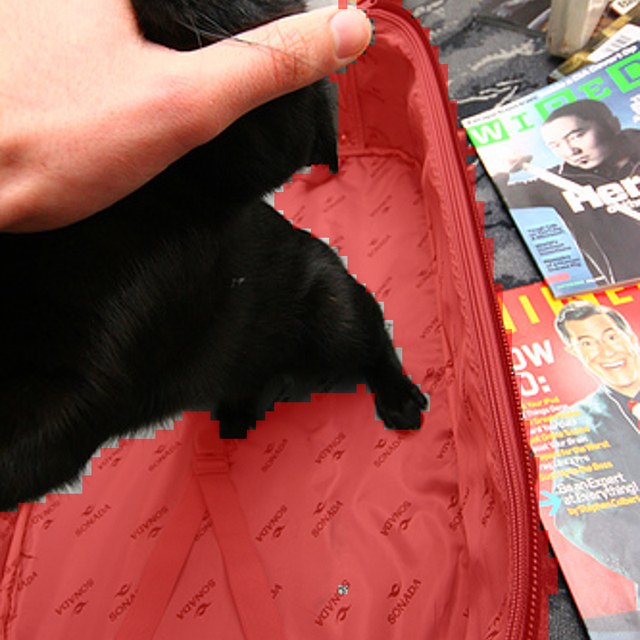} \\
    \noalign{\vspace{2pt}}

    \raisebox{-0.5\height}{\small Input} &
    \raisebox{-0.5\height}{\small GT} &
    \raisebox{-0.5\height}{\small\shortstack{PANC\\Eigen Attn.}} &
    \raisebox{-0.5\height}{\small\shortstack{PANC\\Mask}} \\
\end{tabular}
\end{adjustbox}
\endgroup
\caption{Additional qualitative comparison on the non-rigid MS COCO \textbf{Suitcase} class.}
\label{fig:non_rigid_suitcase}
\end{figure*}

\begin{figure*}[p]
\centering
\begingroup
\setlength{\tabcolsep}{0pt}
\begin{adjustbox}{max width=\textwidth,max totalheight=1.0\textheight,center}
\begin{tabular}{@{}c @{\hspace{2pt}} c @{\hspace{2pt}} c @{\hspace{2pt}} c@{}}
  \raisebox{-0.5\height}{\includegraphics[width=0.24\linewidth]{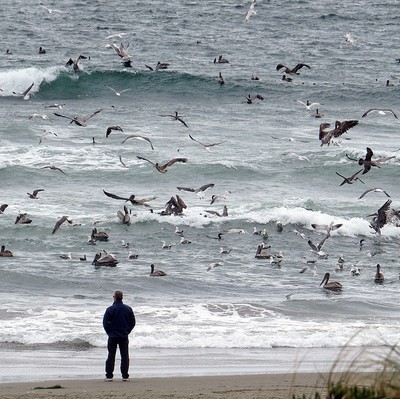}} &
  \raisebox{-0.5\height}{\includegraphics[width=0.24\linewidth]{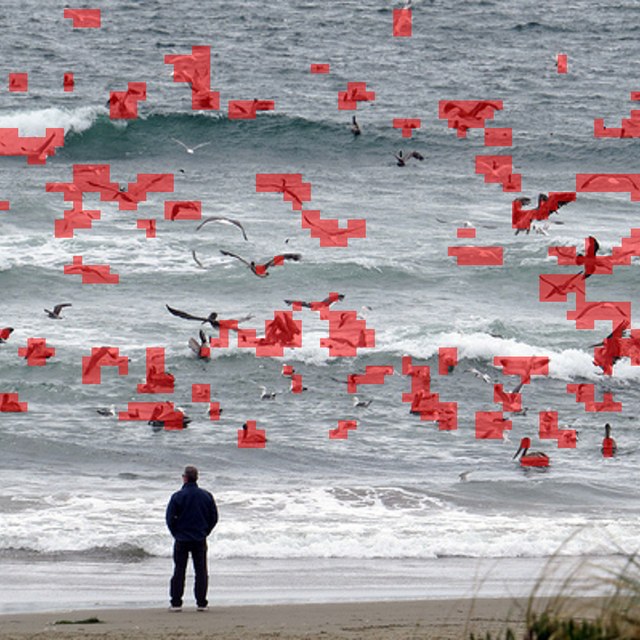}} &
  \raisebox{-0.5\height}{\includegraphics[width=0.24\linewidth]{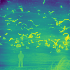}} &
  \raisebox{-0.5\height}{\includegraphics[width=0.24\linewidth]{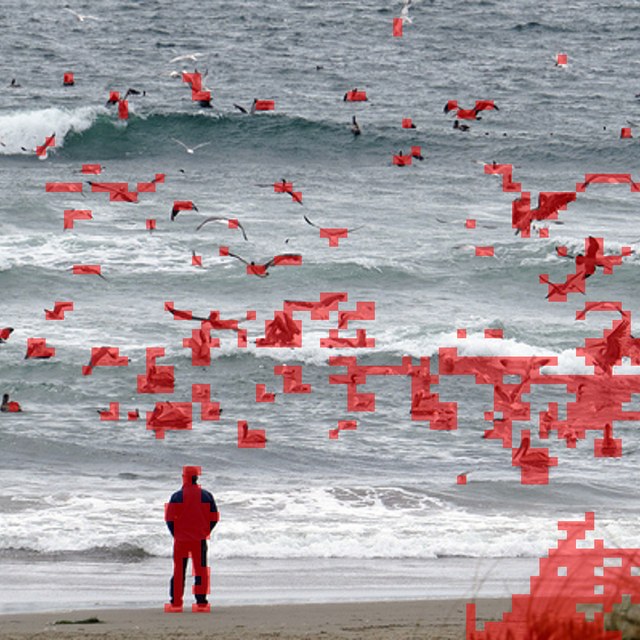}} \\
  \noalign{\vspace{1pt}}

  \raisebox{-0.5\height}{\includegraphics[width=0.24\linewidth]{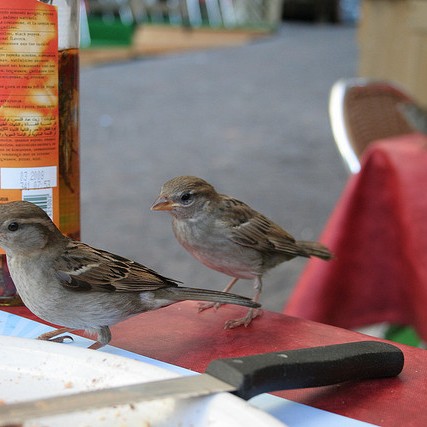}} &
  \raisebox{-0.5\height}{\includegraphics[width=0.24\linewidth]{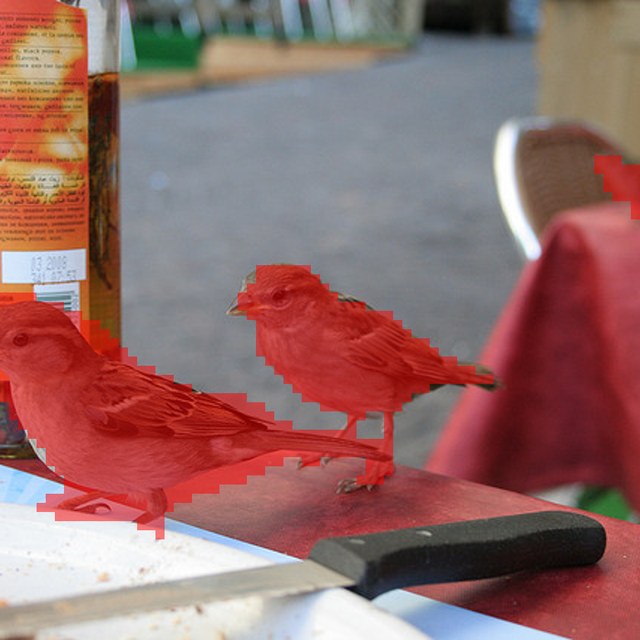}} &
  \raisebox{-0.5\height}{\includegraphics[width=0.24\linewidth]{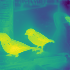}} &
  \raisebox{-0.5\height}{\includegraphics[width=0.24\linewidth]{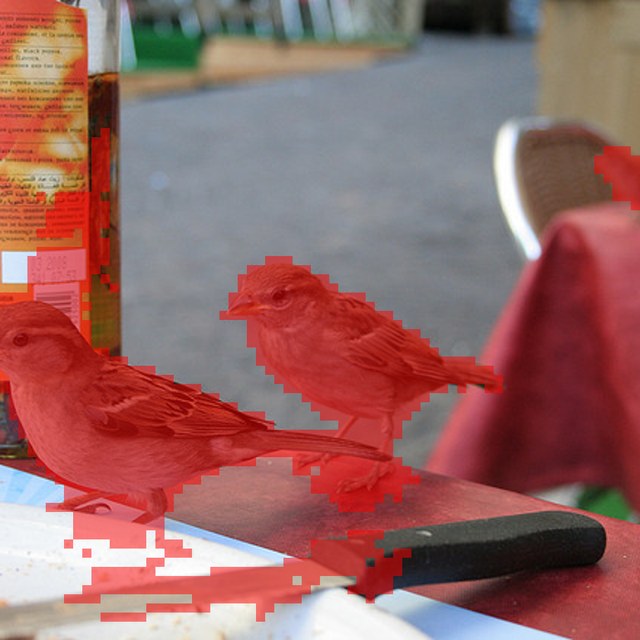}} \\
  \noalign{\vspace{2pt}}

  \raisebox{-0.5\height}{\small Input} &
  \raisebox{-0.5\height}{\small GT} &
  \raisebox{-0.5\height}{\small\shortstack{Eigen\\Attn.}} &
  \raisebox{-0.5\height}{\small Mask} \\
\end{tabular}
\end{adjustbox}
\endgroup
\caption{Transfer learning from the CUB-200-2011 dataset enables bird segmentation in the MS COCO dataset.}
\label{fig:transfer_cub_coco}
\end{figure*}

\begin{figure*}[p]
\centering
\begingroup
\setlength{\tabcolsep}{0pt}
\begin{adjustbox}{max width=\textwidth,max totalheight=1.0\textheight,center}
\begin{tabular}{@{}c @{\hspace{2pt}} c @{\hspace{2pt}} c @{\hspace{2pt}} c @{\hspace{2pt}} c @{\hspace{2pt}} c@{}}
  \qimgsix{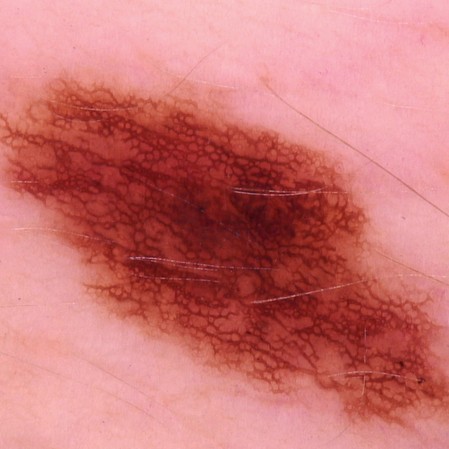} &
  \qimgsix{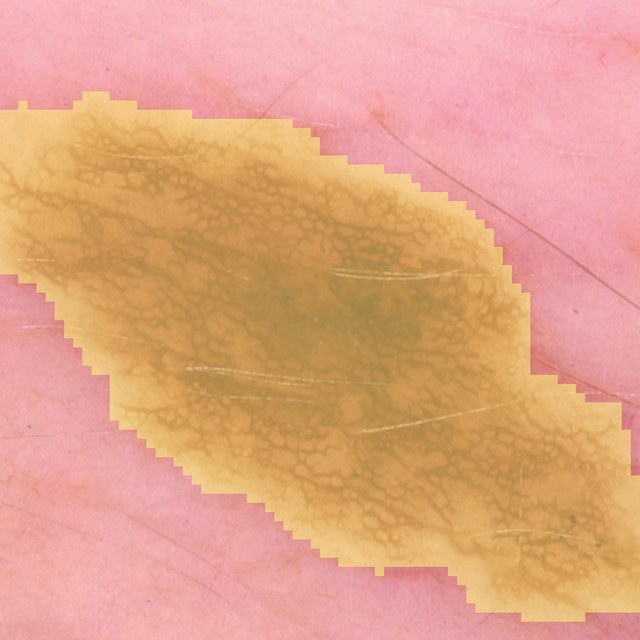} &
  \qimgsix{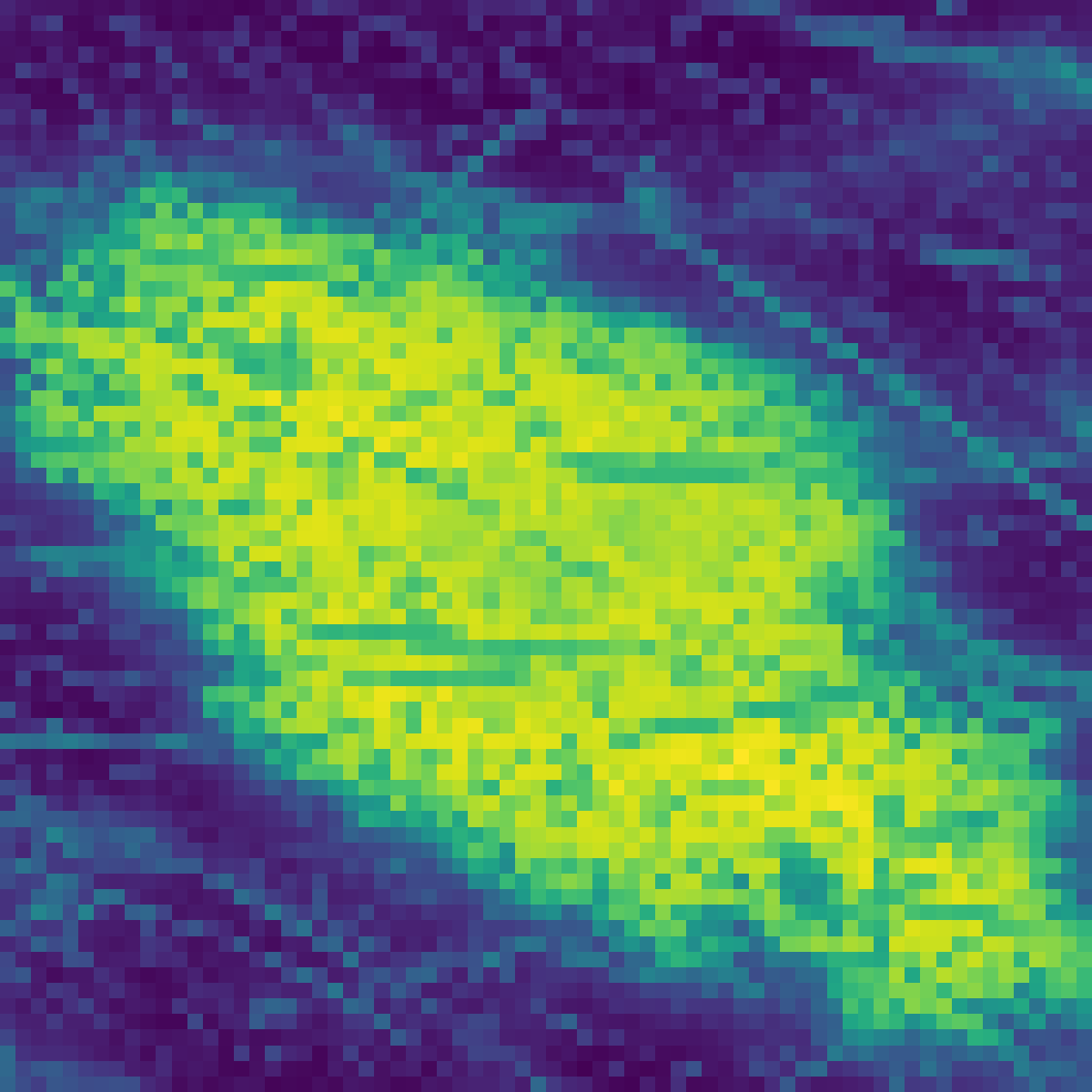} &
  \qimgsix{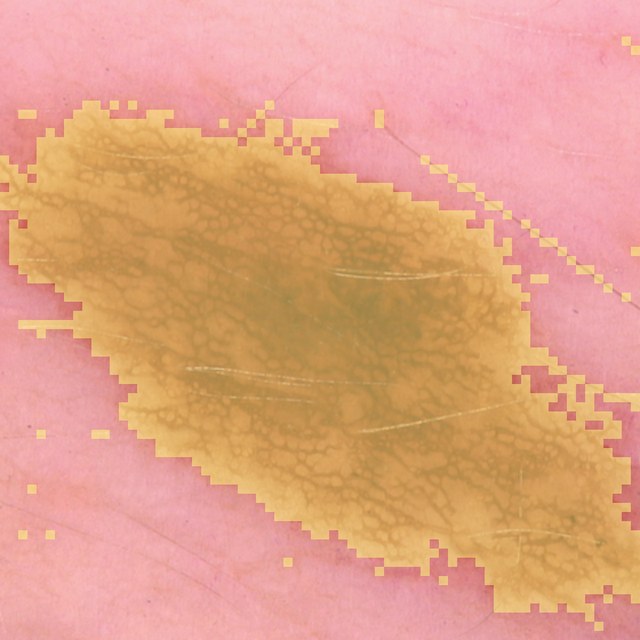} &
  \qimgsix{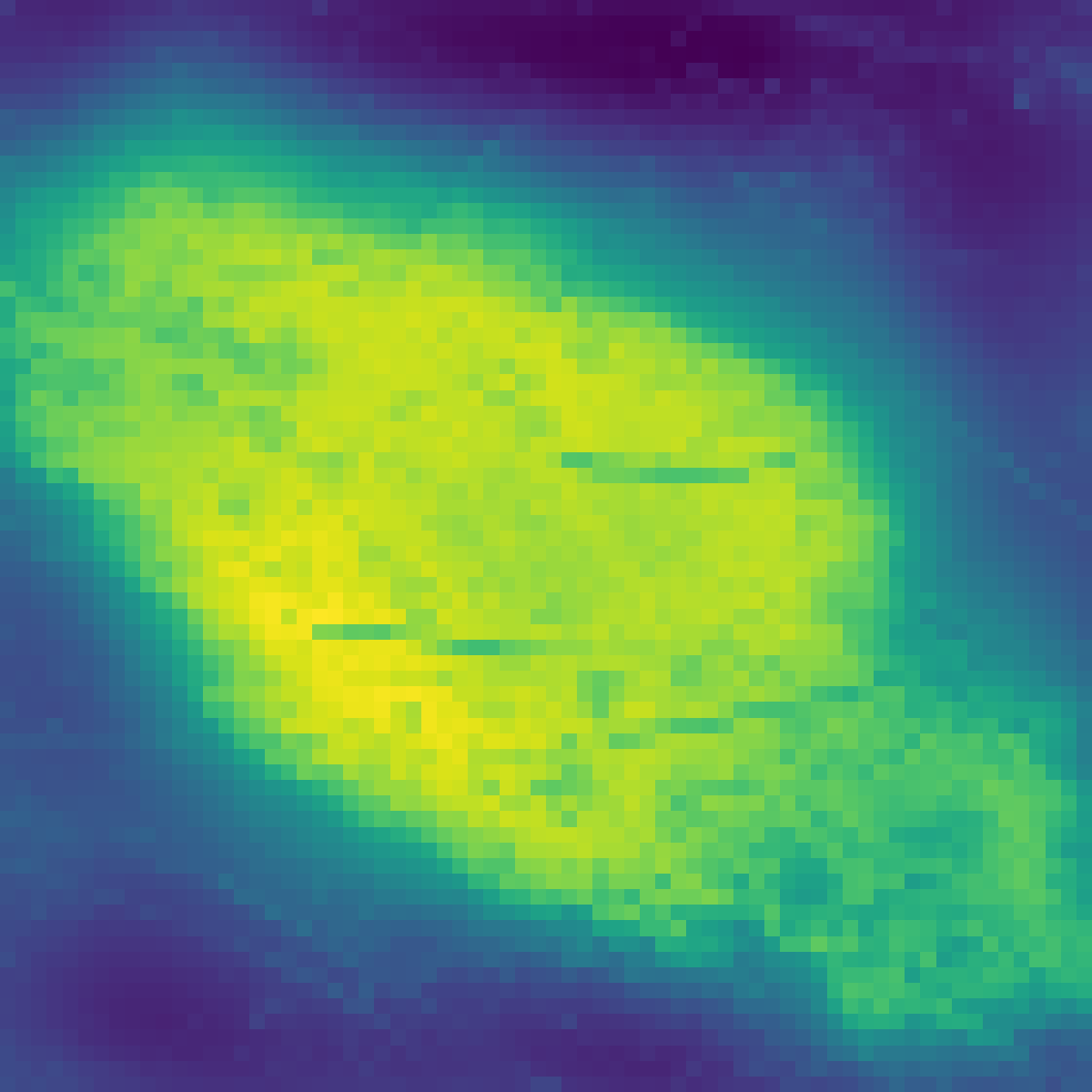} &
  \qimgsix{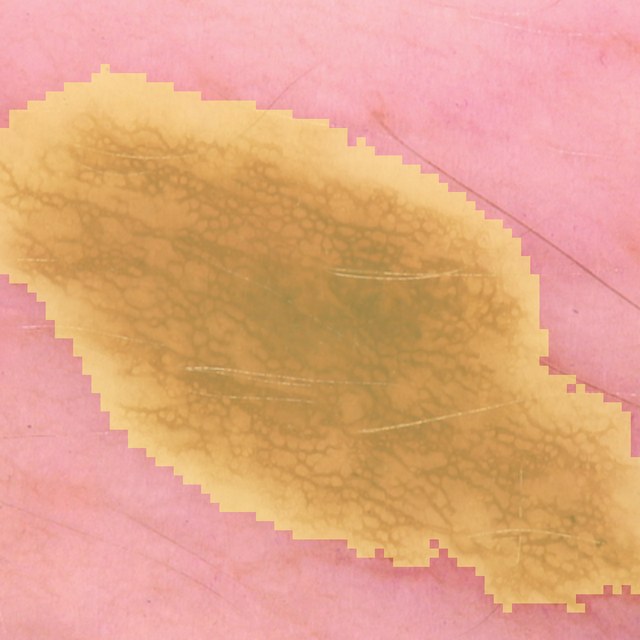} \\
  \noalign{\vspace{1pt}}

  \qimgsix{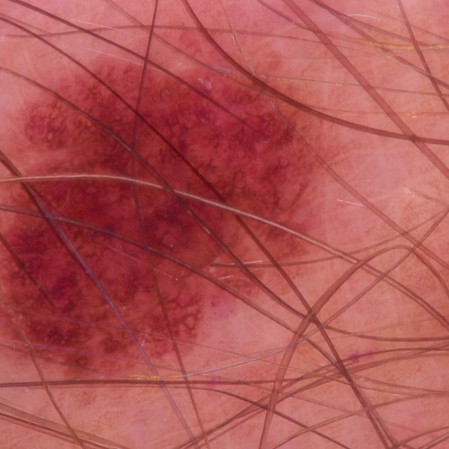} &
  \qimgsix{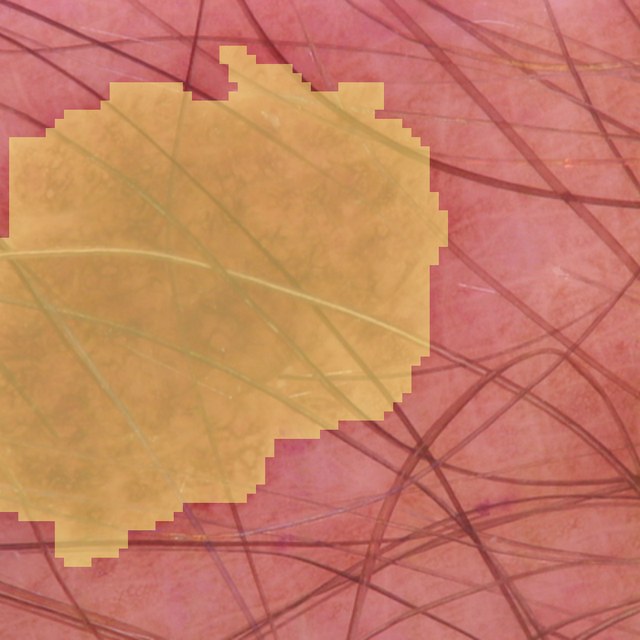} &
  \qimgsix{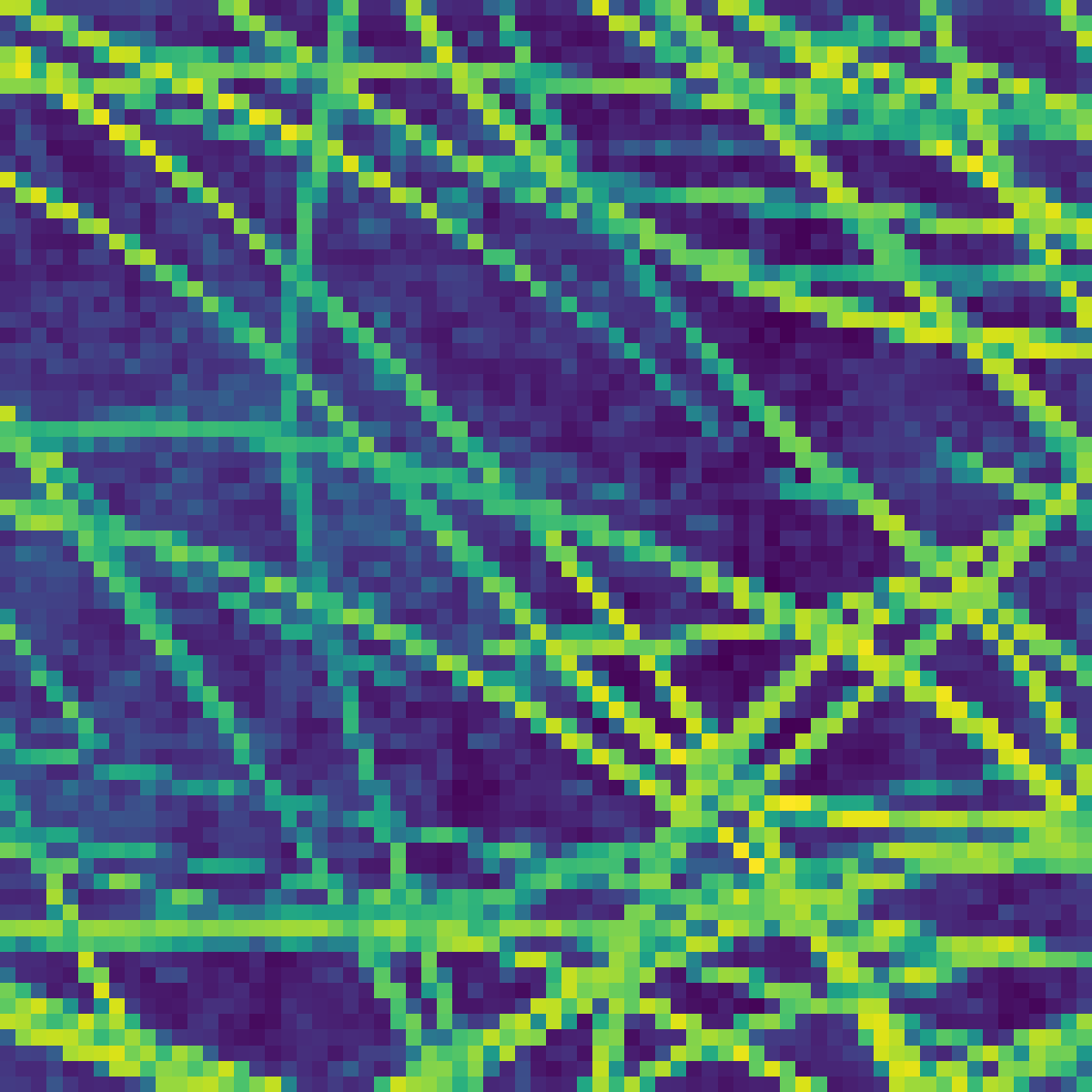} &
  \qimgsix{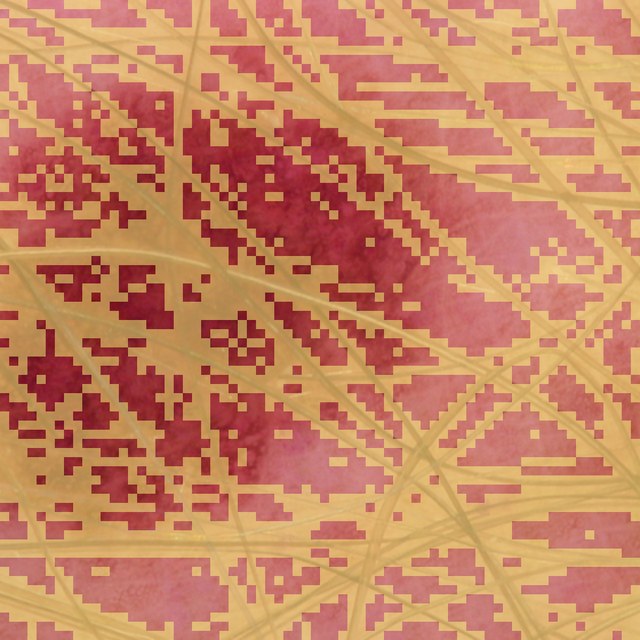} &
  \qimgsix{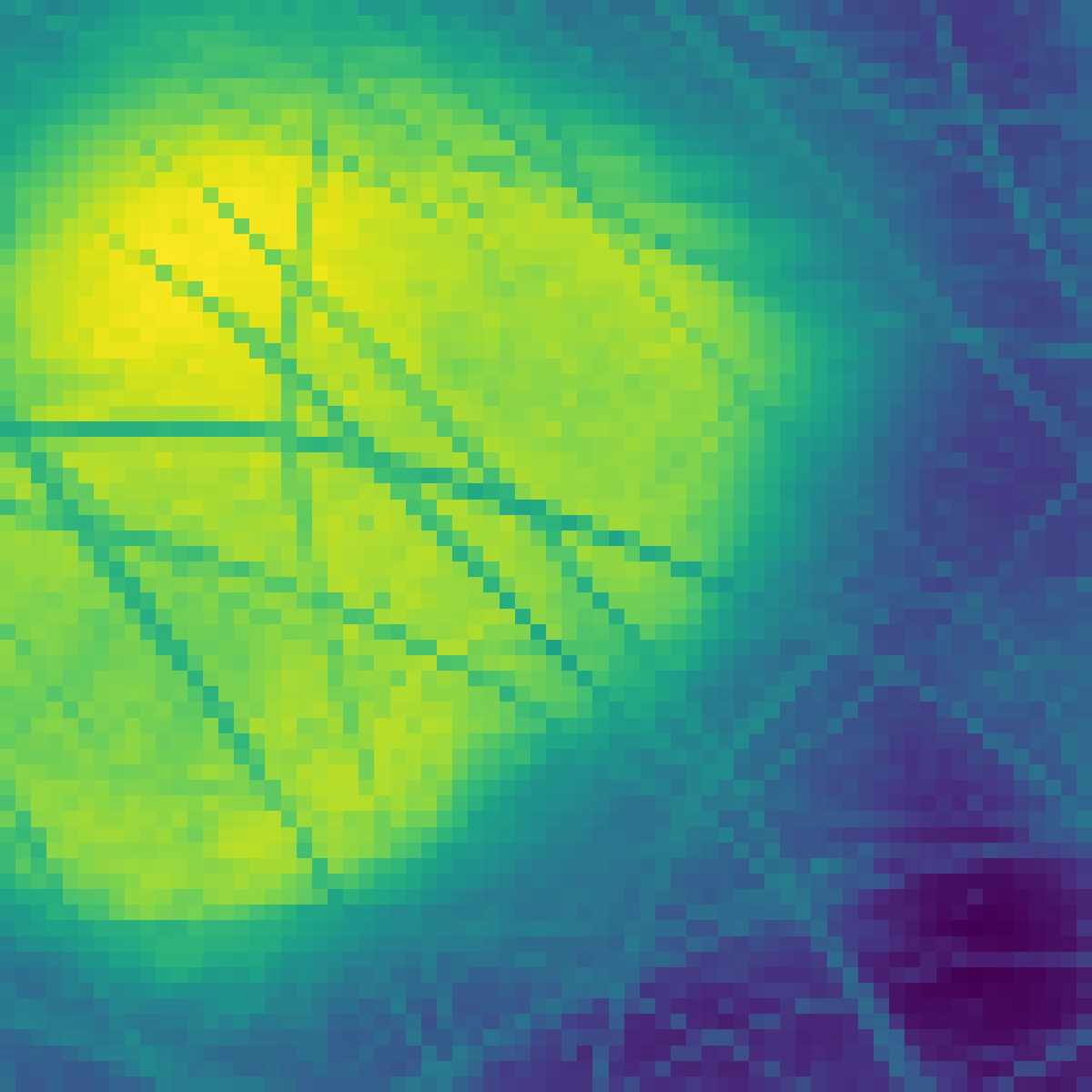} &
  \qimgsix{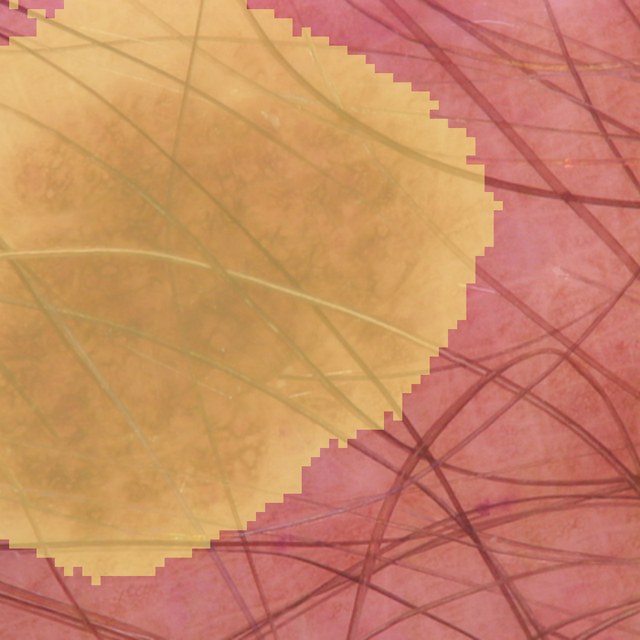} \\
  \noalign{\vspace{1pt}}

  \qimgsix{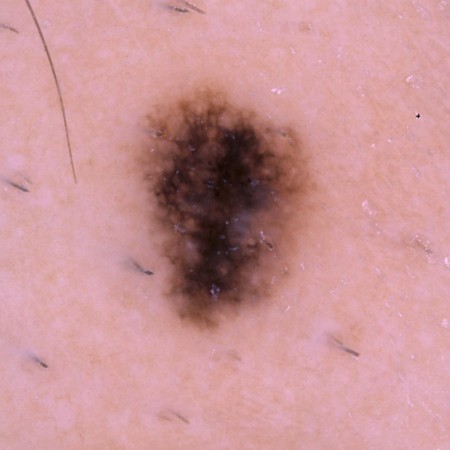} &
  \qimgsix{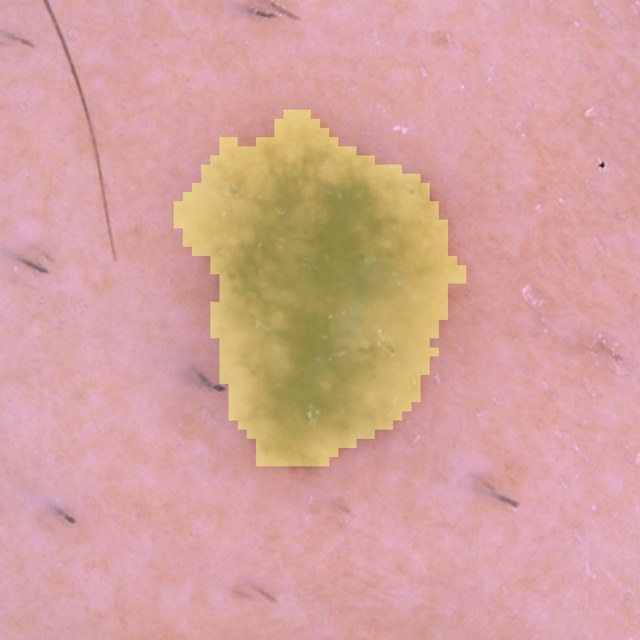} &
  \qimgsix{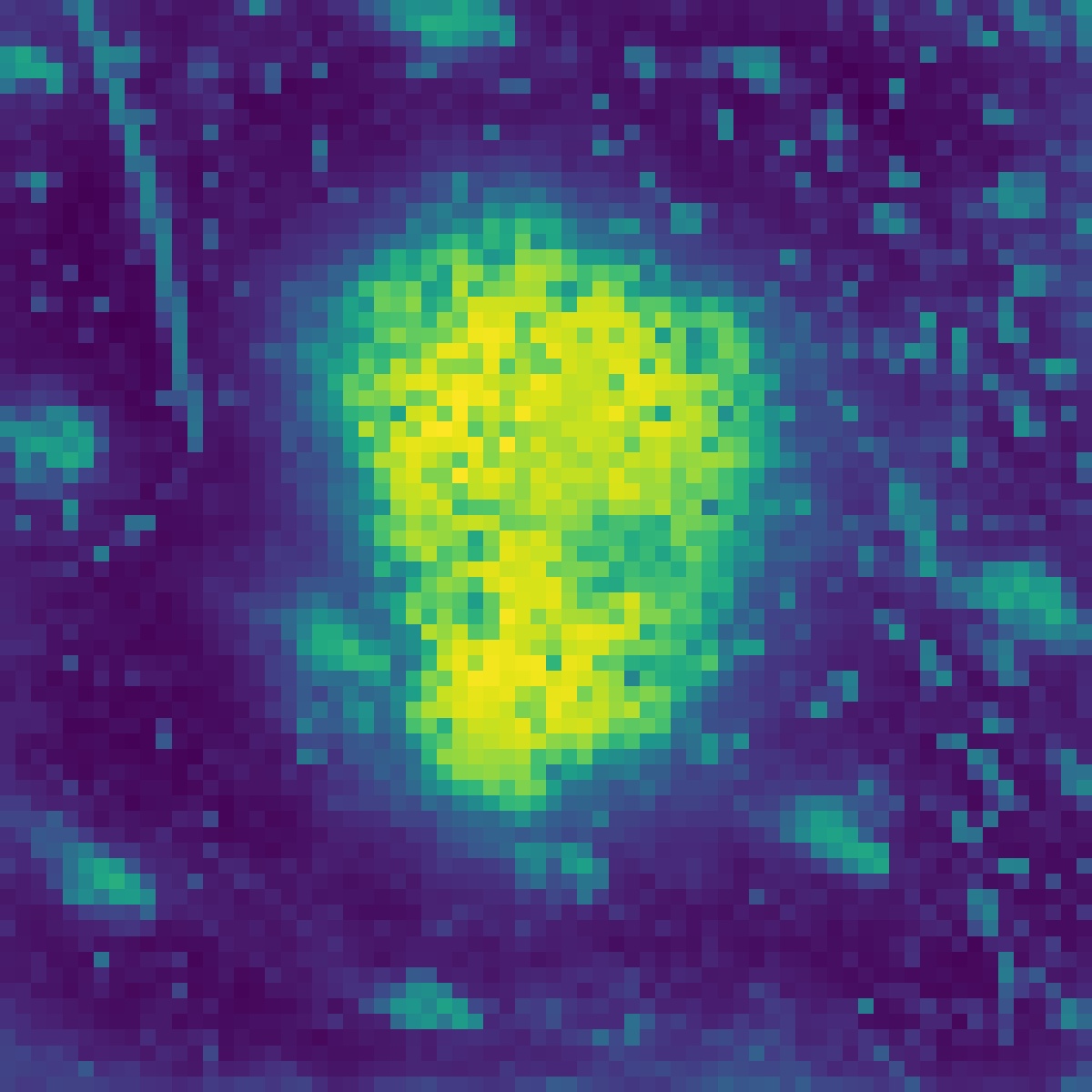} &
  \qimgsix{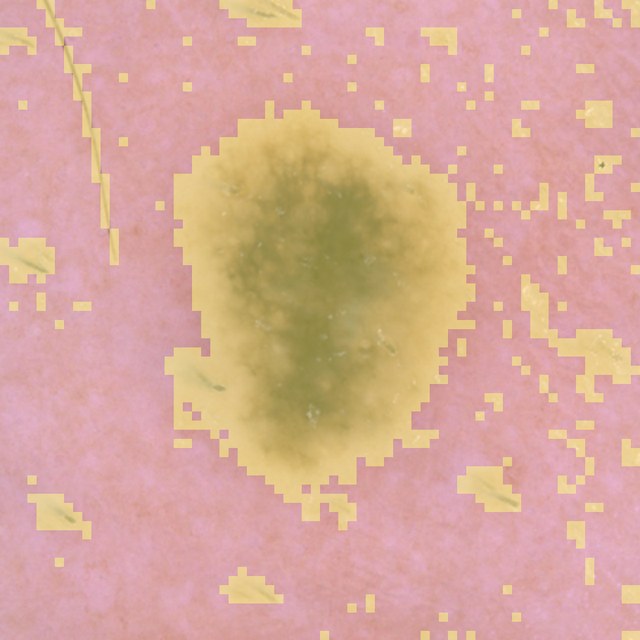} &
  \qimgsix{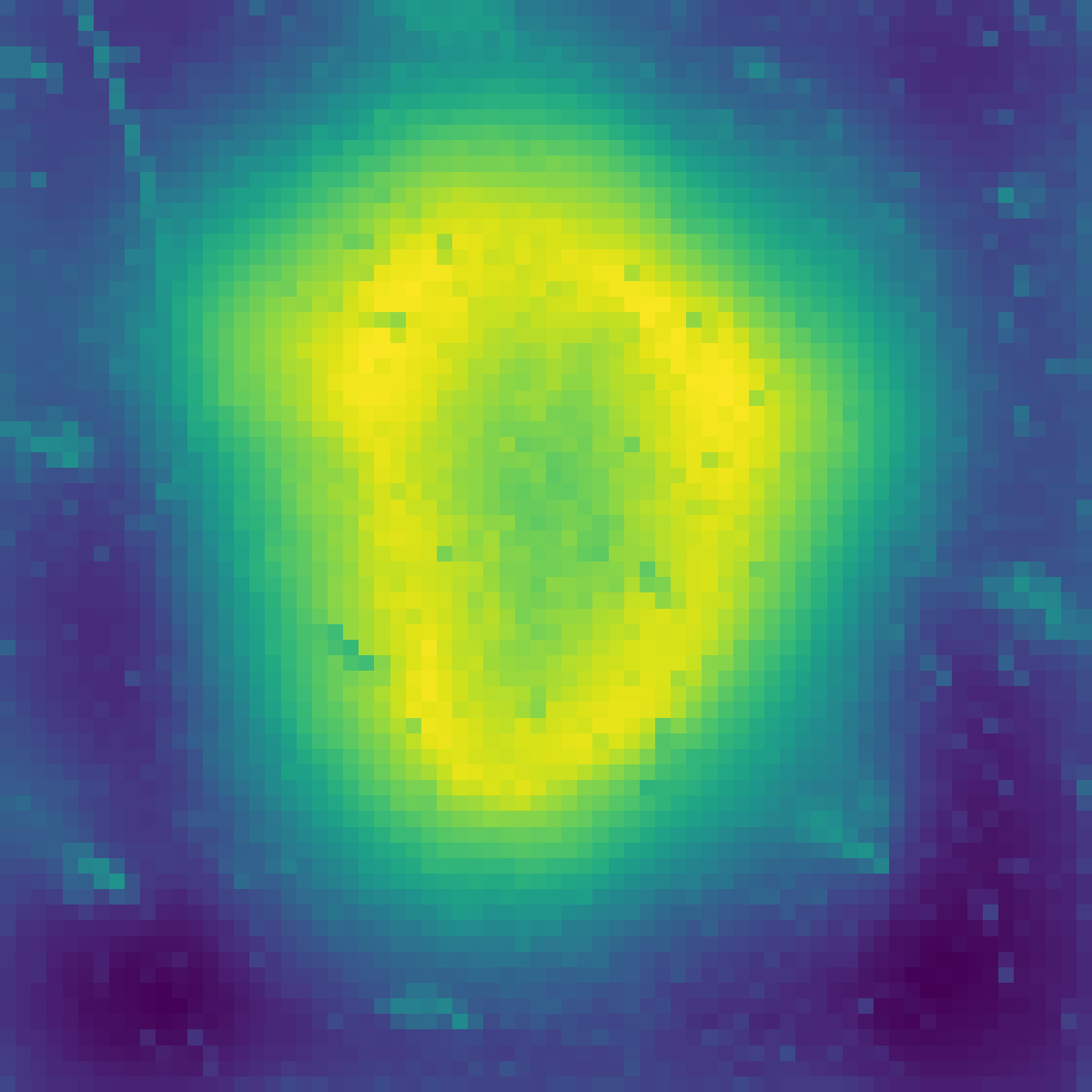} &
  \qimgsix{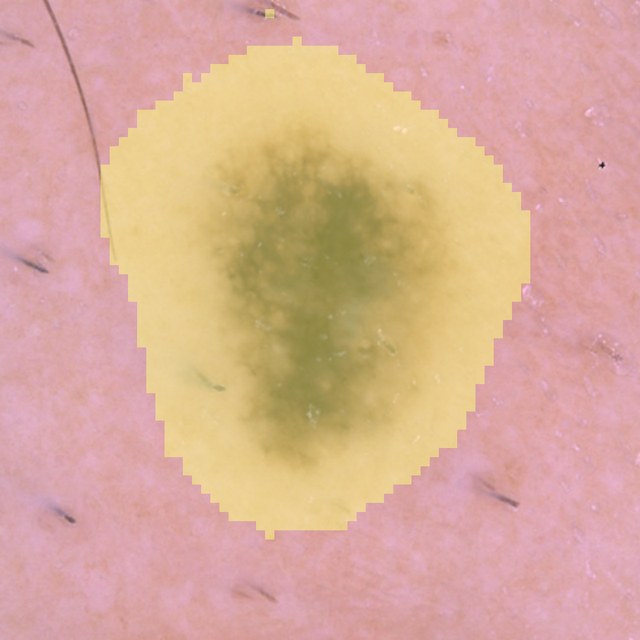} \\
  \noalign{\vspace{1pt}}

  \qimgsix{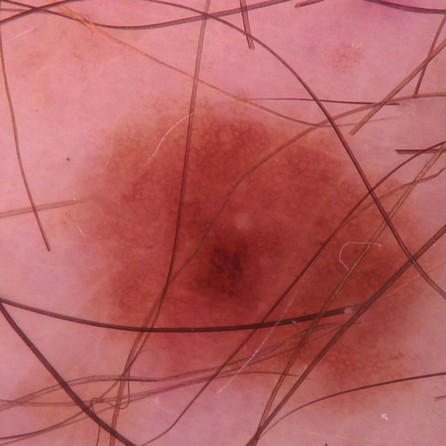} &
  \qimgsix{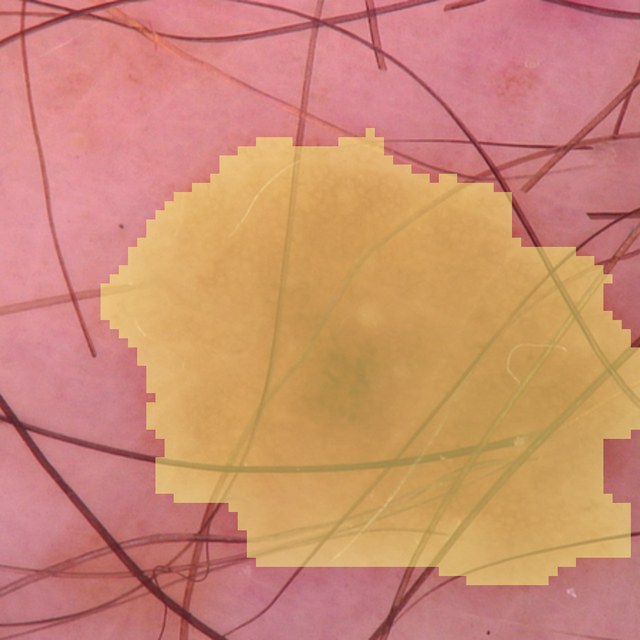} &
  \qimgsix{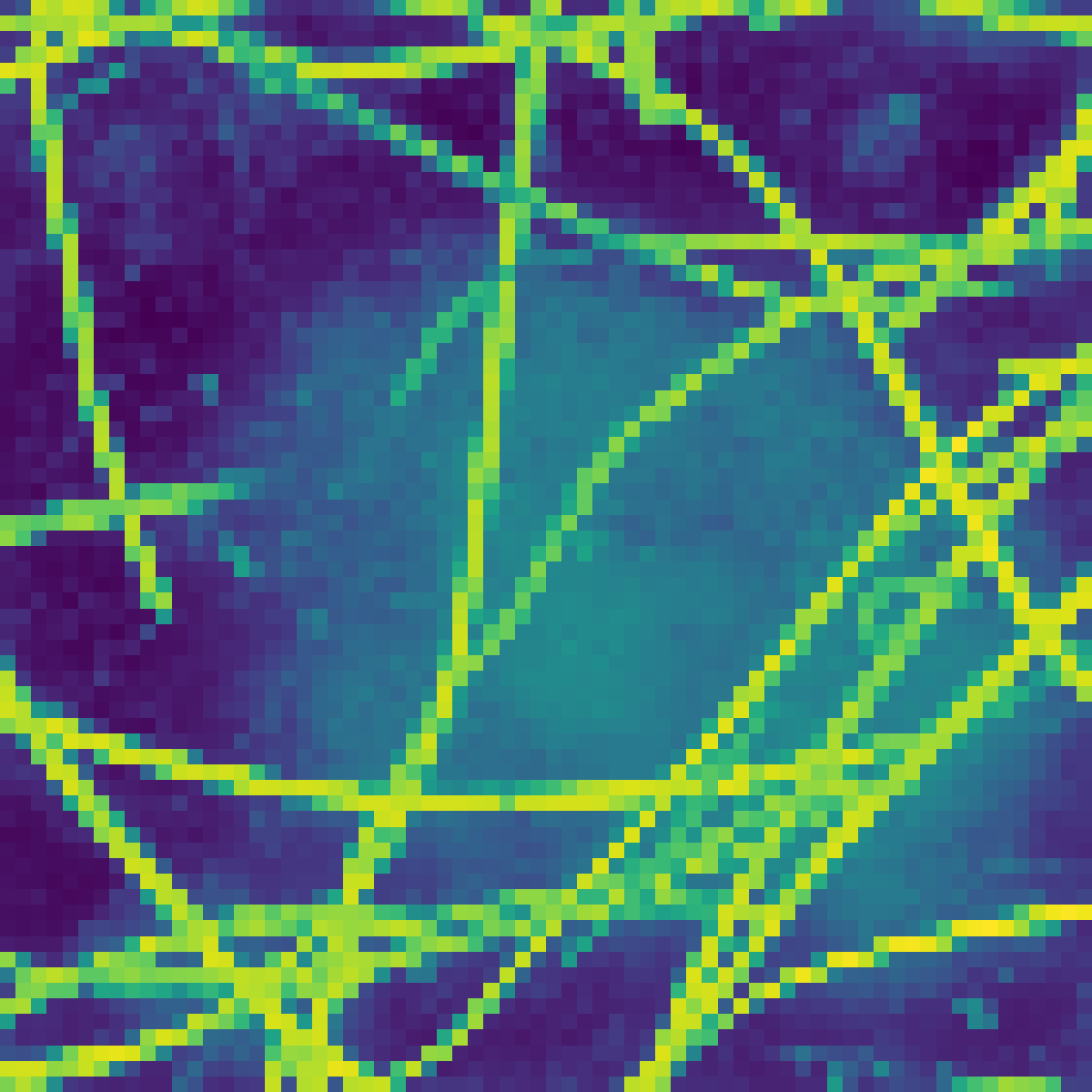} &
  \qimgsix{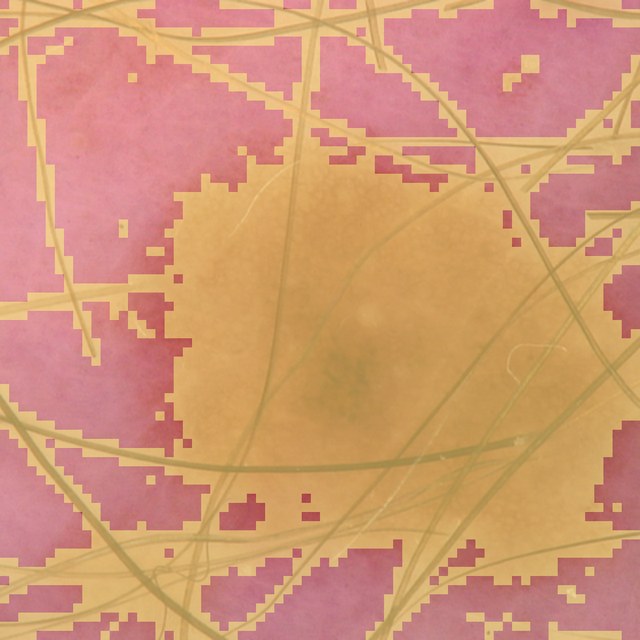} &
  \qimgsix{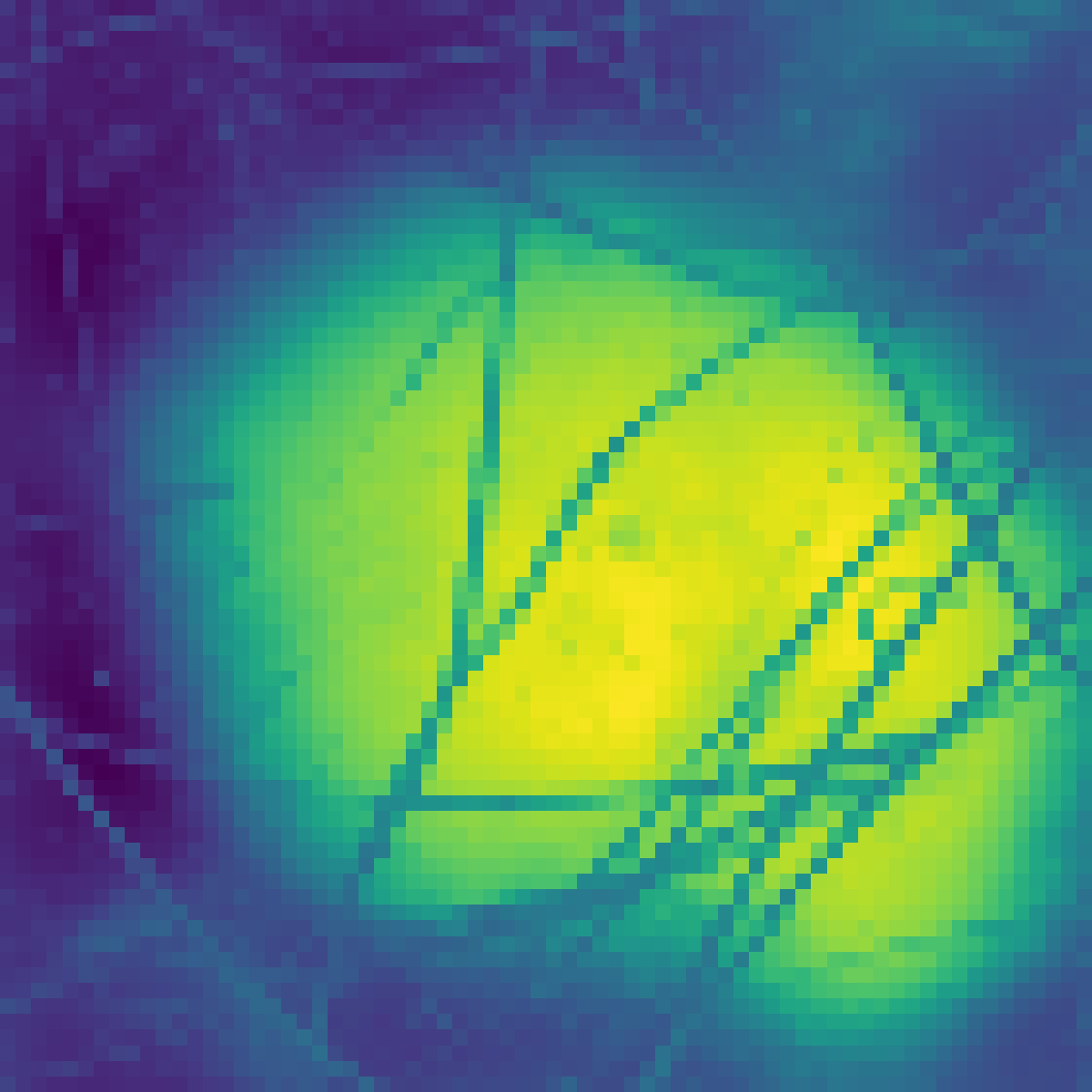} &
  \qimgsix{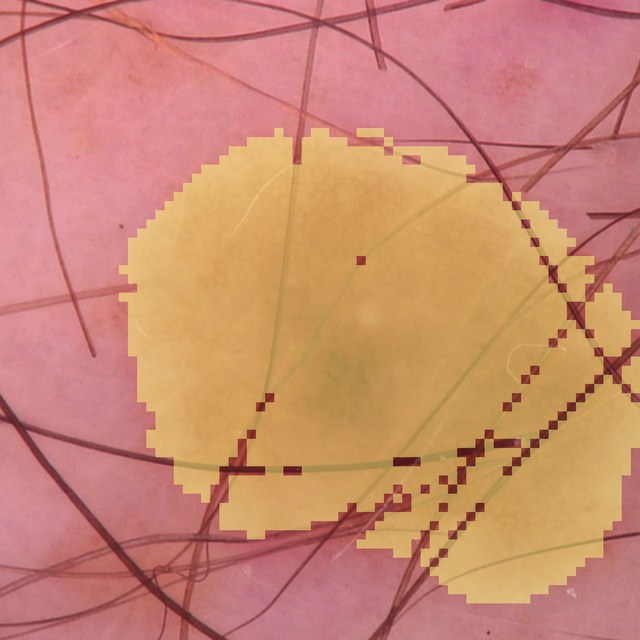} \\
  \noalign{\vspace{1pt}}

  \qimgsix{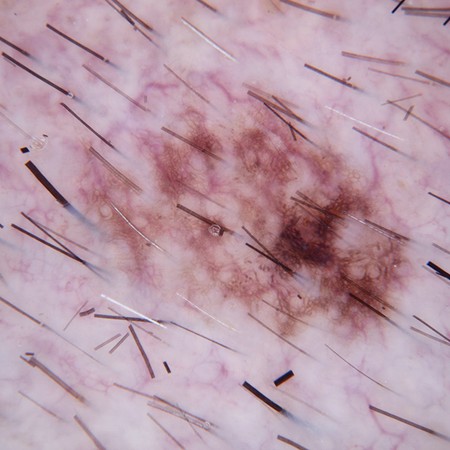} &
  \qimgsix{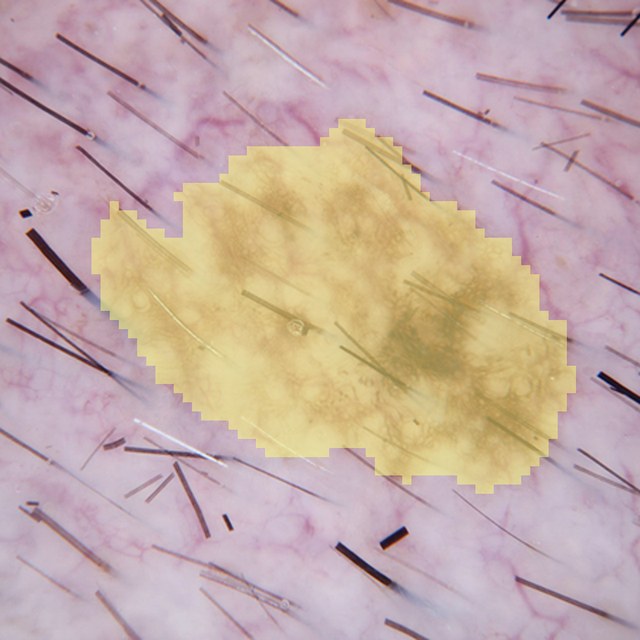} &
  \qimgsix{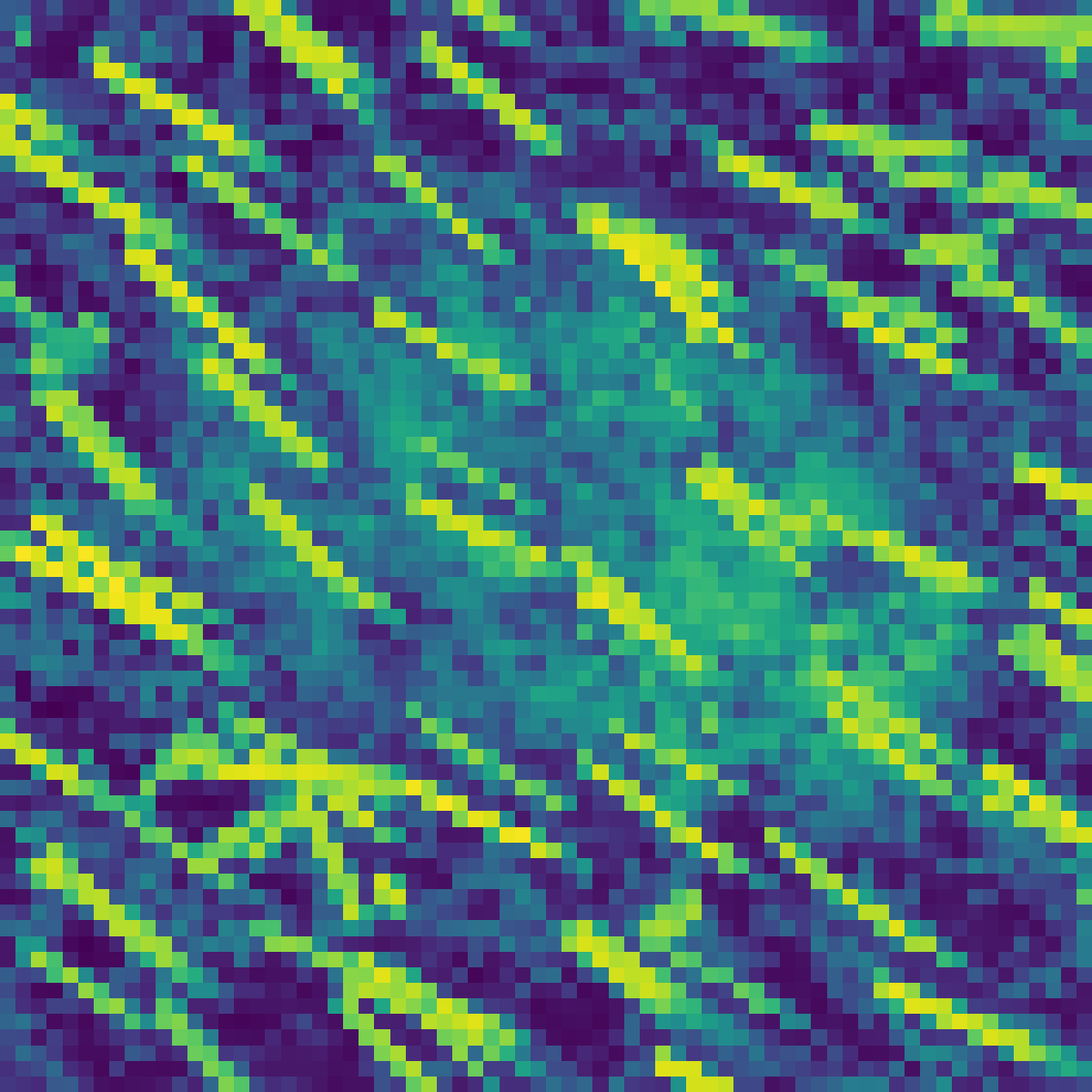} &
  \qimgsix{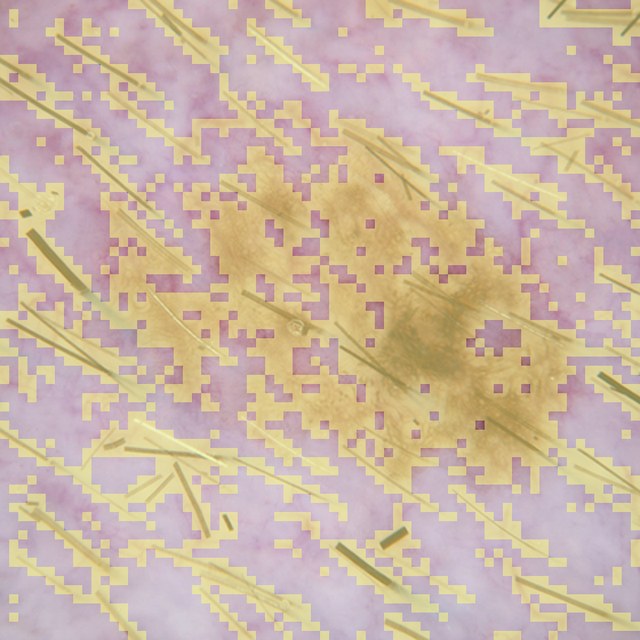} &
  \qimgsix{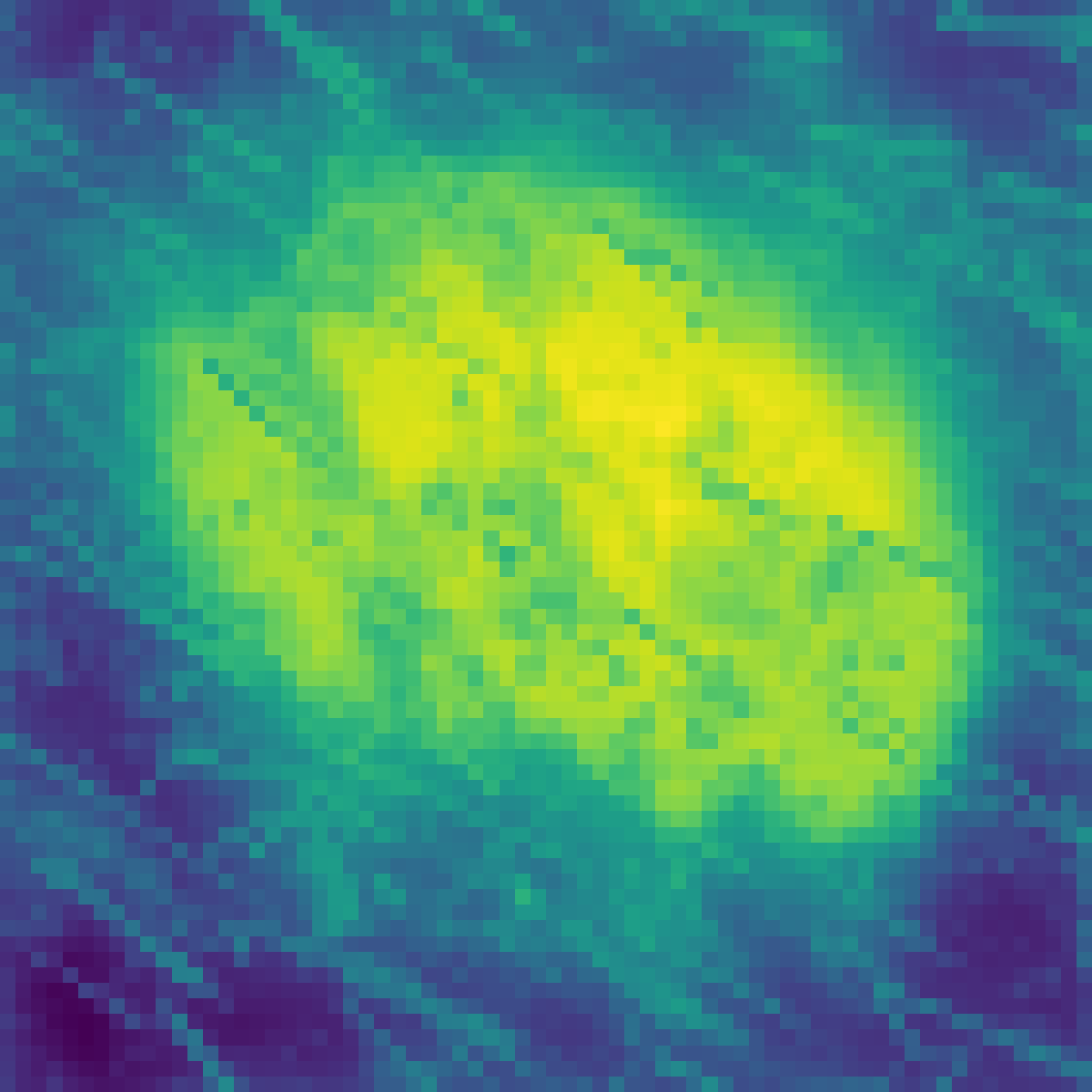} &
  \qimgsix{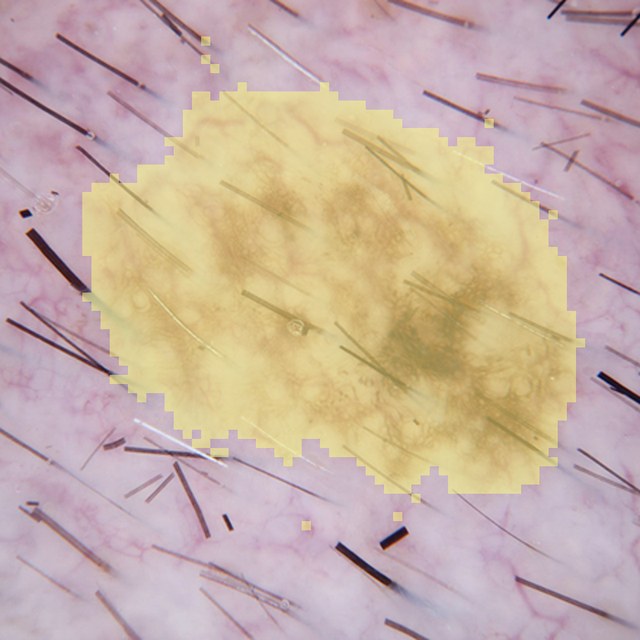} \\
  \noalign{\vspace{1pt}}

  \qimgsix{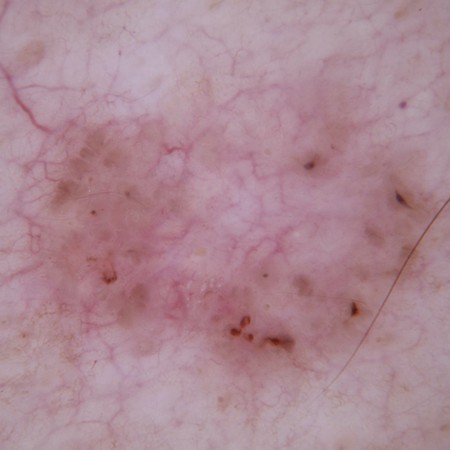} &
  \qimgsix{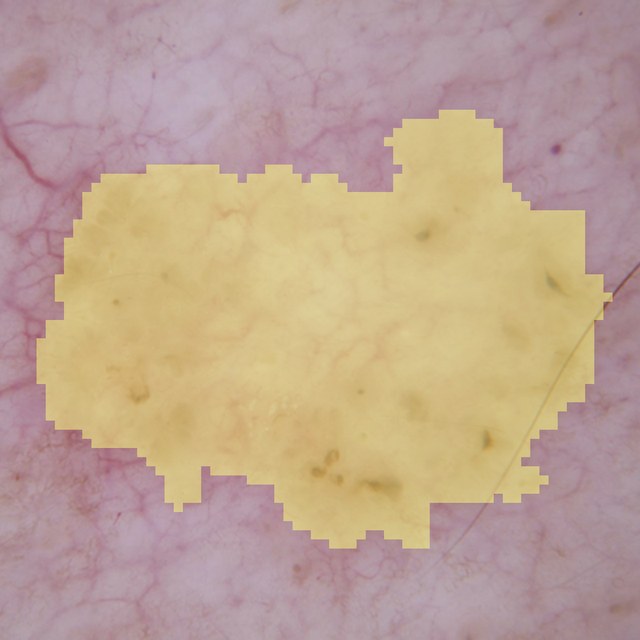} &
  \qimgsix{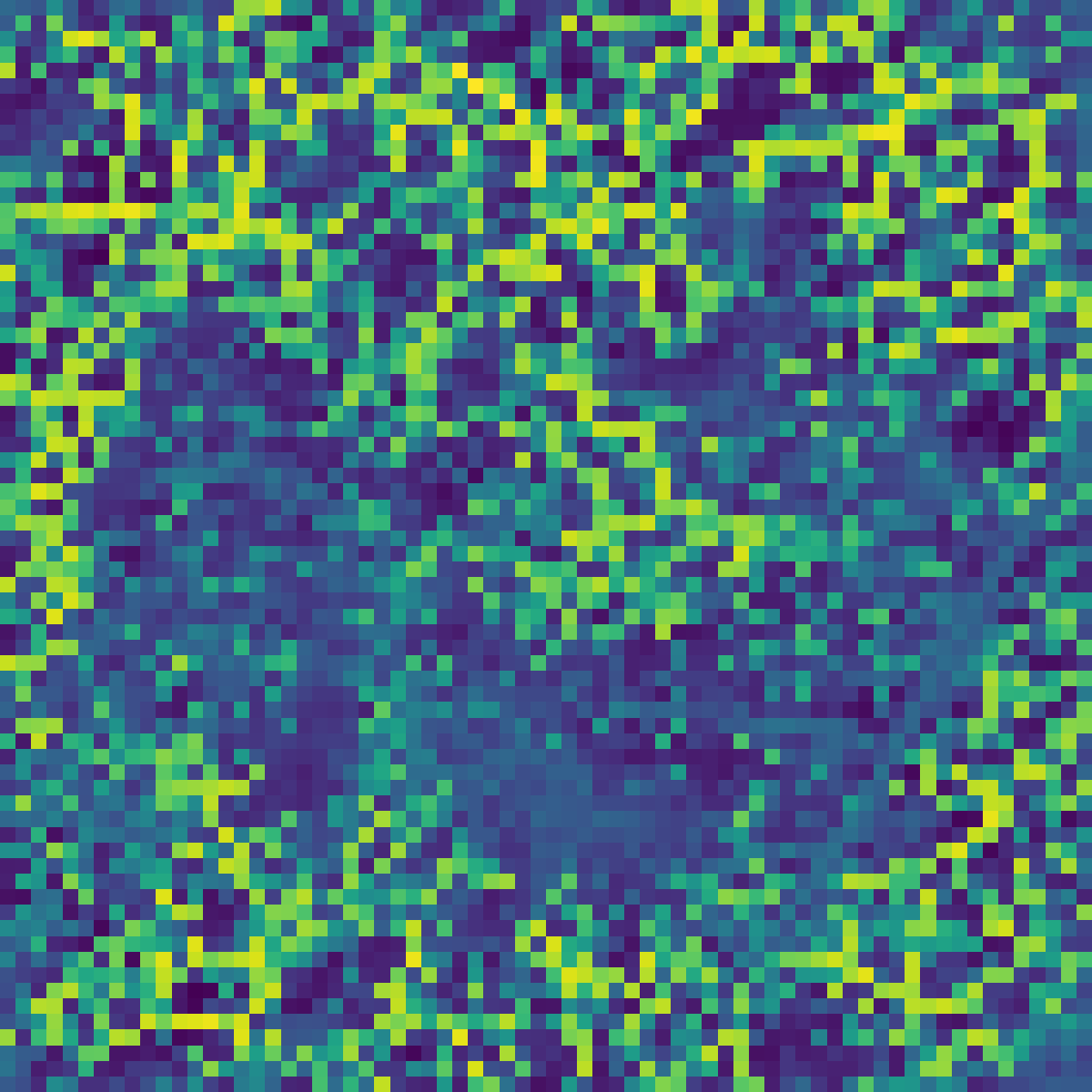} &
  \qimgsix{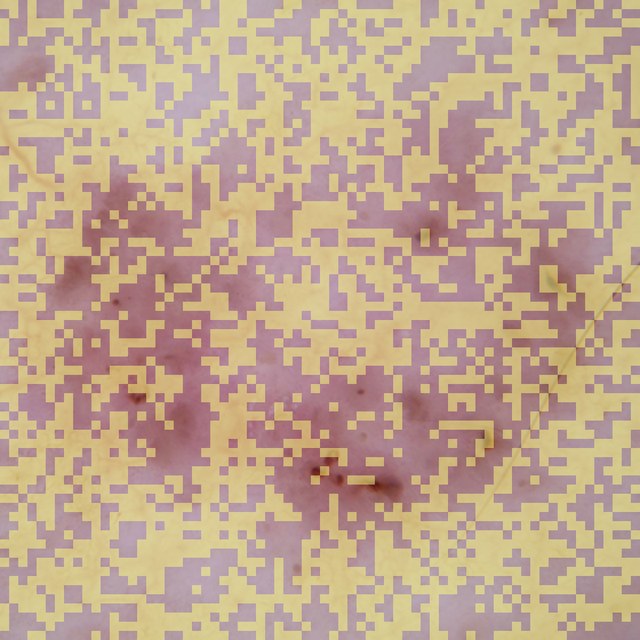} &
  \qimgsix{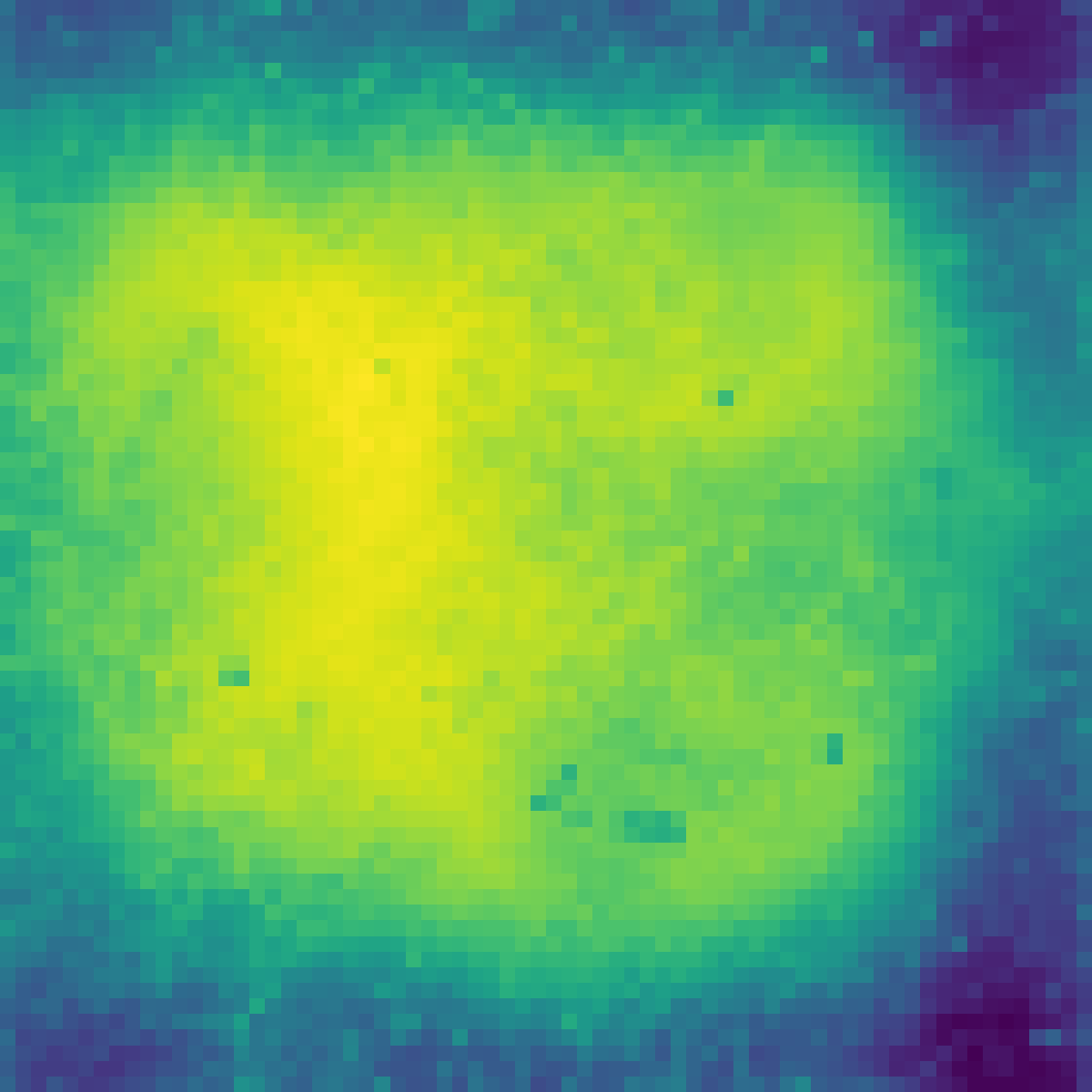} &
  \qimgsix{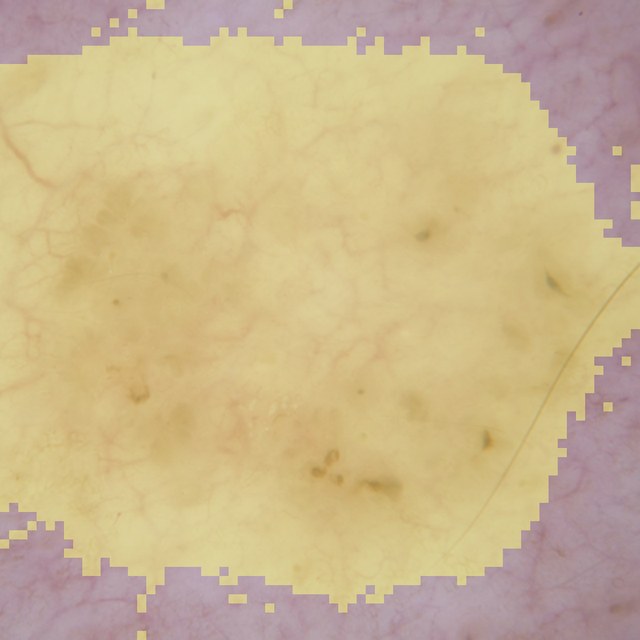} \\
  \noalign{\vspace{1pt}}

  \qimgsix{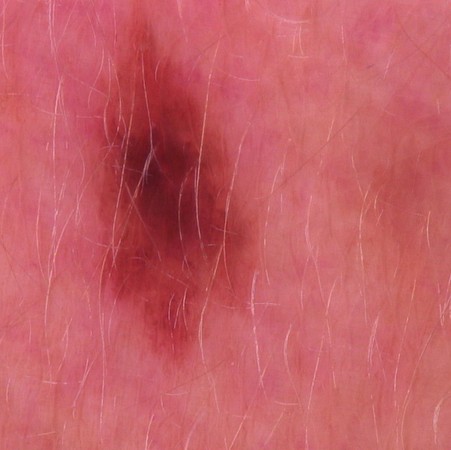} &
  \qimgsix{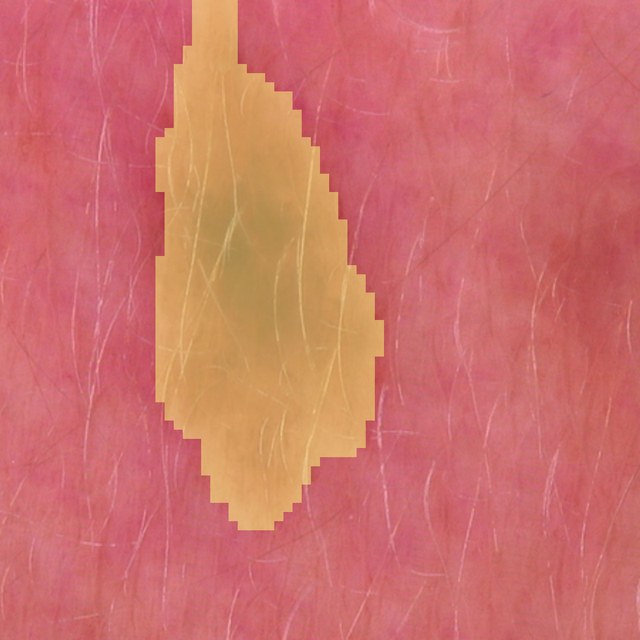} &
  \qimgsix{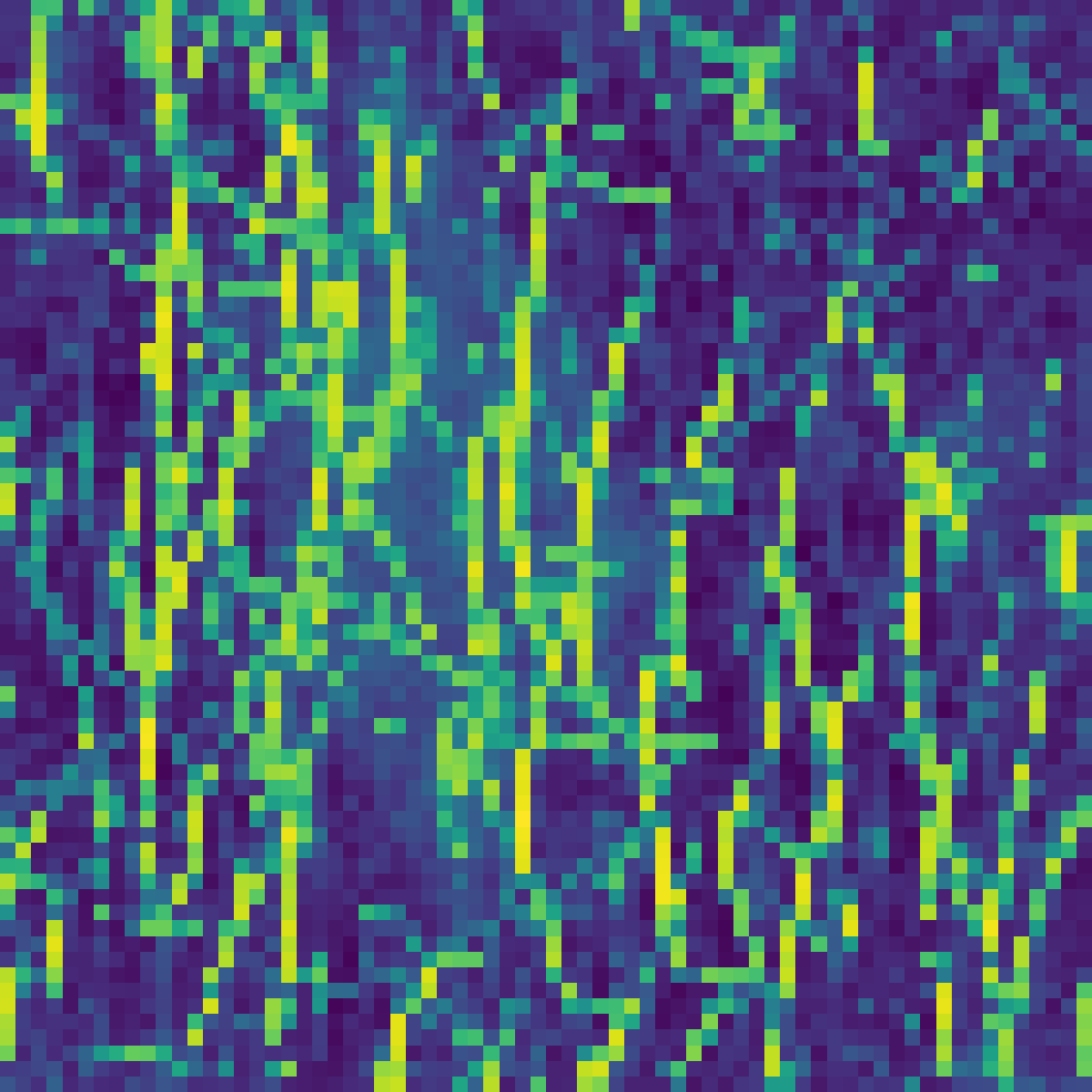} &
  \qimgsix{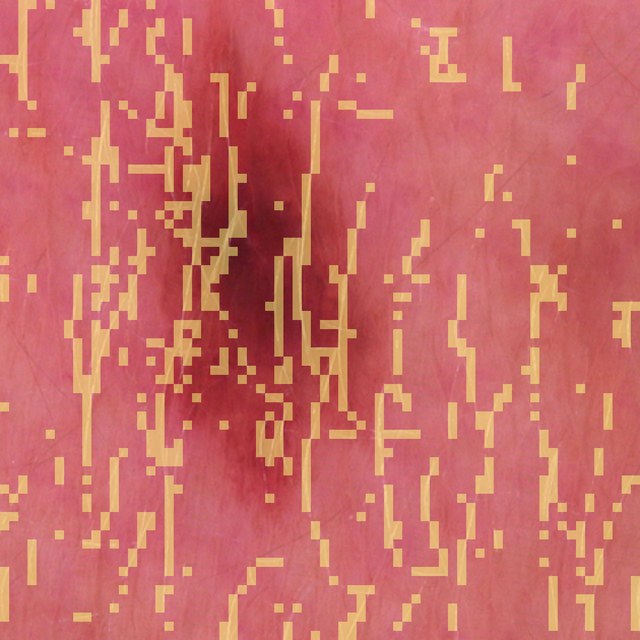} &
  \qimgsix{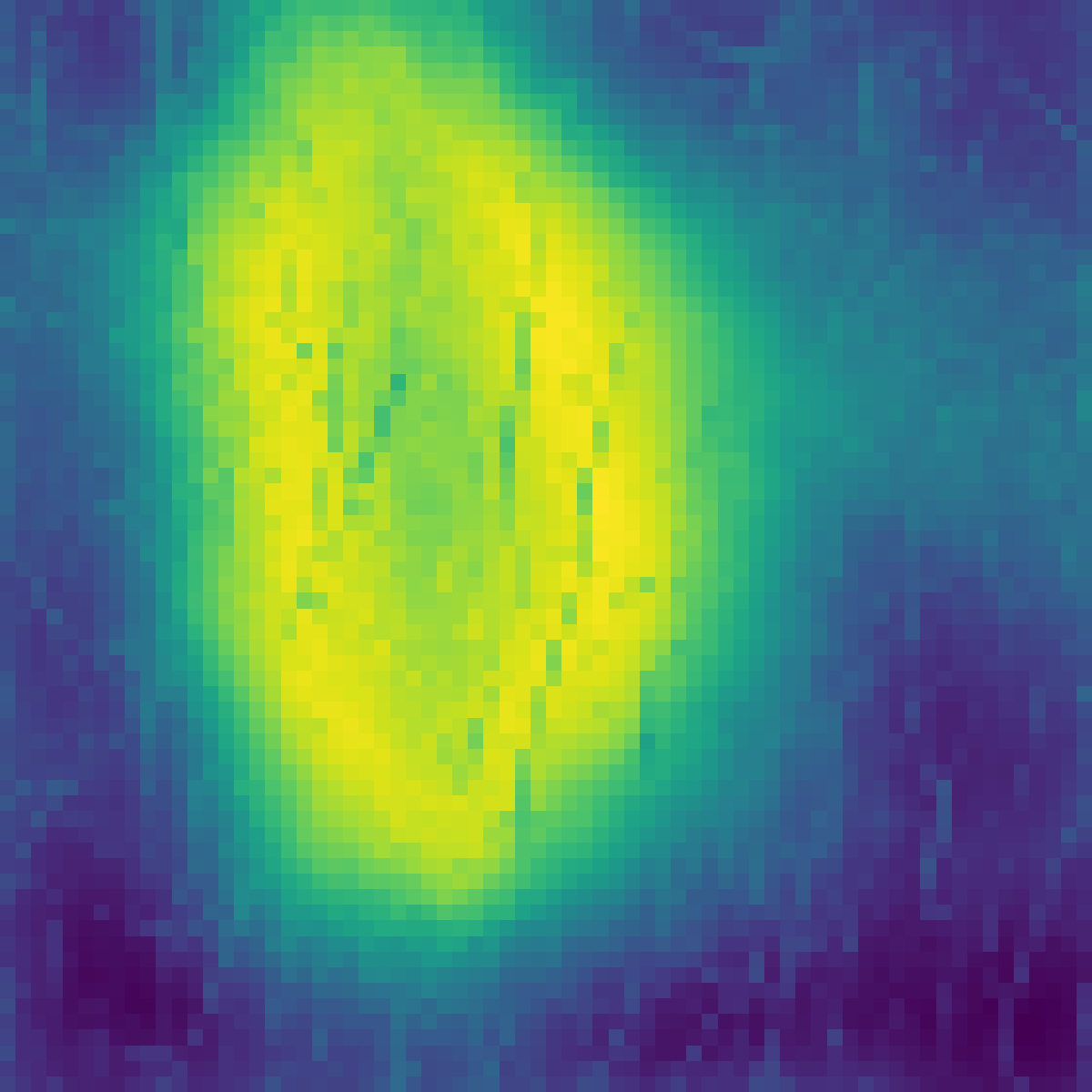} &
  \qimgsix{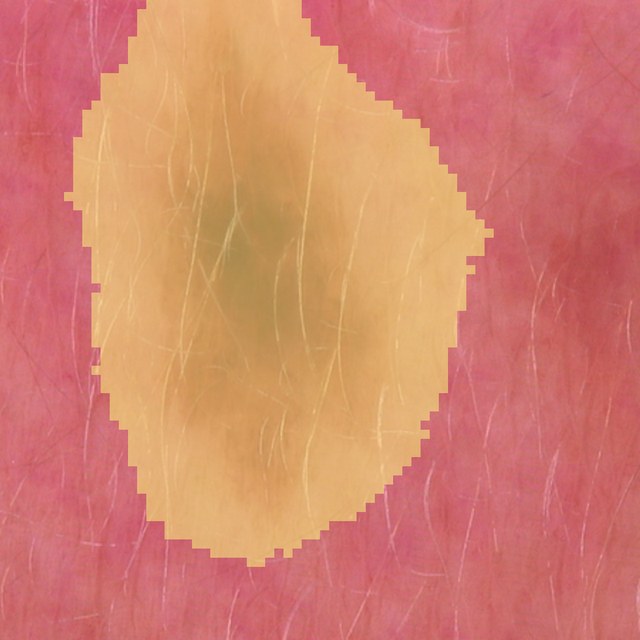} \\
  \noalign{\vspace{2pt}}

  \raisebox{-0.5\height}{\small Input} &
  \raisebox{-0.5\height}{\small GT} &
  \raisebox{-0.5\height}{\small\shortstack{Unsup.\\Eigen Attn.}} &
  \raisebox{-0.5\height}{\small\shortstack{Unsup.\\Mask}} &
  \raisebox{-0.5\height}{\small\shortstack{PANC\\Eigen Attn.}} &
  \raisebox{-0.5\height}{\small\shortstack{PANC\\Mask}} \\
\end{tabular}
\end{adjustbox}
\endgroup
\caption{Additional qualitative comparison on the HAM10000 dataset (yellow mask).}
\label{fig:add_ham_examples}
\end{figure*}

\begin{figure*}[p]
\centering
\begingroup
\setlength{\tabcolsep}{0pt}
\begin{adjustbox}{max width=\textwidth,max totalheight=1.0\textheight,center}
\begin{tabular}{@{}c @{\hspace{2pt}} c @{\hspace{2pt}} c @{\hspace{2pt}} c @{\hspace{2pt}} c @{\hspace{2pt}} c@{}}
  \qimgsix{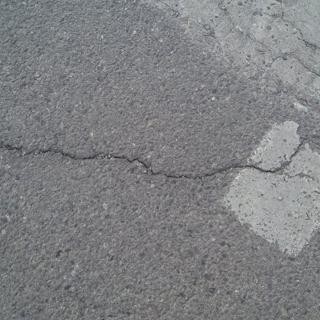} &
  \qimgsix{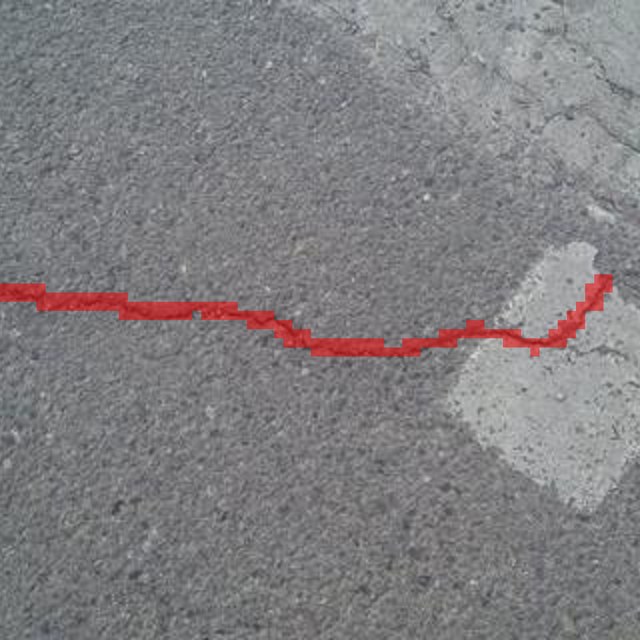} &
  \qimgsix{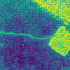} &
  \qimgsix{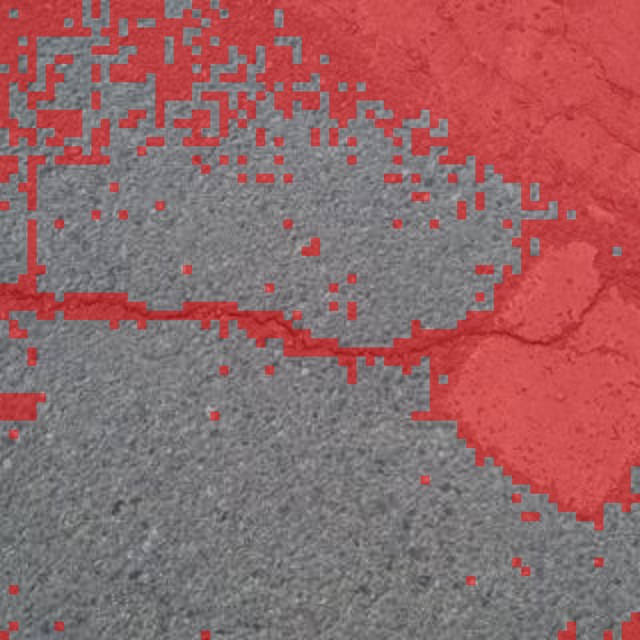} &
  \qimgsix{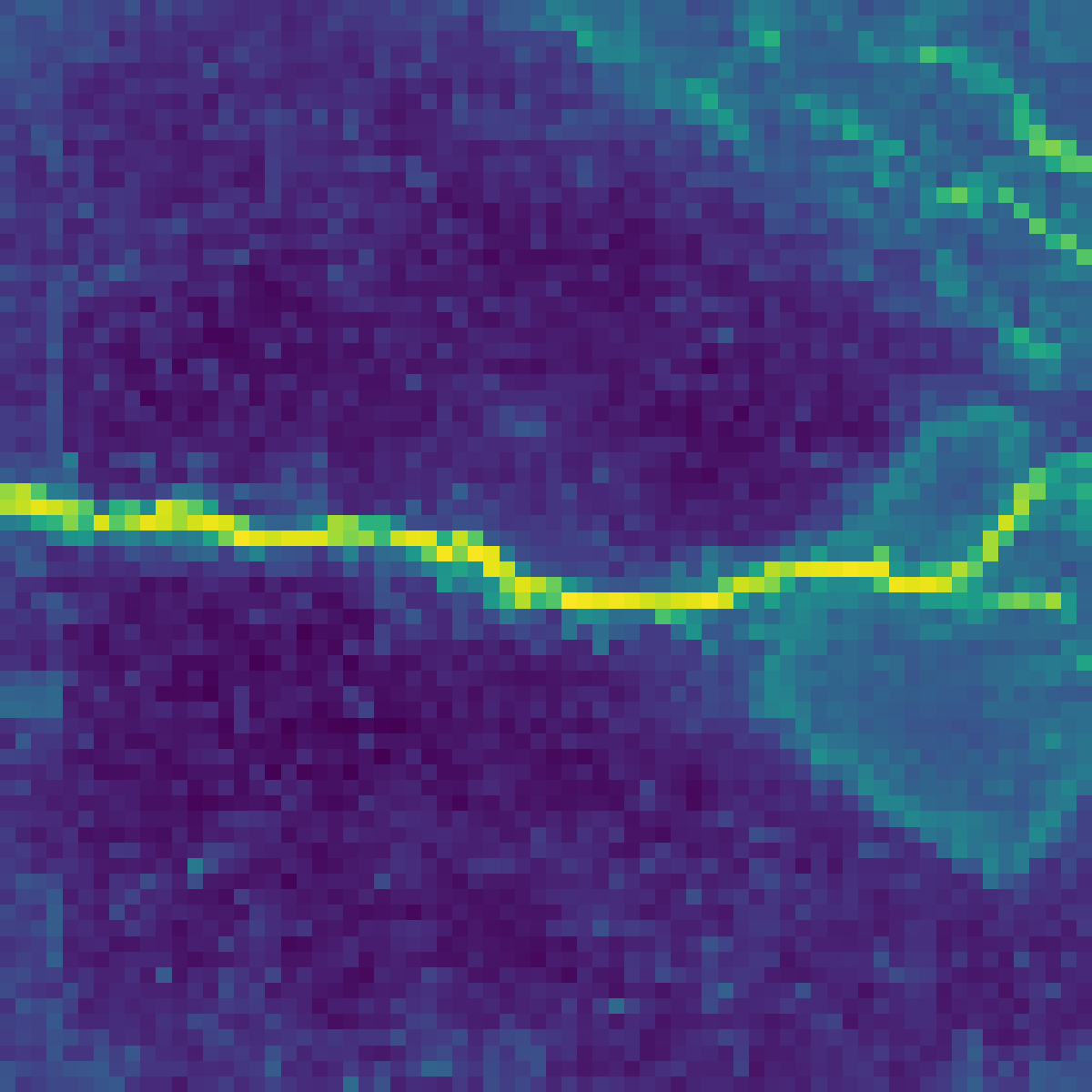} &
  \qimgsix{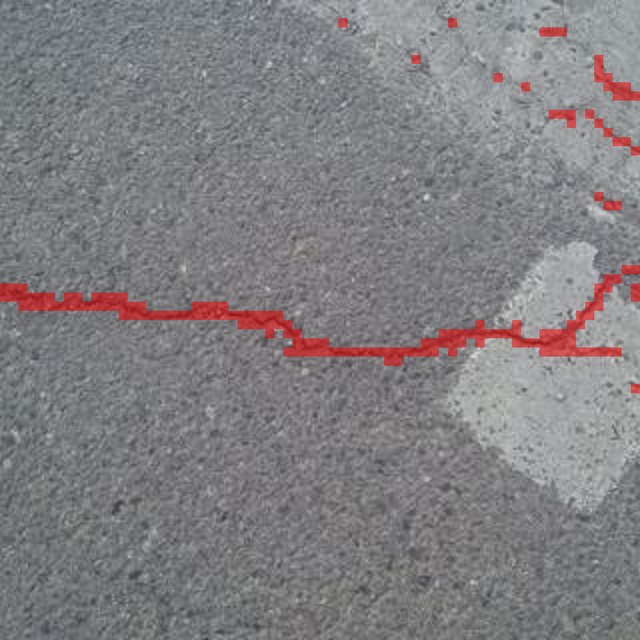} \\
  \noalign{\vspace{1pt}}

  \qimgsix{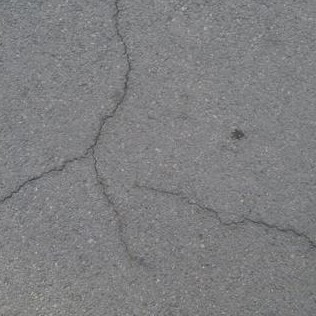} &
  \qimgsix{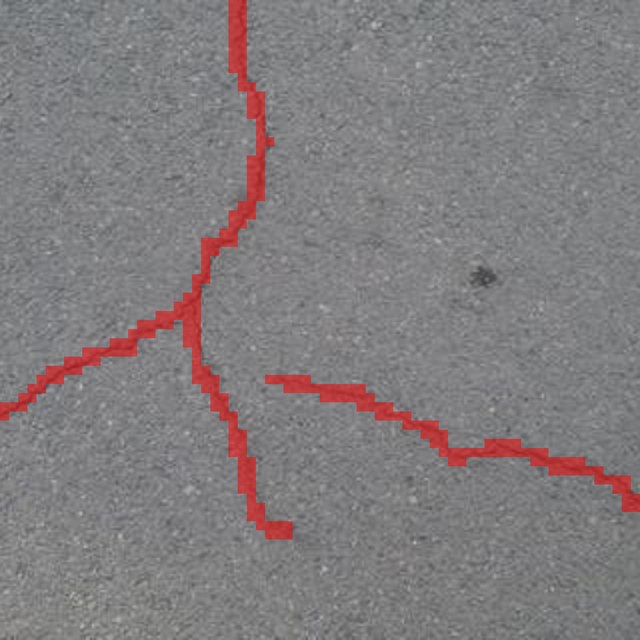} &
  \qimgsix{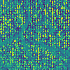} &
  \qimgsix{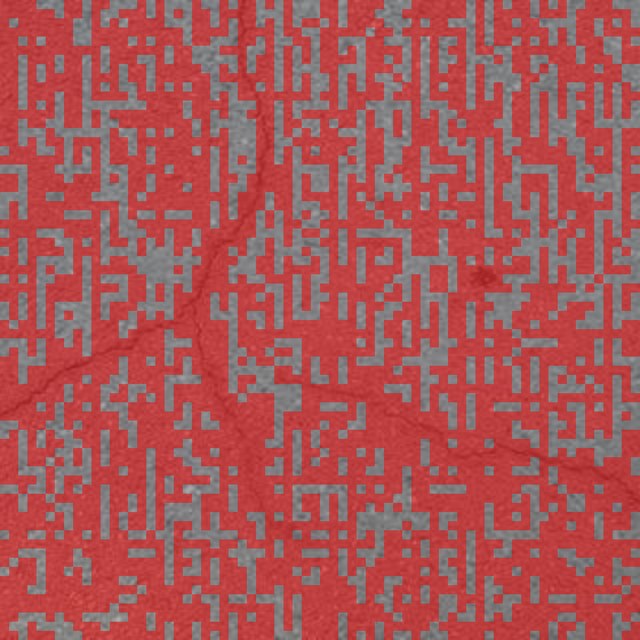} &
  \qimgsix{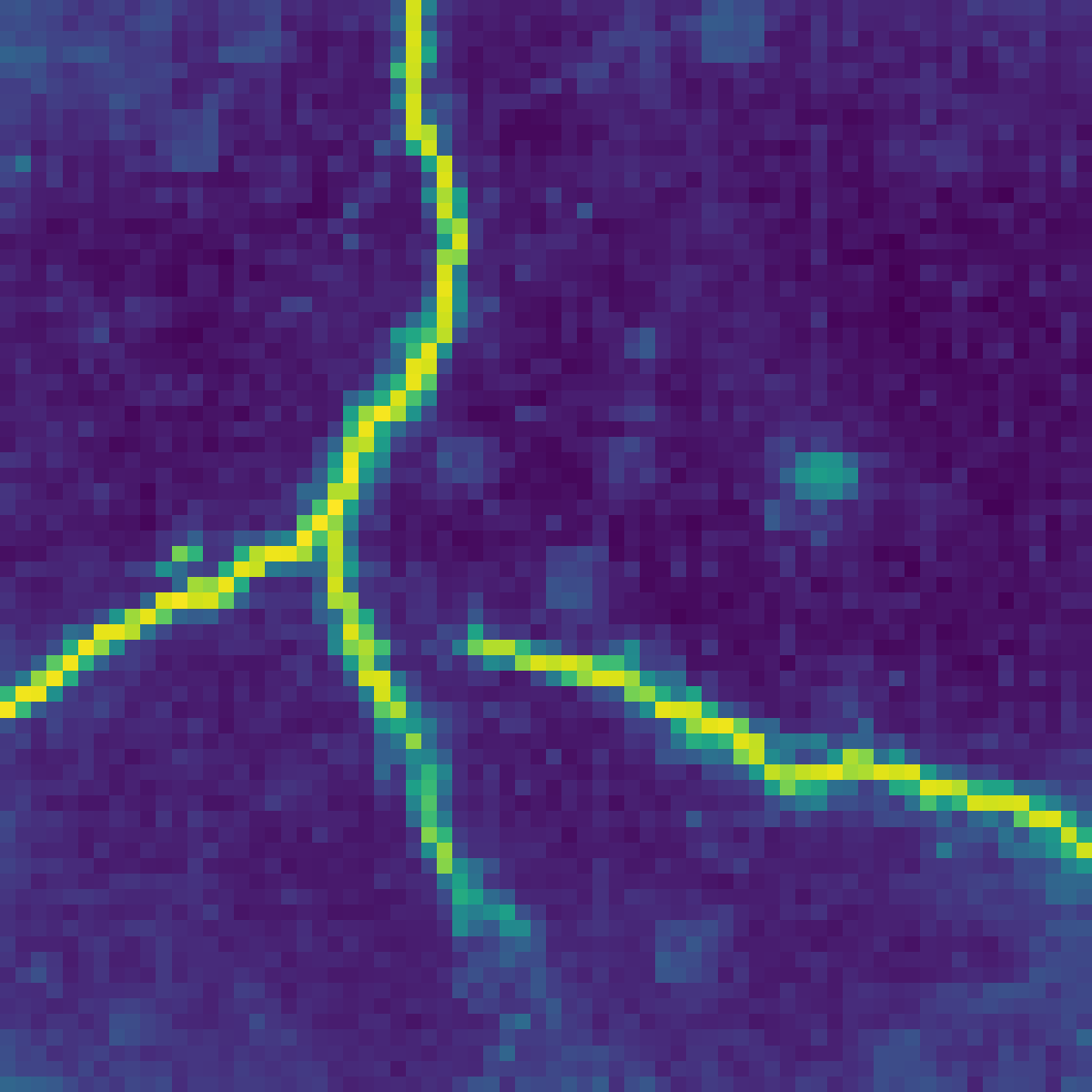} &
  \qimgsix{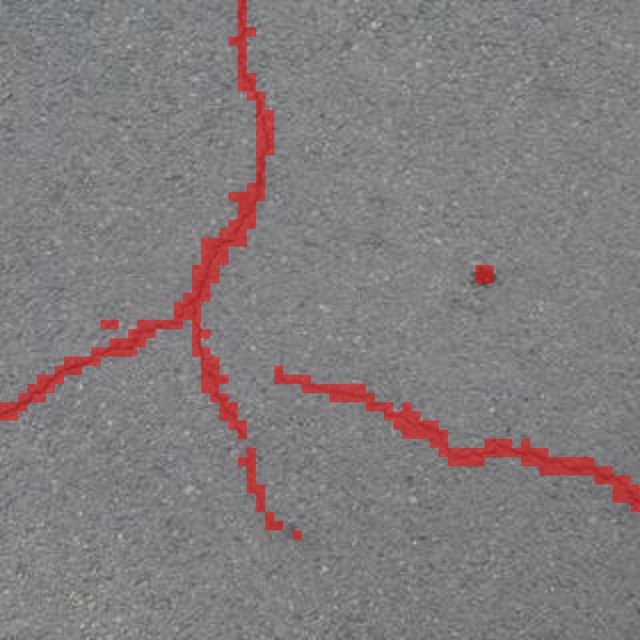} \\
  \noalign{\vspace{1pt}}

  \qimgsix{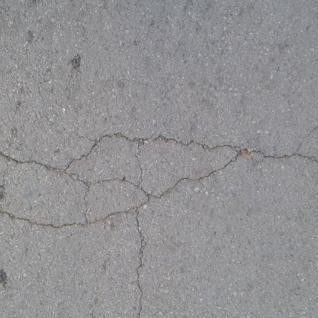} &
  \qimgsix{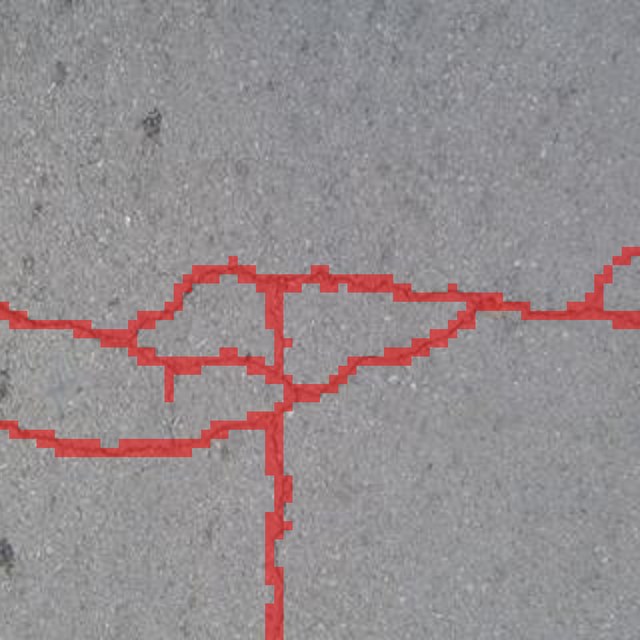} &
  \qimgsix{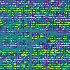} &
  \qimgsix{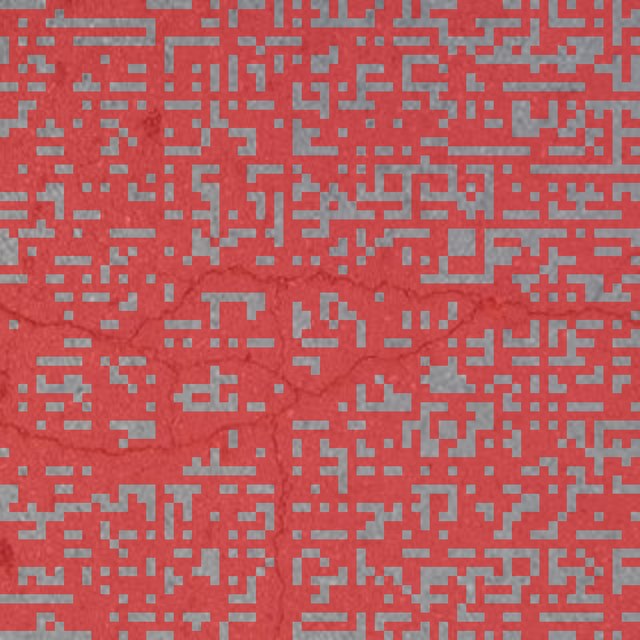} &
  \qimgsix{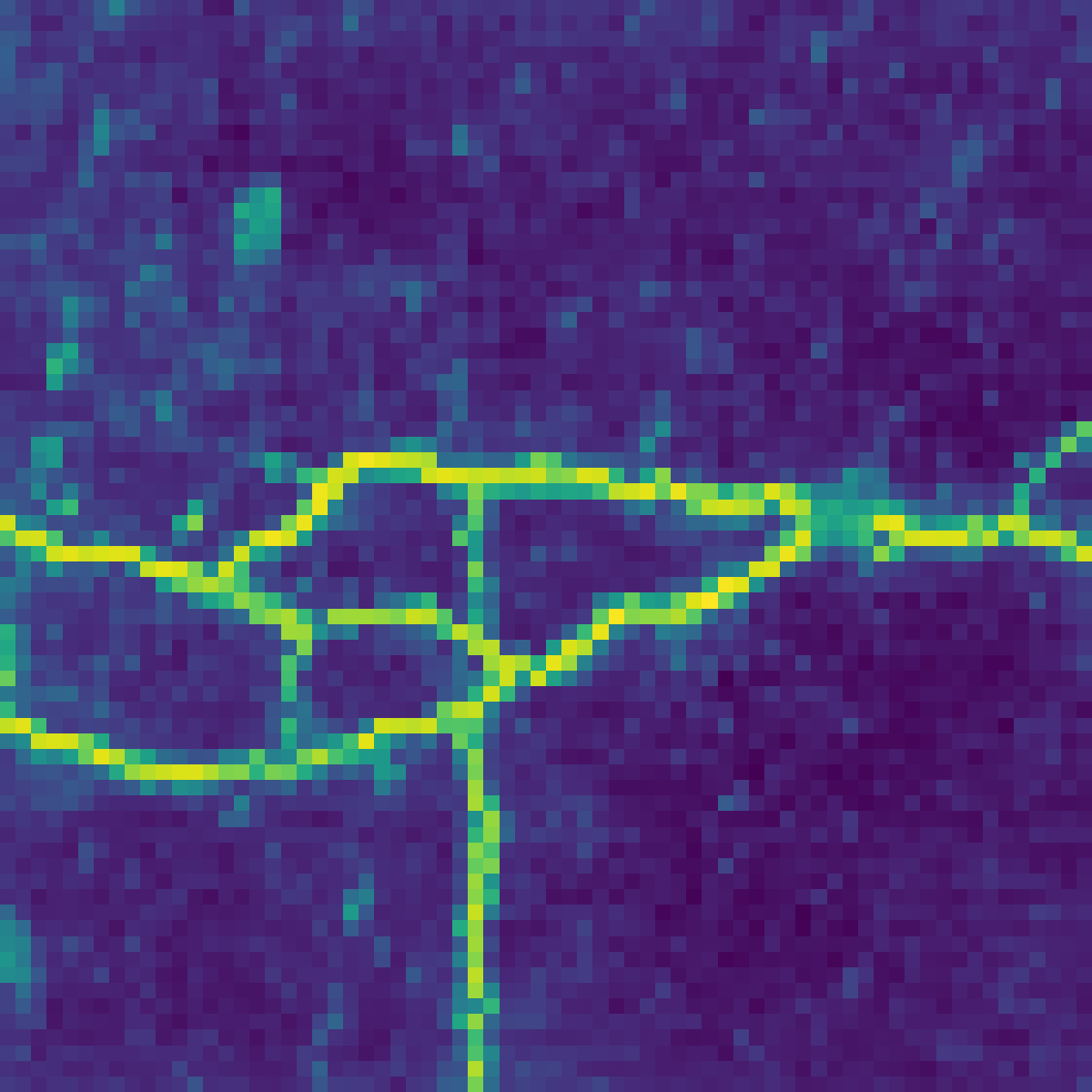} &
  \qimgsix{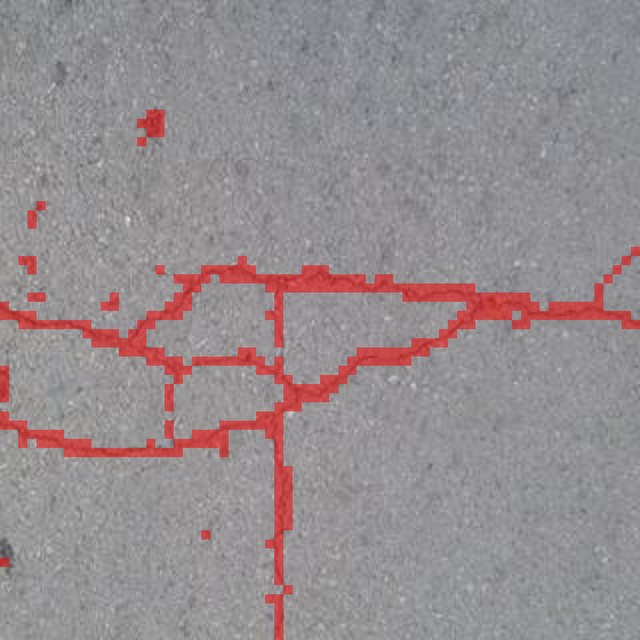} \\
  \noalign{\vspace{1pt}}

  \qimgsix{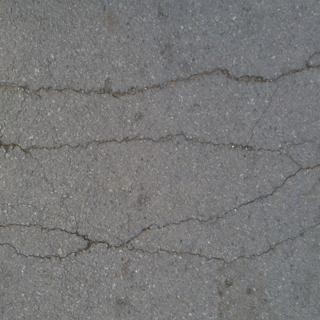} &
  \qimgsix{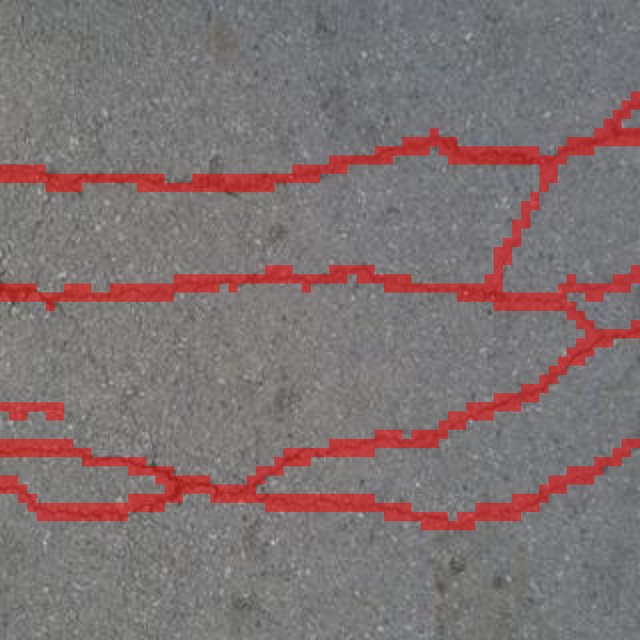} &
  \qimgsix{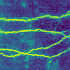} &
  \qimgsix{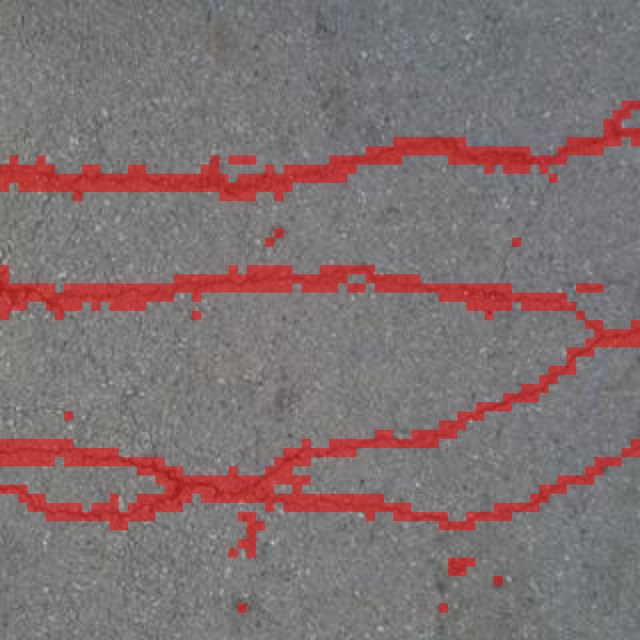} &
  \qimgsix{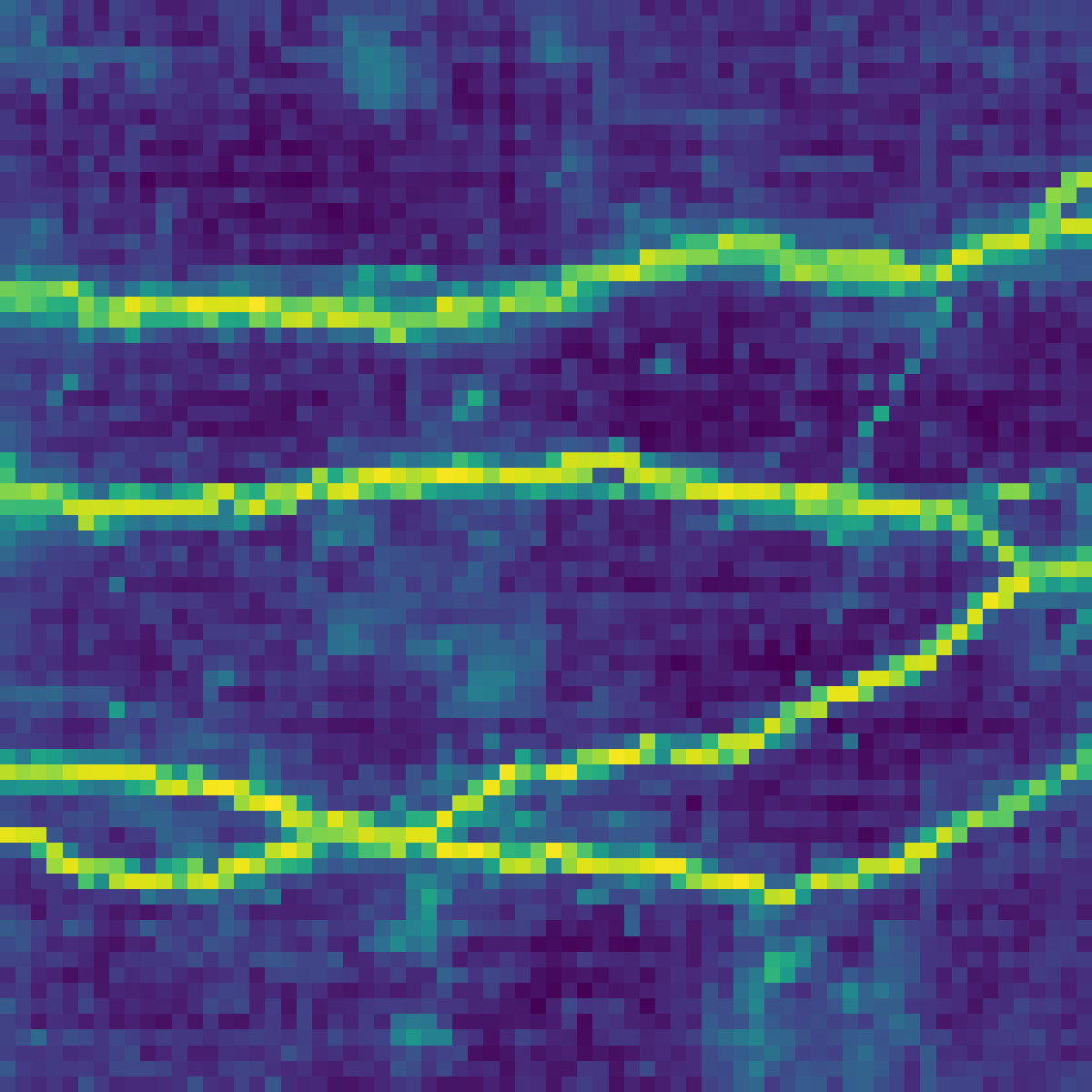} &
  \qimgsix{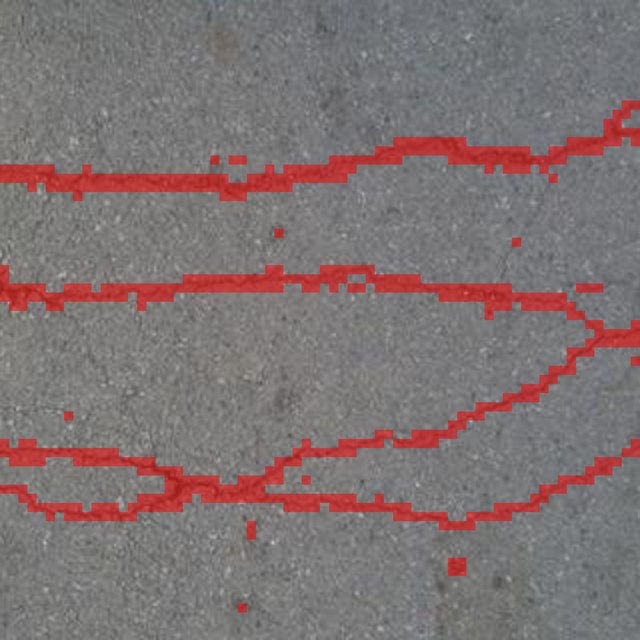} \\
  \noalign{\vspace{1pt}}

  \qimgsix{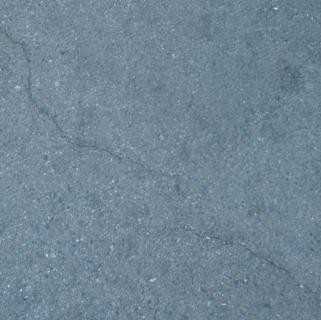} &
  \qimgsix{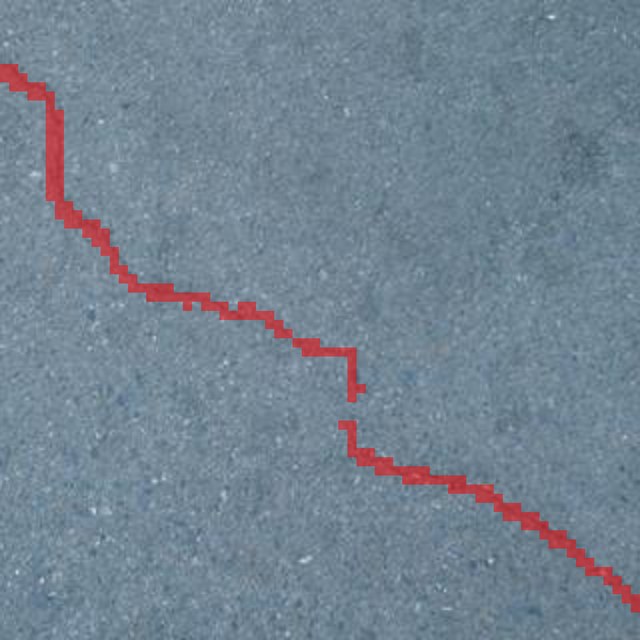} &
  \qimgsix{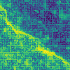} &
  \qimgsix{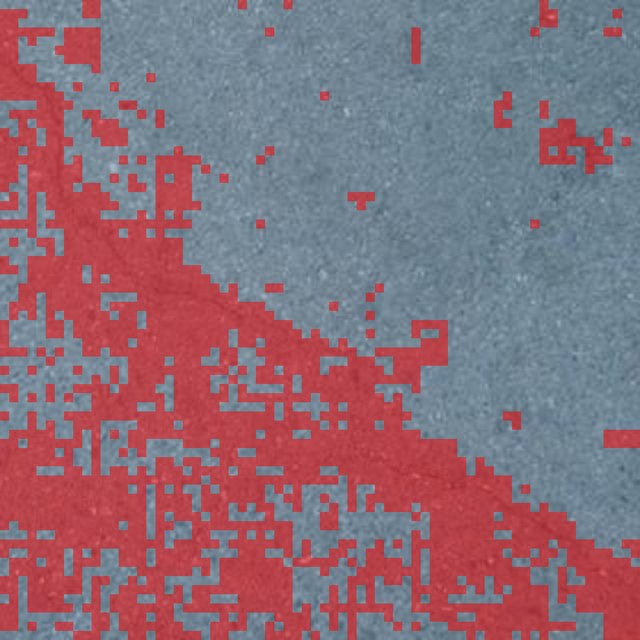} &
  \qimgsix{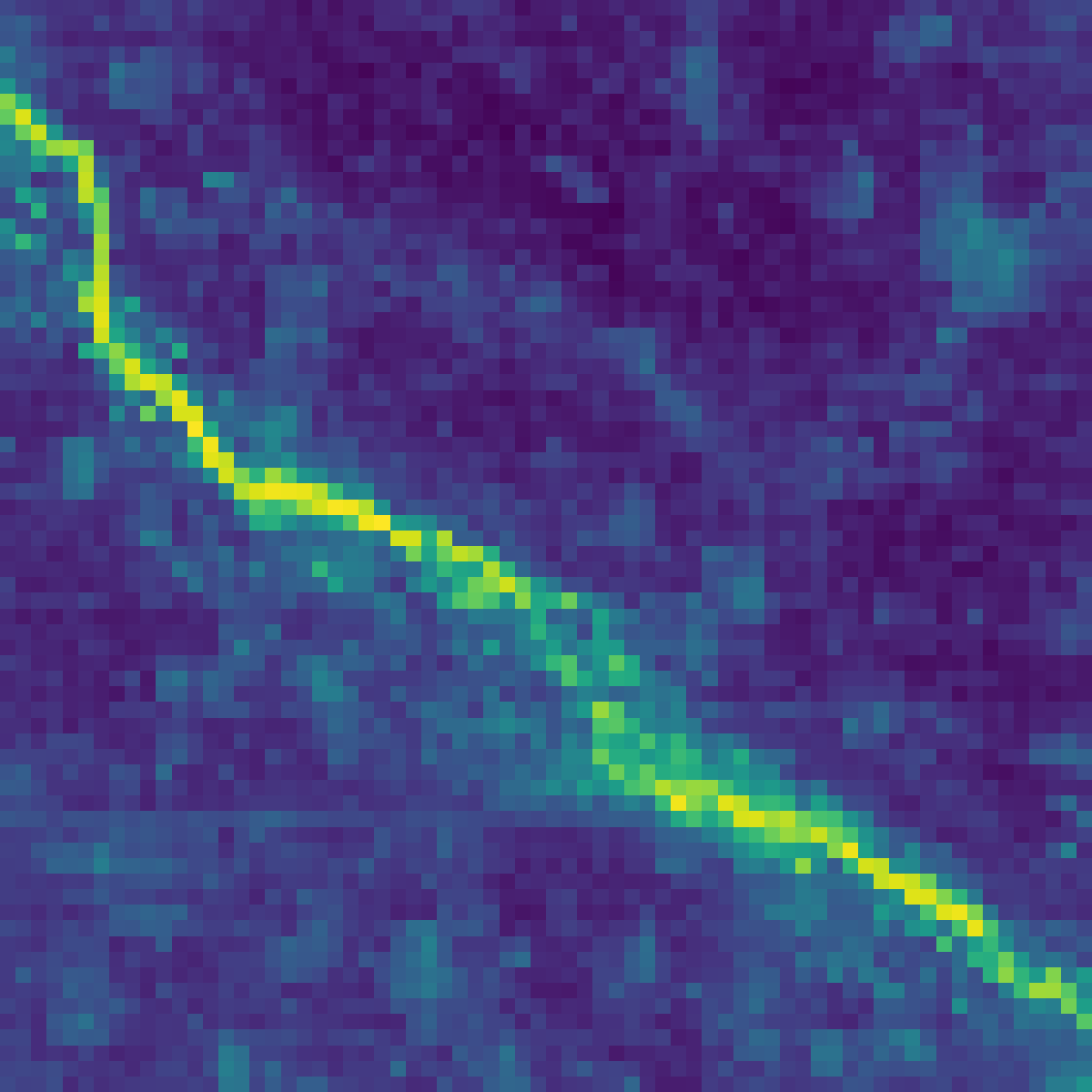} &
  \qimgsix{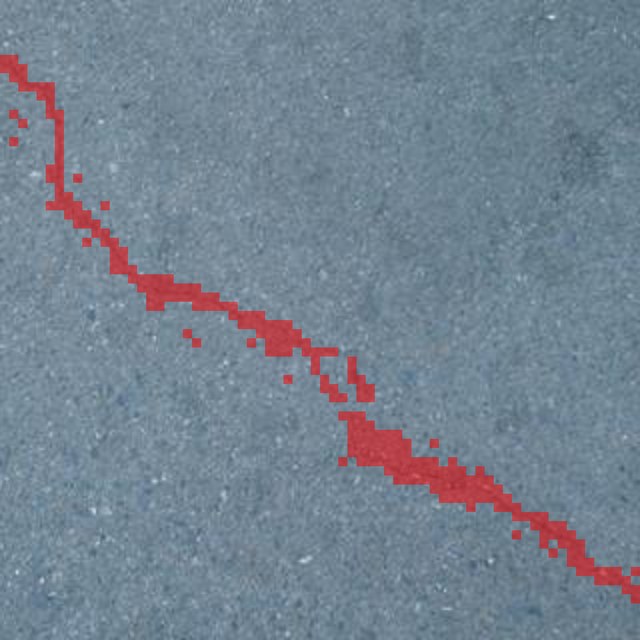} \\
  \noalign{\vspace{1pt}}

  \qimgsix{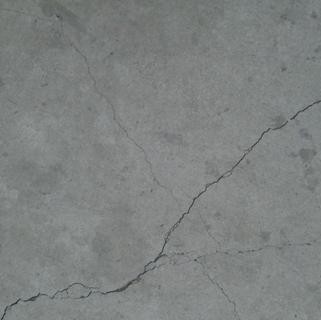} &
  \qimgsix{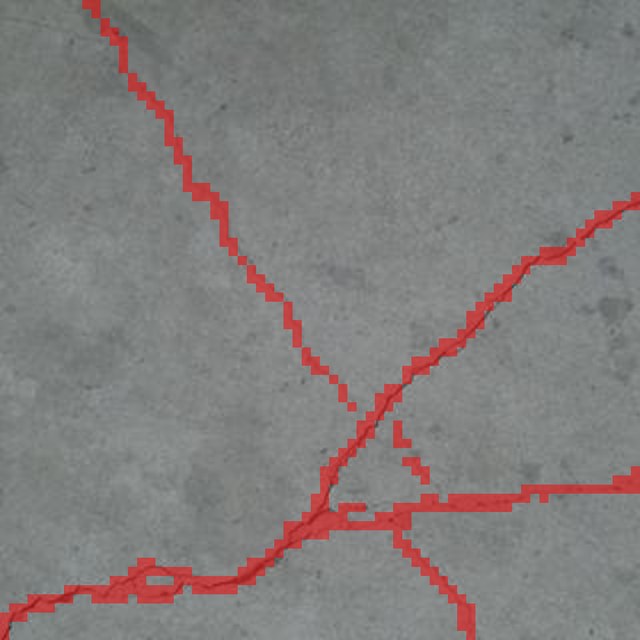} &
  \qimgsix{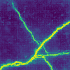} &
  \qimgsix{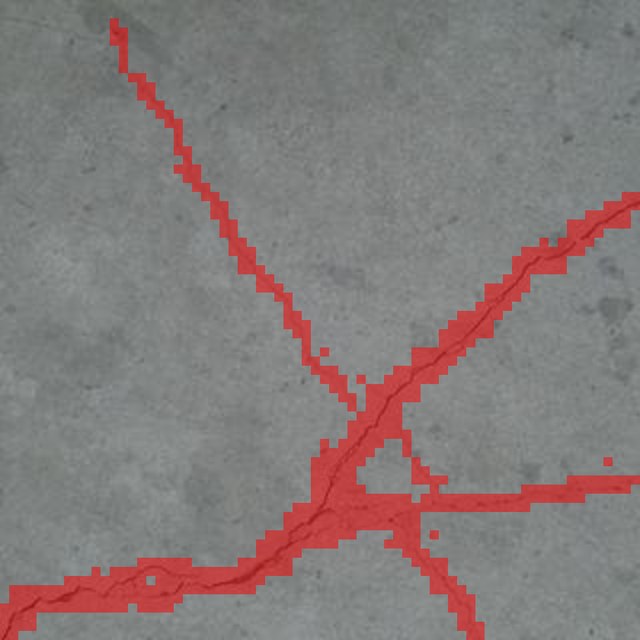} &
  \qimgsix{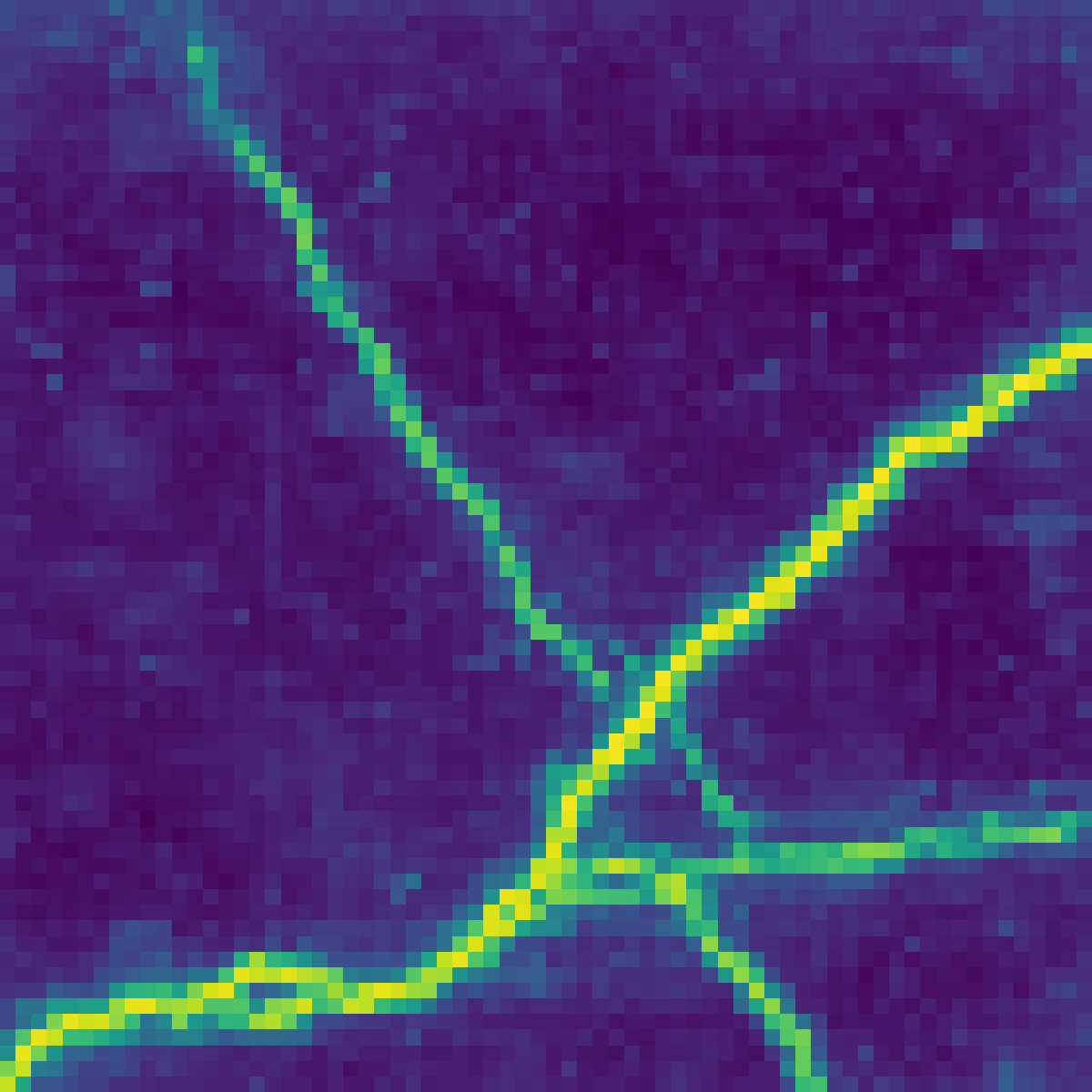} &
  \qimgsix{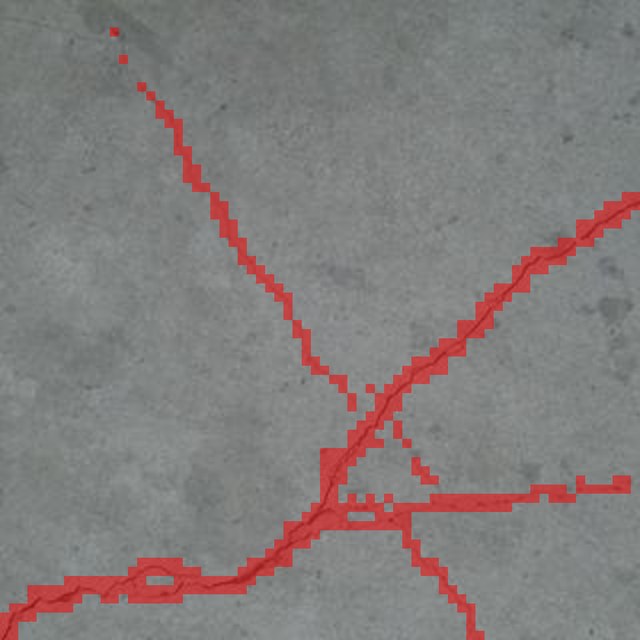} \\
  \noalign{\vspace{1pt}}

  \qimgsix{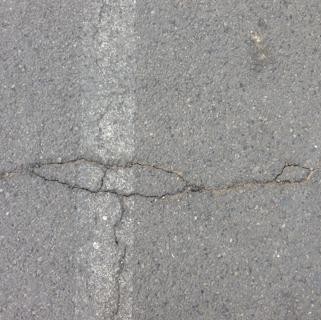} &
  \qimgsix{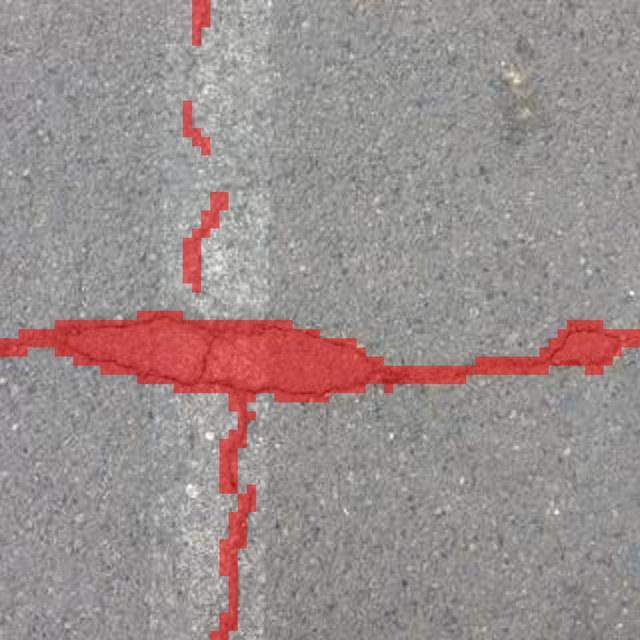} &
  \qimgsix{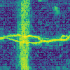} &
  \qimgsix{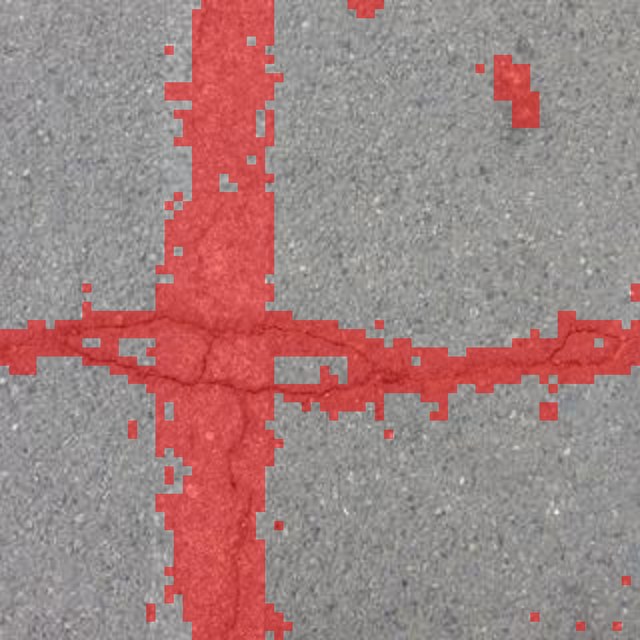} &
  \qimgsix{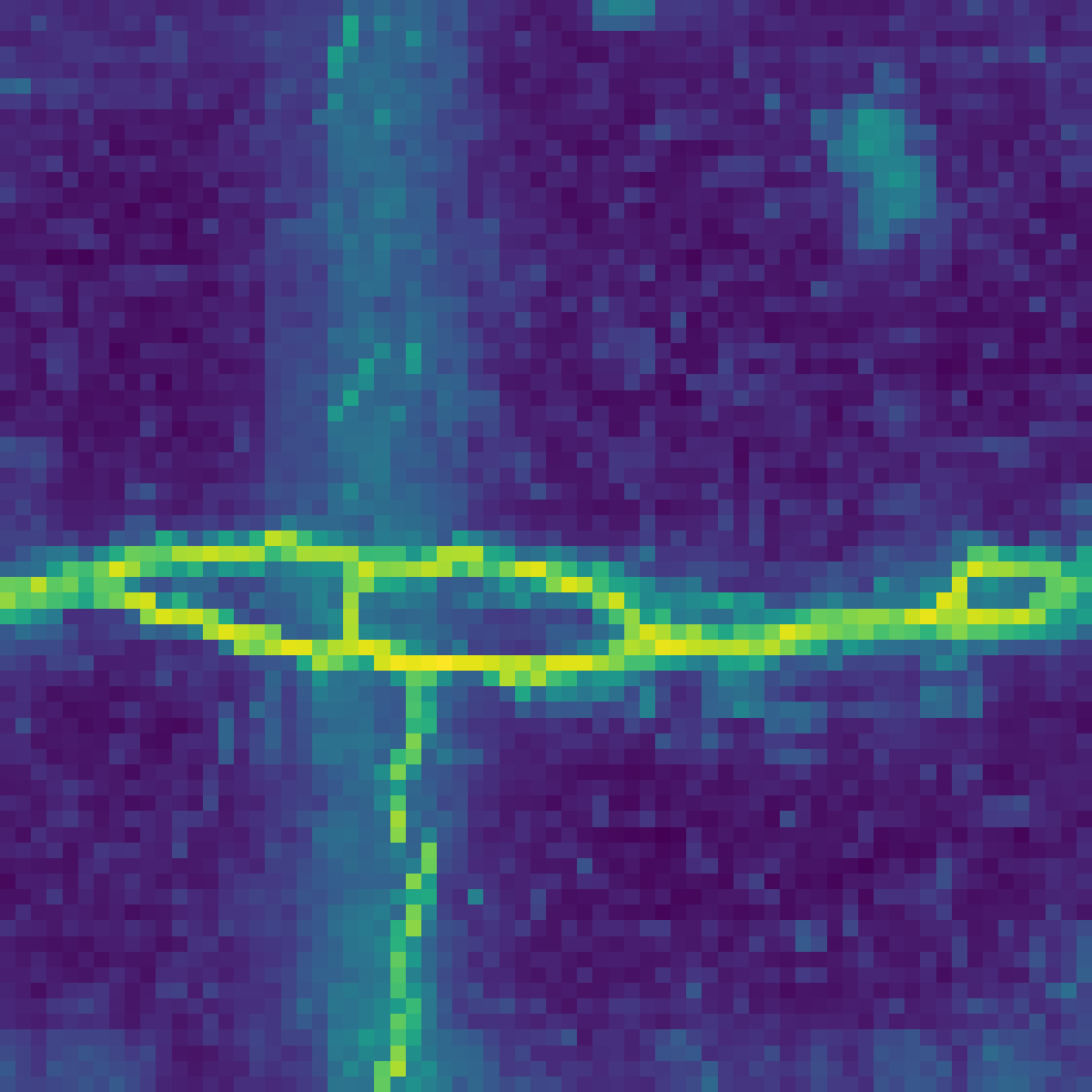} &
  \qimgsix{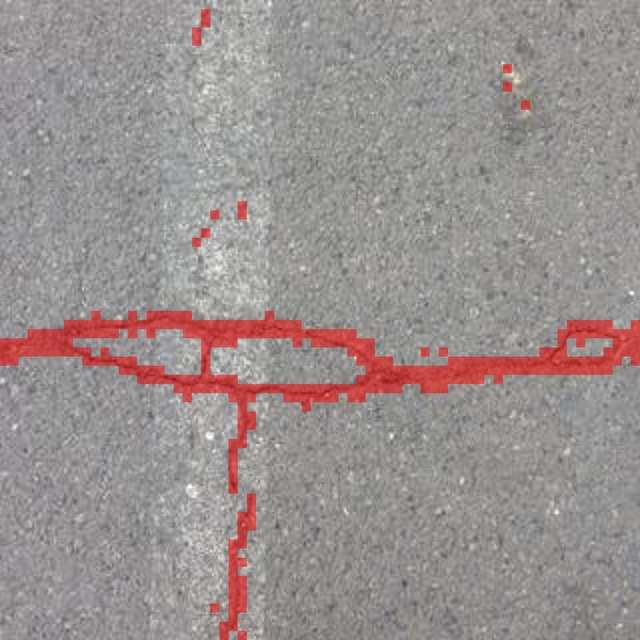} \\
  \noalign{\vspace{2pt}}

  \raisebox{-0.5\height}{\small Input} &
  \raisebox{-0.5\height}{\small GT} &
  \raisebox{-0.5\height}{\small\shortstack{Unsup.\\Eigen Attn.}} &
  \raisebox{-0.5\height}{\small\shortstack{Unsup.\\Mask}} &
  \raisebox{-0.5\height}{\small\shortstack{PANC\\Eigen Attn.}} &
  \raisebox{-0.5\height}{\small\shortstack{PANC\\Mask}} \\
\end{tabular}
\end{adjustbox}
\endgroup
\caption{Additional qualitative comparison on the CrackForest dataset.}
\label{fig:add_cfd_examples}
\end{figure*}

\begin{figure*}[p]
\centering
\begingroup
\setlength{\tabcolsep}{0pt}
\begin{adjustbox}{max width=\textwidth,max totalheight=1.0\textheight,center}
\begin{tabular}{@{}c @{\hspace{2pt}} c @{\hspace{2pt}} c @{\hspace{2pt}} c @{\hspace{2pt}} c @{\hspace{2pt}} c@{}}
  \qimgsix{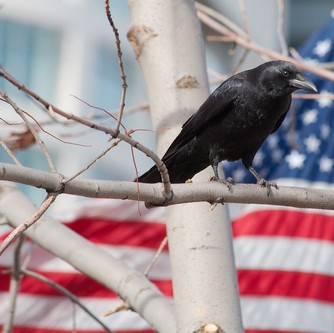} &
  \qimgsix{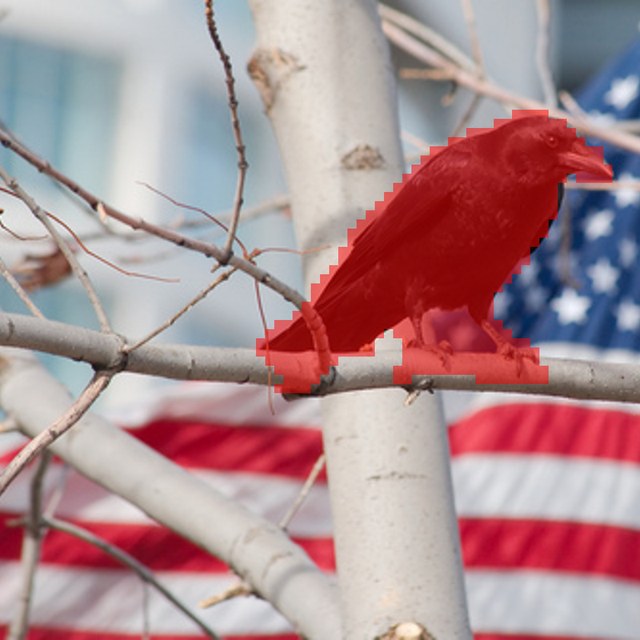} &
  \qimgsix{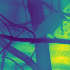} &
  \qimgsix{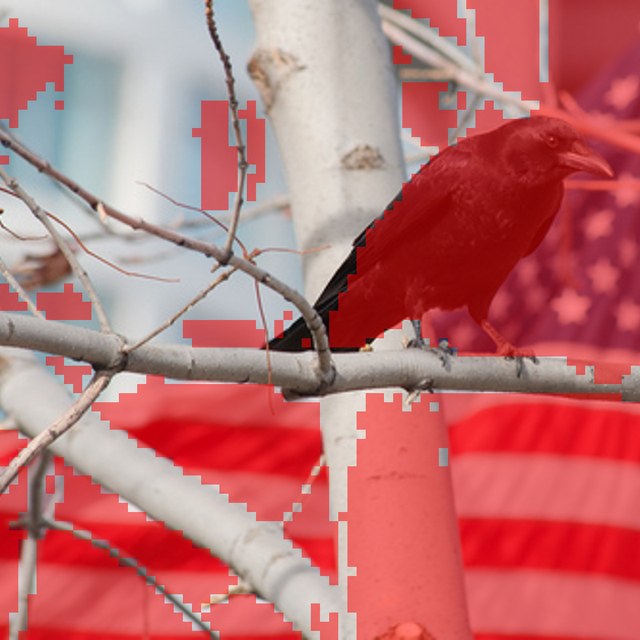} &
  \qimgsix{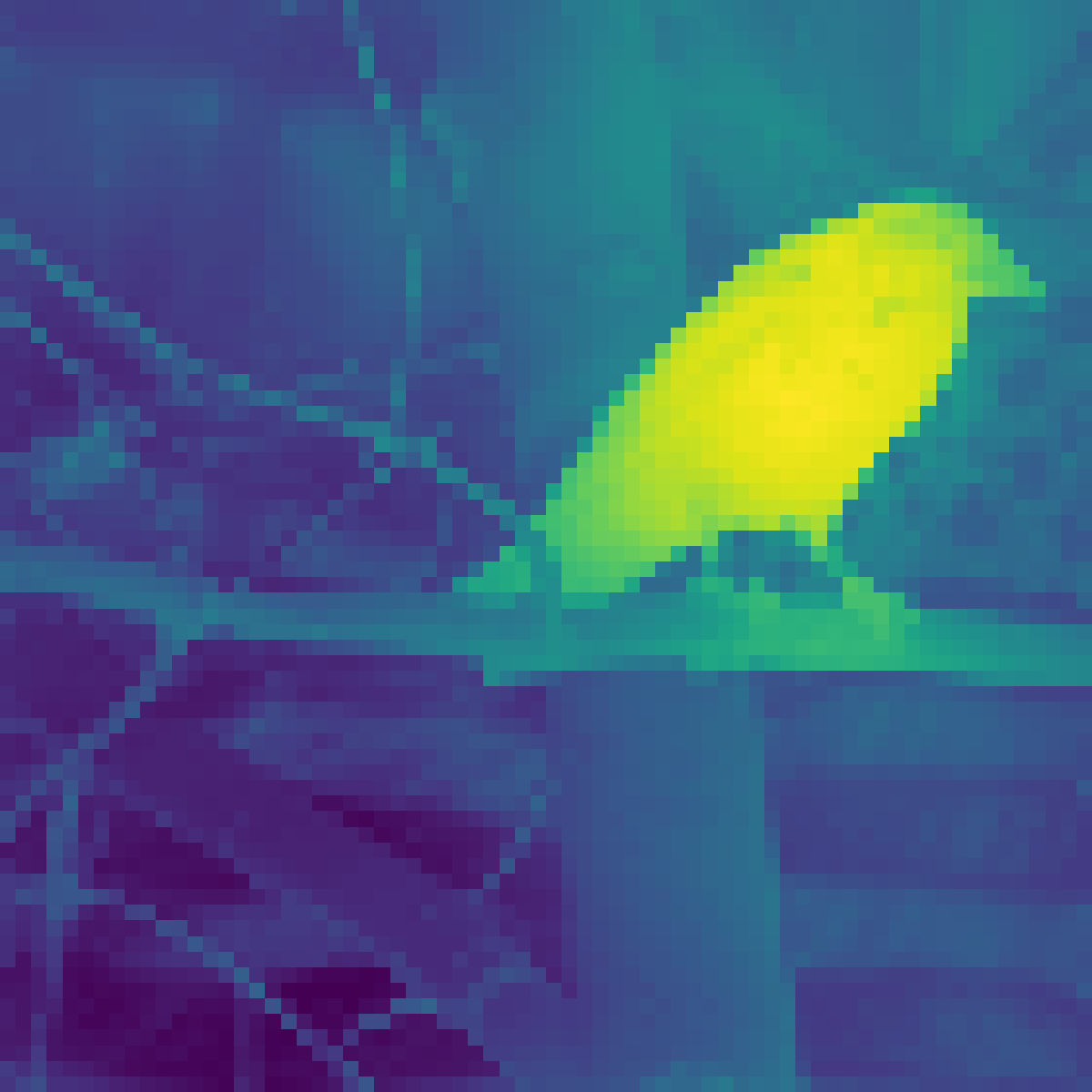} &
  \qimgsix{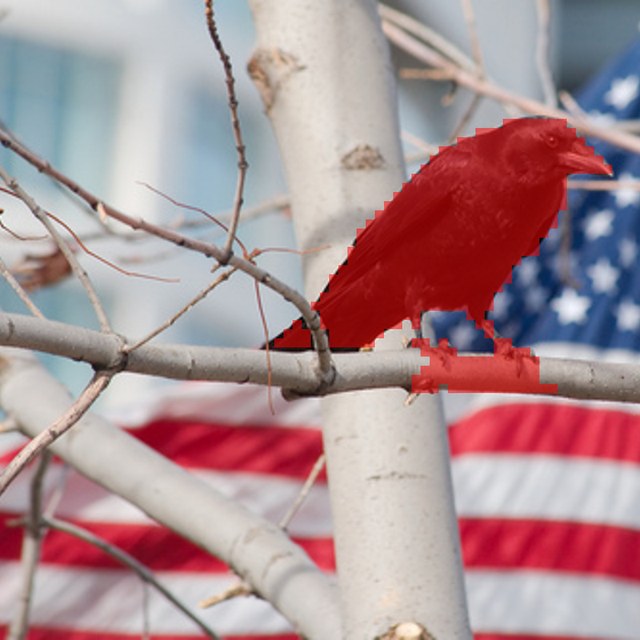} \\
  \noalign{\vspace{1pt}}

  \qimgsix{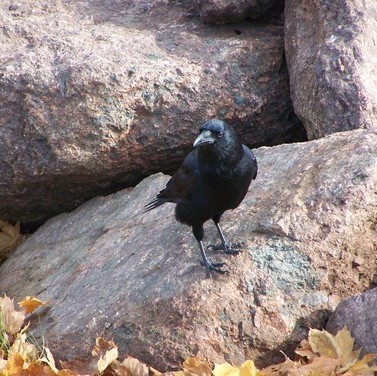} &
  \qimgsix{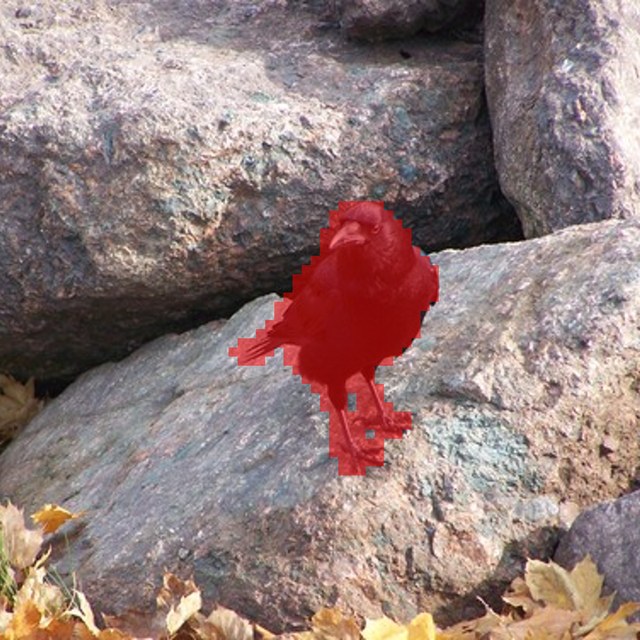} &
  \qimgsix{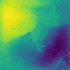} &
  \qimgsix{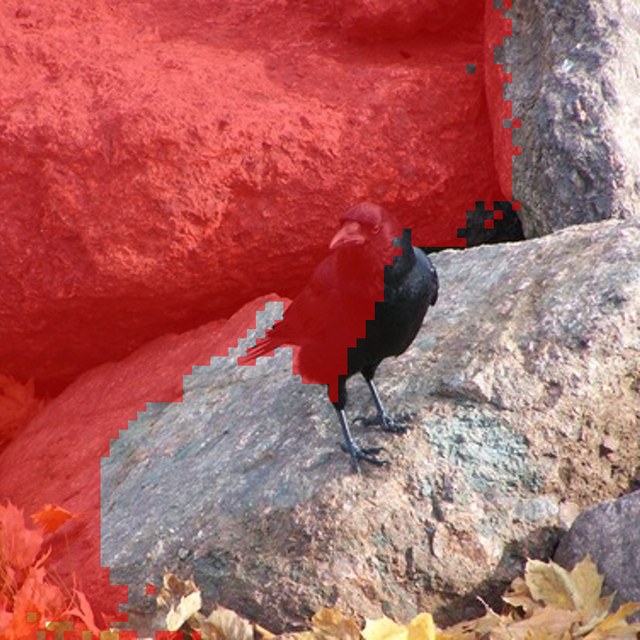} &
  \qimgsix{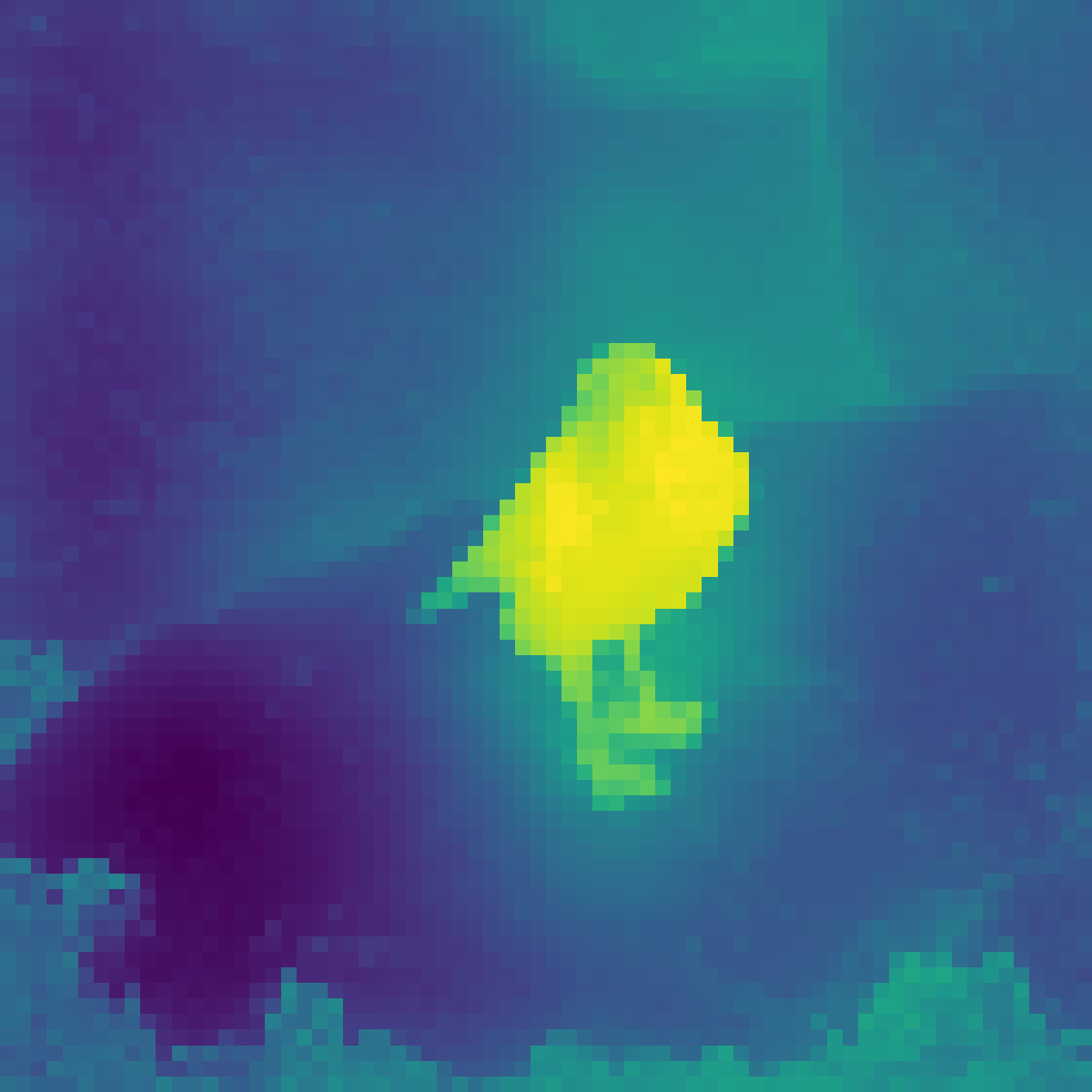} &
  \qimgsix{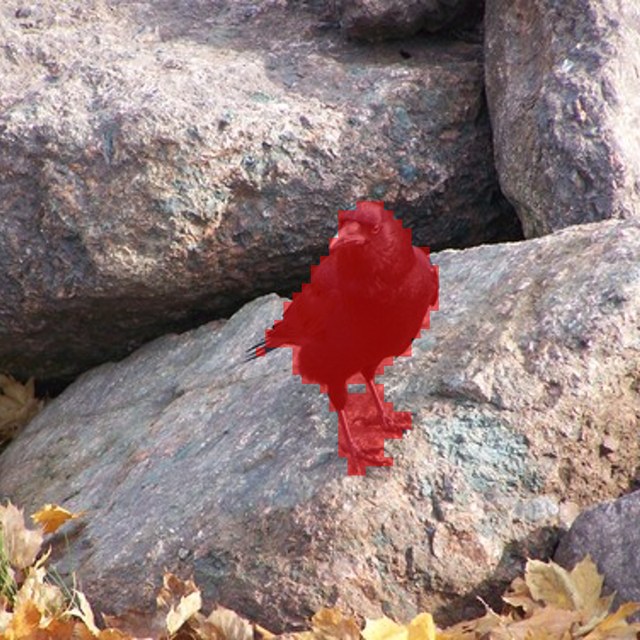} \\
  \noalign{\vspace{1pt}}

  \qimgsix{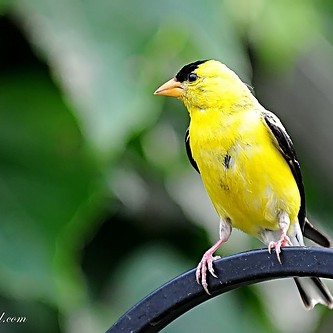} &
  \qimgsix{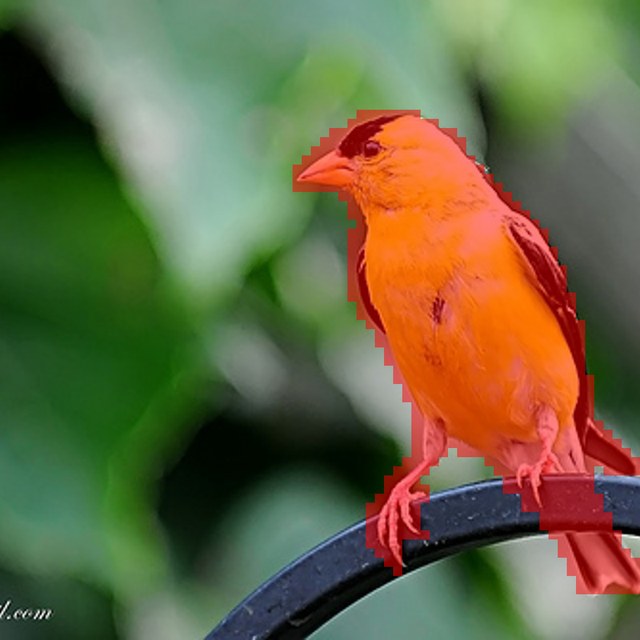} &
  \qimgsix{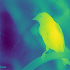} &
  \qimgsix{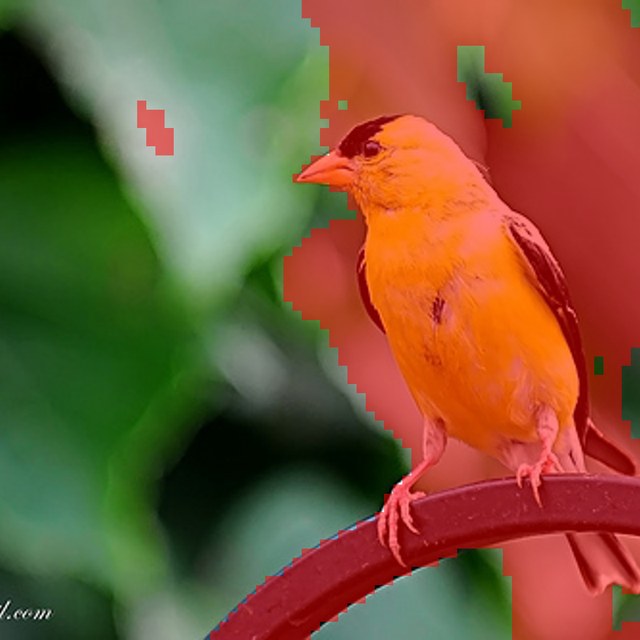} &
  \qimgsix{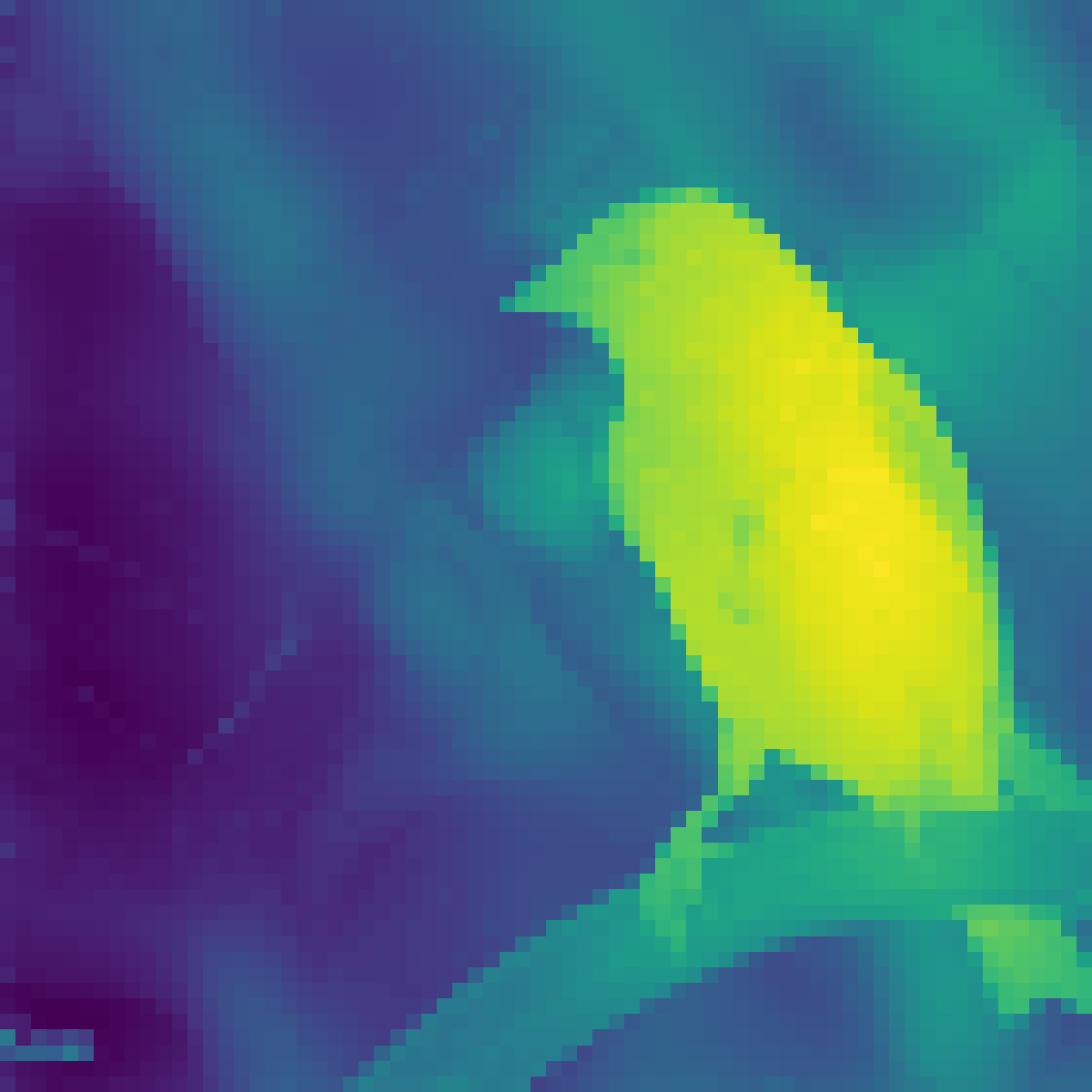} &
  \qimgsix{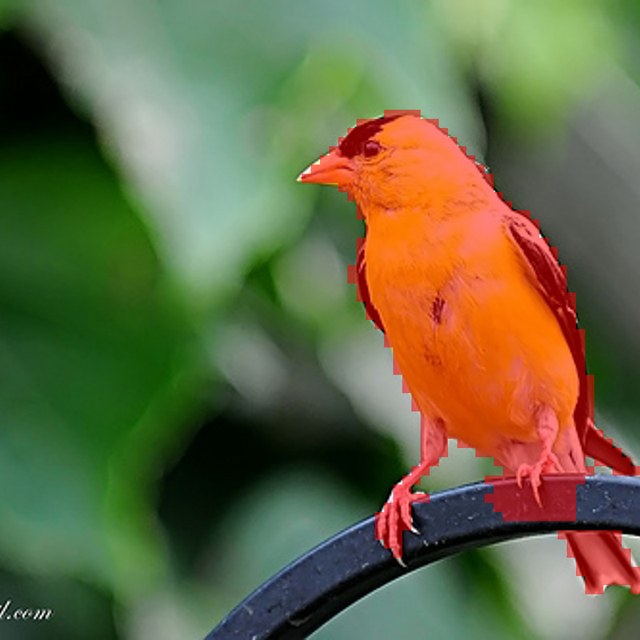} \\
  \noalign{\vspace{1pt}}

  \qimgsix{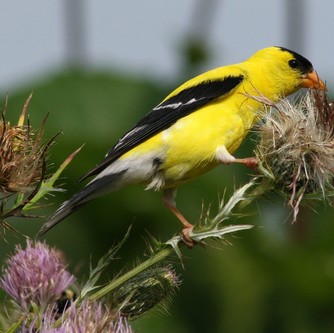} &
  \qimgsix{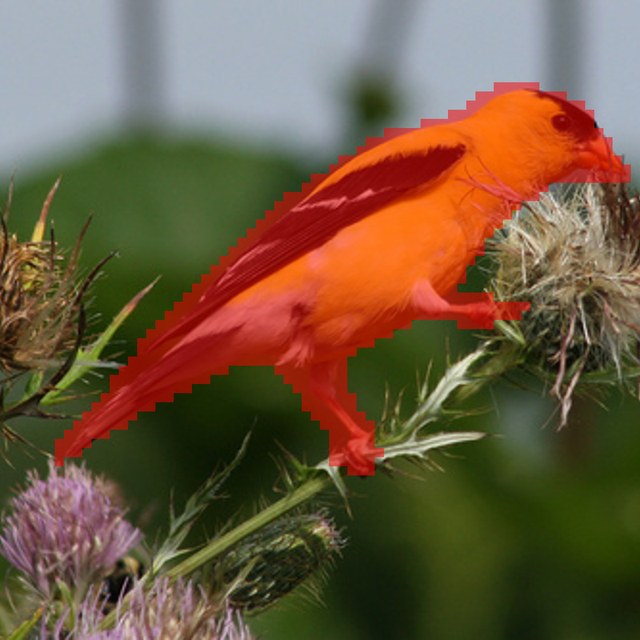} &
  \qimgsix{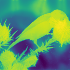} &
  \qimgsix{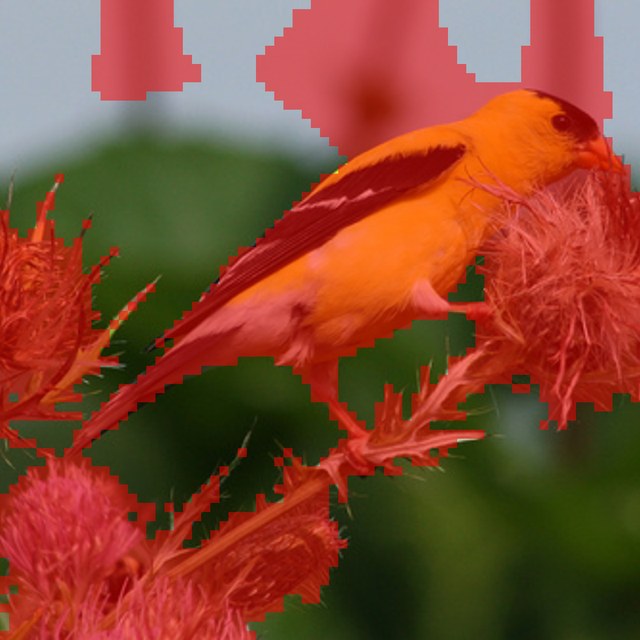} &
  \qimgsix{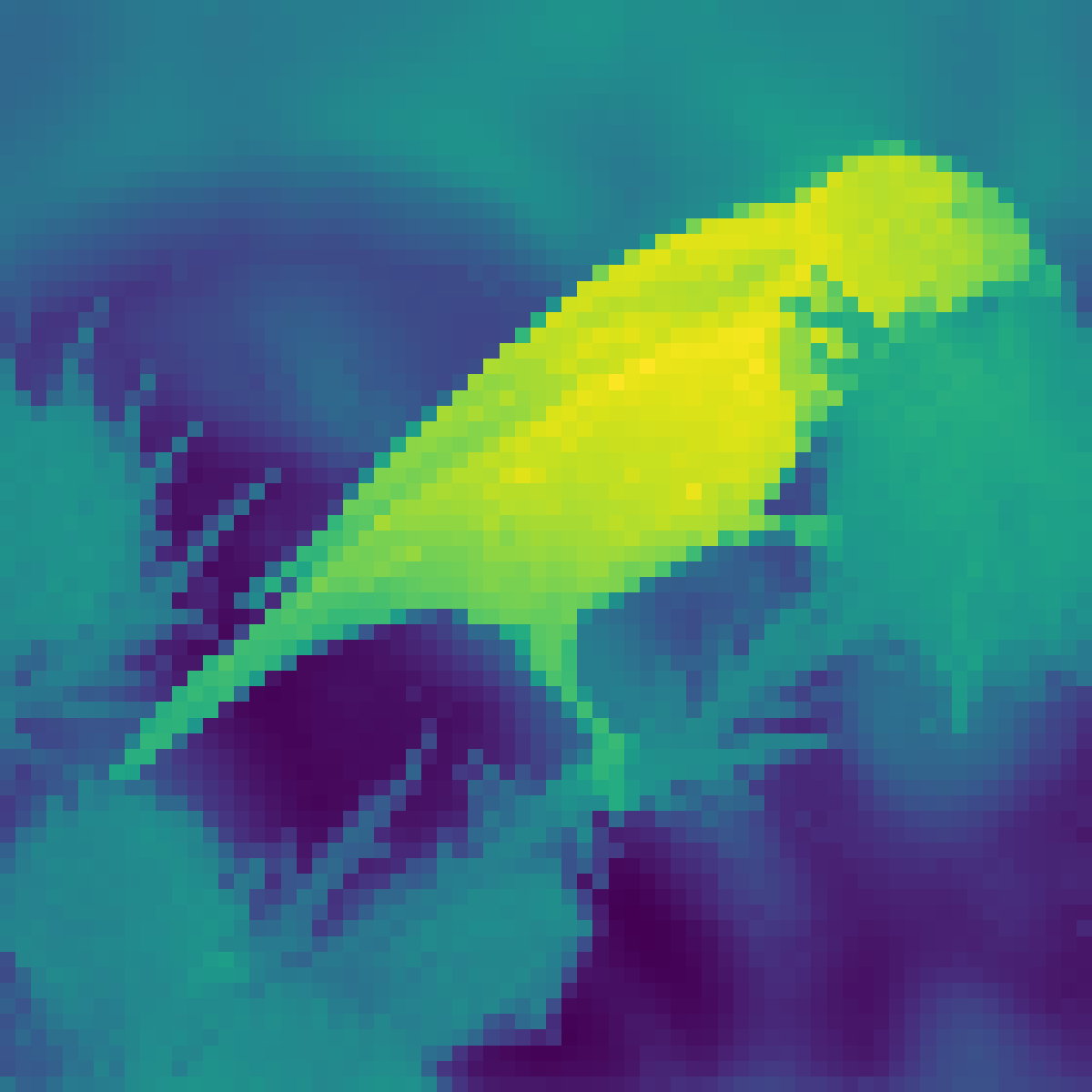} &
  \qimgsix{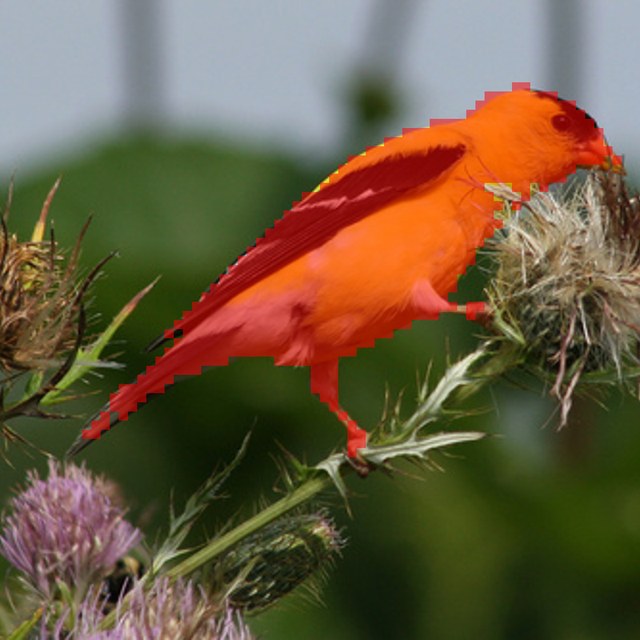} \\
  \noalign{\vspace{1pt}}

  \qimgsix{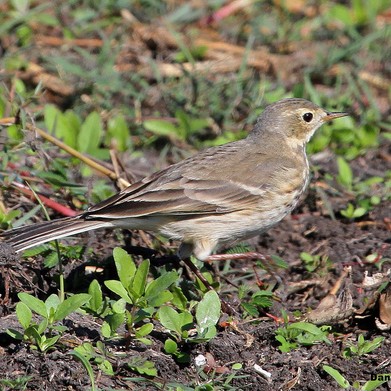} &
  \qimgsix{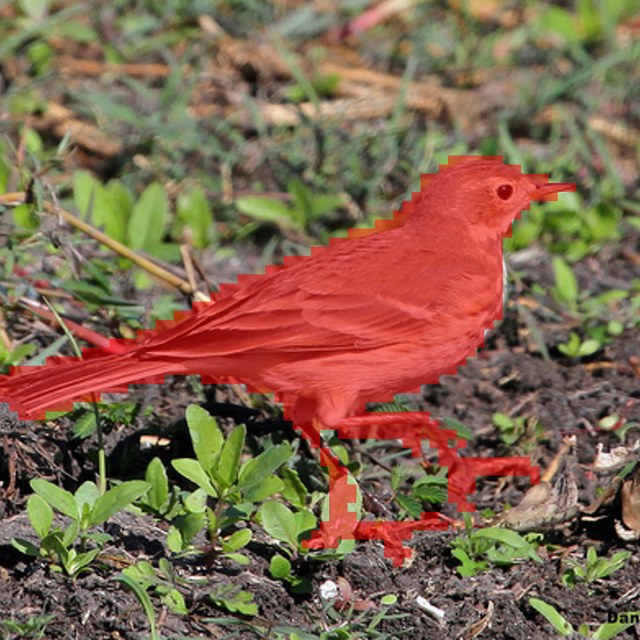} &
  \qimgsix{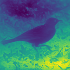} &
  \qimgsix{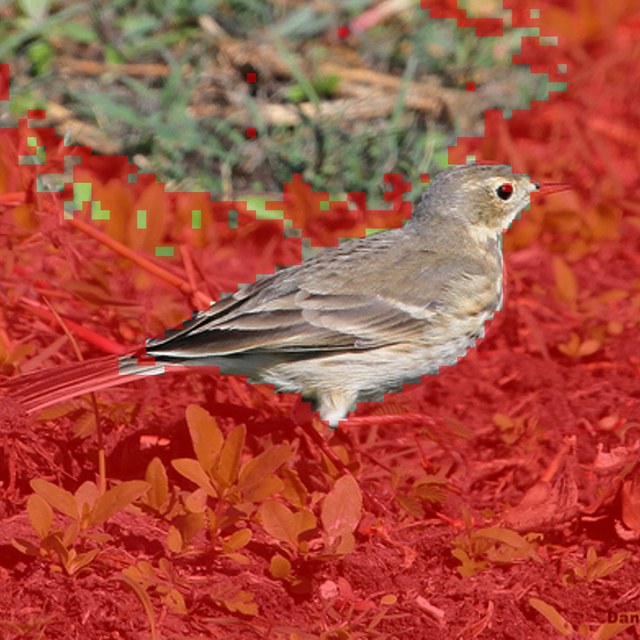} &
  \qimgsix{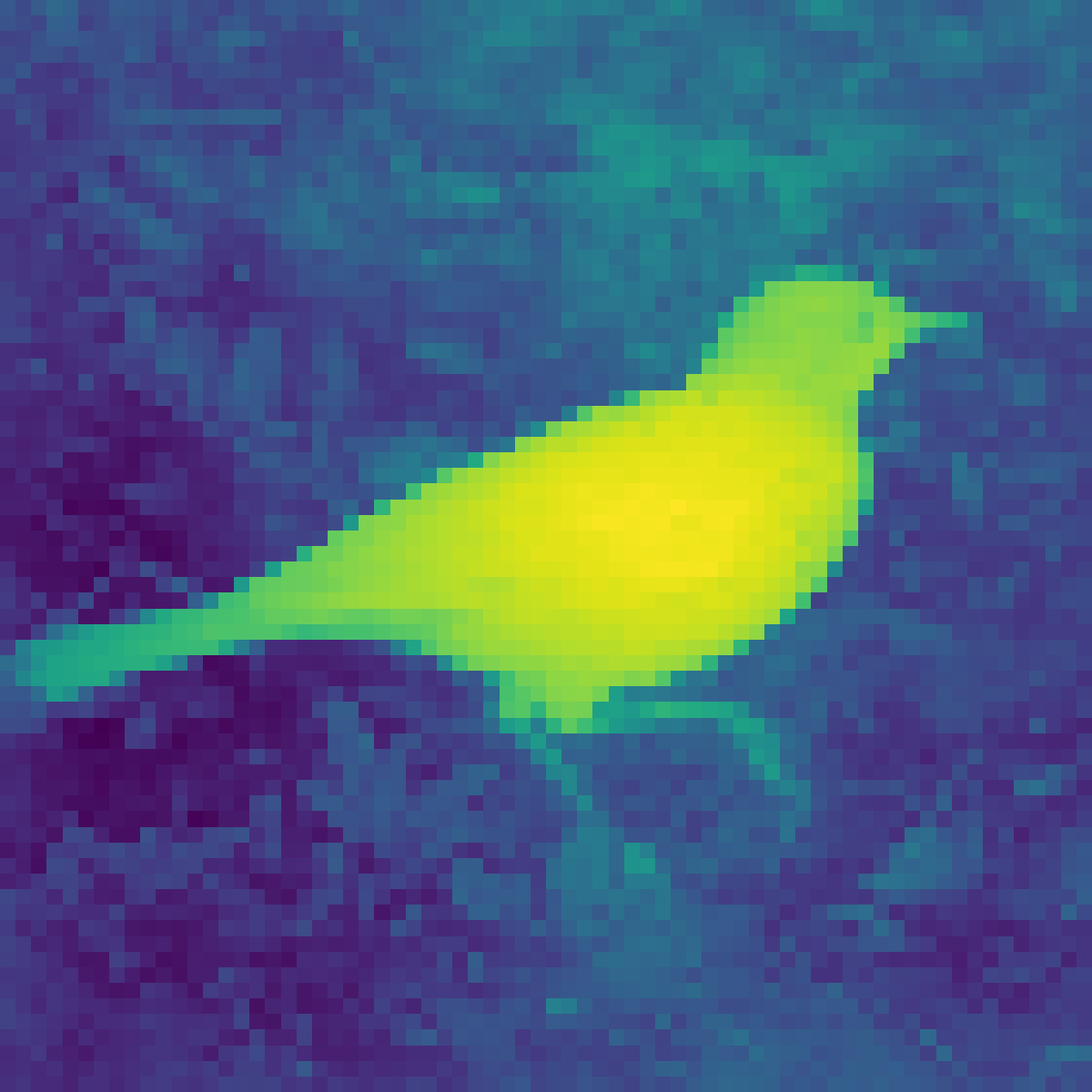} &
  \qimgsix{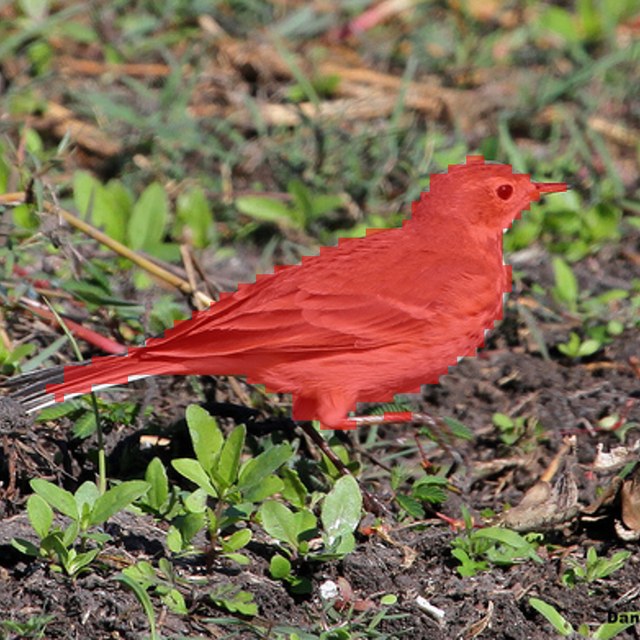} \\
  \noalign{\vspace{1pt}}

  \qimgsix{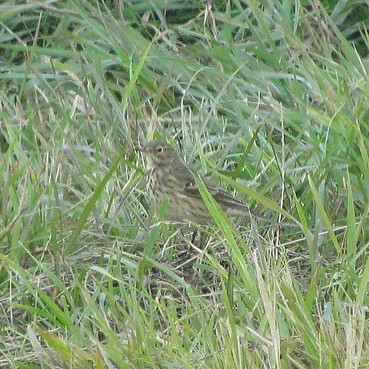} &
  \qimgsix{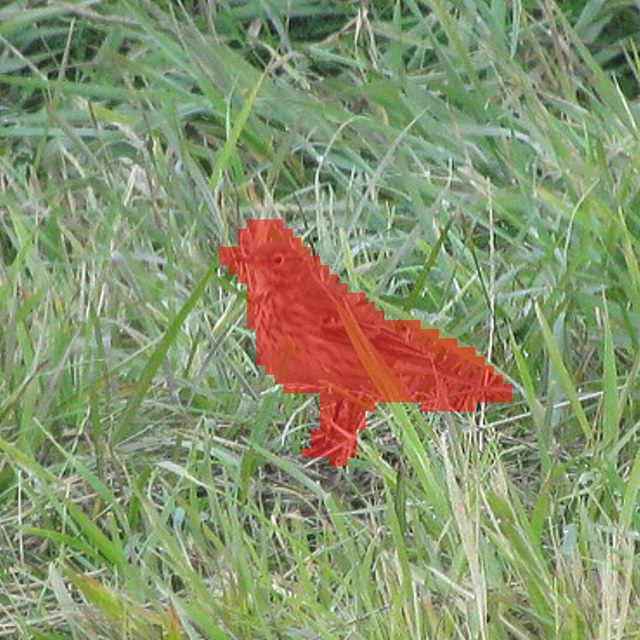} &
  \qimgsix{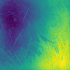} &
  \qimgsix{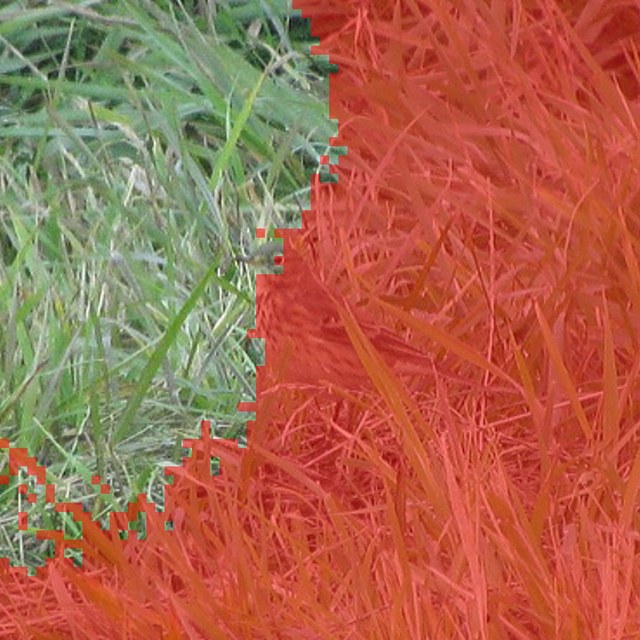} &
  \qimgsix{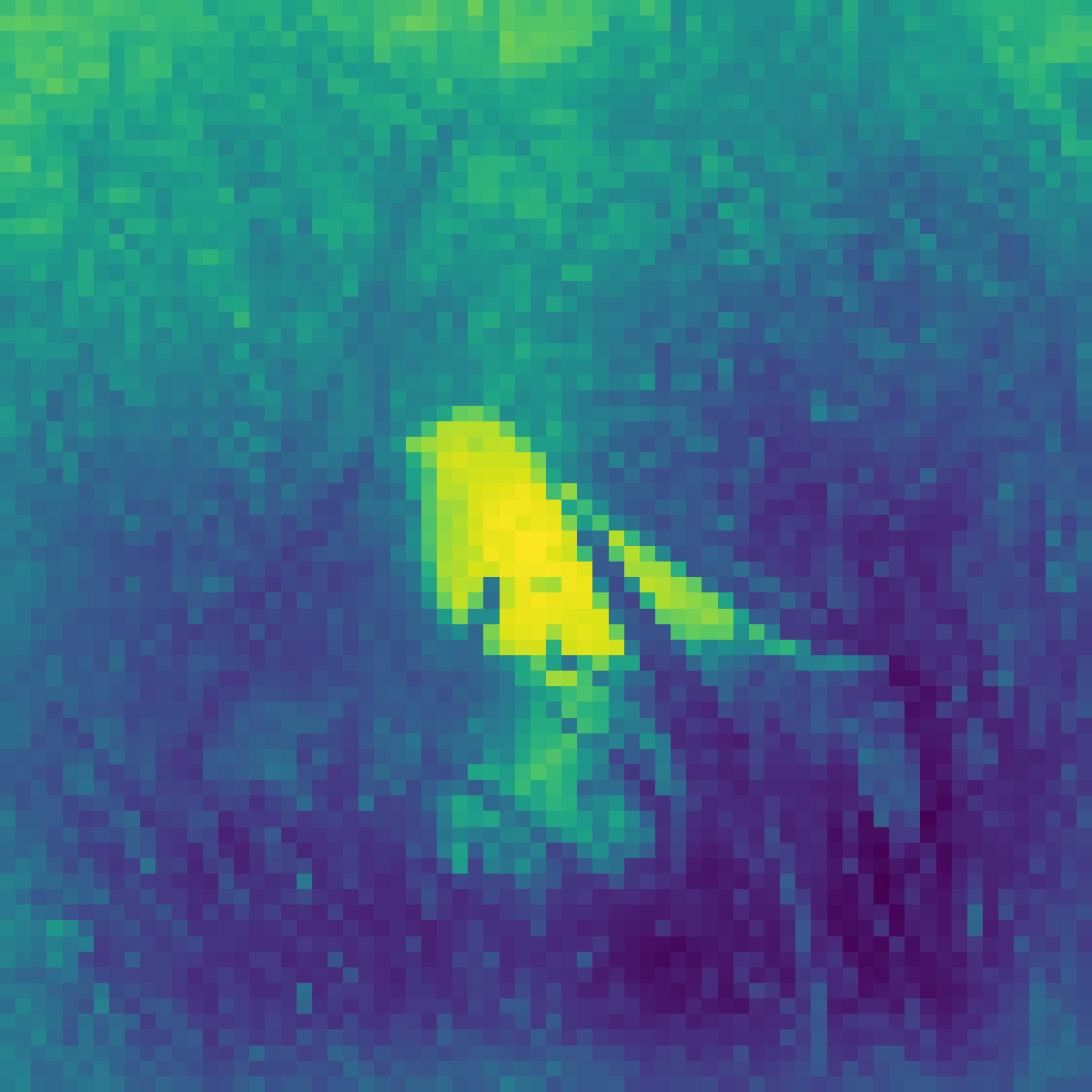} &
  \qimgsix{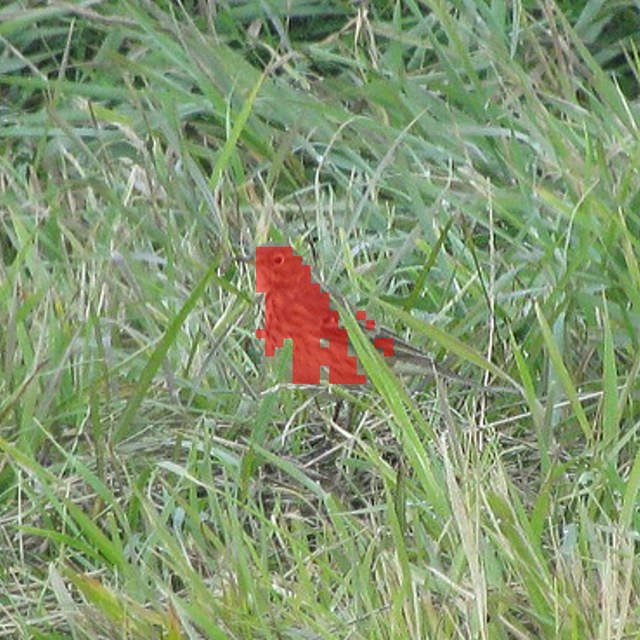} \\
  \noalign{\vspace{1pt}}

  \qimgsix{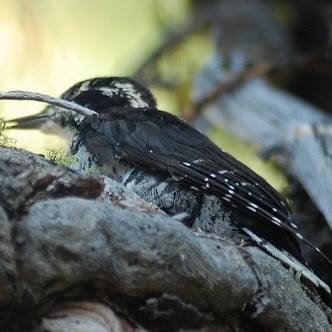} &
  \qimgsix{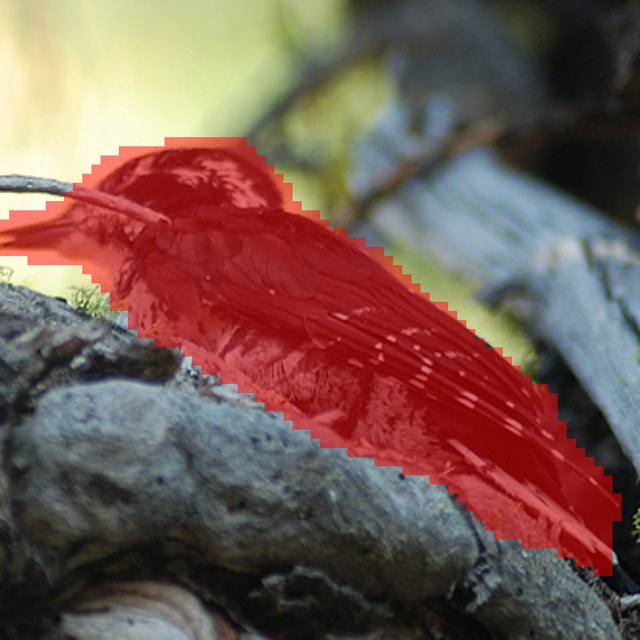} &
  \qimgsix{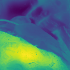} &
  \qimgsix{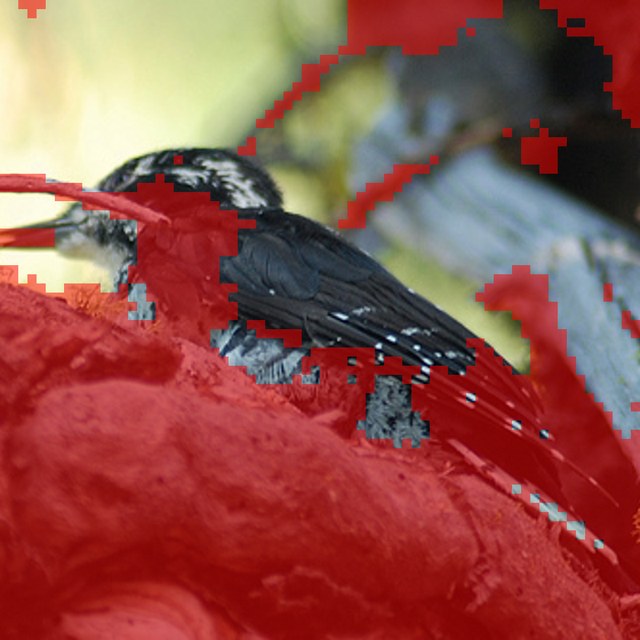} &
  \qimgsix{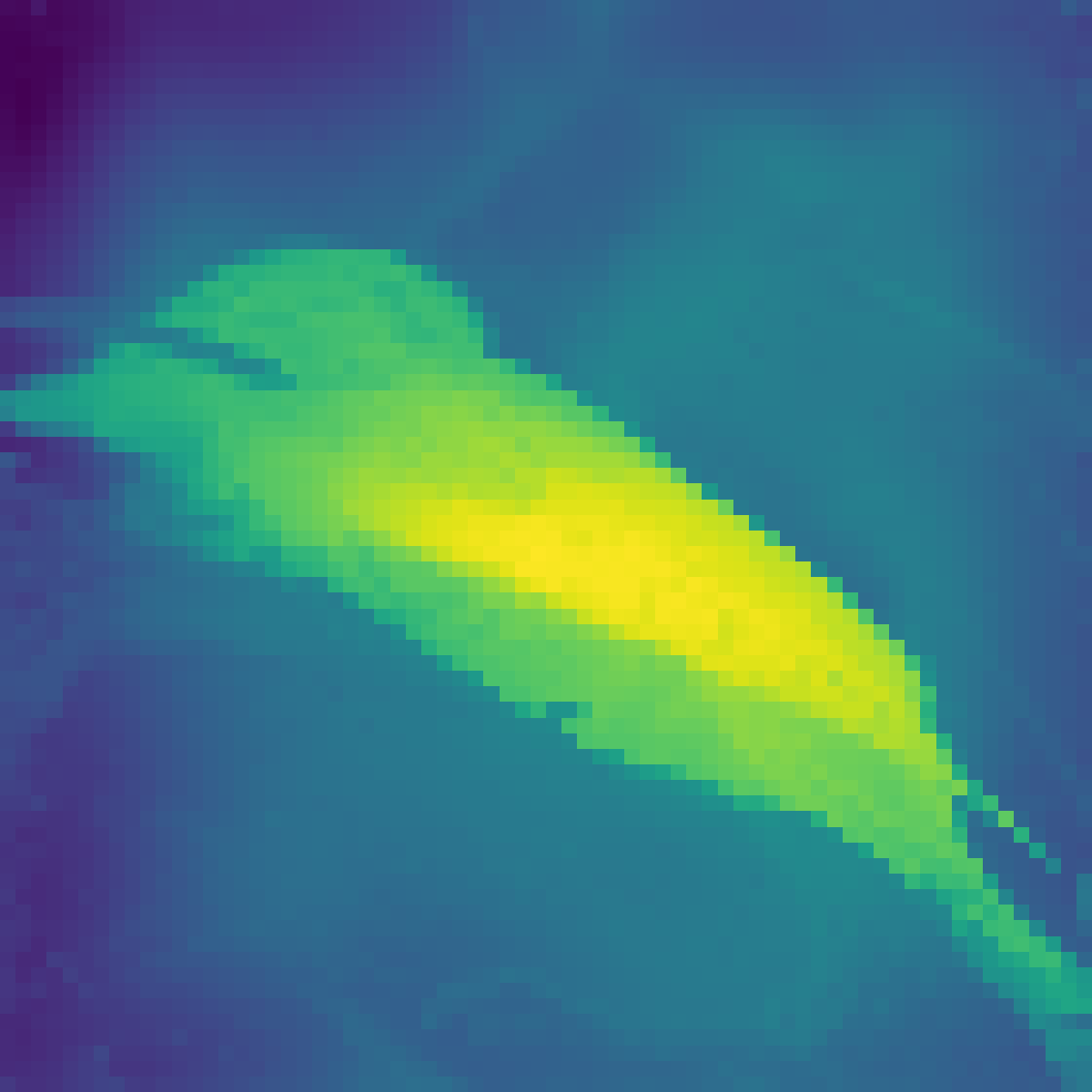} &
  \qimgsix{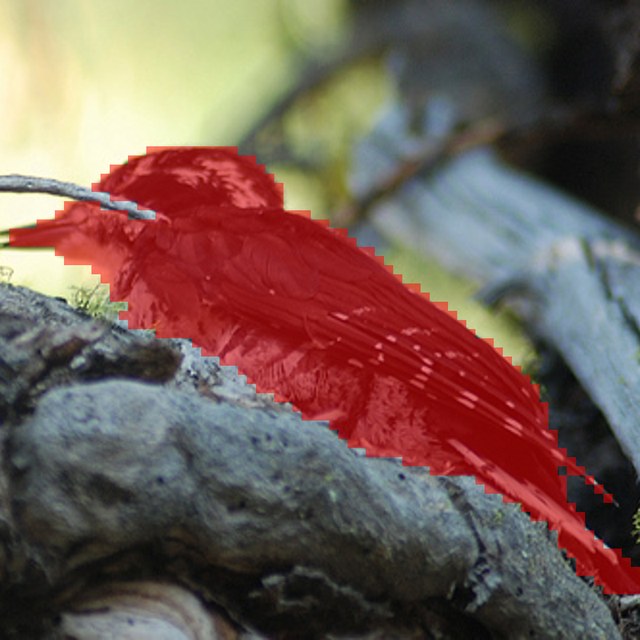} \\
  \noalign{\vspace{2pt}}

  \raisebox{-0.5\height}{\small Input} &
  \raisebox{-0.5\height}{\small GT} &
  \raisebox{-0.5\height}{\small\shortstack{Unsup.\\Eigen Attn.}} &
  \raisebox{-0.5\height}{\small\shortstack{Unsup.\\Mask}} &
  \raisebox{-0.5\height}{\small\shortstack{PANC\\Eigen Attn.}} &
  \raisebox{-0.5\height}{\small\shortstack{PANC\\Mask}} \\
\end{tabular}
\end{adjustbox}
\endgroup
\caption{Additional qualitative comparison on the CUB-200-2011 dataset.}
\label{fig:add_cub_examples}
\end{figure*}

\subsection{Known Weaknesses and Limitations}

We have outlined key limitations of our method for segmentation. A core challenge in our weakly supervised approach is selecting priors for heterogeneous datasets with high variability and annotation errors, hindering collection of reliable representatives for every class.

\begin{figure*}[ht]
\centering
\begin{adjustbox}{max width=\linewidth, max height=0.85\textheight}
\begin{tabular}{c @{\hspace{2pt}} c @{\hspace{2pt}} c @{\hspace{2pt}} c}
  
  \raisebox{-0.5\height}{\includegraphics[width=0.24\linewidth]{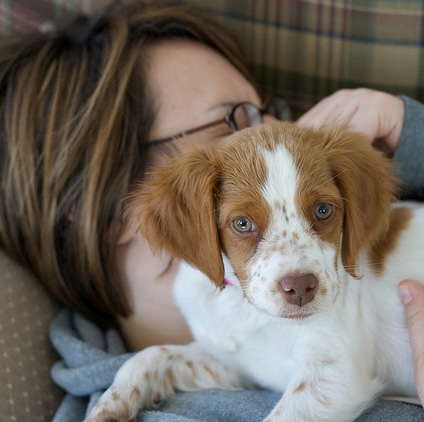}} &
  \raisebox{-0.5\height}{\includegraphics[width=0.24\linewidth]{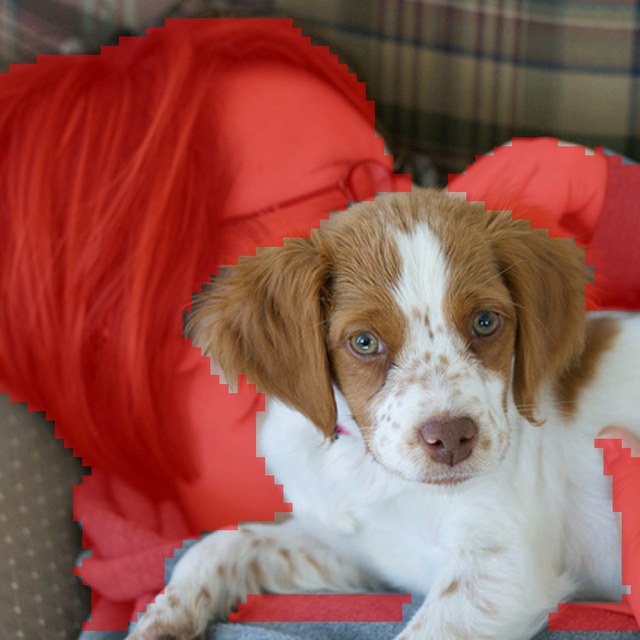}} &
  \raisebox{-0.5\height}{\includegraphics[width=0.24\linewidth]{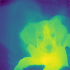}} &
  \raisebox{-0.5\height}{\includegraphics[width=0.24\linewidth]{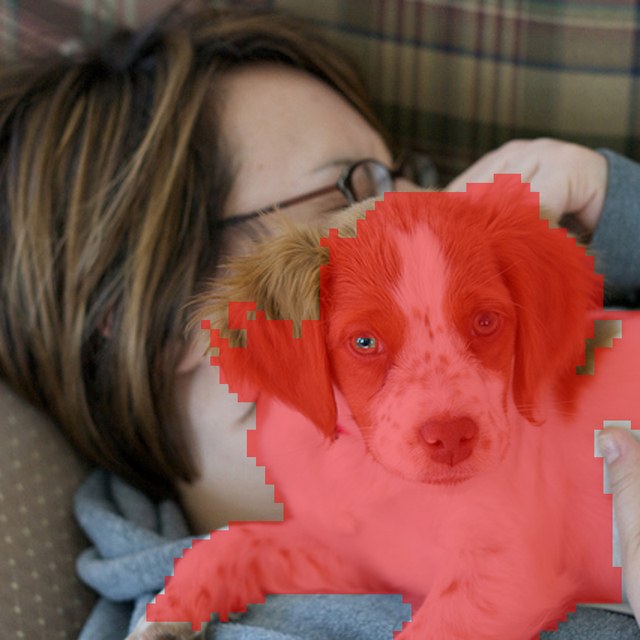}} \\
  \noalign{\vspace{2pt}}

  \raisebox{-0.5\height}{\includegraphics[width=0.24\linewidth]{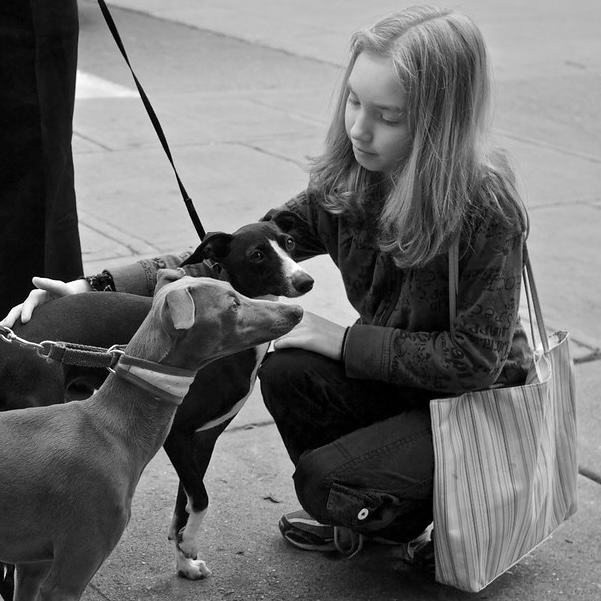}} &
  \raisebox{-0.5\height}{\includegraphics[width=0.24\linewidth]{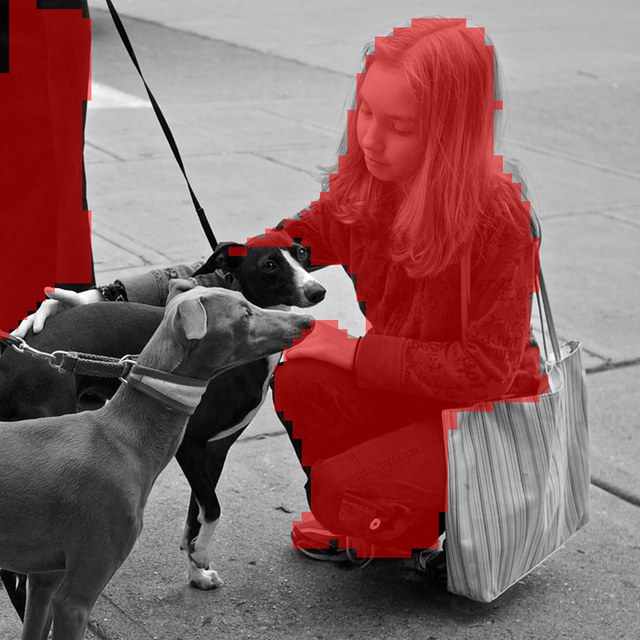}} &
  \raisebox{-0.5\height}{\includegraphics[width=0.24\linewidth]{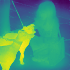}} &
  \raisebox{-0.5\height}{\includegraphics[width=0.24\linewidth]{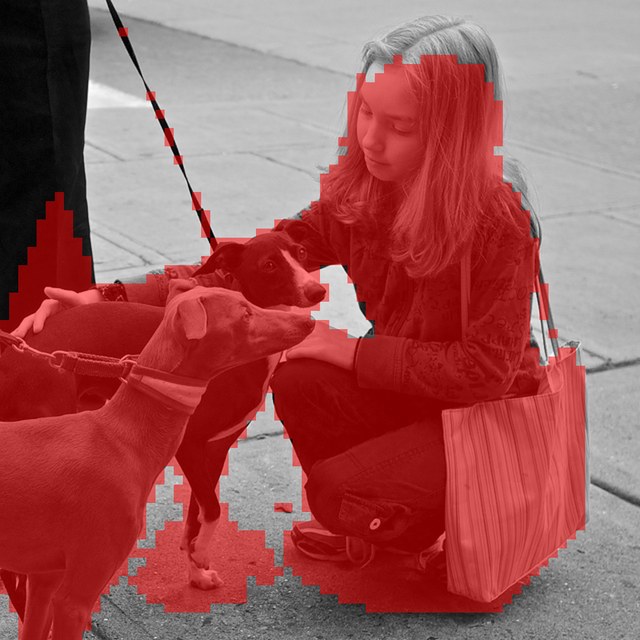}} \\
  \noalign{\vspace{4pt}}

  \raisebox{-0.5\height}{Input} & 
  \raisebox{-0.5\height}{GT} & 
  \raisebox{-0.5\height}{\shortstack{Eigen\\Attn.}} & 
  \raisebox{-0.5\height}{Mask} \\
\end{tabular}
\end{adjustbox}
\caption{Examples of the missed prior selection leading to inaccurate segmentation of the target class.}
\label{fig:wrong_images}
\end{figure*}

A prior bank from $L$ representative classes (e.g., $L=10$, classes ${1, 2, \dots, 10}$) guides attention, but outliers in underrepresented classes may be mis-segmented toward the nearest prior---as shown in Figure~\ref{fig:wrong_images}.

Ultimately, the success of the method relies on prior quality. Consequently, it requires highly reliable annotations. In practice, PANC may contribute to computer-assisted annotation, accelerating workflows. Future work should quantify segmentation time, per-item cost, and scaling with object density/diversity to assess trade-offs.

\end{document}